\documentclass[a4paper,12pt]{article}
\usepackage[margin=1in]{geometry}

\usepackage[utf8]{inputenc}

\usepackage[authoryear, round]{natbib}
\usepackage{amsmath}
\usepackage{hyperref}
\usepackage{bbm}
\usepackage{xspace}
\usepackage{xcolor}
\usepackage{booktabs}
\usepackage{multicol}
\usepackage{algorithm}
\usepackage{algpseudocode}
\usepackage{graphicx}
\usepackage{subcaption}
\usepackage{authblk}
\usepackage{siunitx}
\usepackage{tikz}
\usetikzlibrary{shapes,arrows,positioning,fit,calc}
\newcommand{\elbo}{\ensuremath{L(\boldsymbol{\theta})}\xspace}
\newcommand{\lowerBound}{\ensuremath{\tilde{L}(\boldsymbol{\theta}, \auxiliary)}\xspace}
\newcommand{\targetDist}{\ensuremath{p(\mathbf{x})}\xspace}
\newcommand{\targetDistUnnormalized}{\ensuremath{\tilde{p}(\mathbf{x})}\xspace}
\newcommand{\approximatedDist}{\ensuremath{q(\mathbf{x};\boldsymbol{\theta})}\xspace}
\newcommand{\approximatedDistIter}[1]{\ensuremath{q(\mathbf{x};\boldsymbol{\theta}^{(#1)})}\xspace}
\newcommand{\approximatedDistNoTheta}{\ensuremath{q(\mathbf{x})}\xspace}
\newcommand{\weightsTheta}{\ensuremath{q(o;\boldsymbol{\theta})}\xspace}
\newcommand{\componentTheta}{\ensuremath{q(\mathbf{x}|o;\boldsymbol{\theta})}\xspace}

\newcommand{\responsibilitiesIter}[1]{\ensuremath{q(o|\mathbf{x};\boldsymbol{\theta}^{(#1)})}\xspace}
\newcommand{\auxiliary}{\ensuremath{\tilde{q}(o|\mathbf{x})}\xspace}

\newcommand{\weights}{\ensuremath{q(o)}\xspace}
\newcommand{\component}{\ensuremath{q(\mathbf{x}|o)}\xspace}
\newcommand{\responsibilities}{\ensuremath{q(o|\mathbf{x})}\xspace}

\newcommand{\argmin}{\ensuremath{\arg \min\;}\xspace}
\newcommand{\argmax}{\ensuremath{\arg \max\;}\xspace}
\newcommand{\reward}{\ensuremath{R(\mathbf{x})}\xspace}
\newcommand{\compReward}{\ensuremath{R_o(\mathbf{x})}\xspace}
\newcommand{\compRewardSurrogate}{\ensuremath{\tilde{R}_o(\mathbf{x})}\xspace}
\newcommand{\rewardSurrogate}{\ensuremath{\tilde{R}(\mathbf{x})}\xspace}
\newcommand{\weightReward}{\ensuremath{{R}(o)}\xspace}
\newcommand{\weightRewardApprox}{\ensuremath{\tilde{R}(o)}\xspace}
\newcommand{\entropy}{\ensuremath{\text{H}}}
\newcommand{\gaussian}{\ensuremath{\mathcal{N}}}

\tikzstyle{decision} = [diamond, draw,  
    text width=4.5em, text badly centered, node distance=3cm, inner sep=0pt]
\tikzstyle{block} = [rectangle, draw, 
    text width=5em, text centered, minimum height=4em]
\tikzstyle{line} = [draw, -latex']

\tikzstyle{db} = [draw, cloud, cloud puffs=10,cloud puff arc=120,fill=red!20, node distance=2cm,
    minimum height=2em, text width=2cm, text centered]
\tikzstyle{startstop} = [rectangle, draw, 
    text width=5em, text centered, rounded corners, minimum height=4em] 
\makeatletter
\tikzset{
  link/.style    = {white, double = black, line width = 1.8pt,
                     double distance = 0.8pt },
  channel/.style = {white, double = black, line width = 0.8pt,
                     double distance = 0.6pt },
    database/.style={
        path picture={
            \draw (0, 1.5*\database@segmentheight) circle [x radius=\database@radius,y radius=\database@aspectratio*\database@radius];
            \draw (-\database@radius, 0.5*\database@segmentheight) arc [start angle=180,end angle=360,x radius=\database@radius, y radius=\database@aspectratio*\database@radius];
            \draw (-\database@radius,-0.5*\database@segmentheight) arc [start angle=180,end angle=360,x radius=\database@radius, y radius=\database@aspectratio*\database@radius];
            \draw (-\database@radius,1.5*\database@segmentheight) -- ++(0,-3*\database@segmentheight) arc [start angle=180,end angle=360,x radius=\database@radius, y radius=\database@aspectratio*\database@radius] -- ++(0,3*\database@segmentheight);
        },
        minimum width=2*\database@radius + \pgflinewidth,
        minimum height=3*\database@segmentheight + 2*\database@aspectratio*\database@radius + \pgflinewidth,
    },
    database segment height/.store in=\database@segmentheight,
    database radius/.store in=\database@radius,
    database aspect ratio/.store in=\database@aspectratio,
    database segment height=0.1cm,
    database radius=0.25cm,
    database aspect ratio=0.35,
}
\makeatother

\begin{document}

\title{Trust-Region Variational Inference with Gaussian Mixture Models}
\author[1]{Oleg Arenz}
\author[2]{Mingjun Zhong}
\author[3,4]{Gerhard Neumann}
\affil[1]{Intelligent Autonomous Systems, TU Darmstadt}
\affil[2]{Department of Computing Science, University of Aberdeen}
\affil[3]{Autonomous Learning Robots, Karlsruhe Institute of Technology}
\affil[4]{Bosch Center for Artificial Intelligence, Renningen}

\date{}                     
\setcounter{Maxaffil}{0}
\renewcommand\Affilfont{\itshape\small}

       
\maketitle

\begin{abstract}
Many methods for machine learning rely on approximate inference from intractable probability distributions. Variational inference approximates such distributions by tractable models that can be subsequently used for approximate inference.
Learning sufficiently accurate approximations requires a rich model family and careful exploration of the relevant modes of the target distribution. We propose a method for learning accurate GMM approximations of intractable probability distributions based on insights from policy search by using information-geometric trust regions for principled exploration. For efficient improvement of the GMM approximation, we derive a lower bound on the corresponding optimization objective enabling us to update the components independently. Our use of the lower bound ensures convergence to a stationary point of the original objective. The number of components is adapted online by adding new components in promising regions and by deleting components with negligible weight.
We demonstrate on several domains that we can learn approximations of complex, multimodal distributions with a quality that is unmet by previous variational inference methods, and that the GMM approximation can be used for drawing samples that are on par with samples created by state-of-the-art {\sc{MCMC}} samplers while requiring up to three orders of magnitude less computational resources.
\end{abstract}

\section{Introduction}
Inference from a complex distribution \targetDist is a huge problem in machine learning that is needed in many applications. Typically, we can evaluate the distribution except for the normalization factor $Z$, that is, we can only evaluate the unnormalized distribution \targetDistUnnormalized, where $$\targetDist = \targetDistUnnormalized / Z,$$ with $Z = \int_{\mathbf{x}} \targetDistUnnormalized d\mathbf{x}$. For example, in Bayesian inference \targetDistUnnormalized would correspond to the product of prior and likelihood. As exact inference is often intractable, we have to rely on approximate inference.

Markov chain Monte Carlo ({\sc{MCMC}}) is arguably the most commonly applied technique for approximate inference. Samples are drawn from the desired distribution by building Markov chains for which the equilibrium distribution matches the desired distribution \targetDist. Monte Carlo estimates based on these samples are then used for inference. However, {\sc{MCMC}} can be very inefficient, because it is difficult to make full use of function evaluations of \targetDistUnnormalized without violating the Markov assumption.

Instead, we propose a method based on variational inference, which is another commonly applied technique for approximate inference. In variational inference, the desired distribution \targetDist is approximated by a tractable distribution \approximatedDist which can be used for exact inference instead of \targetDist, or as a more direct alternative to {\sc{MCMC}} for drawing samples for (possibly importance weighted) Monte Carlo estimates. The approximation \approximatedDist is typically found by minimizing the reverse Kullback-Leibler (KL) divergence
\begin{equation}
\text{KL}\left(\approximatedDist||\targetDist\right) 
= \int_{\mathbf{x}} \approximatedDist \log{\left( \frac{\approximatedDist}{\targetDist} \right)} d\mathbf{x} \label{eq:VIKL},
\end{equation}
with respect to the parameters $\boldsymbol{\theta}$ of the approximation. 

By framing inference as an optimization problem, variational inference can make better use of previous function evaluations of \targetDistUnnormalized than {\sc{MCMC}} and is therefore computationally more efficient. However, in order to perform the KL minimization efficiently, \approximatedDist is often restricted to belonging to a simple family of models or is assumed to have non-correlating degrees of freedom~\citep{Blei2017,Peterson1989}, which is known as the mean field approximation. Unfortunately, such restrictions can introduce significant approximation error especially for multimodal target distributions. Comparing {\sc{MCMC}} with variational inference, we can conclude that we should use {\sc{MCMC}} when we require accuracy (due to its asymptotic guarantee of exactness), whereas we should prefer variational inference when we need computationally efficient solutions~\citep{Blei2017}. 

Hence, there is a huge interest in finding computationally efficient solutions with high sample quality. 
Our work aims at learning highly accurate approximations for computationally efficient variational inference methods. We use Gaussian mixture models (GMMs) as model family, because they can be sampled efficiently and are capable of representing any target distribution arbitrarily well if the number of components is sufficiently large. As the required number of components is typically not known a priori, we dynamically add or delete components during optimization.

A major challenge of learning highly accurate approximations of multimodal distributions is to achieve stable and efficient optimization of an intractable objective function. We derive a lower bound on the KL divergence (Equation~\ref{eq:VIKL}) based on a decomposition that is related to the one used by the expectation-maximization procedure for fitting GMMs for density estimation. We can thus optimize the original objective by iteratively maximizing and tightening this lower bound. Maximizing the lower bound decomposes into independent sub-problems for each Gaussian component that are solved, analogously to the policy search method MORE~\citep{Abdolmaleki2015}, based on local quadratic approximations. Due to its strong ties to policy search, we call our method Variational Inference by Policy Search ({\sc{VIPS}}). 

Another major challenge when striving for high quality approximations is to discover the relevant modes of the target distribution. The areas of high density are initially unknown and have to be discovered during learning based on function evaluations of \targetDistUnnormalized. The unnormalized target distribution, however, is typically evaluated at locations that have been sampled from the current approximation \approximatedDist, because these samples are well suited for the optimization, for example for approximating the objective (Equation ~\ref{eq:VIKL}) or its gradient. The current approximation thus serves as search distribution and needs to be adapted carefully in order to avoid erroneously discarding important regions. 
The conflicting goals of moving the approximation towards high density regions and evaluating \targetDistUnnormalized at unexplored regions can be seen as an instance of the exploration-exploitation dilemma that is well-known in reinforcement learning~\citep{SuttonBarto1998} but currently hardly addressed by the variational inference community. 

Our proposed method leverages insights from policy search~\citep{Deisenroth2013}, a sub-field of reinforcement learning, by bounding the KL divergence between the updated approximation and the current approximation at each learning step. This information-geometric trust region serves the dual-purpose of staying in the validity of the local quadratic models as well as ensuring careful exploration of the search space. By finding the best approximation within such information-geometric trust region, we limit the change in search space while making sufficient progress during each iteration. However, information-geometric trust regions only address local exploration in the vicinity of the components of the current approximation and may in practice still discard regions of the search space prematurely. In order to discover modes that are not covered by the current approximation, we dynamically create new mixture components at interesting regions. Namely, we add additional components at regions where the current approximation has little probability mass although we suspect a mode of the target distribution based on previous function evaluations. 

We evaluate {\sc{VIPS}} on several domains and compare it to state-of-the-art methods for variational inference and Markov-chain Monte Carlo. We demonstrate that we can learn high quality approximations of several challenging multimodal target distributions that are significantly better than those learned by competing methods for variational inference. Compared to sampling methods, we show that we can achieve similar sample quality while using several orders of magnitude less function evaluations. Samples from the learned approximation can therefore often be used directly for approximate inference without needing importance weighting. Still, knowing the actual generative model can be a further advantage compared to model-free samplers. 

This work extends previously published work about VIPS~\citep{Arenz2018} by using more efficient sample reuse, by showcasing and fixing a failure case of the previous initialization of covariance matrices, and by several other improvements such as adaptation of regularization coefficients and KL bounds. These modifications lead to a further reduction of sample complexity by approximately one order of magnitude. We will refer to the improved version as {\sc{VIPS++}}.
We evaluate {\sc{VIPS++}} on additional, more challenging domains, namely Bayesian Gaussian process regression and Bayesian parameter estimation of ordinary differential equations applied to the Goodwin oscillator~\citep{Goodwin1965} as well as more challenging variations of the previously published \textit{planar robot} and \textit{Gaussian mixture model} experiments~\citep{Arenz2018}. Furthermore, we now also compare to normalizing flows~\citep{Kingma2016} and black-box variational inference~\citep{Ranganath2014}.

\section{Preliminaries}
In this section we formalize the optimization problem and show its connection to policy search. We further discuss the policy search method {\sc{MORE}}~\citep{Abdolmaleki2015} and show that a slight variation of it can be used for learning Gaussian variational approximations (GVAs) for variational inference. This variant of {\sc{MORE}} is used by {\sc{VIPS}} for independent component updates, which will be discussed in Section~\ref{sec:VIPS}.

\subsection{Problem formulation}
Variational inference is typically framed as an information projection (I-projection) problem, that is, we want to find the parameters $\boldsymbol{\theta}$ of a model \approximatedDist that minimize the KL divergence between \approximatedDist and the target distribution \targetDist,
\begin{align}
\text{KL}\left(\approximatedDist||\targetDist\right) 
&= \int_{\mathbf{x}} \approximatedDist \log{\left( \frac{\approximatedDist}{\targetDist} \right)} d\mathbf{x} \nonumber \\
&= \int_{\mathbf{x}} \approximatedDist \log{\left( \frac{\approximatedDist}{\targetDistUnnormalized} \right)} d\mathbf{x} + \log Z \nonumber
\\
&= -L(\boldsymbol{\theta}) + \log Z. \nonumber
\end{align}
The normalizer Z does not affect the optimal solution for the parameters $\boldsymbol{\theta}$ as it enters the objective function as constant offset and can thus be ignored. Hence, the KL divergence can be minimized by maximizing $L(\boldsymbol{\theta})$, which is a lower bound on the log normalizer due to the non-negativity of the KL divergence. In Bayesian inference, the target distribution \targetDist corresponds to the posterior, the unnormalized distribution \targetDistUnnormalized corresponds to the product of prior and likelihood, and the normalizer corresponds to the evidence. Minimizing the KL divergence thus corresponds to maximizing a lower bound on the (log) evidence, $L(\boldsymbol{\theta})$, which is therefore commonly referred to as the evidence lower bound objective (ELBO, e.g.,~\citealt{Blei2017}).

Although {\sc{VIPS}} is not restricted to the Bayesian setting but aims to approximate intractable distributions in general, we also frame our objective as ELBO maximization because this formulation highlights an interesting connection to policy search.
We treat information projection as the problem of finding a search distribution, $\approximatedDist$, over a parameter space $\mathbf{x}$, that maximizes an expected return $\reward = \log \targetDistUnnormalized$ with an additional objective of maximizing its entropy $\entropy\big(\approximatedDist\big)=-\int_{\mathbf{x}} \approximatedDist \log \approximatedDist d\mathbf{x}$, that is, we aim to solve
\begin{align*}
\underset{\boldsymbol{\theta}}{\argmax}  \left[ \elbo 
= \int_{\mathbf{x}} \approximatedDist \big( \log\targetDistUnnormalized - \log{\approximatedDist} \big) d\mathbf{x} \nonumber 
= \int_{\mathbf{x}} \approximatedDist \reward d\mathbf{x} + \entropy(\approximatedDist)  \right].
\end{align*}
Entropy objectives are also commonly used in policy search for better exploration~\citep{Neu2017, Abdolmaleki2015}. Policy search methods that support such entropy objectives can thus be applied straightforwardly for variational inference. However, 
many policy search methods are restricted to unimodal distributions (typically Gaussians) and are therefore not suited for learning accurate approximations of multimodal target distributions. We will now review one such policy search method, {\sc{MORE}}~\citep{Abdolmaleki2015}, and show that it can be adapted straightforwardly for learning Gaussian variational approximations.

\subsection{Model-Based Relative Entropy Stochastic Search}
Policy search methods start with an initial search distribution $q^{(0)}(\mathbf{x})$ and iteratively update it in order to increase its expected reward.\footnote{Here and in the following, we indicate variables and functions at a given iteration by using superscripts that are set in parentheses.} Areas of high reward are initially not known and have to be discovered based on evaluations of the reward function \reward during learning. Policy search methods, therefore, typically evaluate the reward function on samples from the current search distribution in order to identify regions of high reward, and update the search distribution to increase the likelihood of the search distribution in these areas. 

In order to avoid premature convergence to poor local optima, it is crucial to start with an initial search distribution $q^{(0)}$ with sufficiently high entropy and to ensure that high reward regions are not erroneously discarded due to too greedy updates. This trade-off between further exploring the search space and focusing on high reward areas is an instance of the exploration-exploitation dilemma that several policy search methods address using information-geometric trust regions~\citep{Peters2010,Levine2013,Schulman2015,Abdolmaleki2015, Abdolmaleki2017}. These methods compute each policy update by solving a constrained optimization problem that bounds the KL divergence between the next policy and the current policy.

MORE~\citep{Abdolmaleki2015} additionally limits the entropy loss between subsequent iterations by computing the update as
\begin{equation}
   \label{eq:MORE}
   \begin{aligned}
q^{(i+1)} =  \;\;  \underset{ q}{\argmax}  \;\;  & \int_{\mathbf{x}} q(\mathbf{x}) \reward d\mathbf{x},\\
\text{s.t.}  \quad\quad     &\textrm{KL}\Big(q(\mathbf{x})||q^{(i)}(\mathbf{x})\Big) \le \epsilon, \quad \quad \entropy\Big(q(\mathbf{x})\Big ) \ge \beta^{(i)}, \quad \quad \int_\mathbf{x} q(\mathbf{x}) d\mathbf{x} = 1,
   \end{aligned}
\end{equation}
where the lower bound on the entropy, $\beta^{(i)} = \entropy\Big(q^{(i)}(\mathbf{x})\Big) - \gamma$, is computed at each iteration based on a hyper-parameter $\gamma$ and $\epsilon$ specifies the maximum allowable KL divergence. Hence, at each iteration, the entropy of the search distribution may not decrease by more than $\gamma$.

Introducing Lagrangian multipliers $\eta$, $\omega$ and $\lambda$, the Lagrangian function corresponding to Optimization Problem~\ref{eq:MORE} is given by
\begin{equation*}
   \begin{alignedat}{2}
\mathcal{L}(q,\eta,\beta,\omega) =&  \int_{\mathbf{x}} q(\mathbf{x})  \reward d\mathbf{x} + \eta \left( \epsilon -  \textrm{KL}\Big(q(\mathbf{x})||q^{(i)}(\mathbf{x})\Big) \right) \\
&+ \omega \left( \entropy\Big(q(\mathbf{x})\Big ) - \beta^{(i)}\right)  + \lambda \left( 1 - \int_\mathbf{x} q(\mathbf{x}) d\mathbf{x} \right).
   \end{alignedat}
\end{equation*}

Maximizing the Lagrangian with respect to the search distribution $q$ allows us to express the optimal search distribution $q^{(i+1)}$ as a function of the Lagrangian multipliers, 
\begin{align}
	q^{(i+1)}(\mathbf{x}) \propto {q^{(i)}(\mathbf{x})}^{\frac{\eta}{\eta + \omega}} \exp\left(\reward \right)^{\frac{1}{\eta + \omega}}\label{eq:MOREsolution}.
\end{align}


The update according to Equation~\ref{eq:MOREsolution} can not be computed analytically for general choices of policies $q$ and reward functions \reward. {\sc{MORE}} is therefore restricted to Gaussian search distributions $\approximatedDistIter{i}=\gaussian\big(\mathbf{x};\boldsymbol{\mu}^{(i)},\boldsymbol{\Sigma}^{(i)}\big)$ and optimizes a local, quadratic reward surrogate
\begin{align}
\label{eq:quadSurrogate}
\rewardSurrogate = -\frac{1}{2} \mathbf{x}^\top \mathbf{R}^{(i)} \mathbf{x} + \mathbf{x}^\top \mathbf{r}^{(i)} + \mathrm{const}.
\end{align}
The parameters of the reward surrogate, $\mathbf{R}^{(i)}$ and $\mathbf{r}^{(i)}$,
are learned using linear regression based on samples from the current approximation. For this choice of search distribution and reward surrogate, the updated distribution according to Equation~\ref{eq:MOREsolution} is also Gaussian with natural parameters

\noindent\begin{minipage}{.44\linewidth}
\begin{equation}
\mathbf{Q}(\eta,\omega) = \frac{\eta}{\eta + \omega} 
\mathbf{Q}^{(i)} + \frac{1}{\eta + \omega} \mathbf{R}^{(i)}, 
\label{eq:MOREQuadUpdate}
\end{equation}
\end{minipage}
\begin{minipage}{.1\linewidth}
\hspace{.1\linewidth}
\end{minipage}
\begin{minipage}{.44\linewidth}
\begin{equation}
\mathbf{q}(\eta,\omega) = \frac{\eta}{\eta + \omega} \mathbf{q}^{(i)} + \frac{1}{\eta + \omega} \mathbf{r}^{(i)},
\label{eq:MORELinUpdate}
\end{equation}
\end{minipage}\\

\noindent which directly relate to mean  $\boldsymbol{\mu}=\mathbf{Q}^{-1}\mathbf{q}$ and covariance matrix $\boldsymbol{\Sigma}=\mathbf{Q}^{-1}$. It can be seen from Equation \ref{eq:MOREQuadUpdate} and \ref{eq:MORELinUpdate} that $\eta$ controls the step size, whereas $\omega$ affects the entropy by scaling the covariance matrix without affecting the mean. The optimal parameters $\eta^\star$ and $\omega^\star$ can be learned by minimizing the convex dual objective
\begin{align*}
\mathcal{G}(\eta, \omega) =& \eta \epsilon - \omega \beta^{(i)} + \eta \log Z(\mathbf{Q}^{(i)},\mathbf{q}^{(i)}) - (\eta+\omega) \log Z(\mathbf{Q}(\eta,\omega),\mathbf{q}(\eta,\omega)),
\end{align*}
where $\log Z(\mathbf{X},\mathbf{x}) = -\frac{1}{2} (\mathbf{x}^\top \mathbf{X}^{-1} \mathbf{x} + \log |2\pi\mathbf{X}^{-1}|)$ is the log partition function of a Gaussian with natural parameters $\mathbf{X}$ and $\mathbf{x}$.
This optimization can be performed very efficiently using the partial derivatives\\
\noindent\begin{minipage}{.4\linewidth}
\begin{equation*}
\frac{\partial \mathcal{G}(\eta, \omega)}{\partial \eta} = \epsilon - \text{KL}(q_{\eta,\omega}(\mathbf{x})||\approximatedDistIter{i}),
\end{equation*}
\end{minipage}
\begin{minipage}{.2\linewidth}
\hspace{.2\linewidth}
\end{minipage}
\begin{minipage}{.4\linewidth}
\begin{equation*}
\frac{\partial \mathcal{G}(\eta, \omega)}{\partial \omega} = \entropy(q_{\eta,\omega}(\mathbf{x})) - \beta,
\end{equation*}
\end{minipage}
where 
$q_{\eta,\omega}(\mathbf{x})$ refers to the Gaussian distribution with natural parameters computed according to Equation~\ref{eq:MOREQuadUpdate} and Equation~\ref{eq:MORELinUpdate}. In the next section we introduce a slight variant of {\sc{MORE}} that can be used for variational inference. The derivations of that variant are shown in Appendix~\ref{app:Vips1Derivations} and can be straightforwardly extended to derive the equations shown in this section.

\subsection{Adapting {\sc{MORE}} to Variational Inference}
\label{sec:MORE4VI}
Inspired by policy search methods, we want to use information-geometric trust regions for variational inference in order to achieve efficient optimization while avoiding premature convergence.
Hence, we want to compute each update of the approximation by solving the constrained optimization problem
\begin{equation}
	\label{eq:VIPS1}
   \begin{alignedat}{2}
\boldsymbol{\theta}^{(i+1)} = & \;\; \underset{ \boldsymbol{\theta}}{\argmax}& \;\; & \int_{\mathbf{x}} \approximatedDist  \reward d\mathbf{x} + \entropy(\approximatedDist),\\
&\text{subject to} &      &  \textrm{KL}\Big(\approximatedDist||\approximatedDistIter{i}\Big) \le \epsilon,\\
&                  &      & \int_\mathbf{x} \approximatedDist d\mathbf{x} = 1.
   \end{alignedat}
\end{equation}
Optimization Problem~\ref{eq:VIPS1} is very similar to Optimization Problem~\ref{eq:MORE} solved by {\sc{MORE}} and only differs due to the fact that the entropy of the search distribution does not enter the optimization problem as constraint, but as additional term in the objective. It can be solved analogously to {\sc{MORE}} by introducing Lagrangian multipliers and minimizing the dual problem


\begin{align}
\label{eq:VIPS1_dual}
\mathcal{G}(\eta) =& \eta \epsilon  + \eta \log Z(\mathbf{Q}^{(i)},\mathbf{q}^{(i)}) - (\eta+1) \log Z(\mathbf{Q}(\eta,1),\mathbf{q}(\eta,1)),
\end{align}
using the gradient
\begin{equation}
\label{eq:VIPS1_gradient}
\frac{d \mathcal{G}(\eta)}{d \eta} = \epsilon - \text{KL}(q_{\eta,1}(\mathbf{x})||\approximatedDistIter{i}).
\end{equation}
Here, the natural parameters $\mathbf{Q}(\eta,1)$ and $\mathbf{q}(\eta,1)$ for a given step size $\eta$ are obtained by substituting $\omega=1$ in Equation \ref{eq:MOREQuadUpdate} and \ref{eq:MORELinUpdate}. Please refer to Appendix~\ref{app:Vips1Derivations} for the full derivations.

Hence, a Gaussian variational approximation can be learned analogously to {\sc{MORE}} by iteratively (1) fitting a local, quadratic surrogate $\rewardSurrogate \approx \log \targetDistUnnormalized$, (2) finding the optimal step size $\eta$ by convex optimization and (3) updating the approximation based on Equation \ref{eq:MOREQuadUpdate} and \ref{eq:MORELinUpdate}. The update of a Gaussian variational approximation given a quadratic reward surrogate is shown in Algorithm~\ref{alg::GVAupdate}. 

\begin{algorithm}
   \caption{Updating a Gaussian variational approximation based on surrogate}
   \label{alg::GVAupdate}
\begin{algorithmic}[1]
\Require coefficients of quadratic surrogate $\mathbf{R},\mathbf{r}$ (equation~\ref{eq:quadSurrogate})
\Require current mean and covariance matrix $\boldsymbol{\mu}, \boldsymbol{\Sigma}$
\Require KL bound $\epsilon$
\Function{GVA\_update}{$\boldsymbol{\mu}, \boldsymbol{\Sigma}, \mathbf{R}, \mathbf{r}, \epsilon$}
 \State Compute natural parameters
 \State $\mathbf{Q} \gets \boldsymbol{\Sigma}^{-1}, \quad \mathbf{q} \gets\boldsymbol{\Sigma}^{-1}\boldsymbol{\mu}$
 \State $\eta \gets \text{minimize dual (Equation ~\ref{eq:VIPS1_dual}) using the gradient 
 (Equation~\ref{eq:VIPS1_gradient})}$
 \State Compute new natural parameters
 \State $\mathbf{Q}' \gets \frac{\eta}{\eta + 1} \mathbf{Q} + \frac{1}{\eta + 1} \mathbf{R}, \quad \mathbf{q}' \gets \frac{\eta}{\eta + 1} \mathbf{q} + \frac{1}{\eta + 1} \mathbf{r}$
 \State Compute new search distribution
 \State $\boldsymbol{\Sigma}' \gets \mathbf{Q}'^{-1}, \quad \boldsymbol{\mu}' \gets \mathbf{Q}'^{-1} \mathbf{q}'$
 \State \Return $\boldsymbol{\Sigma}',\boldsymbol{\mu}'$
\EndFunction
\end{algorithmic}
\end{algorithm}

\section{Variational Inference by Policy Search}
\label{sec:VIPS}
We showed in Section~\ref{sec:MORE4VI} that we can learn Gaussian variational approximations using our variant of MORE~\citep{Abdolmaleki2015}. However, Gaussian approximations can lead to high modeling errors, especially for multimodal target distributions. We will now derive {\sc{VIPS++}}, a general-purpose method for learning GMM approximations of an unnormalized target distribution \targetDistUnnormalized. 
In Section~\ref{sec:VIPSBasic} we will show that an I-projection to a GMM can be decomposed into independent I-projections for its Gaussian components using a similar decomposition as used by expectation-maximization. In combination with our variant of {\sc{MORE}}, this result enables us to learn GMM approximations with a fixed number of components.  Sections~\ref{sec:sampleReusage},~\ref{sec:sampleSelection} and~\ref{sec:adapting_num_components} discuss several extensions to this procedure that are critical for efficiently learning high quality approximations in practice. Namely, we will discuss reusing function evaluations from previous iterations, selecting relevant samples and dynamically adapting the number of components.

\subsection{Learning a GMM Approximation}
\label{sec:VIPSBasic}
In order to represent high quality approximations of multimodal distributions, we want to learn a GMM approximation,
\begin{equation*}
\approximatedDist = \sum_o \weightsTheta \componentTheta,
\end{equation*}
where $o$ is the index of the mixture component, $\weightsTheta$ are the mixture weights and $\componentTheta = \mathcal{N}(\mathbf{x}|\boldsymbol{\mu}_o, \boldsymbol{\Sigma}_o)$ is a multivariate normal distribution with mean $\boldsymbol{\mu}_o$ and full covariance matrix $\boldsymbol{\Sigma}_o$. The parameters $\boldsymbol\theta$ of our variational approximation are thus given by the mixture weights, means and covariance matrices. To improve readability we will often omit the parameter $\boldsymbol{\theta}$ when referring to the distribution $q$.

The approximation is learned by maximizing the ELBO 
\begin{align}
\elbo
&= \sum_o \weights \int_{\mathbf{x}} \component \big(\reward   - \log{\approximatedDistNoTheta} \big) d\mathbf{x}  \nonumber \\
&= \sum_o \weights \int_{\mathbf{x}} \component \big(\reward   - \log{\weights} - \log{\component} + \log{\responsibilities} \big) d\mathbf{x}   \nonumber \\
&= \sum_o \weights \Big[ \int_{\mathbf{x}} \component \big(\reward  + \log{\responsibilities} \big) d\mathbf{x} + \entropy{\big(\component\big)} \Big] + \entropy\big( \weights \big) , \label{eq:ElboGMM}
\end{align}
where we used the identity $$\log \approximatedDistNoTheta = \log{\weights} + \log{\component} - \log \responsibilities$$ which can be derived from Bayes' rule.
\subsubsection{Variational Lower Bound}
Unfortunately, the occurrence of the \textit{log responsibilities}, $\log \responsibilities$, in Equation~\ref{eq:ElboGMM} prevents us from optimizing each component independently. However, we can derive a lower bound \lowerBound on the objective by adding and subtracting an auxiliary distribution \auxiliary,
\begin{align}
 L(\boldsymbol{\theta})
=& \sum_o \weights \Big[ \int_{\mathbf{x}} \component \big(\reward  + \log{\responsibilities} \big) d\mathbf{x} + \entropy{\big(\component\big)} \Big] + \entropy\big( \weights \big)   \nonumber \\
=& \sum_o \weights \Big[ \int_{\mathbf{x}} \component \big(\reward  + \log{\auxiliary} + \log{\responsibilities} - \log{\auxiliary} \big) d\mathbf{x} + \entropy{\big(\component\big)} \Big] \nonumber \\
&+ \entropy\big( \weights \big)   \nonumber \\ 
\begin{split}
=& \underbrace{\sum_o \weights \Big[ \int_{\mathbf{x}} \component \big(\reward  + \log{\auxiliary} \big) d\mathbf{x} + \entropy{\big(\component\big)} \Big]+ \entropy\big( \weights \big)}_{\lowerBound}   \\
&+ \int_{\mathbf{x}} \approximatedDistNoTheta \text{KL}\left( \responsibilities || \auxiliary \right) d\mathbf{x}.
\end{split}\label{eq:ElboDecomposition} 
\end{align}
Please note, that the last term in Equation~\ref{eq:ElboDecomposition} corresponds to an expected KL divergence and is therefore non-negative which implies that 
\begin{align*} 
\lowerBound \le \elbo.
\end{align*}
The decomposition in Equation \ref{eq:ElboDecomposition} has already been previously applied in the broad context of variational inference \citep{Agakov2004, Tran2016, Ranganath2016, Maaloe2016}. However, these approaches parameterize the auxiliary distribution and are not well-suited for learning accurate GMM approximations. In contrast, we exploit that the responsibilities \responsibilities can be computed in closed form for Gaussian mixture models, which allows us to exactly tighten the lower bound similar to expectation-maximization~\citep{Bishop2006}. However, whereas EM minimizes the forward KL divergence, $\text{KL}(\targetDist||\approximatedDist)$, for density estimation, our approach can be used for minimizing the reverse KL divergence, $\text{KL}(\approximatedDist||\targetDist)$, in a variational inference setting. The forward KL divergence can be easier optimized when samples from the target distribution are available while the (unnormalized) target density function $\targetDistUnnormalized$ is unavailable and is therefore well suited for density estimation. In contrast, the reverse KL divergence can be more easily optimized based on samples from the model only, when assuming access to the (unnormalized) target density function and is therefore well suited for variational inference.

Following the same reasoning as EM, we can show convergence to a stationary point of the ELBO \elbo by iteratively setting $\auxiliary=\responsibilities$ (analogously to an E-step) and increasing the lower bound \lowerBound (M-step) while keeping the auxiliary distribution fixed. Tightening the lower bound by setting $\auxiliary=\responsibilities$ does not affect the ELBO since the parameters $\boldsymbol{\theta}$ are not changed. Increasing the lower bound increases both the lower bound and the expected KL divergence and thus also increases the ELBO. Such procedure strictly increases the ELBO until we reach a fixed point of the (hierarchical) lower bound optimization, that is, 
\begin{equation*}
    \boldsymbol{\theta}^{(i)} = \underset{\boldsymbol{\theta}}{\argmax} \tilde{L}\left(\boldsymbol{\theta}, q(\mathbf{x},\boldsymbol{\theta}^{(i)})\right).
\end{equation*}
At such fixed point, the gradients of both terms of Equation~\ref{eq:ElboDecomposition} are zero (since they are both at an extremum) and thus the gradient of the ELBO is also zero.

In order to ensure monotonous improvement of the approximation, we need to ensure that the lower bound indeed increases during the M-Step. The lower bound \lowerBound, however, contains intractable integrals that need to be approximated based on samples. In order to keep the resulting approximation errors low, we need to stay close to the current set of samples. We therefore combine the iterative procedure with trust region optimization by bounding the change of each component during the M-step. For sufficiently small step sizes, such trust region updates ensure monotonous improvement~\citep{Akrour2018, Schulman2015}. Furthermore, such constrained maximization does not affect the theoretical guarantees of the iterative procedure as any increase of the lower bound ensures an increase of the ELBO. 

\subsubsection{M-Step for Component Updates}
Maximizing the lower bound $\lowerBound$ with respect to the mean and covariance matrix $\boldsymbol{\theta}_o = [\boldsymbol{\mu}_o, \boldsymbol{\Sigma}_o]$ of an individual component is not affected by the mixture coefficients $\weights$ or the parameters of the remaining components and can be performed independently and in parallel by maximizing the term inside the square brackets of Equation~\ref{eq:ElboDecomposition}, that is, 
\begin{equation}
	\label{eq:VIPS_COMP_UPDATE}
   \begin{alignedat}{2}
 & \;\; \underset{\boldsymbol{\theta}_o}{\argmax}  & \;\; & \int_{\mathbf{x}} q(\mathbf{x}|o;\boldsymbol{\theta}_o) \big(\reward  + \log{\auxiliary} \big) d\mathbf{x} + \entropy{\big(\component\big)},\\
&\text{subject to} &  & \textrm{KL}\Big(q(\mathbf{x}|o;\boldsymbol{\theta}_o)||q(\mathbf{x}|o;\boldsymbol{\theta}^{(i)})\Big) \le \epsilon(o),
   \end{alignedat}
\end{equation}
where we already added the trust region constraint for better exploration and stability. The upper bound on the Kullback-Leibler divergence, $\epsilon(o)$, is adapted during learning. If the Monte-Carlo estimate of the component-specific objective after the component update is smaller than the Monte-Carlo estimate before the update, we decrease $\epsilon(o)$ by multiplying it by \num{0.8}; otherwise we increase it slightly by multiplying it by \num{1.1}. The optimization problem can be solved using our variant of {\sc{MORE}} (Equation~\ref{eq:VIPS1}) with a component specific reward function $\compReward = \reward + \log{\auxiliary}$. As the auxiliary distribution \auxiliary was fixed to the responsibilities $\responsibilitiesIter{i}$ according to the previous mixture model, the component specific part of \compReward penalizes each component for putting probability mass on areas that are already covered by other components.

For applying our variant of {\sc{MORE}}, we need to fit a quadratic reward surrogate \linebreak ${\compRewardSurrogate~\approx~\compReward}$ that approximates the component specific reward \compReward in the vicinity of the respective component $\component$. The surrogate can be fit using ordinary least squares, where the independent variables are samples from the respective component and the dependent variables are the corresponding function evaluations of \compReward. 
However, because we want to use the same set of samples for all component updates as well as the weight update, we use weighted least squares based on importance weights which will be discussed in greater detail in Section $\ref{sec:sampleReusage}$. 
After fitting the surrogate, the optimization problem in Equation~\ref{eq:VIPS_COMP_UPDATE} can be solved efficiently using L-BFGS-B~\citep{Byrd1995} to minimize the dual problem (Equation~\ref{eq:VIPS1_dual}) and using the learned step size $\eta$ to compute the update in closed form as outlined in Section \ref{sec:MORE4VI}.

Drawing the connection to reinforcement learning and investigating the reward function \compReward for a given component reveals that the proposed algorithm treats every component update as a reinforcement learning problem, where the reward is computed based on the achieved log-densities $\log \targetDistUnnormalized$ with a penalty for sampling in regions that are already covered by other components due to low log responsibilities. Moreover, the components strive for high entropy which prevents them from always choosing the same sample.
    
\subsubsection{M-Step for Weight Updates}
After updating the individual components, we can keep the learned means and covariance matrices fixed while updating the mixture coefficients \weights. As shown in previous work~\citep{Arenz2018}, we can also enforce an information-geometric trust region for the weight update. However, in subsequent experiments we could not show a significant effect of such constraint and will therefore only consider the unconstrained optimization. The M-step with respect to the mixture coefficients is thus framed as
\begin{equation}
   \label{eq:VIPS_WEIGHT_UPDATE}
\underset{\weights}{\argmax}  \;\;  \sum_o \weights \weightReward + \entropy\big( \weights \big),
\end{equation}

where the objective for the component update,
\begin{equation}
\label{eq:componentReward}
\weightReward = \int_{\mathbf{x}} \component \big(\reward  + \log{\auxiliary} \big) d\mathbf{x} + \entropy{\big(\component\big)},
\end{equation}
serves as reward for choosing component $o$. The reward $R(o)$ contains an intractable integral, and thus it needs to be approximated from samples. It is to note that $R(o)$ corresponds to a discrete function, which can be represented by a vector, whereas the reward function $\compReward$ used for the component update is a continuous function. It is not beneficial to approximate $R(o)$ based on a quadratic surrogate of \compReward, since we can estimate each element of the vector more efficiently and more accurately using a Monte-Carlo estimate
\begin{equation}
\label{eq:VIPS_MC_estimate}
\weightRewardApprox = \frac{1}{N_o} \sum_{n=1}^{N_o} \big[ R(\mathbf{x}_{o,n}) + \log{\tilde{q}(o|\mathbf{x}_{o,n})} \big] + \entropy(\component),\end{equation} 
where $\mathbf{x}_{o,n}$ refers to the $n$th of $N_o$ samples from component \component. We will discuss in Section \ref{sec:sampleReusage} how we use importance weighting to estimate the reward of each component based on the same set of samples that is used for the component update.

Based on the approximated rewards \weightRewardApprox, the optimal solution of optimization problem in Equation \ref{eq:VIPS_WEIGHT_UPDATE} is given in closed form as
\begin{equation}
\label{eq:optimalWeights}
\weights = \frac{\exp \left( \weightRewardApprox \right)}{\sum_o \exp \left( \weightRewardApprox \right)}.\end{equation}




The weight optimization can also be treated as a reinforcement-learning problem, where actions correspond to choosing components and the agent gets rewarded for choosing components that sample in important regions, that do not interfere with other components and that have high entropy. The agent itself also strives for high entropy and will thus make use of every component.

The complete optimization can be treated as a method for hierarchical reinforcement learning where we learn both, a higher level policy $\weights$ over options and Gaussian lower level policies $\component$. However, since our approach does not consider time series data, it mainly relates to black-box approaches to reinforcement learning that use stochastic optimizers such as {\sc{ARS}}, {\sc{NES}} or {\sc{MORE}}~\citep{Mania2018,Salimans2017,Abdolmaleki2015}. {\sc{HiREPS}} \citep{Daniel2012} already applied black-box optimization for learning GMM policies based on episodic {\sc{REPS}}~\citep{Peters2010}.

The basic variant of our method is shown in Algorithm~\ref{alg:basicVips}. The individual component updates (line 3-8) are performed by sampling from the respective components (line 3), evaluating the samples on the target distribution (line 4), computing the log responsibilities $\log \auxiliary$ according to the previous approximation (line 5), fitting the reward surrogate (line 6-7) and performing the trust region update (line 8). The components can be updated in parallel since the responsibilities are computed based on the same mixture parameters $\boldsymbol{\theta}$.
The weight update (line 11-17) is computed based on Equation~\ref{eq:optimalWeights} (line 17) using the Monte-Carlo estimates of the component rewards (line 15). Updating the parameters of the GMM in between the component updates and the weight update (line 10) is optional and relates to an additional E-Step in EM, which does not affect the theoretical guarantees~\citep{Neal1998}.

\begin{algorithm}
   \caption{Variational Inference by Policy Search (Basic Variant)}
   \label{alg:basicVips}
\begin{algorithmic}[1]
   \Require number of components $N_o$
   \Require initial mixture parameters $\boldsymbol{\theta} = \{ \weights, \boldsymbol{\mu}_{o,\dots,N_o}, \boldsymbol{\Sigma}_{o,\dots,N_o} \}$
   \Require number of iterations $N_i$
   \Require number of samples per component $N_{s}$
   \For{$i = 1 \dots N_i$}
   \For{$o = 1 \dots N_o$}
            \State $\mathcal{X}_o \gets $\sc{sample\_Gaussian}$(\boldsymbol{\mu}_o, 	\boldsymbol{\Sigma}_o, N_s)$
        \State $\tilde{\mathbf{p}}_{o} \gets \log \tilde{p}(\mathbf{\mathcal{X}_o})$ 
\Comment{evaluate target log likelihood for each sample}
            \State $\tilde{\mathbf{q}}_{o|x} \gets \log q(\mathcal{X}_o,o;\boldsymbol{\theta}) - \log{q}(\mathcal{X}_o;\boldsymbol{\theta})$ \Comment{evaluate log responsibilities}
            \State $\mathbf{y}_{o} \gets \tilde{\mathbf{p}}_{o} + \tilde{\mathbf{q}}_{o|x}$ \Comment{Compute targets for ordinary least squares (OLS)}
            \State $\mathbf{R}_o$,$\mathbf{r}_o \gets $\sc{OLS}$(\mathcal{X}_o, \mathbf{y}_o)$ \Comment{learn quadratic surrogate}
            \State $\boldsymbol{\mu}_o', \boldsymbol{\Sigma}_o' \gets $\sc{GVA\_update}$(\boldsymbol{\mu_o}, \boldsymbol{\Sigma_o}, \mathbf{R}_o, \mathbf{r}_o, \epsilon_o)$ \Comment{Algorithm~\ref{alg::GVAupdate}}
        	\EndFor
            \State $\boldsymbol{\theta} \gets $ \sc{update\_components$(\boldsymbol{\theta}, \boldsymbol{\mu}'_{o,\dots,N_o}, \boldsymbol{\Sigma}'_{o,\dots,N_o})$}
            \For{$o = 1 \dots N_o$}
            \State $\mathcal{X}_o \gets $\sc{sample\_Gaussian}$(\boldsymbol{\mu}_o, 	\boldsymbol{\Sigma}_o, N_s)$
        \State $\tilde{\mathbf{p}}_{o} \gets \log \tilde{p}(\mathbf{\mathcal{X}_o})$ 
\Comment{evaluate target log likelihood for each sample}
            \State $\tilde{\mathbf{q}}_{o|x} \gets \log q(\mathcal{X}_o,o;\boldsymbol{\theta}) - \log{q}(\mathcal{X}_o;\boldsymbol{\theta})$ \Comment{evaluate log responsibilities}
            \State $\tilde{R}_o \gets {N_s}^{-1}$
            \sc{sum}$(\tilde{\mathbf{p}}_{o} + \tilde{\mathbf{q}}_{o|x}) + \entropy(\boldsymbol{\Sigma}_o)$
            \Comment{Estimate reward (Equation~\ref{eq:VIPS_MC_estimate})}
			\EndFor
            \State $q'(o) \gets \frac{\exp(\tilde{R}_o)}{\sum_o \exp(\tilde{R}_o)}$ 
            \State $\boldsymbol{\theta} \gets $ \sc{update\_weights$(\boldsymbol{\theta},q'(o))$}
   \EndFor
\end{algorithmic}
\end{algorithm}
\subsection{Sample Reuse by Importance Weighting}
\label{sec:sampleReusage}
VIPS relies on samples for approximating the reward for choosing a given component, \weightReward, and for computing the quadratic surrogates for the component update. These samples need to be evaluated on the unnormalized target distribution \targetDistUnnormalized which may be costly. In order to reduce the number of function evaluations we want to also make use of samples from previous iterations, which can be achieved by using importance weighting. We will now show how importance weights can be used to approximate the rewards for the weight updates and how to learn the quadratic surrogates for the component update based on the same subset $\boldsymbol{\mathcal{X}}_\subset$ of samples.

\subsubsection{Importance Weighting for Updating the Mixture Weights}
Importance sampling is a technique for estimating the expected value $E_q[f(\mathbf{x})]$ of a given function $f(\mathbf{x})$ with respect to a distribution $q(\mathbf{x})$ while using samples from a different distribution $z(\mathbf{x}) \ne q(\mathbf{x})$. Assuming that the support of $z(\mathbf{x})$ covers the support of $q(\mathbf{x})$, we can express the desired expectation as 
\begin{align*}
    E_q[f(\mathbf{x})] &= \int_\mathbf{x} q(\mathbf{x}) f(\mathbf{x}) d\mathbf{x} 
    = \int_\mathbf{x} z(\mathbf{x}) \frac{q(\mathbf{x})}{z(\mathbf{x})} f(\mathbf{x}) d\mathbf{x} =  E_z[w(\mathbf{x}) f(\mathbf{x})],
\end{align*}
using \textit{importance weights} $w(\mathbf{x})=\frac{q(\mathbf{x})}{z(\mathbf{x})}$. 
Hence, the desired expectation can be approximated by using a Monte-Carlo estimate based on $N_z$ samples from the sampling distribution $z(\mathbf{x})$,
\begin{equation}
    \label{eq:standardIS}
    E_q[f(\mathbf{x})] \approx \sum_{i=1}^{N_z} \frac{1}{N_z} w(\mathbf{x}_i) f(\mathbf{x}_i).
\end{equation}
Instead of using the estimator given by Equation~\ref{eq:standardIS}, it is also common to use \textit{self-normalized importance sampling}
\begin{align*}
E_q[f(\mathbf{x})] \approx \sum_{i=1}^{N_z} \bar{w}(\mathbf{x}_i) f(\mathbf{x}_i),
&& \bar{w}(\mathbf{x}_i) = \left( \sum_{i=1}^{N_z} \frac{q(\mathbf{x}_i)}{z(\mathbf{x}_i)}  \right)^{-1} \frac{q(\mathbf{x}_i)}{z(\mathbf{x}_i)}.
\end{align*}
Self-normalized importance sampling introduces a bias that is asymptotically zero since $\underset{N_z \to \infty}{\lim} \sum_{i=1}^{N_z} \frac{q(\mathbf{x}_i)}{z(\mathbf{x}_i)} = N_z$, but it has the advantages that it is consistent for different constant offsets on the function $f(\mathbf{x})$ and that it is also applicable if the target distribution is not normalized.

An important consideration for choosing the sampling distribution is the variance of the estimator. In general, the estimator's variance can be significantly worse than standard Monte-Carlo~\citep{Hesterberg1988}. When using samples from the desired distribution, that is, $z(\mathbf{x}) = q(\mathbf{x})$, the importance weighted estimate and the self-normalized estimate are both equivalent to standard Monte-Carlo. However, it is also possible to obtain lower variance than standard Monte-Carlo, for example, when using the optimal sampling distribution
\begin{equation}
\label{eq:optimalSampling}
z(\mathbf{x}) = \frac{1}{C} q(\mathbf{x}) |f(\mathbf{x}) - c|,
\end{equation}
where $C$ is a normalizing constant and $c=0$ for importance sampling and $c=E_q[f(\mathbf{x})]$ for self-normalized importance sampling~\citep{Hesterberg1988}. If the function $f(\mathbf{x})$ is positive everywhere, the former estimate has even zero variance since
\begin{equation*}
    w(\mathbf{x}_i) f(\mathbf{x}_i) = \frac{q(\mathbf{x}_i)}{z(\mathbf{x}_i)} f(\mathbf{x}_i) = C \frac{q(\mathbf{x}_i)}{q(\mathbf{x}_i) f(\mathbf{x}_i)} f(\mathbf{x}_i) = C = \int_\mathbf{x} q(\mathbf{x}) f(\mathbf{x}) d\mathbf{x} = E_q[f(\mathbf{x})].
\end{equation*}
Although the optimal sampling distributions according to Equation~\ref{eq:optimalSampling} are intractable as they depend on the expectation $E_q[f(\mathbf{x})]$, which is the value of interest, they can be useful for designing appropriate sampling distributions.

In order to estimate the expected reward \weightReward using a subset $\mathcal{X}_\subset$ of the samples from previous iterations $\mathcal{X}$ we need to evaluate the respective sampling distribution $z_\subset(\mathbf{x})$ for computing the importance weights.
For that purpose, we store all samples together with the respective unnormalized target densities and the parameters of the component from which it was sampled in a database
\begin{equation*}
\mathcal{S}=\{(\mathbf{x}_0, \log \tilde{p}(\mathbf{x_0}), \mathcal{N}_{\mathbf{x}_0}),\dots,(\mathbf{x}_N, \log \tilde{p}(\mathbf{x_N}), \mathcal{N}_{\mathbf{x}_N}) \},
\end{equation*}
where $N_\mathbf{x}$ refers to the Gaussian distribution that was used for obtaining the sample $\mathbf{x}$. 
By also storing its respective Gaussian distributions, we can represent the sampling distribution as a Gaussian mixture model $z^\subset(\mathbf{x})$ that contains for each sample $\mathbf{x}_s \in \boldsymbol{\mathcal{X}}_\subset$ the respective Gaussian distribution $\mathcal{N}_{\mathbf{x}_s}(\mathbf{x})$, that is,
\begin{equation*}
z_\subset(\mathbf{x}) = \sum_{\mathbf{x}_s \in \boldsymbol{\mathcal{X}}_\subset} \frac{1}{|\boldsymbol{\mathcal{X}}_\subset|} \mathcal{N}_{\mathbf{x}_s}(\mathbf{x}).
\end{equation*}
Please note, that in practice, we represent the GMM $z_\subset(\mathbf{x})$ more concisely by exploiting that usually several samples were drawn from the same Gaussian distribution. 
We estimate the reward \compReward for each component using self-normalized importance sampling, that is,
\begin{equation*}
\weightRewardApprox =  \sum_{\mathbf{x}_s \in \boldsymbol{\mathcal{X}}_\subset} \bar{w}_o(\mathbf{x}_s) \big[ R(\mathbf{x}_{s}) + \log{\tilde{q}(o|\mathbf{x}_{s})} \big] + \entropy(\component).
\end{equation*}
where the self-normalized importance weights for component $o$ are given by
\begin{align*}
\bar{w}_o(\mathbf{x}_s) = \frac{1}{Z} \frac{q(\mathbf{x}_s|o)}{z_\subset(\mathbf{x}_s)}, &&Z=\sum_{\mathbf{x}_s \in \boldsymbol{\mathcal{X}}_\subset}\frac{q(\mathbf{x}_s|o)}{z_\subset(\mathbf{x}_s)}.
\end{align*}
We could choose different subsets, depending on the component for which we want to estimate the reward \weightReward. However, because we need to evaluate each sample on any component anyway in order to compute the responsibilities \responsibilities, we use the same subset $\boldsymbol{\mathcal{X}}_\subset$ for estimating all component rewards as well as the surrogate models.

\subsubsection{Importance Weighting for Fitting the Quadratic Surrogates}
For updating the individual components we need to learn local quadratic surrogates $\compRewardSurrogate$ in the vicinity of the respective components. {\sc{MORE}} achieves locality by using samples from the respective component \component as independent variables for ordinary least-squares. Learning the surrogate based on samples from a different distribution $z_\subset(\mathbf{x})$ introduces covariate shift, that is, the distribution of the training data $z_\subset(\mathbf{x})$ does not match the distribution of the test data \component. The covariate shift can be accommodated by minimizing a weighted least-squares problem~\citep{Chen2016}
\begin{align*}
    \underset{\boldsymbol{\beta}_o}{\argmin} E_{z_\subset} \left[ \frac{\component}{z_\subset(\mathbf{x})} \left(\compReward - \tilde{R}_o(\mathbf{x};\boldsymbol{\beta}_o) \right)^2\right] = \underset{\boldsymbol{\beta}_o}{\argmin} E_{z_\subset} \left[ \bar{w}_o(\mathbf{x}_s) \left(\compReward - \tilde{R}_o(\mathbf{x};\boldsymbol{\beta}_o) \right)^2\right],
\end{align*}
where the quadratic surrogate $\tilde{R}_o(\mathbf{x};\beta_o)$ is linear in the parameters $\boldsymbol{\beta}_o$.
In practice, we also perform $\ell_2$-regularization with \textit{ridge coefficient} $\kappa_o$.
The optimal parameters are thus given by
\begin{equation*}
\boldsymbol{\beta}_o = (\mathbf{X}^\top \mathbf{W}_o \mathbf{X} + \kappa_o \mathbf{I})^{-1} \mathbf{X}^\top \mathbf{W}_o \mathbf{y},
\end{equation*}
where $\mathbf{X}$ is the design matrix where each row contains the linear and quadratic features for the respective sample $\mathbf{x}_s \in \boldsymbol{\mathcal{X}}_\subset$ as well as a constant feature, $\mathbf{W}_o$ is a diagonal matrix where each element relates to the respective self-normalized importance weight $w_o(\mathbf{x}_s)$, $\mathbf{y}$ is a vector containing the targets $y_s=R(\mathbf{x}_s) + \log q(o|\mathbf{x}_s)$ and $\boldsymbol{\beta}_o$ is a vector containing the elements of $\mathbf{r}_o$ and $\mathbf{R}_o$  as well as a constant offset that can be discarded.
Specifying an appropriate ridge coefficient $\kappa_o$ can be difficult as different components may require different amounts of regularization. We therefore adapt the coefficient during optimization by multiplying it by \num{10} if the matrix inversion failed and by dividing it by \num{2} if it succeeded.

Although we use importance weights for learning the surrogates, we do not aim to estimate an expected value. The minimum-variance sampling distributions given by Equation~\ref{eq:optimalSampling} are in general not useful for learning accurate surrogate models as they focus on bringing the weighted function evaluation $w(\mathbf{x})f(\mathbf{x})$ close to the expected value, rather than aiming to accurately represent the function's landscape. Instead, we aim to construct a sampling distribution $z_{\subset}(\mathbf{x})$ that covers all components of the current approximation well. Such sampling distribution ensures that the importance weighted estimates are not much worse than Monte-Carlo estimations, both, for estimating the expected rewards \weightRewardApprox and for learning locally valid surrogate models \compRewardSurrogate. In the next section, we will discuss a heuristic for constructing such sampling distribution.

\subsection{Sample Selection}
\label{sec:sampleSelection}
 Using all previous samples in each iteration would be computationally costly. Instead, we want to select a small set of samples such that we can get good approximations of all surrogate models and component rewards while requiring only a small number of new samples from each component. A common technique that was used in CMA-ES~\citep{Shirakawa2015}, MORE~\citep{Abdolmaleki2015} and VIPS~\citep{Arenz2018} is to reuse all samples from the $k$ latest iterations, where $k$ is a hyper-parameter to balance between sample efficiency and computational efficiency. As the components that were used for the most recent iterations were similar to the current components, the reused samples can usually provide meaningful information about the target distribution in the vicinity of the respective components. However, we noticed that such procedure can be wasteful when optimizing large GMMs if some component have already converged and others still need to improve. For example, we typically have enough samples in the database to estimate the reward and local surrogate for components that did not significantly change during several iterations even without requiring any new samples; yet, when only using the latest $k$ samples we need to continuously sample from each component during the whole optimization in order to maintain stability.

In order to avoid discarding old samples, we could sub-sample uniformly among the sample database. However, such procedure can result in a large number of irrelevant samples and, furthermore, does not ensure that the relevant samples are evenly distributed among the components of the current approximation. A more sophisticated method was presented by~\citet{Uchibe2018} in the context of policy search. Instead of sub-sampling uniformly, they treat all components in the database as components of a mixture model, $q^\alpha_\text{sampling}(\mathbf{x})$, and optimize the corresponding mixture coefficients $\alpha$ such that the model is close to the optimal sampling distribution given by Equation~\ref{eq:optimalSampling}.
However, the resulting sampling distribution might not be suited for learning the surrogate models, and, furthermore, such approach would be computational intractable because, by optimizing a GMM, {\sc{VIPS}} may add up to several hundreds of components to the database in each iteration and would also need to identify an optimal sampling distribution for each of the respective components.


Furthermore, it is hard to make use of function evaluations such as $\tilde{p}(\mathbf{x}_i)$ or $q(\mathbf{x}_i|o)$ for deciding whether to reuse a given sample $\mathbf{x}_i$ without introducing additional bias in the importance sampling estimate. When such function evaluations influence our decision to use a given sample $\mathbf{x}_i$ for importance weighting, we can no longer consider it as an unbiased draw from $\mathcal{N}_{\mathbf{{x}}_i}(\mathbf{x})$ and computing the importance weights based on the background distribution $z_\subset(\mathbf{x})$ would, thus, not be admissible. 

Instead, we propose to identify for each component \component of the current approximation those components in the database $\mathcal{N}_{\mathbf{x}_i}(\mathbf{x})$ that are close according to a given dissimilarity measure $d\big(\component, \mathcal{N}_{\mathbf{x}_i}(\mathbf{x})\big)$ that is independent of the actual samples drawn from $\mathcal{N}_{\mathbf{x}_i}(\mathbf{x})$. In order to reduce the risk of selecting the same samples in each iteration, which may result in overfitting, we iteratively sample (without replacement) components from our database according to 
\begin{equation}
\label{eq:reuseDistribution}
h(i,o) \propto \exp\left(-d\big(\component, \mathcal{N}_{\mathbf{x}_i}(\mathbf{x})\big) - n_i\right),
\end{equation}
where $n_i$ keeps track of the number of times the samples of distribution $\mathcal{N}_{\mathbf{x}_i}(\mathbf{x})$ have been reused.
We add all samples from the chosen component to the active set of samples $\mathcal{X}_\subset$ and stop sampling distributions when a desired number of reused samples $n_\text{reused}$ is reached.
This process is performed for each component \component of the current approximation.

A natural choice for the dissimilarity is to use the Kullback-Leibler divergence, 
\begin{equation*}
    d_\text{KL}\big(\component, \mathcal{N}_{\mathbf{x}_i}(\mathbf{x})\big) = \text{KL}\big(\component||\mathcal{N}_{\mathbf{x}_i}(\mathbf{x})\big),
\end{equation*} which favors sampling distributions $\mathcal{N}_{\mathbf{x}_i}(\mathbf{x})$ that cover the respective mixture component \component well. However, even though the KL divergence between two Gaussian distributions can be computed in closed form, computing it for every component in the current approximation with respect to every component in the database can quickly become the computational bottleneck of the whole optimization.

Instead, {\sc{VIPS++}} computes the dissimilarity as the negative Mahalanobis distance of the mean $\mu_i$ of the sampling distribution $\mathcal{N}_{\mathbf{x}_i}(\mathbf{x})$ with respect to the given component \component, that is,
\begin{equation*}
    d_\text{Mahalanobis}\big(\component, \mathcal{N}_{\mathbf{x}_i}(\mathbf{x})\big) = -\log p(\boldsymbol{\mu}_i|o).
\end{equation*}
While neglecting the covariance matrix of the sampling distribution may appear too crude, we argue that it is necessary to stay within a reasonable computational budget for selecting relevant samples. We demonstrate in Section~\ref{sec:Ablations} that the proposed selection strategy is able to identify relevant samples for each component \component among all previous samples without adding significant computational overhead. We also compare the Mahalanobis distance to different dissimilarity measures, namely, forward and reverse KL, as well as uniform selection in Appendix~\ref{app:sampleSelection}.
Pseudo-code for identifying relevant samples is shown in Appendix~\ref{app:sample_selection}.

\subsubsection{Drawing new samples}
\label{sec:drawing_new_samples}
After selecting the set $\boldsymbol{\mathcal{X}}_\subset$ of samples to be reused during the current iteration, we need to draw new samples from those components that are not sufficiently covered. 
A useful diagnostic for monitoring the quality of the chosen sampling distribution is the \textit{effective sample size}
\begin{equation*}
n_\text{eff}(o) = \Big(\sum_{\mathbf{x}_s \in \boldsymbol{\mathcal{X}}_\subset} \bar{w}_o(\mathbf{x}_s)^2 \Big)^{-1},
\end{equation*}
which approximates the number of samples that standard Monte-Carlo would require to achieve the same variance as the importance sampling estimate~\citep{Kong1994,Djuric2003}. 

Hence, we compute for each component the number of effective samples,
and draw $n_\text{new}(o)=n_\text{des} - \lfloor n_\text{eff}(o) \rfloor$ new samples, such that its effective sample size should approximately match a specified desired number of effective samples $n_\text{des}$. These samples are added to the database and to the set of active samples $\boldsymbol{\mathcal{X}}_\subset$ as illustrated in Algorithm~\ref{alg:SampleWhereNeeded}.

\begin{algorithm}
   \caption{Ensure that every component has sufficiently many effective samples.}
   \label{alg:SampleWhereNeeded}
\begin{algorithmic}[1]
\Require database $\mathcal{S}=\{(\mathbf{x}_0, \log \tilde{p}(\mathbf{x_0}), \mathcal{N}_\mathbf{x_0}),\dots,(\mathbf{x}_N, \log \tilde{p}(\mathbf{x_N}), \mathcal{N}_\mathbf{x_N}) \}$
\Require Set of chosen samples $\boldsymbol{\mathcal{X}}_\subset$, respective self-normalized importance weights $w_o(\mathbf{x})$
\Require desired number of effective samples per component $n_\text{des}$ 
\Function{sample\_where\_needed}{}
 \For{$o = 1 \dots N_{o}$}
	\State $n_\text{eff}(o) \gets \Big(\sum_{\mathbf{x}_s \in  \boldsymbol{\mathcal{X}}_\subset} w_o(\mathbf{x}_s)^2 \Big)^{-1}$
    \State $n_\text{new}(o) \gets n_\text{des} - \lfloor n_\text{eff}(o) \rfloor$ 
    \State $\boldsymbol{\mathcal{X}}_{\textup{new},o} \gets $ \sc{sample\_Gaussian}$(\boldsymbol{\mu}_o, \boldsymbol{\Sigma_o}, n_\textup{new}(o))$
    \For{$\mathbf{x}_s$ \textbf{in}  $\boldsymbol{\mathcal{X}}_{\text{new},o}$}
    	\State $\mathcal{S} \gets \mathcal{S} \cup \{(\mathbf{x}_s, \log \tilde{p}(\mathbf{x}_s), \mathcal{N}_{\mathbf{x}_s})\}$
    \EndFor
    \State $\boldsymbol{\mathcal{X}}_\subset \gets \boldsymbol{\mathcal{X}}_\subset \cup \boldsymbol{\mathcal{X}}_{\textup{new},o}$
 \EndFor
 \State \Return $\boldsymbol{\mathcal{X}}_\subset$
\EndFunction
\end{algorithmic}
\end{algorithm}

\subsection{Adapting the Number of Components}
\label{sec:adapting_num_components}
The component optimization (Algorithm~\ref{alg::GVAupdate}) is a local optimization, and the component will typically converge to a nearby mode (although the trust region constraint may help to traverse several poor optima). The quality of the learned approximation thus depends crucially on the initialization of the mixture model. However, the modes of the target distribution are often not known a priori and have to be discovered during optimization.
We therefore adapt the number of components dynamically by adding new components in promising regions and by deleting components with very low weight. The number of components is adapted at the beginning of each learning iteration before obtaining new samples. By always assigning low weight to newly added components and by only deleting components that have low weight, the effect on the approximation is negligible and the stability of the optimization is thus not affected. 

\subsubsection{Deleting Bad Components}
Components that have been initialized at poor locations may converge to irrelevant modes of the target distribution and get very low weights such that they do not affect the approximation in practice. As keeping such components would add unnecessary computational overhead, we delete any component that had low weight for a given number of iterations, $n_{del}$, and that further did not increase its expected reward $\tilde{R}(o)$ during that period.

\subsubsection{Initializing the Mean of New Components}
By adding components to the mixture model, we can increase the representational power and thus improve the quality of the approximation. Furthermore, adding components affects the search distribution and can thus be used for exploration.
In either case, we want the new components to eventually contribute to the approximation and hence achieve high weight $\weights \propto \exp\left({R}(o)\right)$ and thus high reward ${R}(o)$. We treat every sample $\mathbf{x}_s$ in the database as candidate for the initial mean of the new component and then select the most promising candidate according to an estimate of its initial reward. As we will discuss in Section~\ref{sec:CovInit}, we will decide on the initial entropy irrespective of the initial mean, but we will choose the exact initial covariance only after deciding for an initial mean. Hence, in the following we will derive an estimate of the initial reward that depends on the initial mean and initial entropy $\entropy_\text{init}$, but not on the covariance matrix. 

Let $q_{\mathbf{x}_s}(\mathbf{x}|o_n)$ denote the new component $o_n$ assuming that its mean was initialized at location $\boldsymbol{\mu}_n=\mathbf{x}_s$ and let $q_{\mathbf{x}_s}(\mathbf{x}) = (1-q(o_n)) \approximatedDistNoTheta + q(o_n) q_{\mathbf{x}_s}(\mathbf{x}|o_n)$ denote the GMM approximation after adding the new component with initial weight $q(o_n)$.
According to Equation~\ref{eq:componentReward} the initial reward of the new component $R_{\mathbf{x}_s}(o_n)$ would be given by
\begin{align}
R_{\mathbf{x}_s}(o_n) =& \int_{\mathbf{x}} q_{\mathbf{x}_s}(\mathbf{x}|o_n) \big(\reward  + \log q_{\mathbf{x}_s}(o_n|\mathbf{x}) \big) d\mathbf{x} + \entropy{\big(q_{\mathbf{x}_s}(\mathbf{x}|o_n)\big)} \label{eq:approximate_reward_1} \\
=& \int_{\mathbf{x}}   q_{\mathbf{x}_s}(\mathbf{x}|o_n) \big(\reward  + \log q(o_n) + \log  q_{\mathbf{x}_s}(\mathbf{x}|o_n)- \log  q_{\mathbf{x}_s}(\mathbf{x}) \big) d\mathbf{x} + \entropy{\big( q_{\mathbf{x}_s}(\mathbf{x}|o_n)\big)} d\mathbf{x} \nonumber\\
=& \log  q(o_n) + \int_\mathbf{x}  q_{\mathbf{x}_s}(\mathbf{x}|o_n) \big( \reward - \log q_{\mathbf{x}_s}(\mathbf{x}) \big) d\mathbf{x} \nonumber \\
\label{eq:futureReward}
=& \log  q(o_n) + \int_\mathbf{x}  q_{\mathbf{x}_s}(\mathbf{x}|o_n) \big( \reward - \log \big( (1- q(o_n)) q(\mathbf{x}) +  q(o_n)  q_{\mathbf{x}_s}(\mathbf{x}|o_n) \big) \big) d\mathbf{x}.
\end{align}
Based on Equation~\ref{eq:futureReward} we can estimate the initial reward depending on the initial weight of the new component $q(o_n)$, the \emph{current} mixture model $q(\mathbf{x})$, the target distribution \reward, and the new component $q_{\mathbf{x}_s}(\mathbf{x}|o_n)$. The first term can be ignored because we choose the initial weight of the new component irrespective of its mean and a constant offset does not affect which initial mean achieves the maximum initial reward. The integral is intractable but can be approximated based on the sample $\mathbf{x}_s = \boldsymbol{\mu}_n$ as
\begin{equation}
   \label{eq:approximateReward}
   \tilde{R}_{\mathbf{x}_s}(o_\text{n}) = R(\mathbf{x}_s) -  \log \bigg( (1- q(o_n)) q(\mathbf{x}_s) +  q(o_n) \exp\Big(\frac{1}{2} D - \entropy_\text{init}\Big) \bigg) 
\end{equation}
where we exploit that the Gaussian density at its mean can be computed based on its entropy $\entropy_\text{init}$ and the number of dimensions $D$, that is, $\log q_{\mathbf{x}_s}(\mathbf{x}_s|o_n)=\frac{1}{2} D - \entropy_\text{init}$. As the function evaluations $R(\mathbf{x}_s)$ of the target distribution are stored in the database, we only need to evaluate the current mixture model $q(\mathbf{x})$ on all candidate samples $\mathbf{x}_s$ to estimate the initial reward for these locations.

To investigate the approximated reward in Equation~\ref{eq:approximateReward} we note that the second term corresponds to a log-sum-exp (LSE), that is,
\begin{align}
    \tilde{R}_{\mathbf{x}_s}(o_\text{n}) &= R(\mathbf{x}_s) -  \log \big( (1- q(o_n)) q(\mathbf{x}_s) +  q(o_n) q_{\mathbf{x}_s}(\mathbf{x}_s|o_n) \big) \nonumber \\
    &=  R(\mathbf{x}_s) -  \text{LSE}\big(\log(1- q(o_n)) + \log q(\mathbf{x}_s) ,  \log q(o_n) + \log q_{\mathbf{x}_s}(\mathbf{x}_s|o_n) \big) \nonumber \\
    \label{eq:addingHeuristicMax}
    &\approx  R(\mathbf{x}_s) -  \max\left( \log q(\mathbf{x}_s),  \log q(o_n) + \log q_{\mathbf{x}_s}(\mathbf{x}_s|o_n)\right),
\end{align}
where we exploit that $\text{LSE}(a_1, a_2, \dots, a_n) = \log \sum_{i=1}^n \exp(a_i)$ behaves similar to a maximum and that $(1-q(o_n)) q(\mathbf{x}) \approx q(\mathbf{x})$, since we initialize the new component with negligible weight, $q(o_n) \approx 0$. Although the effect of the initial weight on the first operand of the log-sum-exp is negligible, it may have considerable effect on the second operand because the logarithm of small values is a large negative value. Hence, the initial weight that we choose for a new component may affect its approximated reward, which can be explained by its effect on the responsibilities $q_{\mathbf{x}_s}(o_n|\mathbf{x}_s)$ in Eq.~\ref{eq:approximate_reward_1}.

If we would add the new components with an initial weight of zero, the maximum-operator would always return the first operand and the proposed estimate (which ignores the constant offset $\log q(o_n)=-\infty$ in Eq.~\ref{eq:approximateReward}) of the initial reward would return the amount of missing log probability-density, $\tilde{R}_{\mathbf{x}_s}(o_\text{n} | q(o_n) = 0) = R(\mathbf{x}_s) - \log q(\mathbf{x}_s)$.
Adding a new component at the location where our current approximation misses most log probability density seems sensible. However, the problem of such heuristic becomes evident when considering target distributions with heavy tails. In such cases, the amount of missing log probability density increases the farther we move away from the current approximation. The new component might, thus, be added in a region where the target distribution has low probability density, since the current approximation might have even lower probability density. 

This failure case is a direct consequence of ignoring the effect of the new component on the mixture model. When considering non-zero weights, the log-responsibilities of the new component are finite and tend to increase the farther we move away from the current approximation. Yet, they saturate at $\log q_{\mathbf{x}_s}(o_n|\mathbf{x}_s) \approx 0$ for every candidate location $\mathbf{x}_s$ that is sufficiently far from the current approximation, that is, where $q(\mathbf{x}_s) \approx 0$. This behavior is reflected by the log-sum-exp in Equation~\ref{eq:approximateReward}, which provides additional reward based on the negative log probability density $-\log q(\mathbf{x}_s)$ of the current approximation but never much more than $-(\log q(o_n) + \log q_{\mathbf{x}_s}(\mathbf{x}_s|o_n))$. 

The proposed heuristic has different effects depending on the choice of the initial weight, which upper-bounds the benefit of adding a component far from the current approximation to $-(\log q(o_n) + \log q_{\mathbf{x}_s}(\mathbf{x}_s|o_n))$ (Eq.~\ref{eq:addingHeuristicMax}). Small initial weights increase this threshold and, thus, the proposed heuristic becomes more explorative by tending to initialize new components far from the current approximation. However, a benefit of the proposed heuristic is that it often does not rely on a specific threshold to propose useful candidate locations. For example, when a candidate is very close to a mode of the target distribution that is currently not covered by the approximation, the heuristic will often choose it for a large range of different thresholds that might vary across several orders or magnitude. 
If there is no clear winner, the choice of $\log q(o_n)$ typically affects the proposed location.
For relatively large initial weights, we will create the component at a location where $R(\mathbf{x}_s)$ is close to the best values that we have discovered and therefore often close to an existing component. Such component will improve our approximation with high probability by allowing the mixture model to approximate the mode more accurately, but is not likely to discover a new mode. 
Estimating the initial reward based on a small initial weight, in contrary, is more likely to place the component far from the current mixture model at locations where $R(\mathbf{x}_s)$ may be significantly worse than the best discovered values. Such component might converge to an irrelevant mode, that is, a local maximum of the target distribution that is still significantly worse than the best mode. The component will then get a very low weight, such that its effect on the approximation is negligible and the computational time (e.g., function evaluations) that was spent for improving this component was mainly wasted. If, however, such component discovers a new relevant mode, it will turn out much more valuable than a component that was added close to an existing mode.

In our experiments, we always add component with an initial weight of $\num{1e-29}$ which results in $\log q(o_n) \approx -66.77$. However, this value is quite arbitrary because adding a new component with initial weight of $\num{1e-300}$ would result in essentially the same mixture model and $\log q(o_n) \approx -690.78$. Hence, we do not estimate the initial reward based on the actual initial weight, but instead choose a value in place of $\log q(o_n)$ and vary it in the range of $[-1000, -50]$. By varying the (assumed) initial weight we can maintain exploration and avoid only adding components at irrelevant locations. Please refer to Appendix~\ref{app:deltaSensitivity} for a sensitivity analysis and for details on how the initial reward in Equation~\ref{eq:approximateReward} is approximated.

\subsubsection{Initializing the Covariance Matrix of New Components}
\label{sec:CovInit}
The initialization of the covariance matrix of the new component is performed in two steps. In the first step, we decide on the initial entropy; in the second step, we decide on the initial correlations.

A possible option for choosing the initial entropy is to use the same entropy that was used when initializing the mixture at the beginning of the optimization, which would typically be relatively large in accordance with an uninformed prior. Such an initialization has the benefit of maintaining broad exploration during the whole optimization, and is not very sensitive to the initialization of the mean. However, initializing new components with high entropy can also be very wasteful as it will typically take a long time until they can contribute to the approximation. Furthermore, smaller initial entropies in combination with our heuristic for initializing the mean will result in a more directed exploration of promising regions. Hence, we initialize the new component with an entropy that is similar to those of the best components in the current model, namely we choose $\entropy_{init} = \sum_o \weights \entropy(\component)$ as initial entropy. The entropy of the best components will typically decrease during optimization until it reaches a problem specific level. Hence, the exploration of new components will also become more local, without falling below a reasonable level.

For deciding on the correlations among the different dimensions, we can consider restarting the local search from scratch by choosing an isotropic covariance matrix $\boldsymbol{\Sigma}_{iso} = c_{iso} \boldsymbol{I}$, and making use of the existing components by averaging their covariance matrices, that is, $\boldsymbol{\Sigma}_{avg} = c_{avg} \sum_o p(o|\boldsymbol{\mu}_{new}) \boldsymbol{\Sigma}_{o}$, where $c_{iso}$ and $c_{avg}$ are appropriately chosen to obtain the desired entropy $\entropy_{init}$ as shown in Appendix~\ref{app:GaussianEntropy}. In VIPS we always averaged the covariance matrices, which can be sensible when adding components close to existing ones, or when similar correlations occur at different locations. However, we noticed that such initialization can impair exploration and, thus, degrade performance in one of our new experiments as shown in Section~\ref{sec:ablations:covinit}.
As it is often difficult to predict, whether the curvature at the most responsible components is similar to the curvature at the new component, we perform a line search over a step size $\alpha \in [0,1]$ to find the best interpolation
\begin{equation*}
\boldsymbol{\Sigma}_\alpha = \alpha \boldsymbol{\Sigma}_{iso} + (1-\alpha) \boldsymbol{\Sigma}_{avg}
\end{equation*}
between both candidate covariance matrices with respect to the expected reward
\begin{equation*}
R_\text{new}(\alpha) = \int_{\mathbf{x}} \mathcal{N}(\mathbf{x}|\boldsymbol{\mu}_{new}, \boldsymbol{\Sigma}_\alpha) \log \targetDistUnnormalized d\mathbf{x}.
\end{equation*}
The expected reward can be approximated using an importance weighted Monte Carlo estimate based on samples from the mixture
\begin{equation*}
z(\mathbf{x}) = 0.5 \mathcal{N}(\mathbf{x}|\boldsymbol{\mu}_{new}, \boldsymbol{\Sigma}_{iso}) + 0.5 \mathcal{N}(\mathbf{x}|\boldsymbol{\mu}_{new}, \boldsymbol{\Sigma}_{avg}).
\end{equation*}
These samples and the respective function evaluations are also stored in the database $\mathcal{S}$ and can thus be reused during subsequent learning iterations.

\begin{figure}
    \centering
	\begin{subfigure}{.3\textwidth}
		\resizebox{.8\textwidth}{!}{
\begin{tikzpicture}[node distance = 2cm, auto]
    \node [startstop] (init) {Initialize Mixture};
    \node [block] (DrawComp) [fill=green!10, text width=3cm, below=1.5cm of init]  {Draw Samples};
    \node [block] (RespC) [text width=3cm, below=1cm of DrawComp]  {Compute Responsibilities};
    \node [block] (updateComp) [below=1cm of RespC, text width=2cm] {update Component (OLS)};
    \node [draw=red!80, ultra thick, label=above:\large For each component $o$, fit=(updateComp)(RespC)(DrawComp)] (compUpdate) {};
    \node [block] (DrawWeight) [fill=green!10, text width=3cm, below=1.5cm of compUpdate] {Draw Samples};
    \node [block] (RespW) [text width=3cm, below=1cm of DrawWeight]  {Compute Responsibilities};
    \node [block] (Ro) [below=1cm of RespW, text width=3cm] {compute $\tilde{R}(o)$ (MC)};
    \node [draw=blue!80, ultra thick, label=above:\large For each component $o$, fit=(Ro)(RespW)(DrawWeight)] (rewCompute) {};
    \node [block] (updateWeights) [fill=yellow!10, below=0.7cm of rewCompute] {Update Weights};
    \node [decision] (isDone) [below=0.7cm of updateWeights] {Max Iterations Reached?};
   \node [startstop] (done) [below=0.7cm of isDone] {Done};

    \path [line] (init.south) -- ++(0,-15pt) -- ++(-65pt,0pt) |- (compUpdate.west);
    \path [line] (DrawComp) -- (RespC);
    \path [line] (RespC) -- (updateComp);
    
    \path [line] (compUpdate.south) -- ++(0,-15pt) -- ++(-65pt,0pt) |- (rewCompute.west);
    \path [line] (DrawWeight) -- (RespW);
    \path [line] (RespW) -- (Ro);

    \path [line] (rewCompute.south) -- (updateWeights);
    \path [line] (updateWeights) -- (isDone);
    \path [line] (isDone) -- node [near start] {yes} (done);
    \path [line] (isDone.west) -- ++(-40pt,0pt) |- node [near start] {no} ($(compUpdate.west) + (0pt, -5pt)$);
\end{tikzpicture}
}
\end{subfigure}
	\begin{subfigure}{.69\textwidth}
			\resizebox{.7\textwidth}{!}{
\begin{tikzpicture}[node distance = 2cm, auto]
    \node [startstop] (init) {Initialize Mixture};
    \node [block] (AddDelete) [below=1cm of init, text width=3cm] {Add / Delete Components};    
    \node [block] (ReuseSamples) [below=1cm of AddDelete] {Reuse Samples}; 
    \node [block] (DrawSamples) [fill=green!10, below=1cm of ReuseSamples] {Draw Samples};
    \node [block] (RespW) [text width=3cm] at (-1.5,-10.5) {Compute Responsibilities};
    \node [block] (Ro) [right=1cm of RespW] {compute $\tilde{R}(o)$ (IS)};
    \node [draw=blue!80, ultra thick, label=above:\large For each component $o$, fit=(RespW)(Ro)] (rewCompute) {};
    \node [block] (updateWeights) [fill=yellow!10, below=1cm of rewCompute] {Update Weights};
    \node [block] (RespC) [text width=3cm] at (-2, -16) {Compute Responsibilities};
    \node [block] (updateComp) [right=1cm of RespC, text width=2cm] {Update Component (WLS)};
    \node [database, label=below:\huge Database] (SampleDB) [right=2cm of DrawSamples] {};
    \node [database, label=below:\huge Active Samples] (ActiveSamples) [left=2.5cm of DrawSamples] {};
    \node [draw=red!80, ultra thick, label=above:\large For each component $o$, fit=(RespC)(updateComp)] (compUpdate) {};
    \node [decision] (isDone) [below=1cm of compUpdate] {Max Iterations Reached?};
    \node [startstop] (done) [below=1cm of isDone] {Done};

    \path [line] (init) -- (AddDelete);
    \path [line] (AddDelete) -- (ReuseSamples);
    \path [line] (ReuseSamples) -- (DrawSamples);
    \path [line] (DrawSamples.south) -- ++(0,-15pt) -- ++(-120pt,0pt) |- (rewCompute.west);
    \path [line] (RespW) -- (Ro);
    \path [line] (rewCompute) -- (updateWeights);
    \path [line] (updateWeights.south) -- ++(0,-15pt) -- ++(-120pt,0pt) |- (compUpdate.west);
    \path [line] (RespC) -- (updateComp);
    \path [line] (compUpdate) -- (isDone);
    \path [line] (isDone) -- node [near start] {yes} (done);
    \path [line] (isDone.east) -- ++(108pt,0pt) |- node [near start] {no} (AddDelete.east);
    \path [line, <->, dashed] ($(AddDelete.south) + (5pt, 0pt)$) -- ++(0pt, -5pt) -| ($(SampleDB.north) + (5pt, 0pt)$); 

    \path [line, dashed] (SampleDB.north) |- (ReuseSamples.east);
    \path [line, dashed] (ReuseSamples.west) -|  node [near start, above] {set}  (ActiveSamples.north);
    \path [line, dashed] (DrawSamples) --  node [above] {add} (ActiveSamples);
    \path [line, dashed] (DrawSamples) -- (SampleDB);

    \path [line, dashed] (ActiveSamples.west) -- ($(ActiveSamples.west) + (-60pt, 0pt)$) |- ($(rewCompute.west) + (0pt, -5pt)$);
    \path [line, dashed]  ($(ActiveSamples.west) + (-60pt, -90pt)$) |- ($(compUpdate.west) + (0pt, -5pt)$);
\end{tikzpicture}
}
\end{subfigure}
\caption{We show flow charts for the basic variant (left) and {\sc{VIPS++}} (right). The basic variant updates the individual components by learning surrogates using ordinary least-squares (OLS) and uses Monte-Carlo (MC) for estimating the component's reward $\tilde{R}(o)$. {\sc{VIPS++}}  adapts the number of component and uses the same set of samples for computing the components' reward using importance sampling (IS) and for updating the individual components using weighted least squares. The order of the weight and component updates has been swapped on the right flow chart to match the actual implementation.}
        \label{fig:flowchart}
\end{figure}
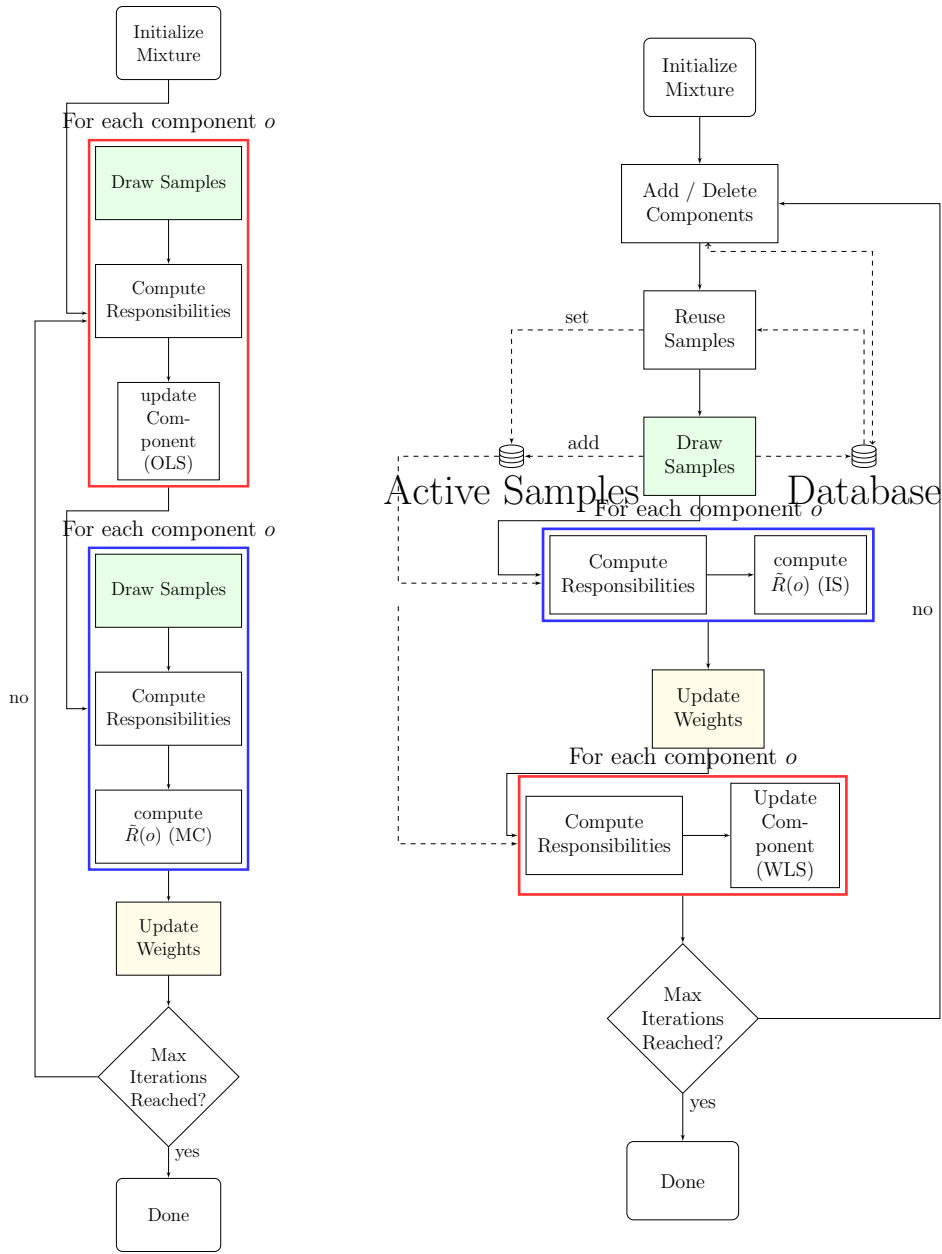

Flow charts for the basic variant and the modified version are shown in Figure~\ref{fig:flowchart}.
An open-source implementation is available online\footnote{The implementation can be found at \url{https://github.com/OlegArenz/VIPS}.}.
In comparison to {\sc{VIPS}}, {\sc{VIPS++}} makes better use of previous function evaluations and initializes new components based on a line search. Furthermore, {\sc{VIPS++}} uses fewer hyper-parameters by automatically adapting the bounds on the KL-divergences for the individual component updates and the regularization coefficients for fitting the reward surrogates. The number of hyper-parameters was further reduced by simplifications of the algorithms; namely, by performing an unconstrained optimization for the weight updates and by performing a single EM-like iteration on a given set of samples.

\section{Related Work}

We will now discuss related work in the fields of variational inference, sampling and policy search.

\subsection{Variational Inference}
Traditionally, variational inference was applied for learning coarse approximations of high dimensional distributions, typically by assuming that the individual dimensions of the random variable are uncorrelated---the so-called mean-field assumption---and by choosing the variational distribution based on the target distribution. For example, \citet{Saul1996} approximated the hidden nodes of sigmoid belief networks with Bernoulli distributions, enabling them to maximize a lower bound on the ELBO in closed form. An iterative procedure was used for improving this lower bound. As such approach can only model unimodal distributions, it was later extended to mixtures of mean field distributions \citep{Jaakkola1998, Bishop1998}.

However, relying on a variational distribution that can be fitted in closed form can be restrictive and the necessary derivations can be a major burden when applying such variational inference approaches to different models. Hence, \citet{gershman2012} introduced non-parametric variational inference ({\sc{NPVI}}), a black-box approach to variational inference that can be applied to any twice-differentiable target distribution. {\sc{NPVI}} is restricted to GMMs with uniform weights and isotropic components that are iteratively optimized using first-order and second-order Taylor approximations. Although such variational approximation can in principle approximate any target distribution arbitrarily well, NPVI is in general not suited for learning highly accurate approximations with a reasonable number of components as shown in our comparisons. 

Similar to {\sc{VIPS}}, several black box approaches to variational inference rely on function evaluations of the target distributions that are chosen by sampling the variational approximation. \citet{Ranganath2014} apply the log-derivative trick, which is well-known in reinforcement learning~\citep{Williams1992}, to variational inference in order to estimate the gradient of the ELBO with respect to the policy parameters. The gradient estimation does not require the gradient of the reward $\log \targetDistUnnormalized$ but typically suffers from high variance. \citet{Ranganath2014}, thus, suggest control-variates and Rao-Blackwellization (for which they assume a mean-field approximation) for variance reduction.
If the target distribution is differentiable and the variational approximation is reparameterizable, it is usually preferable to estimate the gradient with the reparameterization trick~\citep{Kingma2014, Rezende2014} which typically has much lower variance. Such approach can, for example, be used to train normalizing flows~\citep{Dinh2014}. Normalizing flows are likelihood-based models that transform a simple distribution through one or several non-linear mappings. The probability density of the transformed distribution can be evaluated using the change-of-variables formula, which requires that the transformations are invertible and that the (log-)determinants of their Jacobians can be efficiently computed. \citet{Rezende2015} proposed transformations that contract the density with respect to a learned hyperplane or to a point. The expressiveness of these planar and radial flows is rather limited and thus many flows has to be stacked to obtain rich approximations. However, several more expressive flows have been recently proposed~\citep{Kingma2016, Kingma2018, Dinh2017, Papamakarios2017, Huang2018, Grathwohl2019}. Most of these flows make use of autoregressive transformations. For example, \textit{inverse autoregressive flows} \citep[{\sc{IAF}},][]{Kingma2016} shift and scale each dimension of an input, $x_i$, by quantities that are computed based on the previous input dimensions $x_{j<i}$. As the resulting Jacobian matrices are triangular, the log determinants can be efficiently computed based on the diagonal elements. Rich approximations can be learned by stacking several such flows and shuffling the dimensions in-between based on fixed random or learned~\citep{Kingma2018} permutations, which can also be seen as normalizing flows. In order to ensure the autoregressive property, {\sc{IAF}}s use a technique that was previously used for autoregressive auto-encoders~\citep{Germain2015}. Namely, a mask is applied to a fully connected neural network in order to cut weights such that each output $y_i$ is only connected to inputs $x_j$ if $j<i$. Although such flows are invertible by construction, computing the inverse can be expensive because the different dimensions have to be inverted sequentially. Hence, evaluating the probability density of a sample that was produced by different distribution can be inefficient. \textit{Masked autoregressive flows}~\citep{Papamakarios2017}, thus, parameterize the inverse transformation (compared to {\sc{IAFs}}) which makes them more efficient for density estimation at the cost of less efficient sampling. In general, normalizing flows are very popular nowadays, because they scale to high dimensions, allow for rich representations and are reparameterizable whenever the initial distribution is reparameterizable. However, we argue that such purely gradient-based optimization is not suited for learning accurate approximations of multimodal target distributions due to insufficient exploration. In our experiments, we compare against {\sc{IAFs}}, which are well-suited for variational inference because we only need to evaluate the density of samples that were drawn from the normalizing flow.

Hessian-free stochastic Gaussian variational inference ({\sc{HFSGVI}}, \citealt{Fan2015}) and TrustVI \citep{Regier2017} can be used for learning Gaussian variational approximations.
{\sc{HFSGVI}} \citep{Fan2015} learns GVAs with full covariance matrices using fast second order optimization. This idea has been extended by \citet{Regier2017} to trust region optimization. However, in difference to our approach, a euclidean trust region is used in parameter space of the variational distribution. Such approach requires the computation of the Hessian of the objective which is only tractable for mean-field approximations of single Gaussian distributions. In contrast, we use the trust regions directly on the change of the distributions instead of the change of the parameters of the distribution. The information geometric trust regions in this paper allow for efficient estimation of GMMs with full covariance matrices without requiring gradient information from $\targetDistUnnormalized$. 

Information geometric trust regions and related methods such as certain proximal point methods as well as methods based on natural gradient descent have already been applied to variational inference.
\citet{Salismans2013} derive a fixed point update of the natural parameters of a distribution from the exponential family that corresponds to a Monte-Carlo estimate of the gradient of Equation~\ref{eq:VIKL} preconditioned by the inverse of their empirical covariance. By making structural assumptions on the target distribution, they extend their method to mixture models and show its applicability to bivariate GMMs.
\citet{Hoffman2013} consider mean-field variational inference and assume a certain structure on the target distribution. Namely, they consider models that consist of a product of conditionally independent distributions parameterized by local parameters that are correlated through global parameters. Furthermore, all distributions are assumed to belong to the exponential family and the distribution of the global parameters is assumed to be conjugate for computational reasons. They show that the natural gradient of the corresponding mean-field approximation can be efficiently computed, and approximated from mini-batches. \citet{Theis2015} extended their approach by enforcing a trust-region based on the KL-divergence for better exploration. \citet{Khan2015} consider slightly more general models where optimizing the ELBO can be computationally expensive. They propose to apply the proximal point method by adding a penalty to the ELBO based on the reverse Kullback-Leibler divergence to the current iterate. They decompose the ELBO into easy and difficult parts and linearize the difficult parts. The derivations where extended by~\citet{Khan2016} to other divergences and to stochastic gradients making the approach applicable to posterior approximations based on mini-batches. \citet{Altosaar2018} propose a slightly more general framework that can penalize derivations from a moving average instead of derivations from the last iterate, which can further help in avoiding bad local optima.  

Several methods use the same hierarchical bound as {\sc{VIPS}} in the broad context of variational inference. The first usage seems to date back to 2004, where \citet{Agakov2004} proposed the bound for learning an optimal weighting between several mean-field approximations. \citet{Ranganath2016} proposed Hierarchical variational methods ({\sc{HVM}}) where the lower-level distributions $q(\mathbf{x}|\mathbf{o})$ where again mean-field distributions. In their setting, the latent variable $\mathbf{o}$ corresponds to a parameter vector that fully specifies the mean-field distribution. They learned complex priors $q(\mathbf{o})$ over these parameters, namely GMMs and normalizing flows, in order to allow for rich variational approximations. However, in contrast to the responsibilities in {\sc{VIPS}} the conditional $q(\mathbf{o}|\mathbf{x})$ is not tractable and thus has to be approximated and learned along the variational distribution. Our EM-inspired approach based on exact tightening of the hierarchical lower bound would thus not be applicable in their setting. Although \citet{Ranganath2016} learned Gaussian mixture models to model the upper-level distribution $q(\mathbf{o})$, they did not apply the hierarchical bound for this, but optimized the parameters directly using stochastic gradient descent. As we will show in our experiments, such black-box approach is not suited for learning variational GMM approximations.
\citet{Tran2016} consider a similar setup for their \textit{variational Gaussian process}. For the mean-field factors $p(x_n|o_n)$ of their lower-level components they consider degenerated point masses specified by their scalar parameter value $o_n$. As \citet{Ranganath2016}, they optimize the hierarchical lower bound with respect to the prior distribution $q(\mathbf{o})$ and the conditional ${q}(\mathbf{o}|\mathbf{x})$. Their main contribution is the representation of the prior distribution. Each parameter value $o_n$ is sampled by evaluating a Gaussian process~\citep[GP,][]{Rasmussen2006} on an input that was sampled from a fixed distribution. The parameters of the prior are given by the kernel hyper-parameters of the GP as well as the \textit{variational data} that is interpolated by the GP.
Whereas all these methods only consider mean-field distributions for the lower-level components that are fully specified by the latent variable, \citet{Maaloe2016} represent them using inference networks, that is, neural networks that take a data point as input and output the parameters of a (typically diagonal) Gaussian distribution. They consider variational autoencoders~{VAE} and aim to learn more expressive approximations of the latent code $z$. They also introduce an additional latent variable representing class labels in order to train a classifier end-to-end while optimizing the variational autoencoder in semi-supervised fashion. 
In contrast to these previous applications of the hierarchical lower bound, {\sc{VIPS}} shows that it can also be used to learn accurate variational approximations without having to approximate the inverse model $p(o|\mathbf{x})$. This enables us to optimize the ELBO by alternately maximizing and (exactly) tightening the hierarchical lower bound.

Closely related to our work are two recent approaches for variational inference that concurrently explored the idea of applying boosting to make the training of GMM approximations tractable~\citep{Miller2017, Guo2016}. These methods start by minimizing the ELBO objective for a single component and then successively add and optimize new components and learn an optimal weighting between the previous mixture and the newly added component. However, because these methods can not adapt previously added components or their relative weighting, they can require an unnecessary large number of components to learn accurate approximations. Furthermore, they do not use information-geometric trust regions to efficiently explore the sample space and therefore have problems finding all the modes as well as accurate estimates of the covariance matrices. 
GMMs are also used by \citet{zobay2014} where an approximation of the GMM entropy is used to make the optimization tractable. The optimization is gradient-based and does not consider exploration of the sample space. It is therefore limited to rather low dimensional problems.

The work of \citet{Weber2015} already explored the use of reinforcement learning for VI by formalizing VI as sequential decision problem. However, only simple policy gradient methods have been proposed in this context which are unsuitable for learning GMMs.

\subsection{Sampling}
Although {\sc{MCMC}} samplers can not directly be used for approximating distributions, they are for many applications the main alternative to VI. Especially, when applying {\sc{VIPS}} as a model-based sampler, that is, if we do not have direct interest in learning a GMM approximation, it should be compared to other zero-order sampling methods that do not need gradient information from the 
target density. The most prominent methods to use here are {\sc{MCMC}} methods such as slice sampling \citep{Neal2003}, elliptical slice sampling \citep{Murray2010} or generalized elliptical slice sampling \citep{Nishihara2014}. {\sc{MCMC}} methods define a Markov chain for the sampling process, that is, the current sample defines the state of the chain, and we define a conditional distribution how to generate new samples from the current state. 

Slice sampling introduces an auxiliary variable $y$ to define this conditional distribution. The variable $y$ is always sampled between $0$ and the unnormalized target density of the current sample. The random variable $\mathbf x$ is only accepted if the new target density is larger than $y$. In case of rejection, the area where a new $\mathbf x$ sample is generated is reduced to limit the number of rejections. However, the sampling process is still very inefficient for higher dimensional random variables.
Elliptical slice sampling \citep{Murray2010} is a special case of slice sampling and defines the slice by an ellipse defined by the current state $\mathbf x$ and a random sample from a Gaussian prior (with origin $0$). Such ellipse allows for more efficient sampling and rejection in high dimensional spaces but relies on a strong Gaussian prior.

If the gradient of the target distribution is available, Hamiltonian MCMC~\citep{duane1987} and the Metropolis-adjusted Langevin algorithm~\citep{Roberts2002} are also popular choices. The No-U-Turn sampler (NUTS)~\citep{Hoffman2014} is a notable variant of Hamiltonian {\sc{MCMC}} that is appealing for not requiring hyper-parameter tuning. 

While many of these {\sc{MCMC}} methods have problems with multimodal distributions in terms of mixing time, other methods use multiple chains and can therefore better explore multimodal sample spaces \citep{earl2005, Neal1996,Nishihara2014,Calderhead2014}. Parallel tempering MCMC~\citep{earl2005} runs multiple chains, where each chain samples the target distribution at a different temperature. Each step consists either of updating each chain independently, or swapping the state between two neighboring chains which allows for more efficient mixing between isolated modes. However, because only one chain samples the target distribution at the correct temperature, {\sc{PTMCMC}} can be inefficient if the number of chains and their respective temperatures are not adequately tuned for the sampling problem.
Generalized elliptical slice sampling \citep{Nishihara2014} uses multiple Markov chains simultaneously using massive parallel computing. The current state of the Markov chains is used to learn a more efficient proposal distribution, where either Student-t distributions or Gaussian mixture models can be used. Yet, learning such distributions in high dimensional spaces using maximum likelihood is prone to overfitting and the GMM approach has not been evaluated on practical examples. Moreover, the approach requires a massive amount of sample evaluations. In this paper, we want to minimize the amount of sample evaluations. 

\citet{Rainforth2018} explicitly consider the exploration-exploitation trade-off. They use a method similar to Monte-Carlo tree search~\citep{Coulom2006} to build a tree for partitioning the search space. By covering regions where the target distribution has high density more finely, the resulting \textit{inference trees} ({\sc{IT}}) are well-suited for inference on multimodal distributions, for example, in combination with sequential Monte-Carlo~\citep{Doucet2001}.

Stein variational gradient descent ({\sc{SVGD}}) \citep{Liu2016} is a sampling method that closely relates to variational inference. However, instead of optimizing the parameters of a model, {\sc{SVGD}} directly optimizes an initial set of particles. By framing sampling as optimization problem, {\sc{SVGD}} inherits the computational advantages of variational inference and because it is non-parametric, it is capable of approximating multimodal distributions. However, this method requires to construct the Gram matrix of the particles and is thus not suitable for drawing large number of samples. Furthermore, defining appropriate kernels can be challenging for high-dimensional problems.

\subsection{Policy Search}
Our algorithm shares a lot of ideas with information-geometric policy search algorithms such as {\sc{REPS}} \citep{Peters2010}, {\sc{HiREPS}} \citep{Daniel2016a} and {\sc{MORE}} \citep{Abdolmaleki2015}. In difference to policy search, where we want to maximize an average reward objective, we want to minimize the KL-divergence to a target distribution. {\sc{REPS}} introduces the first time information-geometric policy updates, while the {\sc{MORE}} algorithm introduces closed form updates for single Gaussians using compatible function approximation and additional entropy regularization terms that yields an optimization problem similar to KL minimization. 

The {\sc{HiREPS}} \citep{Daniel2016a} and {\sc{LaDiPS}} \citep{End2017} algorithms extended the {\sc{REPS}} and {\sc{MORE}} ideas to mixture distributions such that multiple modes can be represented. However, the used updates were based on approximations or heuristics and can not optimize the entropy of the complete mixture model. 

\section{Experiments}
\label{sec:experiments}
In this section we will evaluate {\sc{VIPS++}} with respect to the quality of the learned approximation and relate it to a variety of state-of-the-art methods in variational inference and Markov chain Monte Carlo. We start with a description of the considered sampling problems in Section~\ref{sec:testbed}. The effects of the most important hyper-parameters and algorithmic choices are examined in Section~\ref{sec:Ablations}. Section~\ref{sec:illustrativeExperiment}
contains an illustrative experiment to show how {\sc{VIPS++}} approximates a two-dimensional, multimodal target distribution by starting with a single component and iteratively adding more components according to our heuristic. The selected methods for our comparisons, and the selection of their hyper-parameters are discussed in Section~\ref{sec:consideredMethods} and Section~\ref{sec:hyperParameters}. The results of the quantitative experiments are presented and discussed in Section~\ref{sec:exp:results}.

\subsection{Sampling Problems}
\label{sec:testbed}
We will evaluate {\sc{VIPS++}} on typical sampling problems such as Bayesian logistic regression, Bayesian Gaussian process regression and posterior sampling of a multi-level Poisson generalized linear model. We further approximate the posterior distribution over the parameters of a system of ordinary differential equations known as the Goodwin model, which can be used for modeling oscillating gene-protein interaction.
As these problems tend to have concentrated modes, we devised several more challenging problems that require careful exploration of the sampling space. Namely, we consider sampling from unknown GMMs with distant modes and sampling the joint configurations of a planar robot such that it reaches given goal positions.

\subsubsection{Bayesian Logistic Regression}
 We perform two experiments for binary classification that have been taken from~\citet{Nishihara2014} using the \textit{German credit} and \textit{breast cancer} data sets~\citep{UCI}. The \textit{German credit} data set has twenty-five parameters and 1000 data points, whereas the \textit{breast cancer} data set is thirty-one dimensional and contains 569 data points. We standardize both data sets and perform linear logistic regression where we put zero-mean Gaussian priors with variance 100 on all parameters. 

\subsubsection{Multi-Level Poisson GLM}
We also took an experiment from the related work {\sc{VBOOST}}~\citep{Miller2017}. For this experiment we want to sample the posterior of a hierarchical Poisson GLM on the 37-dimensional \textit{stop-and-frisk} data set, where we refer to~\cite{Miller2017} for the description of the hierarchical model. 

\subsubsection{GP Regression}
We perform Bayesian Gaussian process regression on the ionosphere data set~\citep{UCI} as described by~\citet{Nishihara2014}. Namely, we use 100 data points and want to sample the hyper-parameters of a squared exponential kernel where we put a gamma prior with shape $1$ and rate $0.1$ on the 34 length-scale hyper-parameters. We initialize {\sc{VIPS}} with a single Gaussian component, $\mathcal{N}(\mathbf{x}|\mathbf{0}, \mathbf{I})$ and sample in log-space to ensure positive values for the hyper-parameters. 

\subsubsection{Goodwin Model}
Similar to~\citet{Calderhead2009}, we want to sample the posterior over the parameters of a Goodwin oscillator~\citep{Goodwin1965} based on noisy observations. The Goodwin oscillator is a system of nonlinear ordinary differential equations (ODE) that models the oscillatory behavior between protein expression and mRNA transcription in enzymatic control processes. We consider a Goodwin oscillator with ten unknown parameters and put a Gamma prior with shape $2$ and rate $1$ on each of these. The likelihood of $41$ observations is computed by numerically integrating the ODE and assuming Gaussian observation noise with zero mean and variance $\sigma^2=0.2$. Please refer to Appendix~\ref{app:Goodwin} for more details on the ODE and the experimental setup.

\subsubsection{Gaussian Mixture Model \label{sec:GmmExperiment}}
In order to evaluate how {\sc{VIPS++}} can explore and approximate multimodal probability distributions with distant modes, we consider the problem of approximating an unknown GMM comprising 10 components. We consider different number of dimensions, namely $D=20$, $D=40$ and $D=60$.
For each component, we draw each dimension of the mean uniformly in the interval $[-50,50]$. The covariance matrices are given by $\boldsymbol{\Sigma} = \mathbf{A}^\top\mathbf{A} + \mathbf{I}_{D}$ where each entry of the $D\times D$-dimensional matrix $\mathbf{A}$ is sampled from a normal distribution with mean $0$ and standard deviation $0.1D$. Note that each component of the target distribution can have a highly correlated covariance matrix, which is even a problem for the tested {\sc{MCMC}} methods. 

\subsubsection{Planar Robot}
In order to test {\sc{VIPS++}} on a multimodal problem with non-Gaussian modes we devised a challenging toy task where we want to sample the joint configurations of a planar robot with \num{10} links of length \num{1} such that it reaches desired goal positions. The robot base is at position $(0,0)$ and the joint configuration describes the angles of the links in radian. In order to induce smooth configurations, we put a zero mean Gaussian prior on the joint configurations where we use a variance of \num{1} for the first joint and a variance of \num{4e-2} for the remaining joints.
Deviations from the nearest goal position are penalized based on a likelihood that is given by a  Gaussian distribution in the Cartesian end-effector space, with a variance of \num{1e-4} in both directions.
We consider two experiments that differ in the number of goal positions. For the first experiment, we want to reach a single goal-position at position $x=7$ and $y=0$. For the second experiment we want to reach four goal positions at positions $(7,0)$, $(0,7)$, $(-7,0)$ and $(0, -7)$. Please refer to Appendix~\ref{app:planarRobot} for details on how the target distribution is computed.

\begin{figure}
    \centering
	\begin{subfigure}{.494\textwidth}
		\centering
		\includegraphics[width=.99\linewidth]{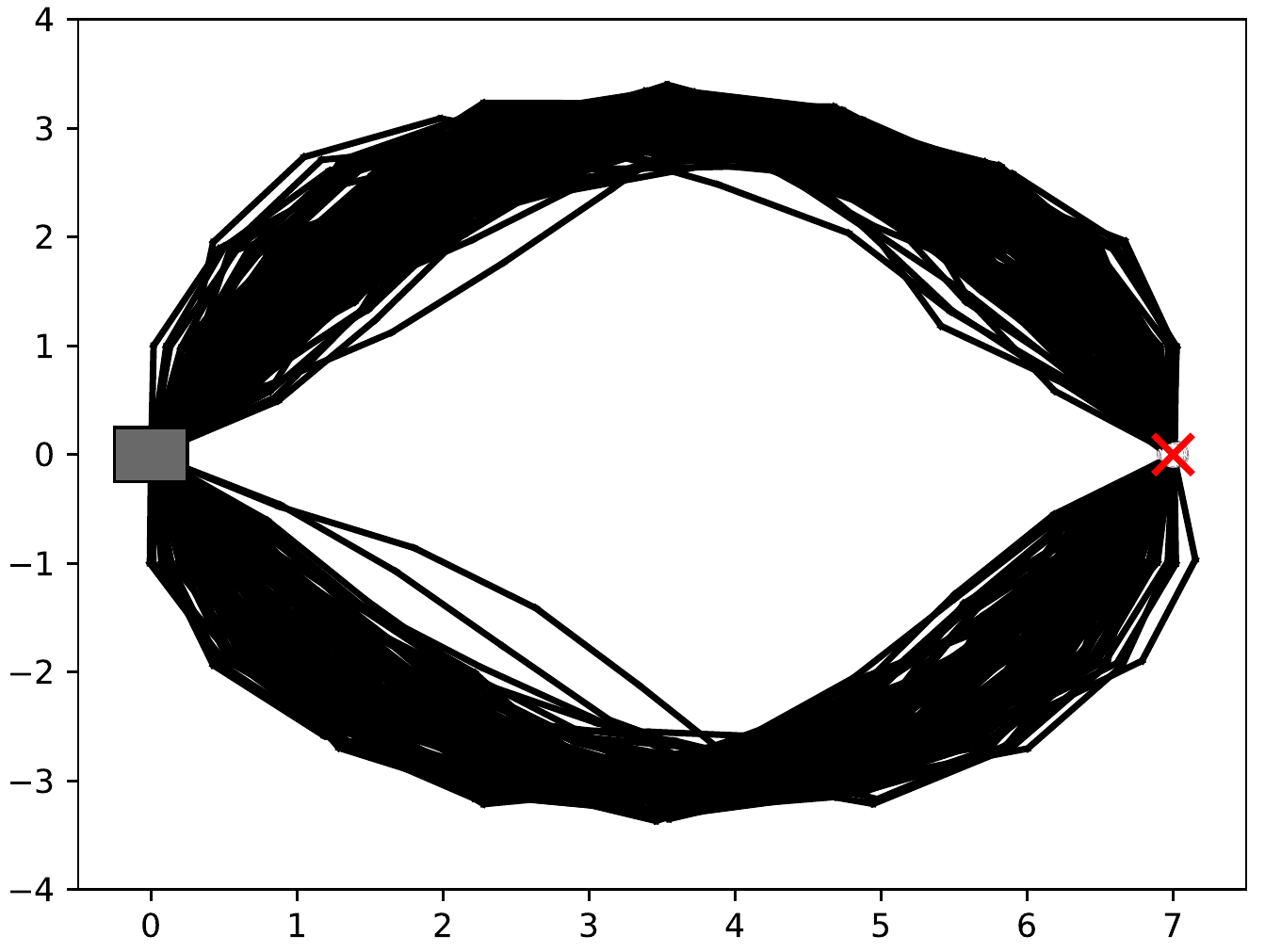}
	\end{subfigure}
	\begin{subfigure}{.495\textwidth}
		\centering
		\includegraphics[width=.99\linewidth]{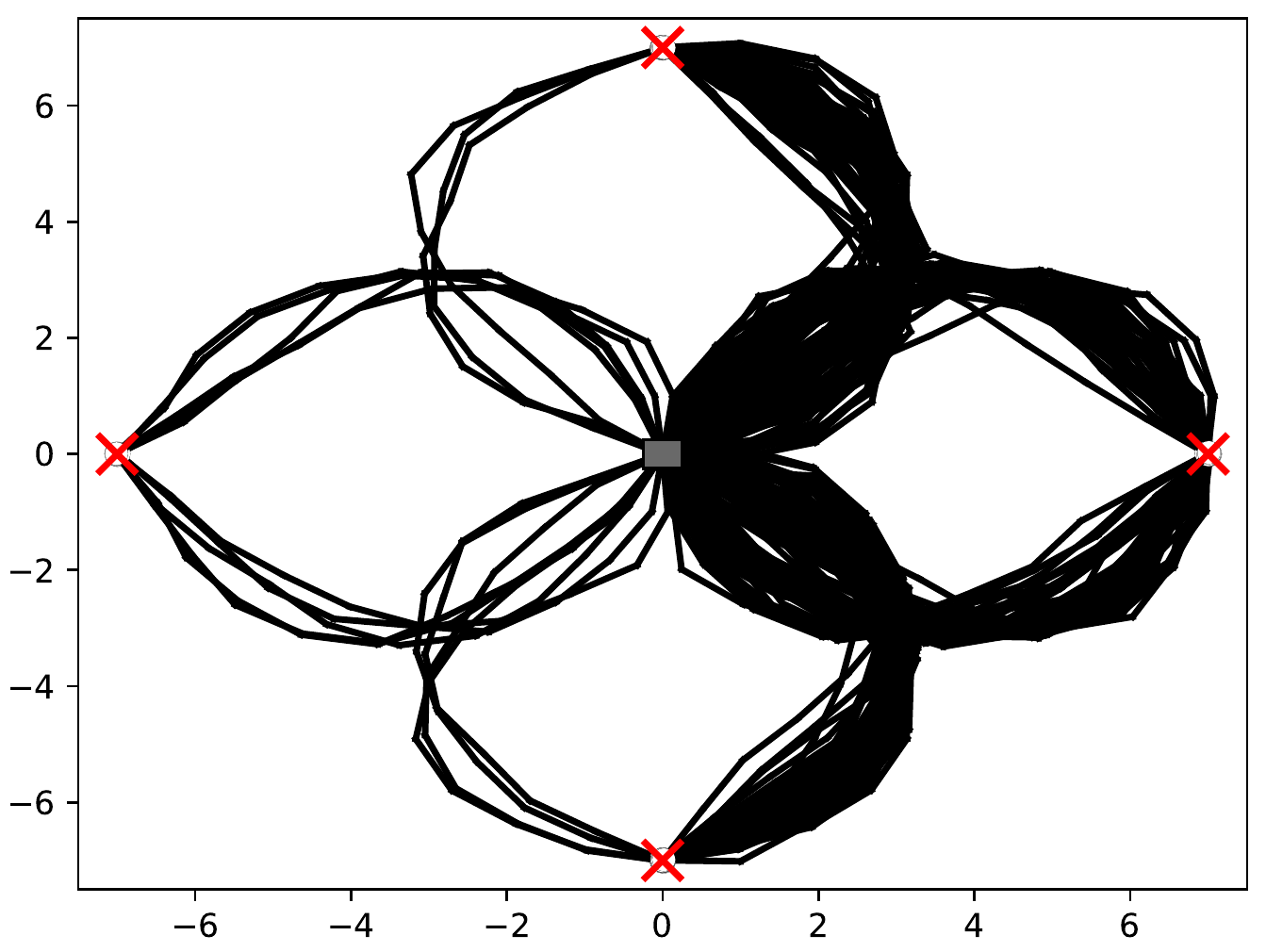}
	\end{subfigure}%
		\caption{The plots show $200$ ground-truth samples for both $\textit{planar robot}$ experiments that have been generated using generalized elliptical slice sampling. The base of the planar robot is shown as a gray box and the end-effector positions are shown as circles.}
		\label{fig:planar_groundtruth}
	\end{figure}

Ground-truth samples for both experiments are shown in Figure~\ref{fig:planar_groundtruth}. Each goal position can be reached from two different sides, either up and down, or left and right. Other configurations that would reach the goal position, for example some zig-zag configurations, are not relevant due to the smoothness prior and can create poor local optima.
Although there are only two relevant ways for reaching each goal position, closely approximating these modes can require many mixture components, because the small variance of the Cartesian likelihood term enforces components with small variance. We therefore also evaluate slightly different hyper-parameters for {\sc{VIPS++}}, where we add a new component at every iteration.
	
\subsection{Ablations}
\label{sec:Ablations}
In this subsection we will evaluate the effects of some algorithmic choices. Namely, we will show that adapting the number of components can be crucial for discovering relevant modes of multimodal target distributions, that the previously proposed initializing of covariance matrices can have detrimental effects, and that the sample reusage of {\sc{VIPS++}} can significantly increase sample efficiency.

\subsubsection{Adapting the number of components}
\label{sec:ablations:covinit}
\begin{figure}[p]
	\begin{subfigure}{.48\textwidth}
		\centering
		\includegraphics[width=.99\linewidth]{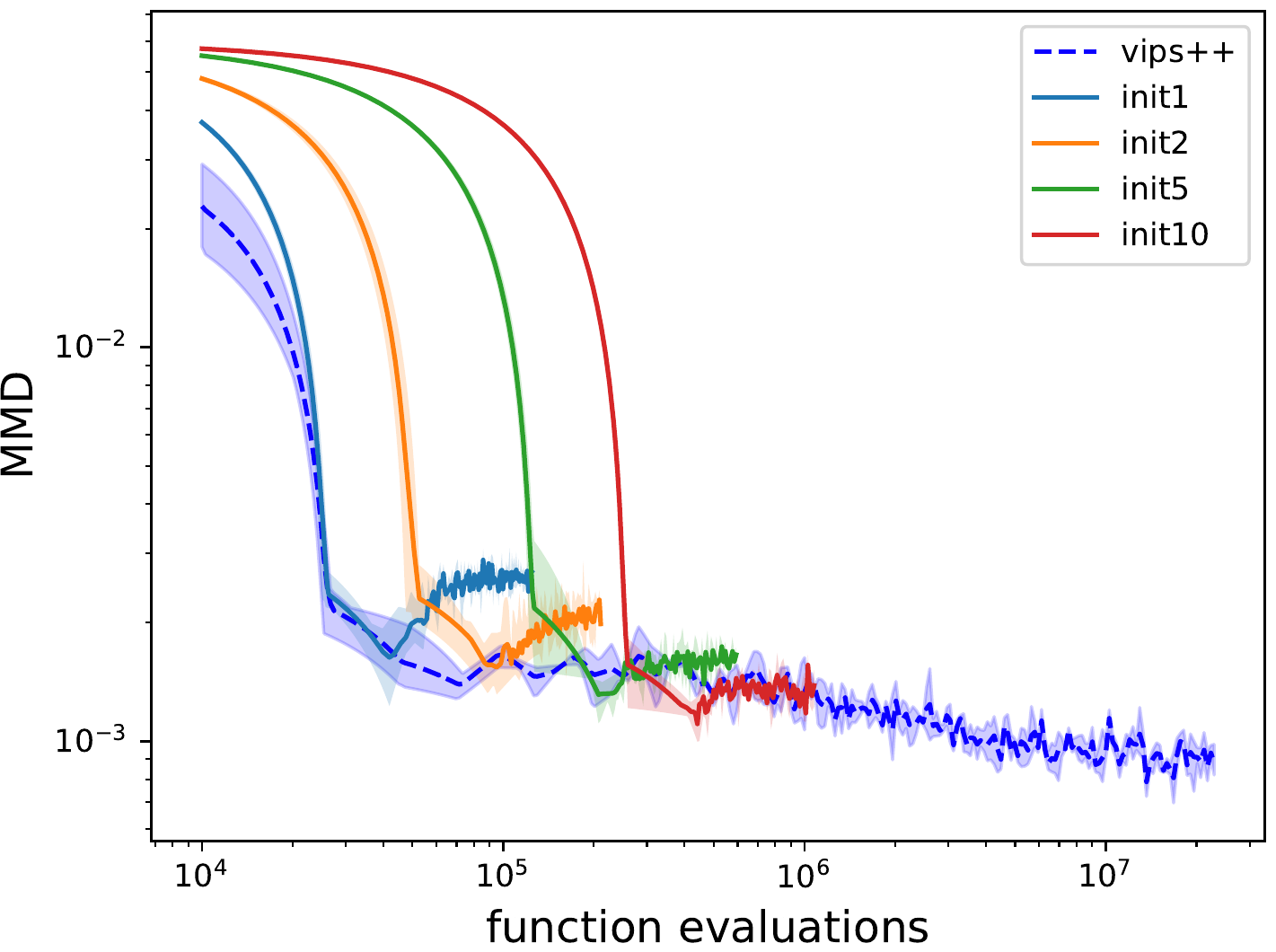}
		\caption{breast cancer}
	\end{subfigure}
		\begin{subfigure}{.48\textwidth}
		\centering
		\includegraphics[width=.99\linewidth]{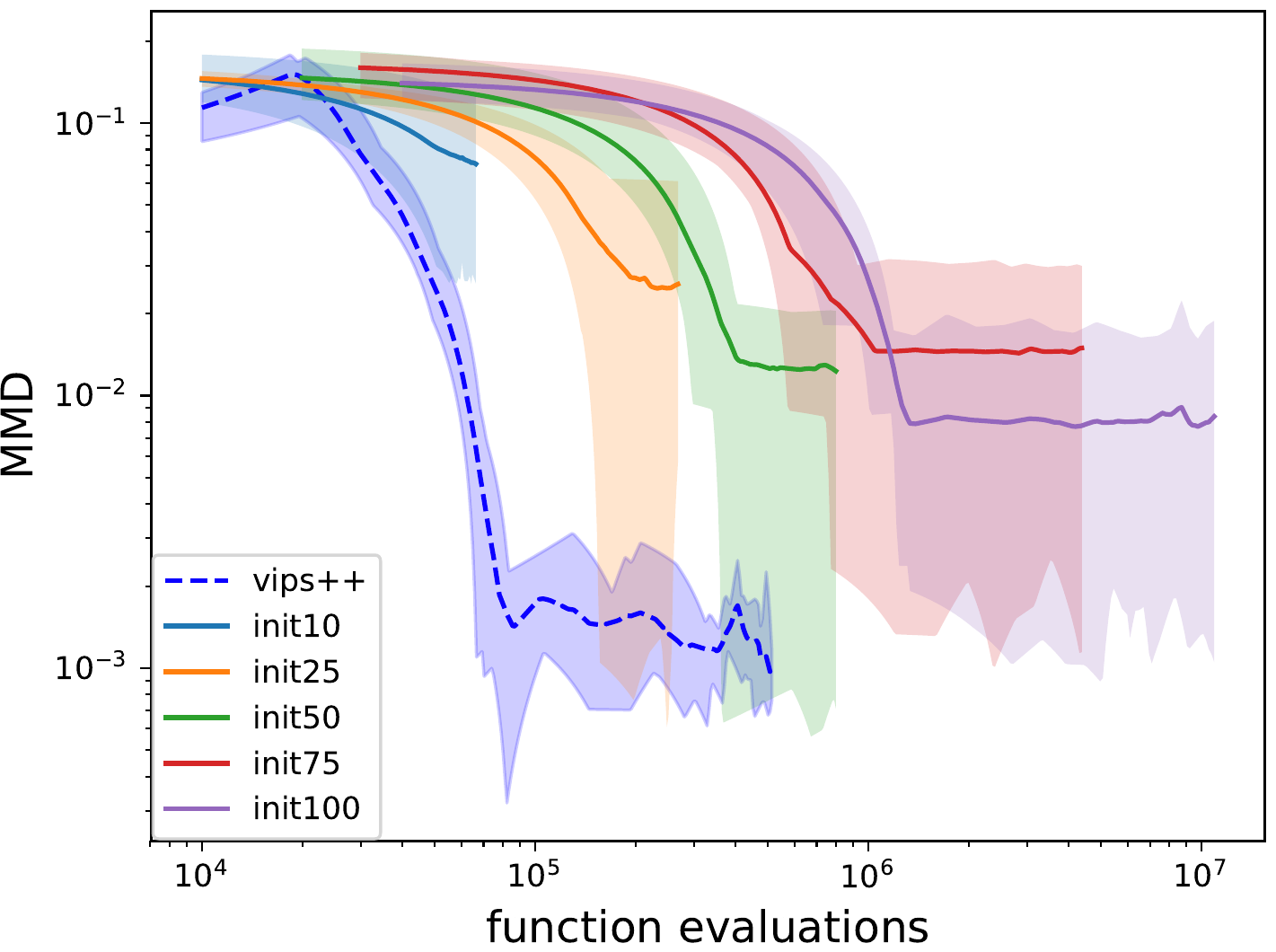}
		\caption{GMM}
	\end{subfigure}
		\caption{We compare VIPS++ with a variant that does not add or delete components. On the \textit{breast cancer} experiment, {\sc{VIPS++}} converges to a good approximation as fast as the variant that learns a single component, but it refines the approximation by adding more components. When not adapting the number of components on the \textit{GMM} experiment, the quality of the approximation strongly depends on the initialization and even 100 initial components would sometimes fail to detect all modes.}
		\label{fig:gmmAblationFixedNum}
	\end{figure}

As discussed in Section~\ref{sec:adapting_num_components}, {\sc{VIPS}} automatically adapts the number of components during learning for better exploration, which enables it to improve on local optimal solutions. We evaluate the effect of this adaptation by comparing {\sc{VIPS++}} with a variant that keeps the number of components fixed on the \textit{breast cancer} experiment and the \num{20}-dimensional \textit{GMM} experiment.
We initialize the non-adaptive variant with different numbers of initial components, where each mean is drawn from an isotropic Gaussian $\mathcal{N}(\mathbf{0},\alpha \boldsymbol{I})$. We use $\alpha=100$ for the \textit{breast cancer} experiment and $\alpha=1000$ for the \textit{GMM} experiments. For {\sc{VIPS++}} we start with a single component with mean $\mathbf{0}$. All covariance matrices are initialized as $\Sigma = \alpha \boldsymbol{I}$. The achieved MMDs are shown in Figure~\ref{fig:gmmAblationFixedNum}. The non-adaptive variant converges to better approximations when increasing the number of components on the \textit{breast cancer} experiment. However, the required number of function evaluations until convergence scales approximately linearly with the number of components. {\sc{VIPS++}} can learn good approximations with few function evaluations and further improves while increasing the size of the mixture model. On the \textit{GMM} experiment, all tested variants would in principle be able to model the target distribution exactly. However, depending on the initialization, several components may converge to the same mode which results in bad local optima. We therefore needed at least \num{25} initial components for occasionally learning good approximations during this experiment and even when initializing with \num{100} components the non-adaptive variant would sometimes fail to discover all true modes. In contrast, by adaptively adding new components at interesting regions {\sc{VIPS++}} reliably discovers all ten modes. Please refer to Appendix~\ref{app:numComponents} for a plot of the average number of components that are learned by {\sc{VIPS++}} for all experiments in the test bed.

\subsubsection{Initializing the covariance matrices}
\begin{figure}[p]
	\begin{subfigure}{.48\textwidth}
		\centering
		\includegraphics[width=.99\linewidth]{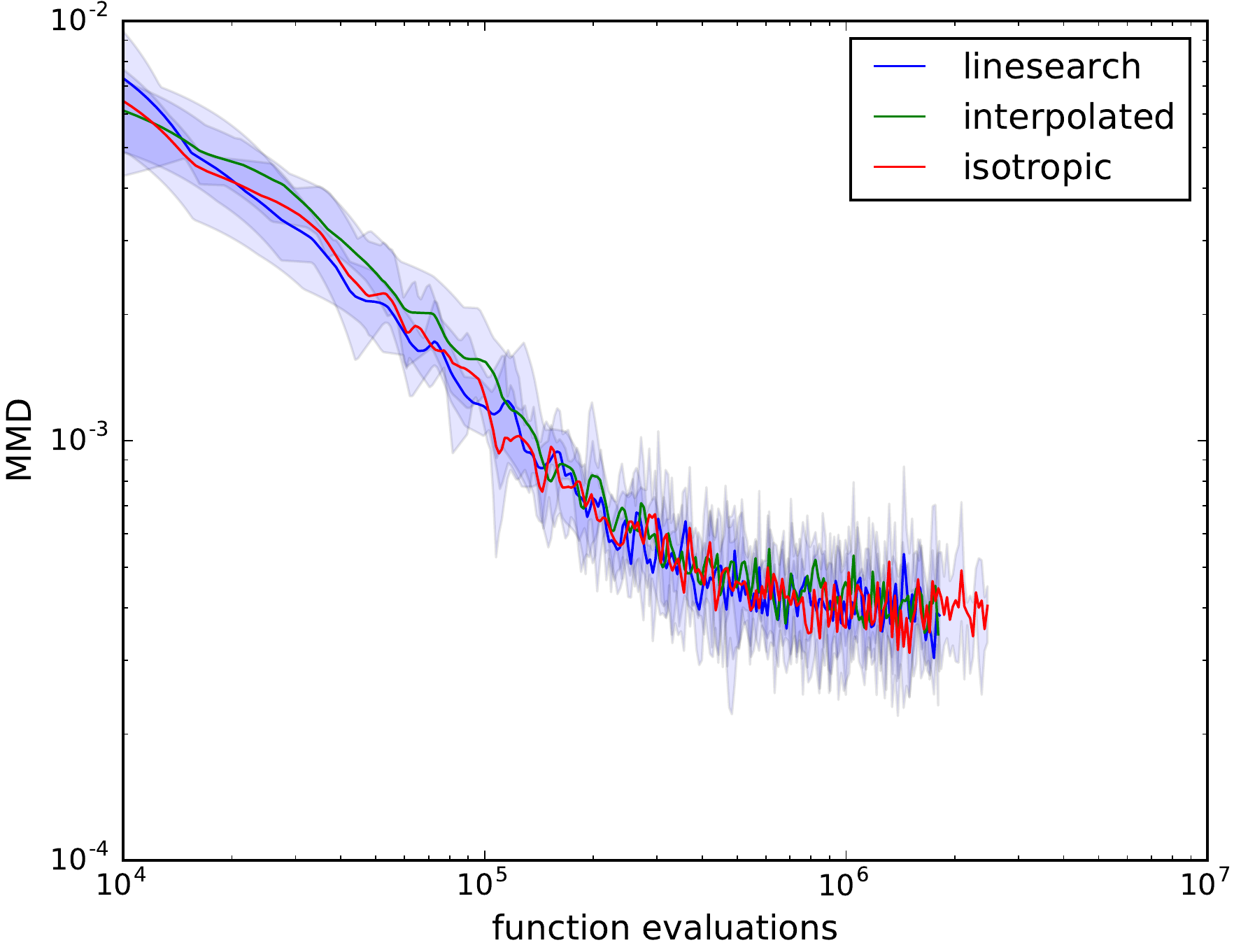}
		\caption{Goodwin model}
	\end{subfigure}
	\begin{subfigure}{.48\textwidth}
		\centering
		\includegraphics[width=.99\linewidth]{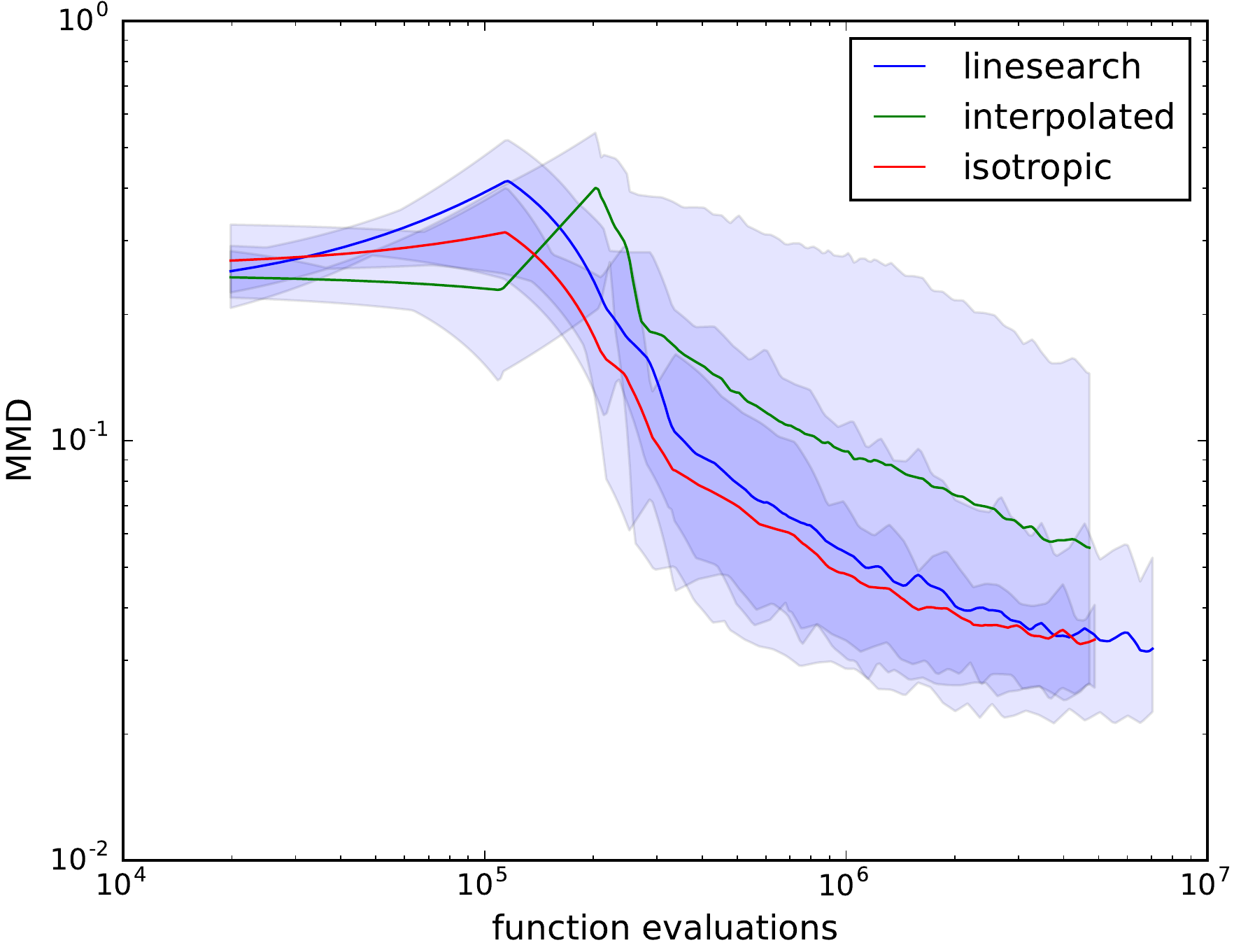}
		\caption{planar robot}
	\end{subfigure}%
		\caption{We compare different strategies for initializing the covariance matrices of newly added components. Interpolating the covariance matrices of the current model based on the responsibilities can have detrimental effects as shown in the planar robot experiment. }
		\label{fig:covAblations}
	\end{figure}

We also evaluate the different strategies for initializing the covariance matrix of a newly added component, which were discussed in Section~\ref{sec:adapting_num_components}. We compare the proposed line search used by {\sc{VIPS++}} with the interpolation used by {\sc{VIPS}} as well as an isotropic initialization. Figure~\ref{fig:covAblations} compares the different strategies on the \textit{Goodwin} experiment and the \textit{planar robot} experiment (with four goal positions). The \textit{planar robot} experiment shows, that interpolating based on the responsibilities can seriously impair the performance on multimodal problems. We believe that interpolating based on the responsibilities can lead to highly anisotropic initial covariance matrices that do not sufficiently explore along relevant directions which would explain the detrimental effects. Although we could not show a benefit of the line search compared to the isotropic initialization, we opted for the line search for the quantitative experiments, because it seems sensible and did not perform significantly worse in our experiments. 

\subsubsection{Sample Reusage}
\begin{figure}
	\begin{subfigure}{.48\textwidth}
		\centering
		\includegraphics[width=.99\linewidth]{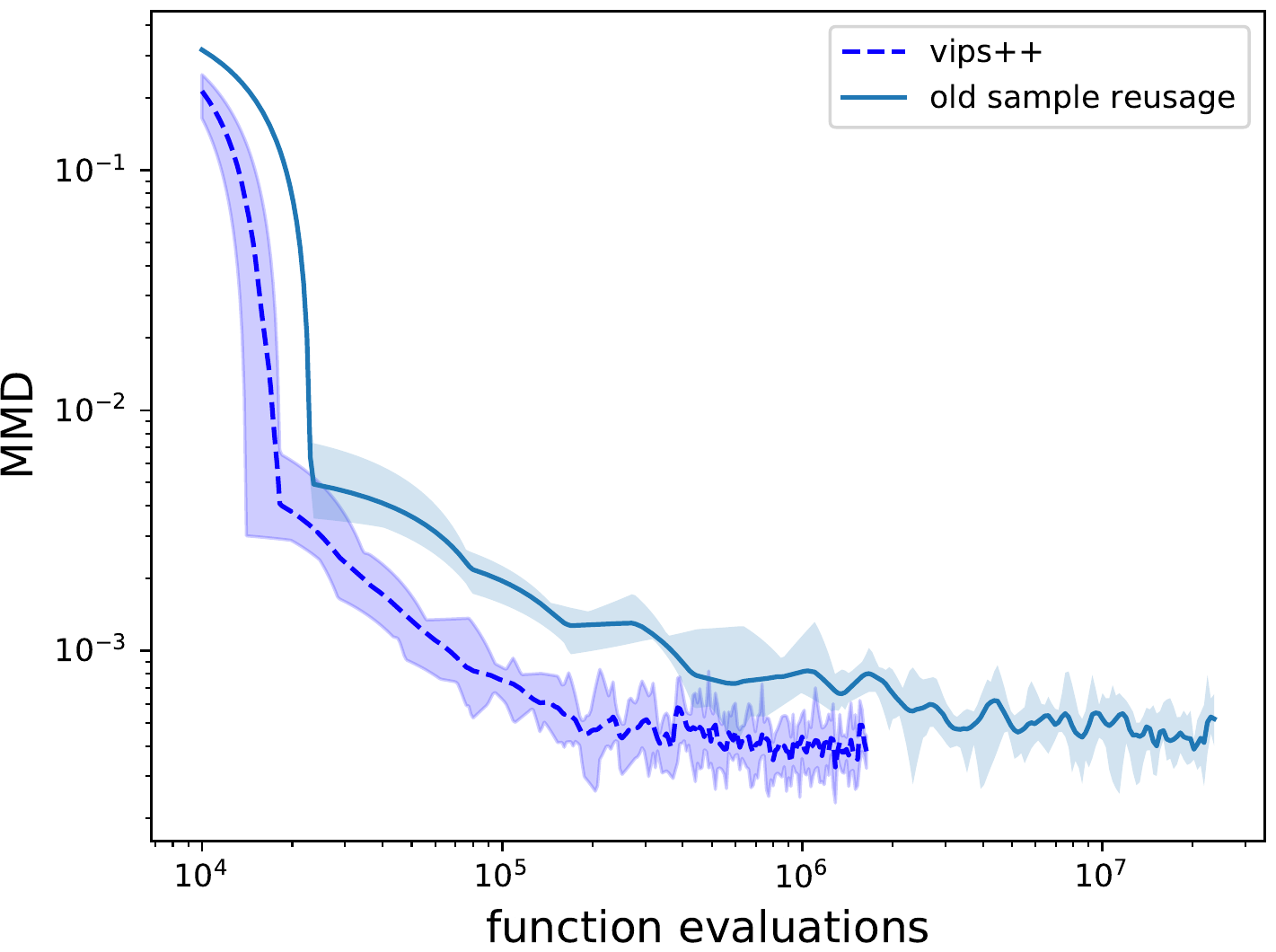}
		\caption{Goodwin model}
	\end{subfigure}
	\begin{subfigure}{.48\textwidth}
		\centering
		\includegraphics[width=.99\linewidth]{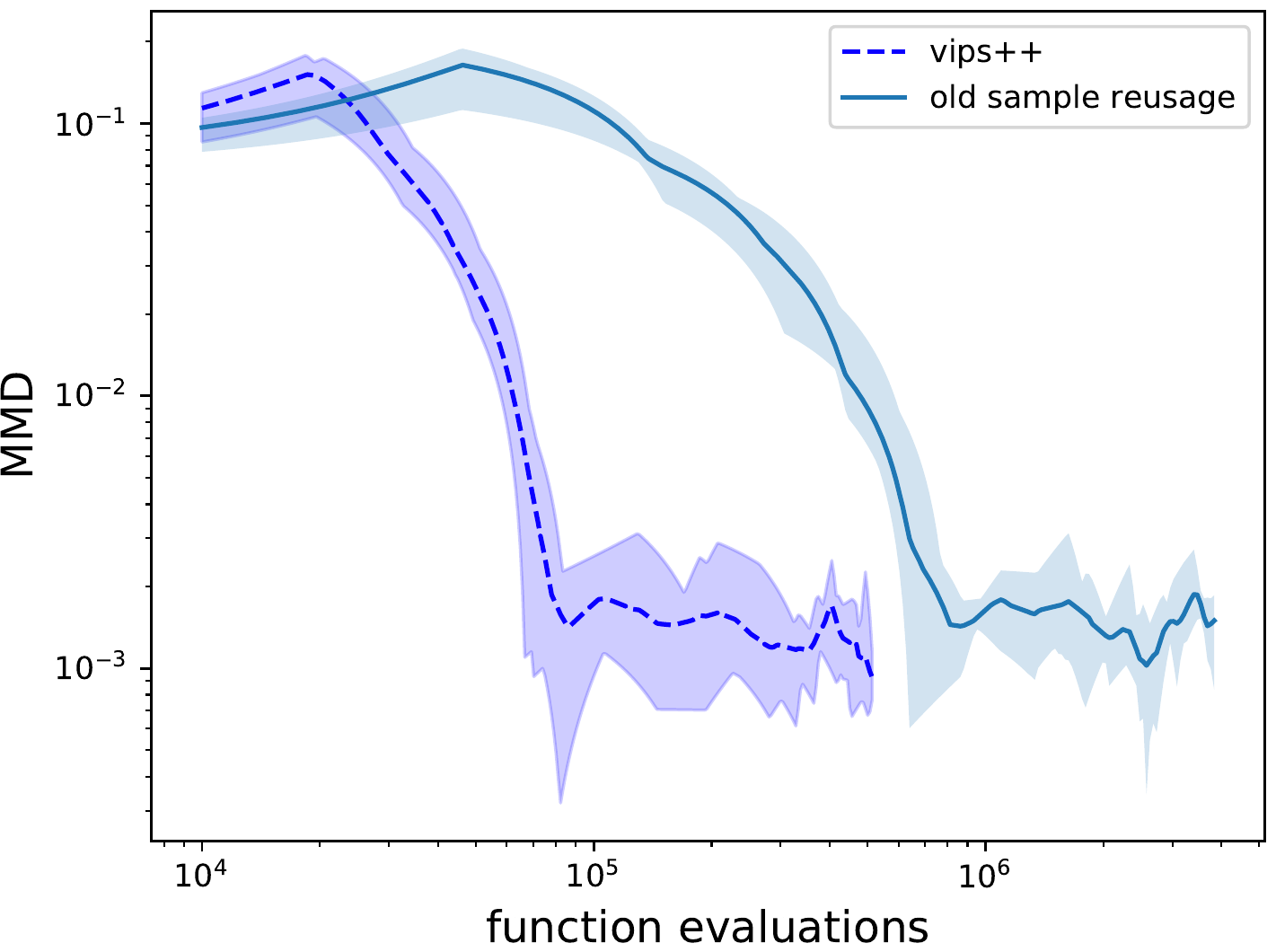}
		\caption{20-dimensional GMM}
	\end{subfigure}%
		\caption{The sample reusage of {\sc{VIPS++}} is approximately one order of magnitude more efficient than the sample reusage of {\sc{VIPS}}. }
		\label{fig:sampleReusage}
	\end{figure}

Compared to {\sc{VIPS}}, {\sc{VIPS++}} uses a more sophisticated method for reusing samples from previous iteration---as detailed in Section~\ref{sec:sampleReusage} and~\ref{sec:sampleSelection}---by identifying relevant samples among all previous function evaluations and by controlling the number of new samples from each component based on its number of effective samples. We compare the new sample strategy with the previously employed method of always using the samples of the three most recent iterations. Figure~\ref{fig:sampleReusage} evaluates the different strategies on the \textit{Goodwin} experiment and the \num{20}-dimensional \textit{GMM} experiment. The proposed strategy of {\sc{VIPS++}} significantly outperforms the previous method by reducing the sample complexity by approximately one order of magnitude.

\subsection{Illustrative Experiment}
\label{sec:illustrativeExperiment}
	\begin{figure}
    \centering
	\begin{subfigure}{.277\textwidth}
		\centering
		\includegraphics[width=\linewidth]{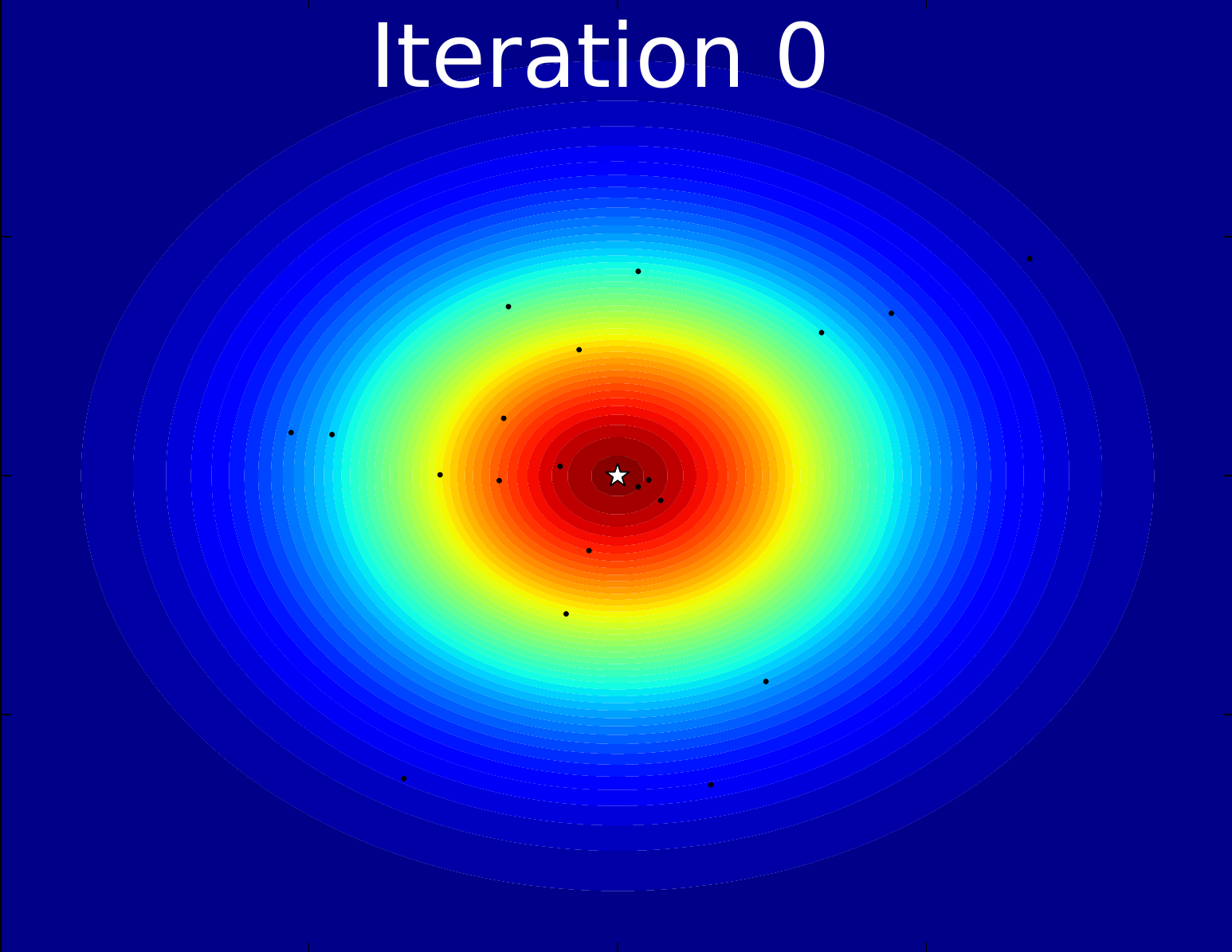}
        \end{subfigure}
        	\begin{subfigure}{.277\textwidth}
		\centering
		\includegraphics[width=\linewidth]{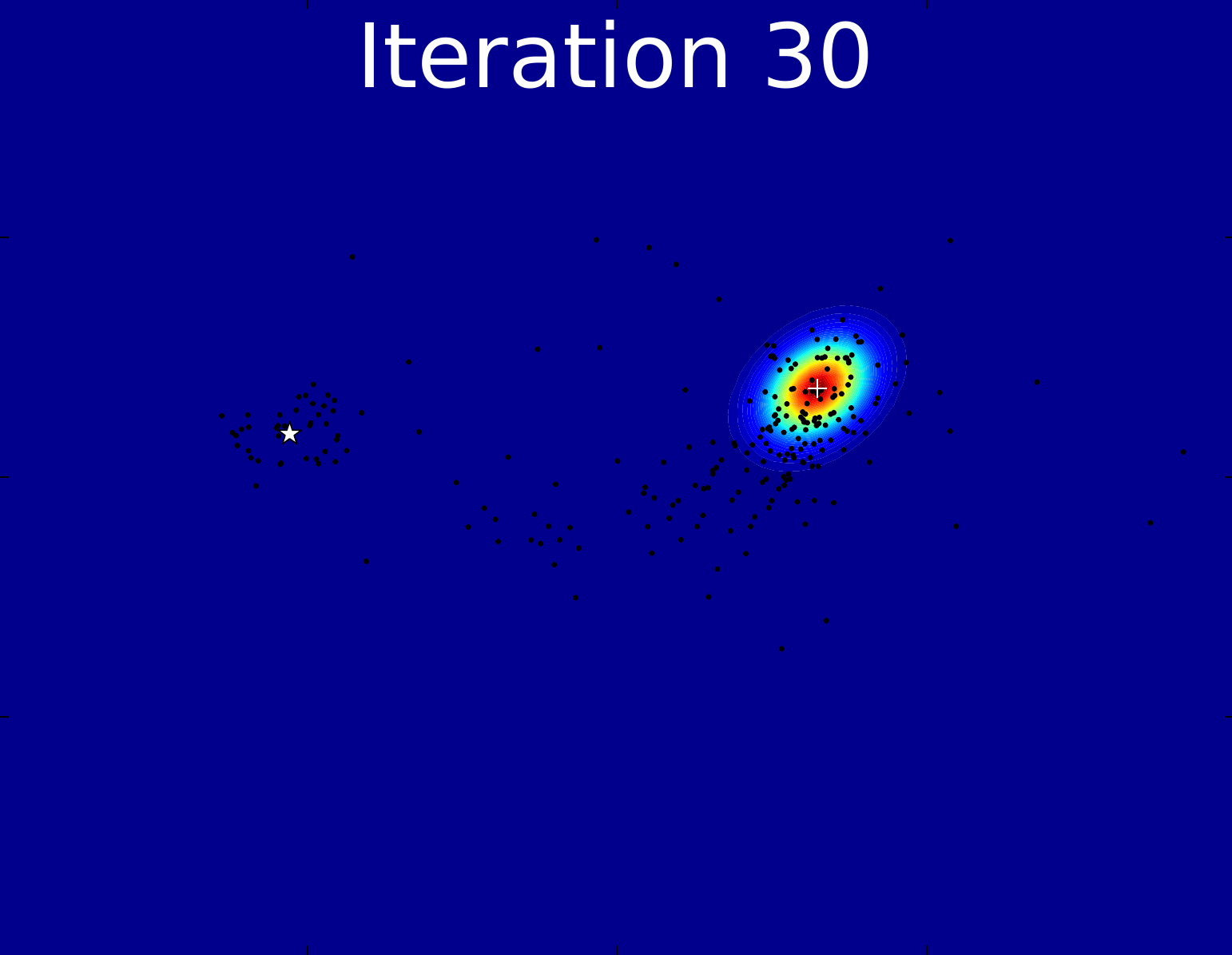}
        \end{subfigure}
        	\begin{subfigure}{.277\textwidth}
		\centering
		\includegraphics[width=\linewidth]{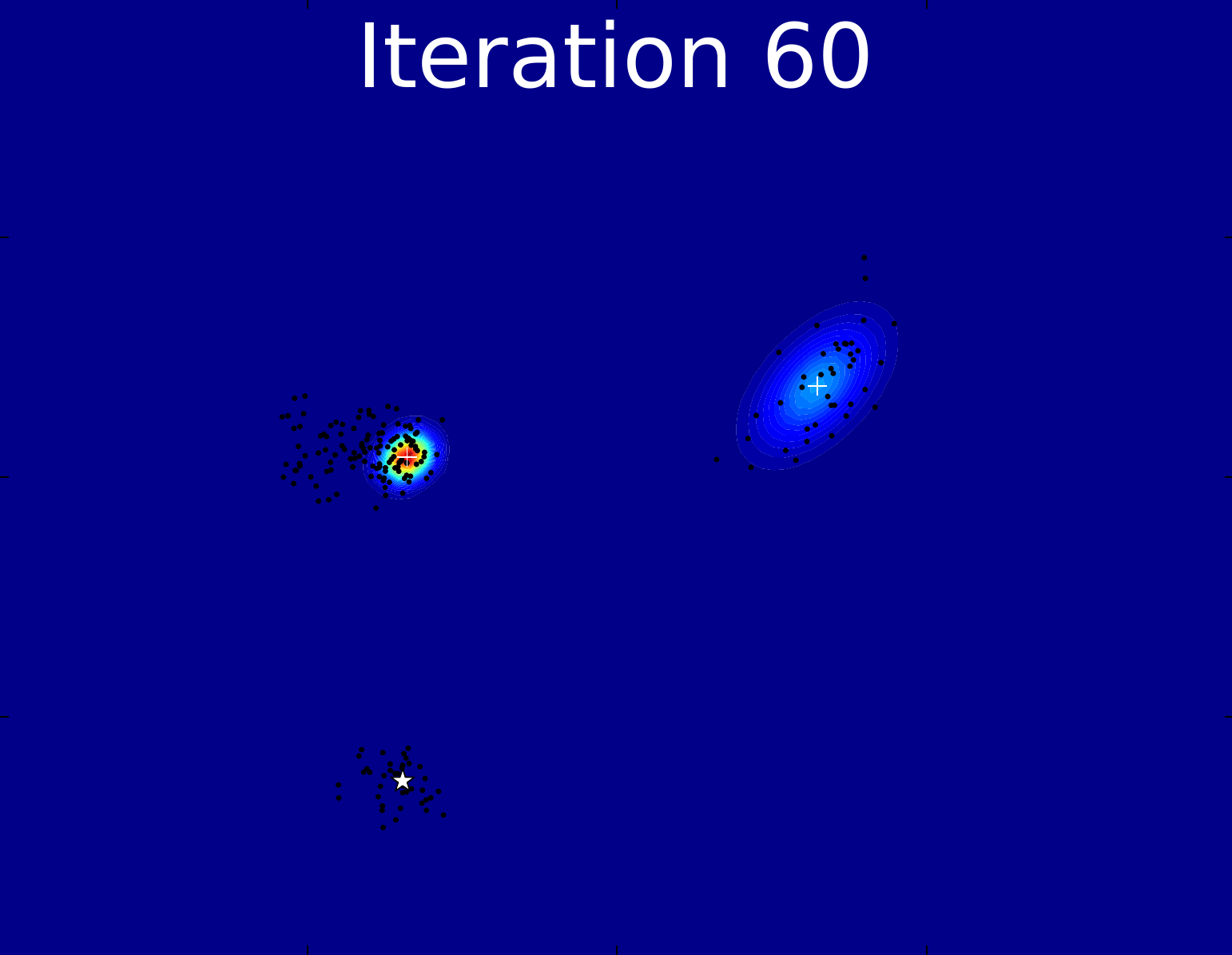}
        \end{subfigure}
        
        	\begin{subfigure}{.277\textwidth}
		\centering
		\includegraphics[width=\linewidth]{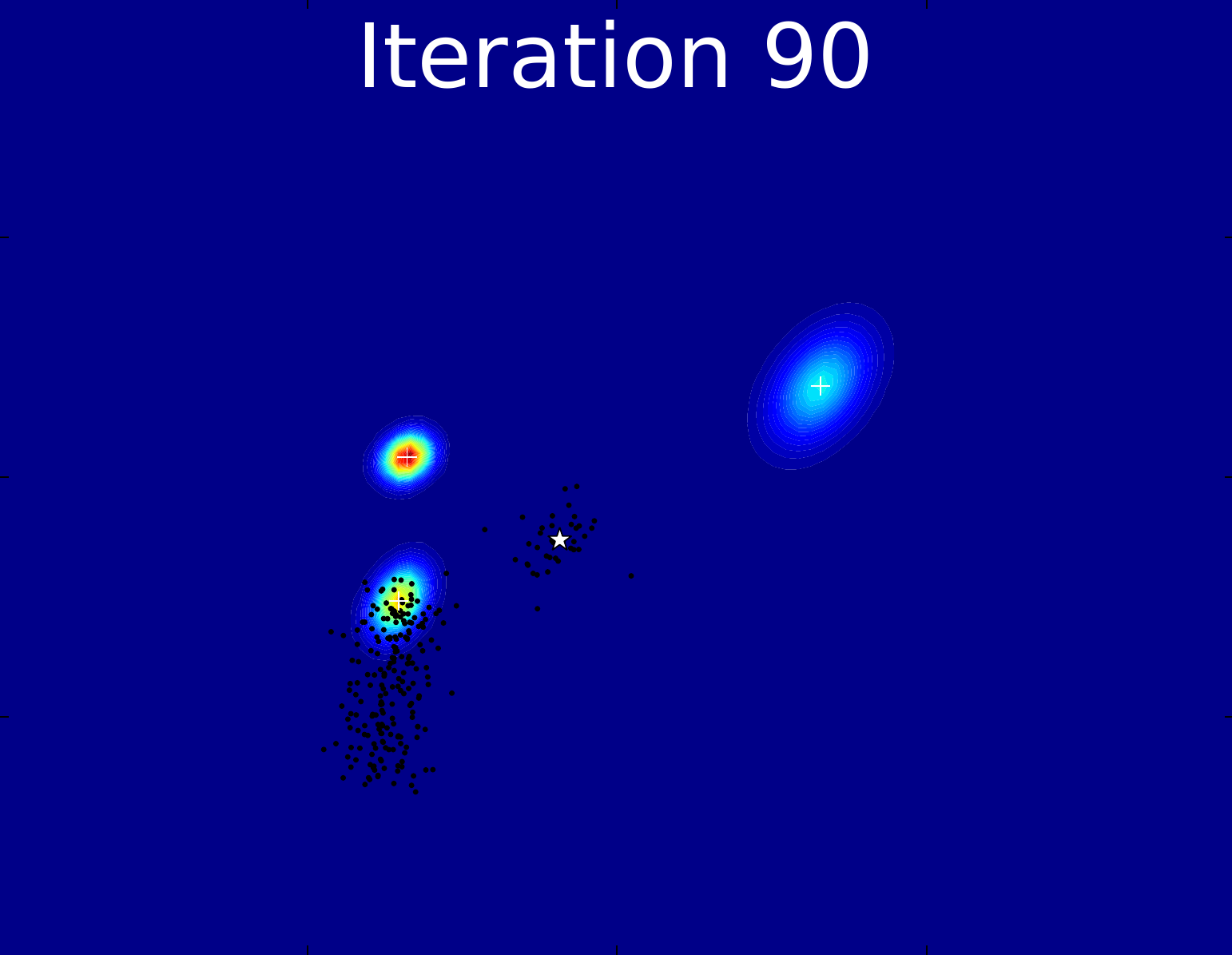}
        \end{subfigure}        
        \begin{subfigure}{.277\textwidth}
		\centering
		\includegraphics[width=\linewidth]{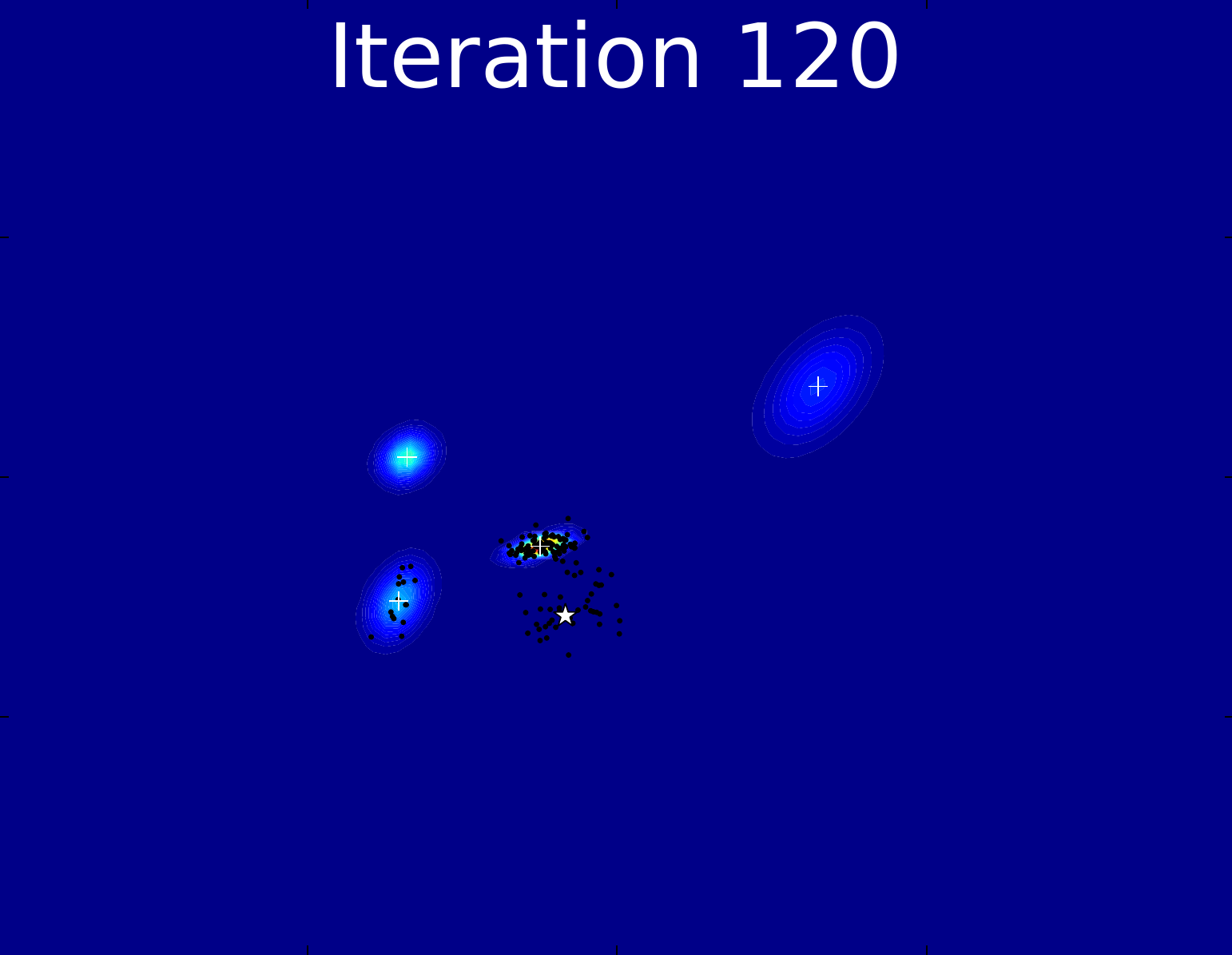}
        \end{subfigure}
        	\begin{subfigure}{.277\textwidth}
		\centering
		\includegraphics[width=\linewidth]{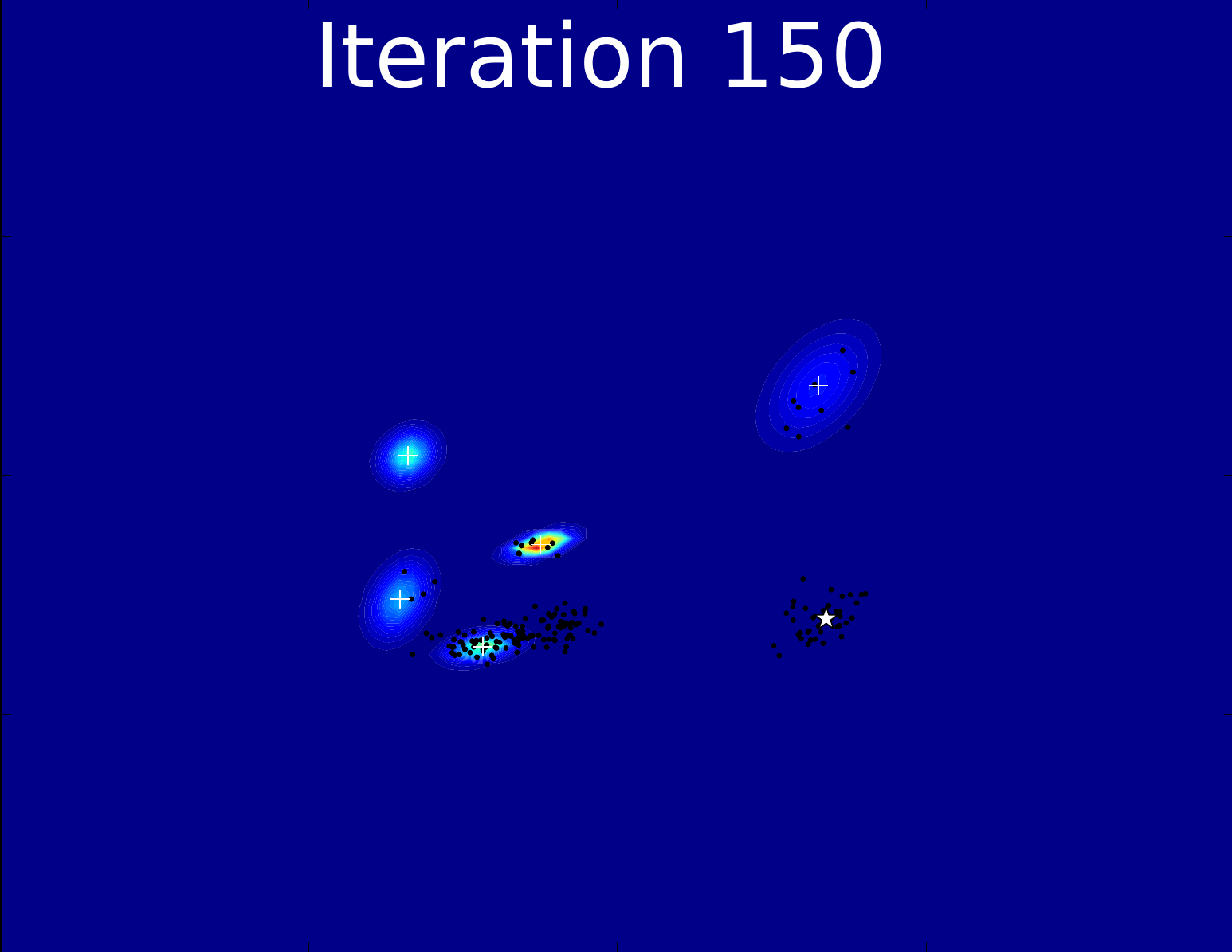}
        \end{subfigure}
        
        	\begin{subfigure}{.277\textwidth}
		\centering
		\includegraphics[width=\linewidth]{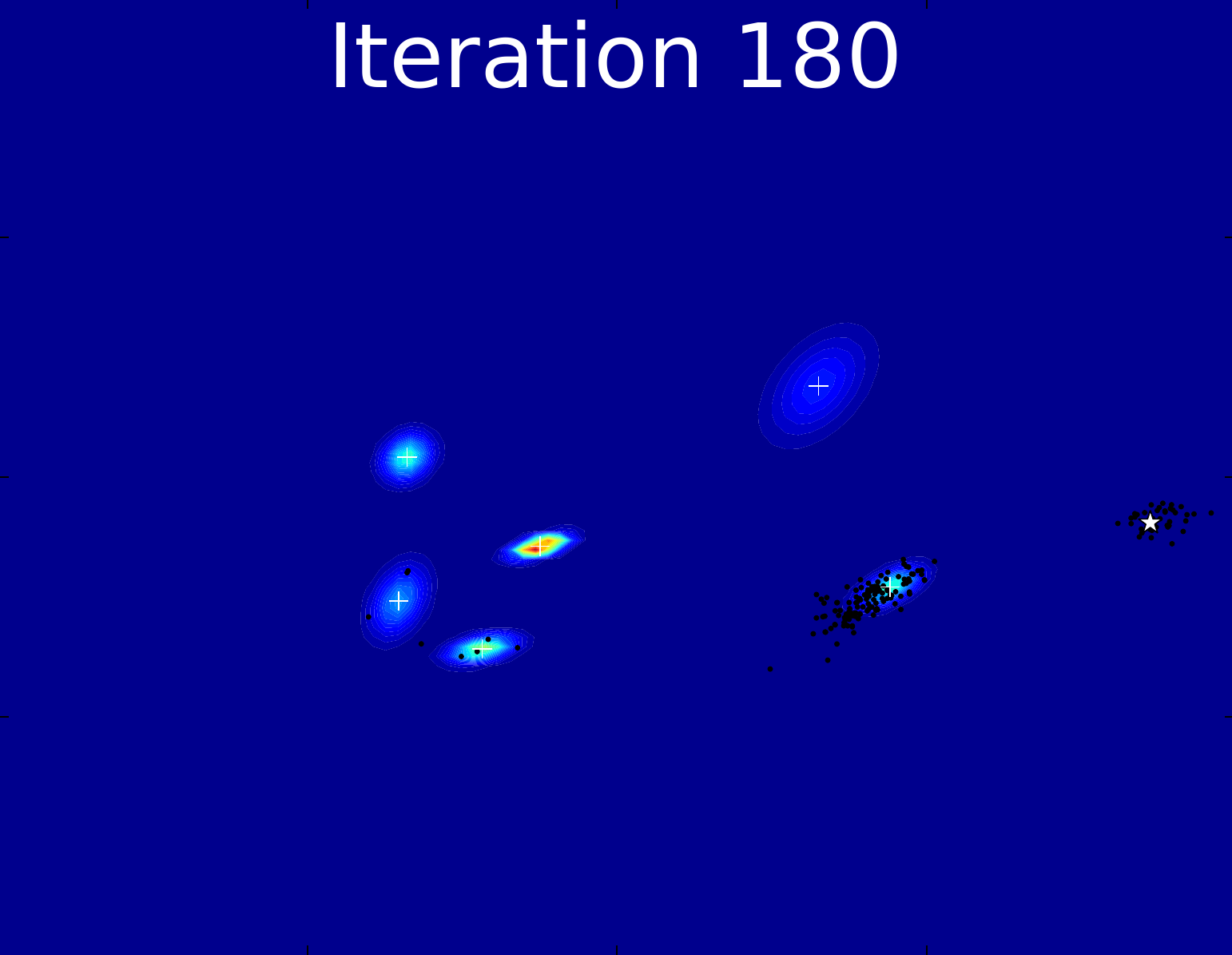}
        \end{subfigure}
        	\begin{subfigure}{.277\textwidth}
		\centering
		\includegraphics[width=\linewidth]{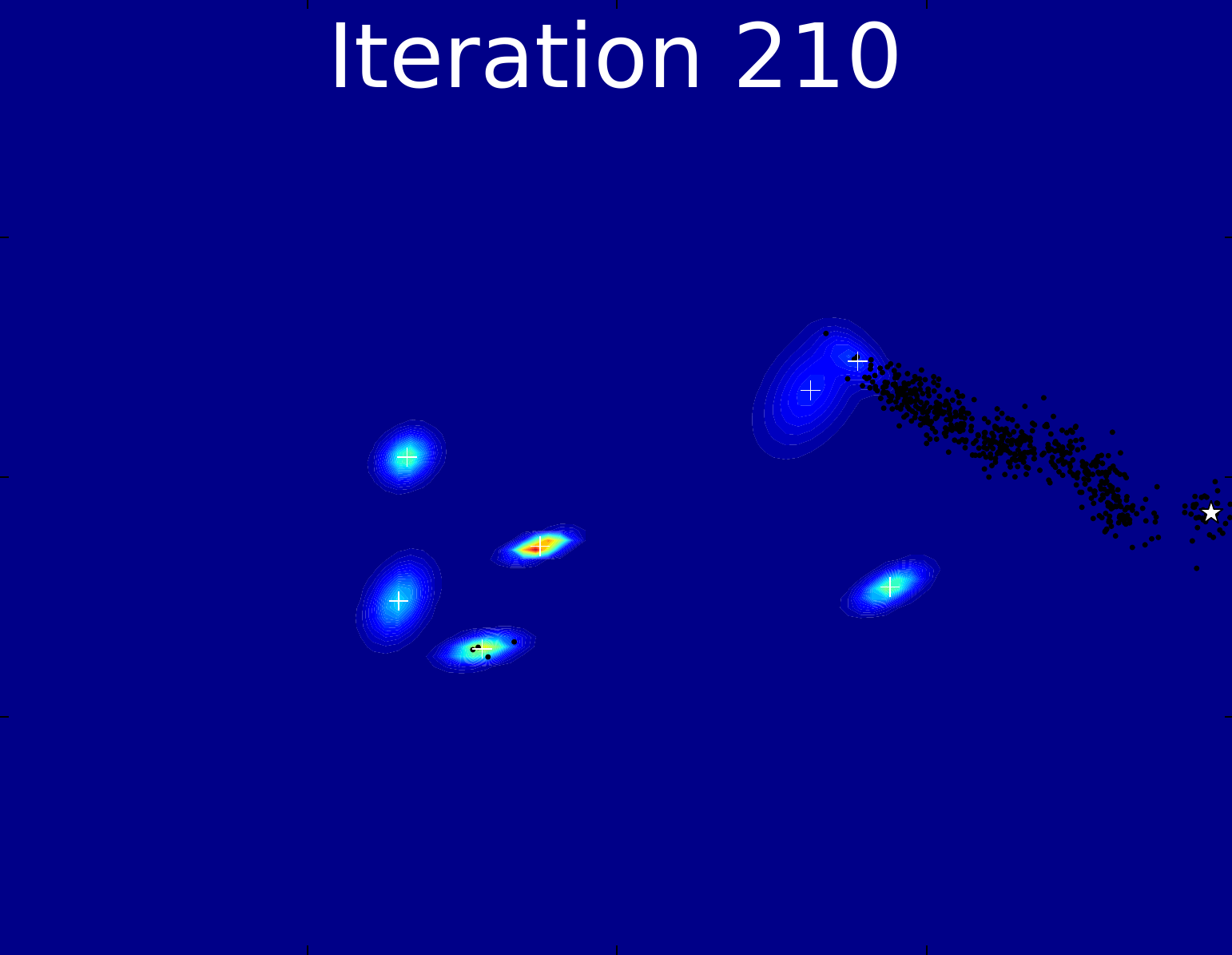}
        \end{subfigure}        
        	\begin{subfigure}{.277\textwidth}
		\centering
		\includegraphics[width=\linewidth]{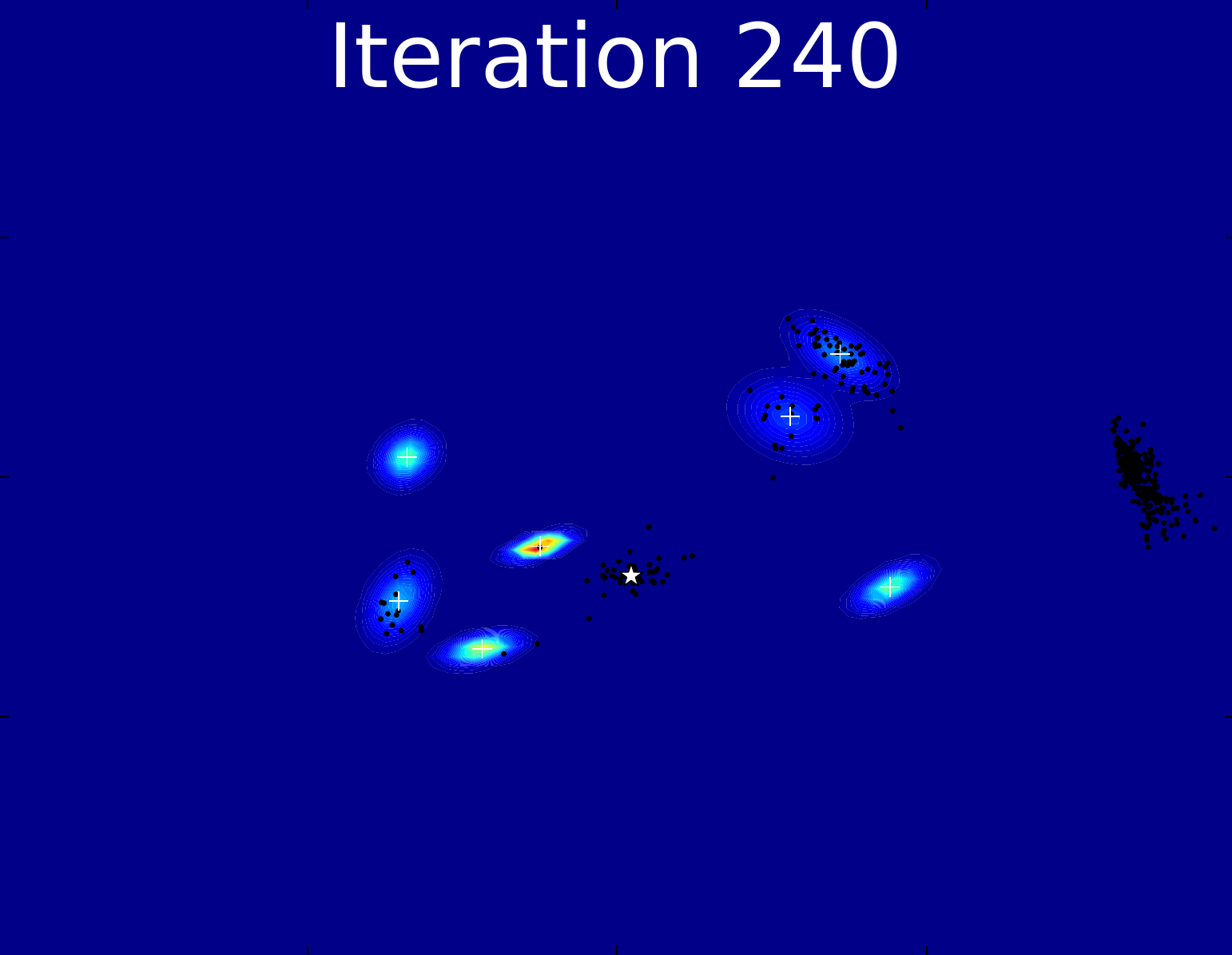}
        \end{subfigure}
        
        	\begin{subfigure}{.277\textwidth}
		\centering
		\includegraphics[width=\linewidth]{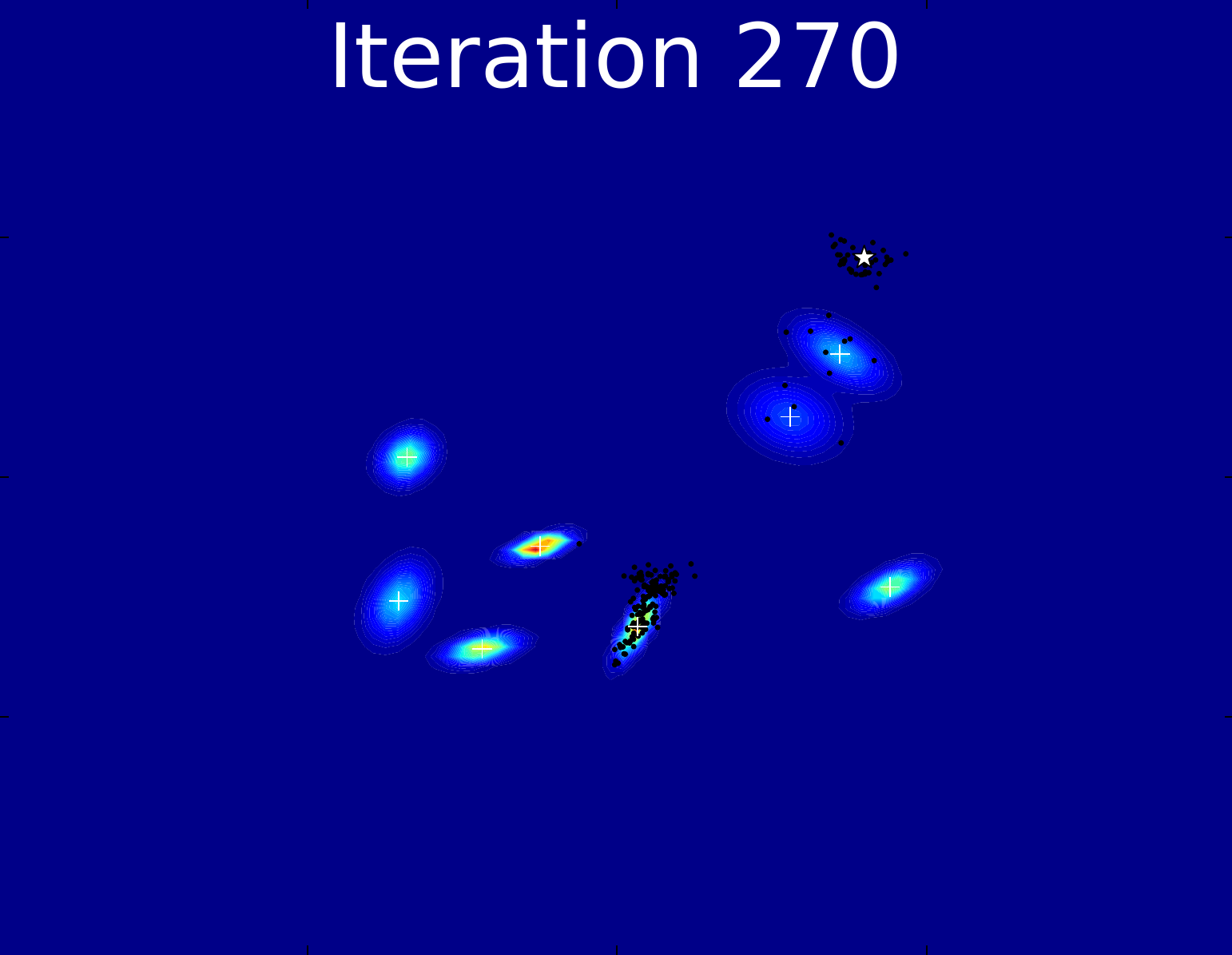}
        \end{subfigure}
        	\begin{subfigure}{.277\textwidth}
		\centering
		\includegraphics[width=\linewidth]{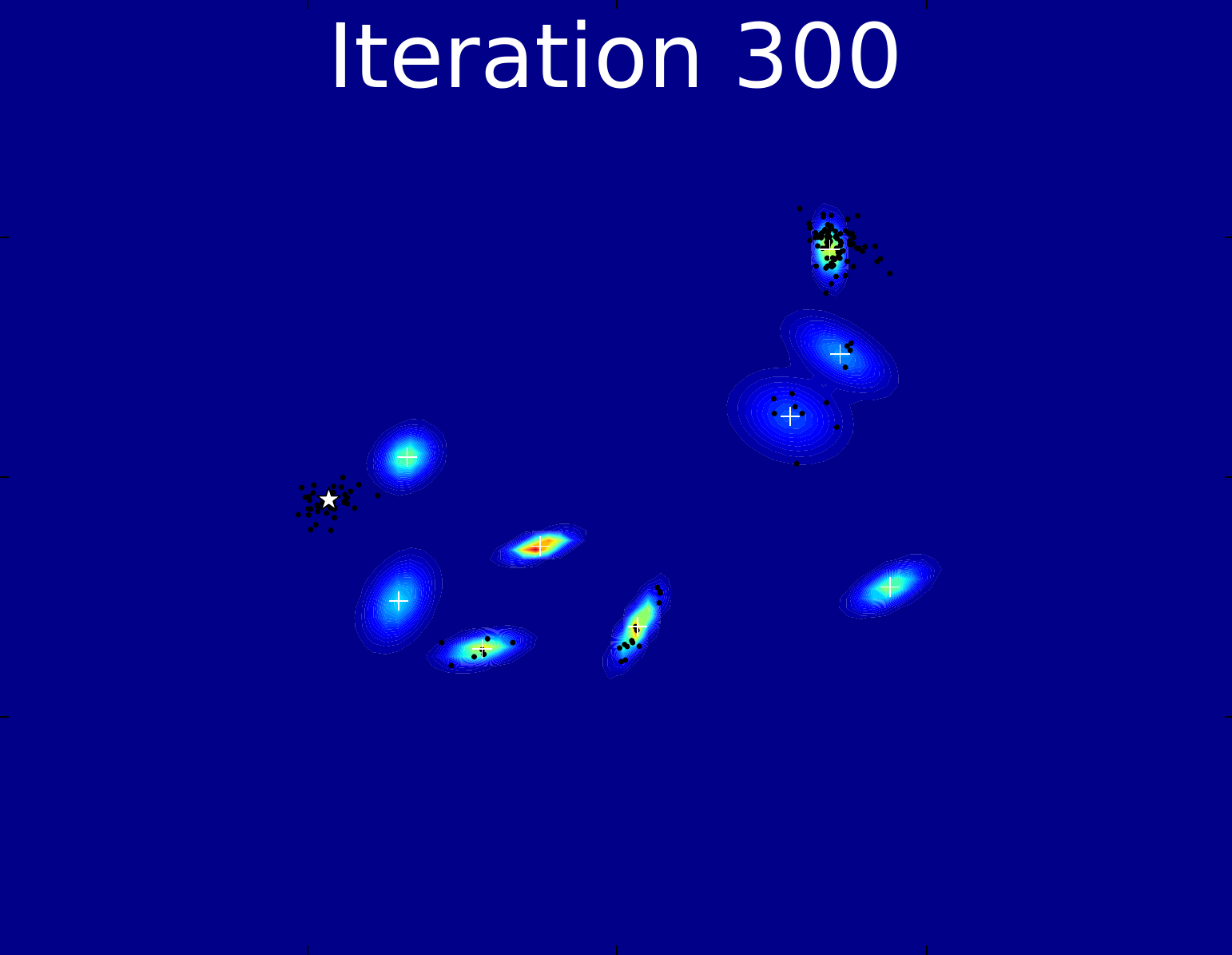}
        \end{subfigure}        
        	\begin{subfigure}{.277\textwidth}
		\centering
		\includegraphics[width=\linewidth]{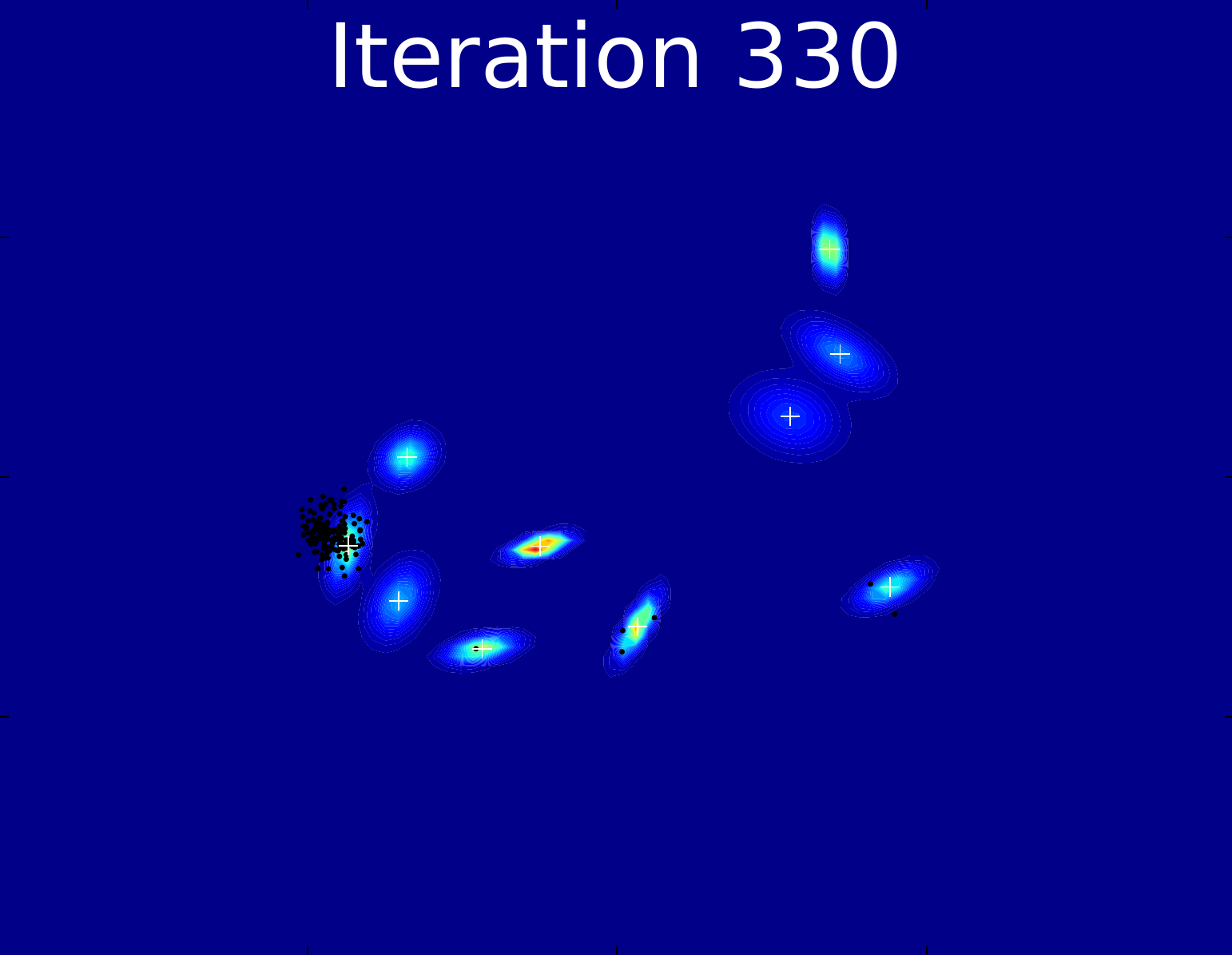}
        \end{subfigure}
        
        \centering
        \begin{subfigure}{.4025\textwidth}
		\centering
		\includegraphics[width=\linewidth]{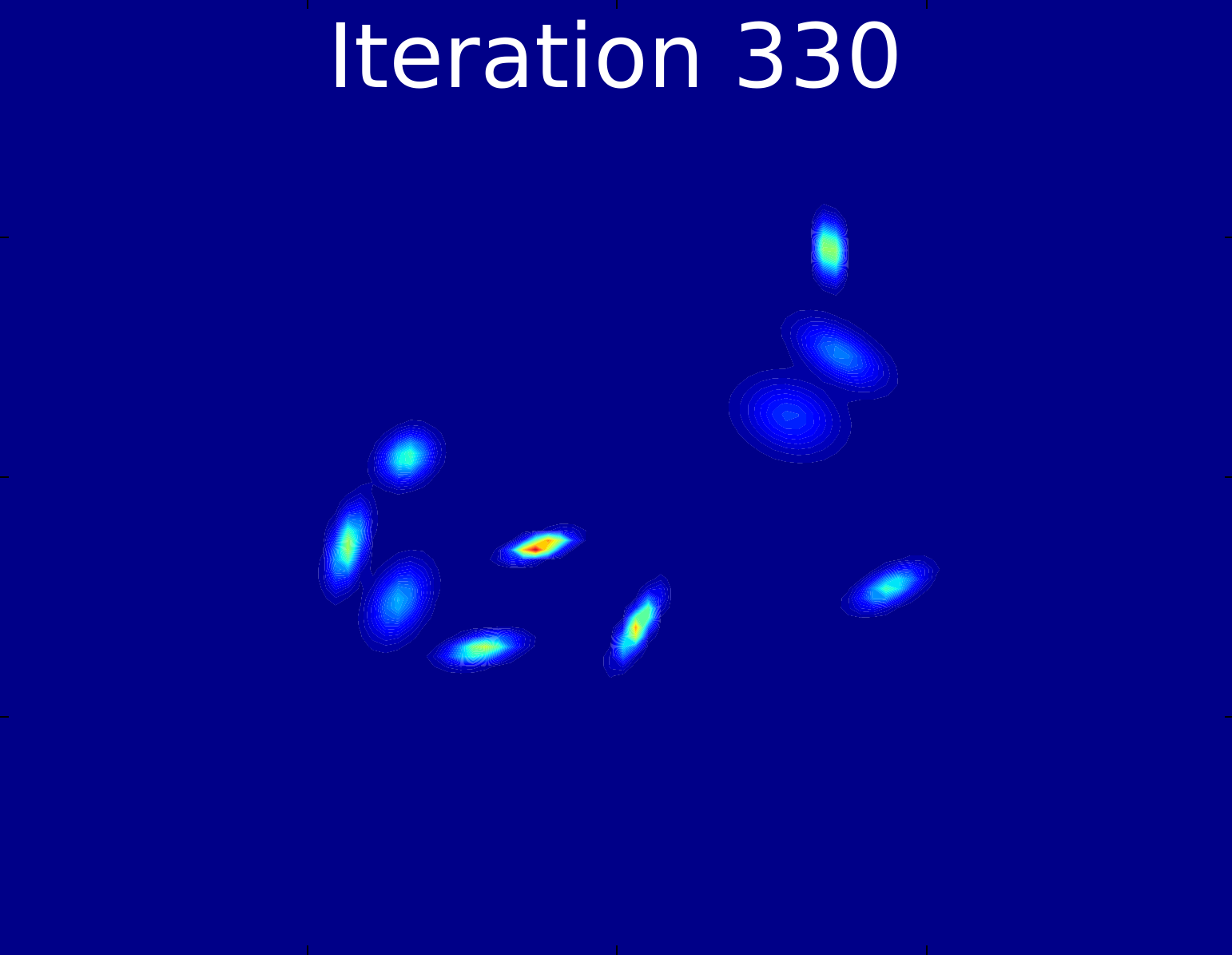}
        \end{subfigure}
        \hspace{11pt}
        	\begin{subfigure}{.4025\textwidth}
		\centering
		\includegraphics[width=\linewidth]{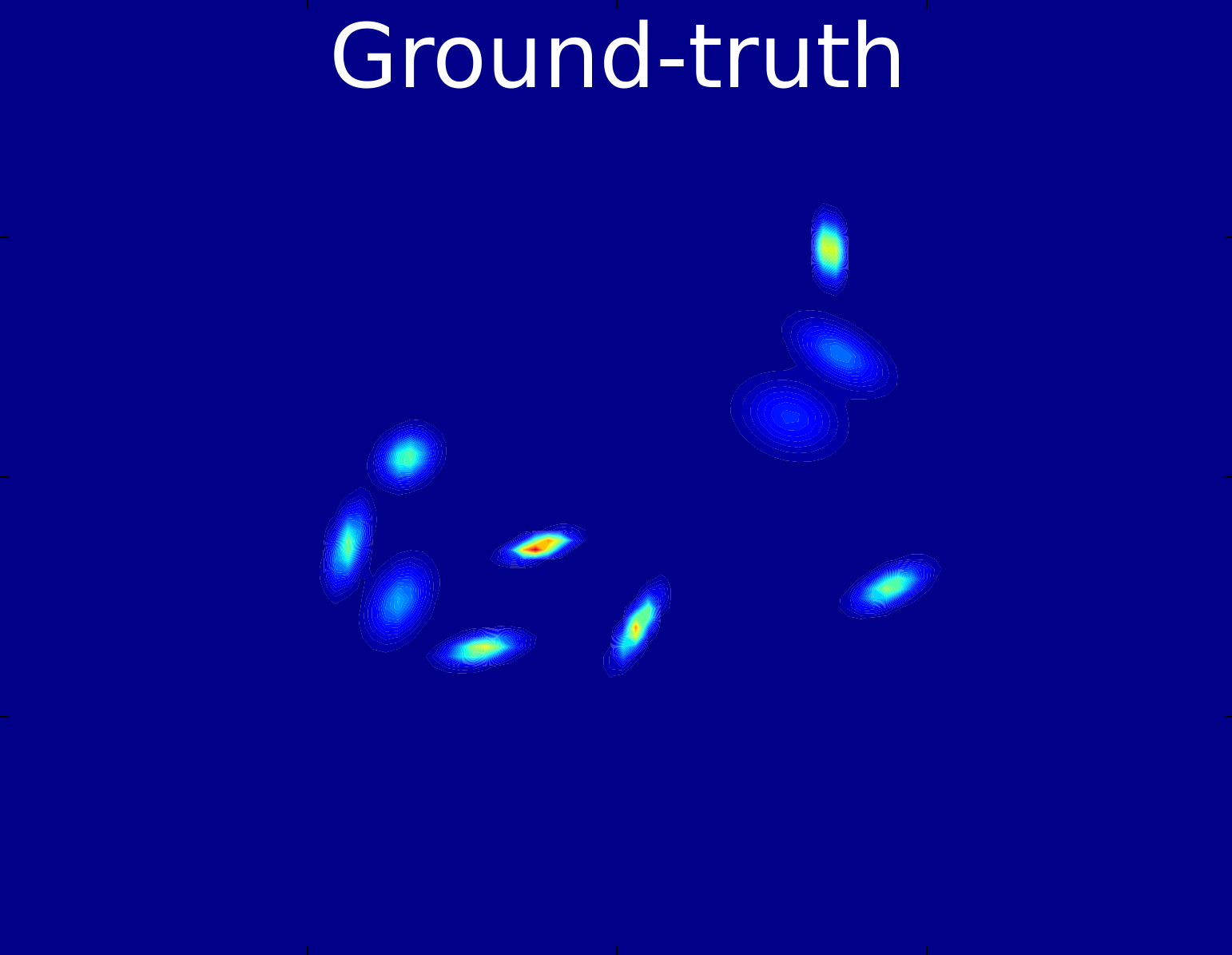}
        \end{subfigure}
        \caption{The first 12 plots show the learned approximation for the illustrative experiment every 30 iterations, directly after adding a new component. The means of the Gaussian mixture model are indicated with a white plus except for the newest component which is marked by a star. Black dots indicate all samples that have been drawn except for those that have already been shown at previous plots. The last two plots compare the learned approximation and the target distribution.}
        \label{fig:gmm_preliminary}
	\end{figure}
We start with a qualitative two-dimensional experiment to illustrate the sample reusage and the adaptation of the number of components. The target distribution is given by a Gaussian mixture model with ten components similar to the higher-dimensional GMM experiments. We use the same hyper-parameters as in the remaining experiments and start with a single component. Figure~\ref{fig:gmm_preliminary} shows the target distribution as well as the learned approximation directly after adding each new component. The new components are often added close to missing modes and components are typically not sampled after they have converged. 
The learned model closely approximates the target distribution. 

\subsection{Considered Competitors}
\label{sec:consideredMethods}
We compare {\sc{VIPS++}} to the closely related methods variational boosting \citep[{\sc{VBOOST}},][]{Miller2017} and non-parametric variational inference ~\citep[{\sc{NPVI}},][]{gershman2012} as well as state-of-the-art methods in variational inference and {\sc{MCMC}}, namely inverse autoregressive flows \citep[{\sc{IAF}},][]{Kingma2016}, Stein variational gradient descent~\citep[{\sc{SVGD}},][]{Liu2016}, Hamiltonian Monte Carlo~\citep[{\sc{HMC}},][]{duane1987}, elliptical slice sampling~\citep[{\sc{ESS}},][]{Murray2010}, parallel tempering {\sc{MCMC}}~\citep[{\sc{PTMCMC}},][]{earl2005} and slice sampling~\citep{Neal2003}. We also compare to naive gradient based optimization of a Gaussian mixture model (with fixed but tuned number of components). As GMMs are not exactly reparameterizable, we compute their stochastic gradients using black-box variational inference ({\sc{BBVI}}).
Please refer to Appendix~\ref{app:implementations} for details on the specific implementations. Due to the high computational demands, we do not compare to every method on each experiment but rather select promising candidates based on the sampling problem or on the preliminary experiments that we had to conduct for hyper-parameter tuning. 
We present our justification for each omitted experiment in Appendix \ref{app:consideredExperiments}, where we also present a table that shows the competitors we compared against on each test problem.

Instead of using a variant of {\sc{MORE}} (which we denote as~{\sc{VIPS1}}), it would also be possible to update the individual components using the reparameterization trick~\citep{Kingma2014, Rezende2014}---which assumes that the target distribution is differentiable---or black-box variational inference~\citep{Ranganath2014}. We evaluated these options by comparing {\sc{VIPS1}}, black-box variational inference and the reparameterization trick for learning Gaussian variational approximations on the \textit{breast cancer} experiment and the \textit{planar robot} experiment. The results are presented in Appendix~\ref{app:GVAs} and show that {\sc{VIPS1}} is not only more efficient than black-box variational inference, but also one to two orders of magnitude more efficient than the reparameterization trick.

\subsection{Hyper-Parameters}
\label{sec:hyperParameters}
For the competing methods, we tuned the hyper-parameters independently for each test problem. We typically tuned the hyper-parameters based on our test metric, the maximum mean discrepancy (MMD). However, in all our experiments black-box variational inference and inverse autoregressive flows collapsed to single modes on multimodal test problems which increased the MMD. In these cases, we tuned the hyper-parameters with respect to the ELBO, rather than setting the learning rate to zero which would perform better on our test metric. 
For {\sc{VIPS++}}, we use the same set of hyper-parameters on all experiments. However, for the planar robot experiment which can profit from large GMMs with several hundred components, we add a new component at every iteration. Learning such large mixture models for simpler, unimodal problems would be wasteful, and we thus use a slower adding rate $n_\text{add}=30$ for the remaining experiments. The remaining hyper-parameters are shown in Appendix~\ref{app:hyperparameters}. 

\subsection{Results}
\label{sec:exp:results}
We compare the different methods in terms of efficiency, regarding both, the number of function evaluations and wall clock time, and in terms of sample quality which we assess by computing the maximum mean discrepancy~\citep[MMD,][]{Gretton2012} with respect to ground-truth samples. The MMD is a nonparametric divergence between mean embeddings in a reproducible kernel Hilbert space. Please refer to Appendix~\ref{app:MMD} on how the MMD and the ground-truth samples are computed. 
\begin{figure}
    \centering
	\begin{subfigure}{.328\textwidth}
		\centering
		\includegraphics[width=\linewidth]{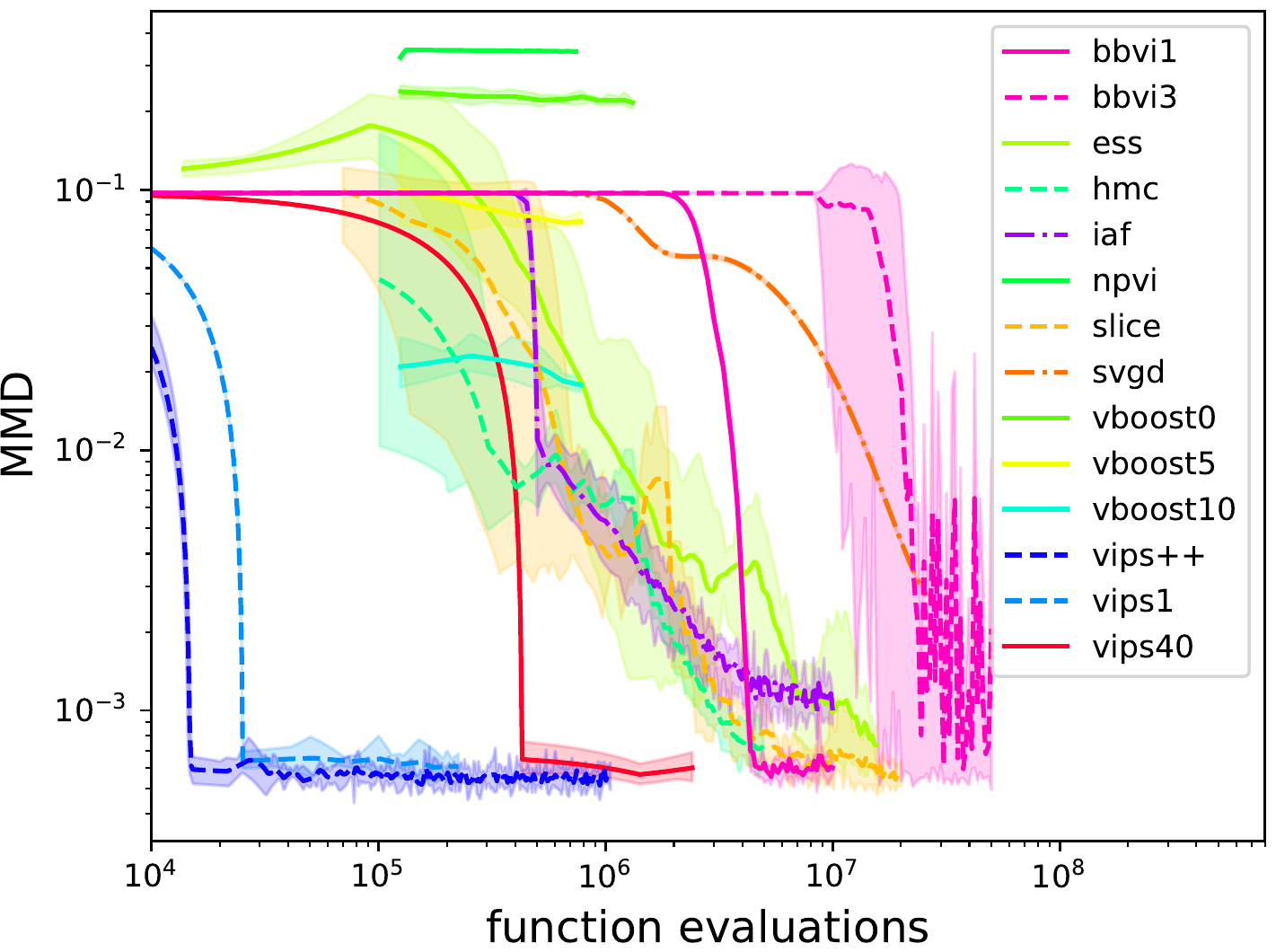}
        \caption*{German credit (Log. Reg.)}
        \end{subfigure}
        	\begin{subfigure}{.328\textwidth}
		\centering
		\includegraphics[width=\linewidth]{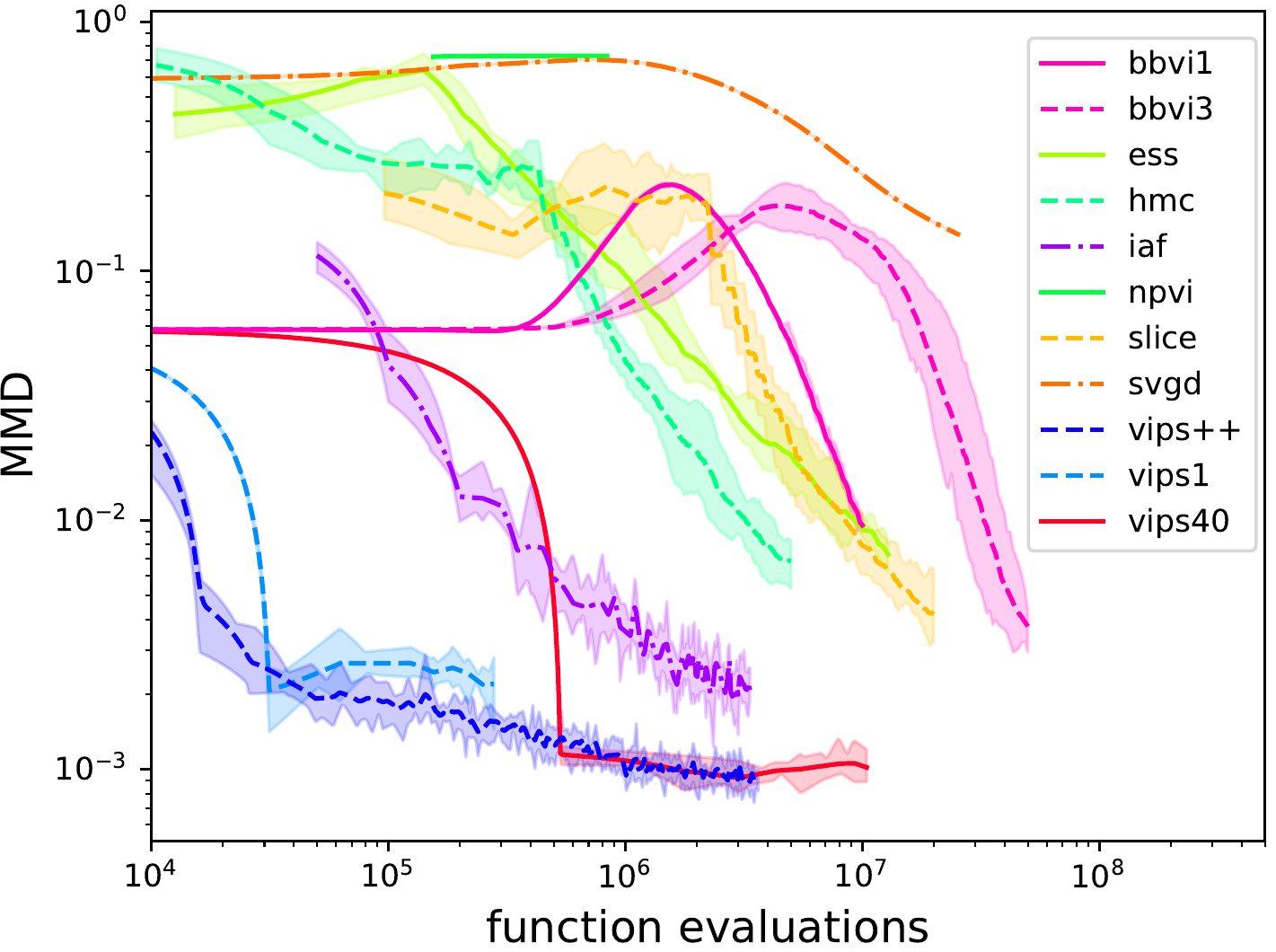}
                \caption*{breast cancer (Log. Reg.)}
        \end{subfigure}
        	\begin{subfigure}{.328\textwidth}
		\centering
		\includegraphics[width=\linewidth]{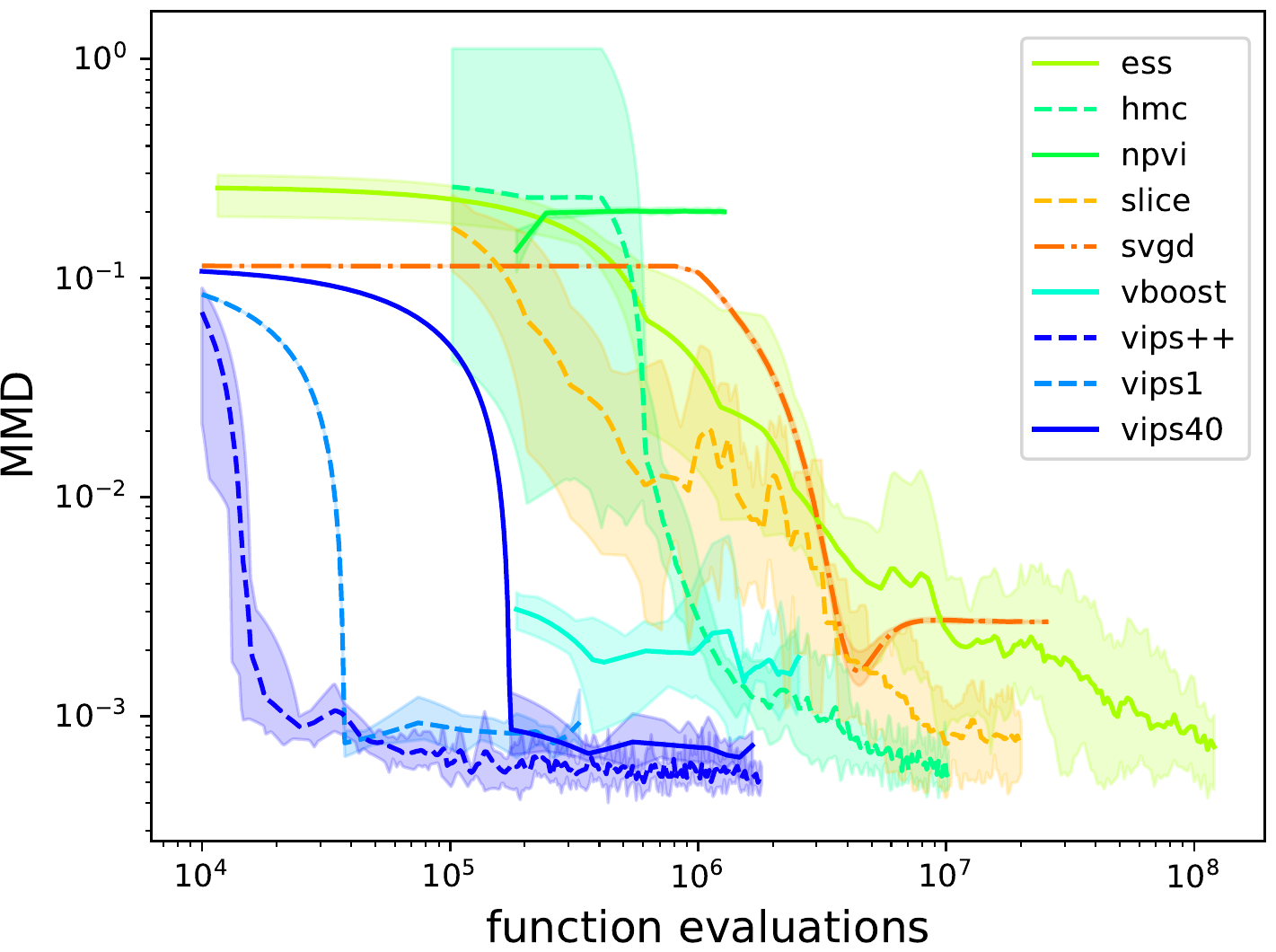}
            \caption*{stop-and-frisk (Poisson GLM)}
        \end{subfigure}
        	\begin{subfigure}{.328\textwidth}
		\centering
		\includegraphics[width=\linewidth]{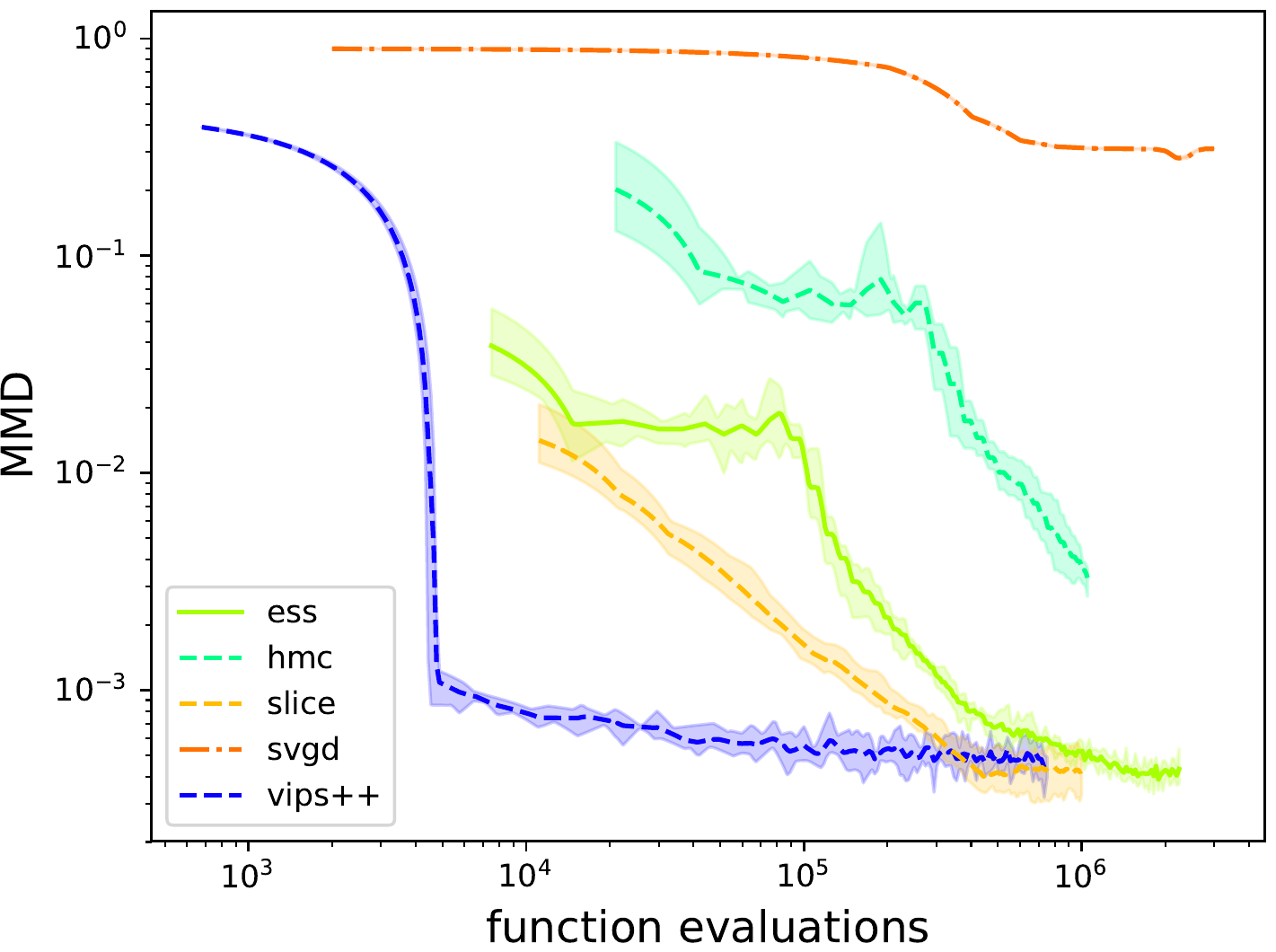}
                \caption*{ionosphere (GP Regression)}
        \end{subfigure}        
        \begin{subfigure}{.328\textwidth}
		\centering
		\includegraphics[width=\linewidth]{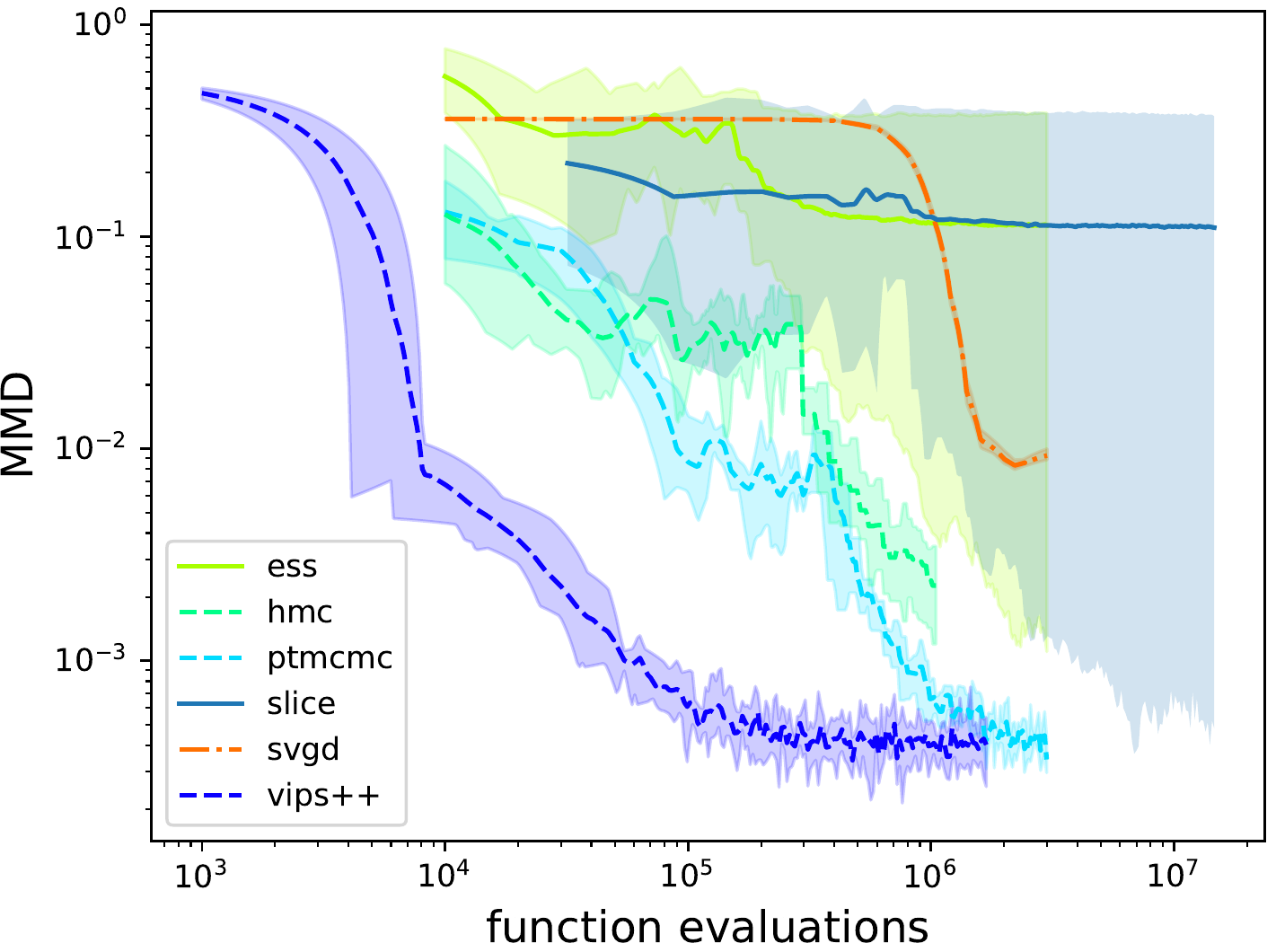}
        \caption*{Goodwin model (ODE)}
        \end{subfigure}
        \begin{subfigure}{.328\textwidth}
		\centering
		\includegraphics[width=\linewidth]{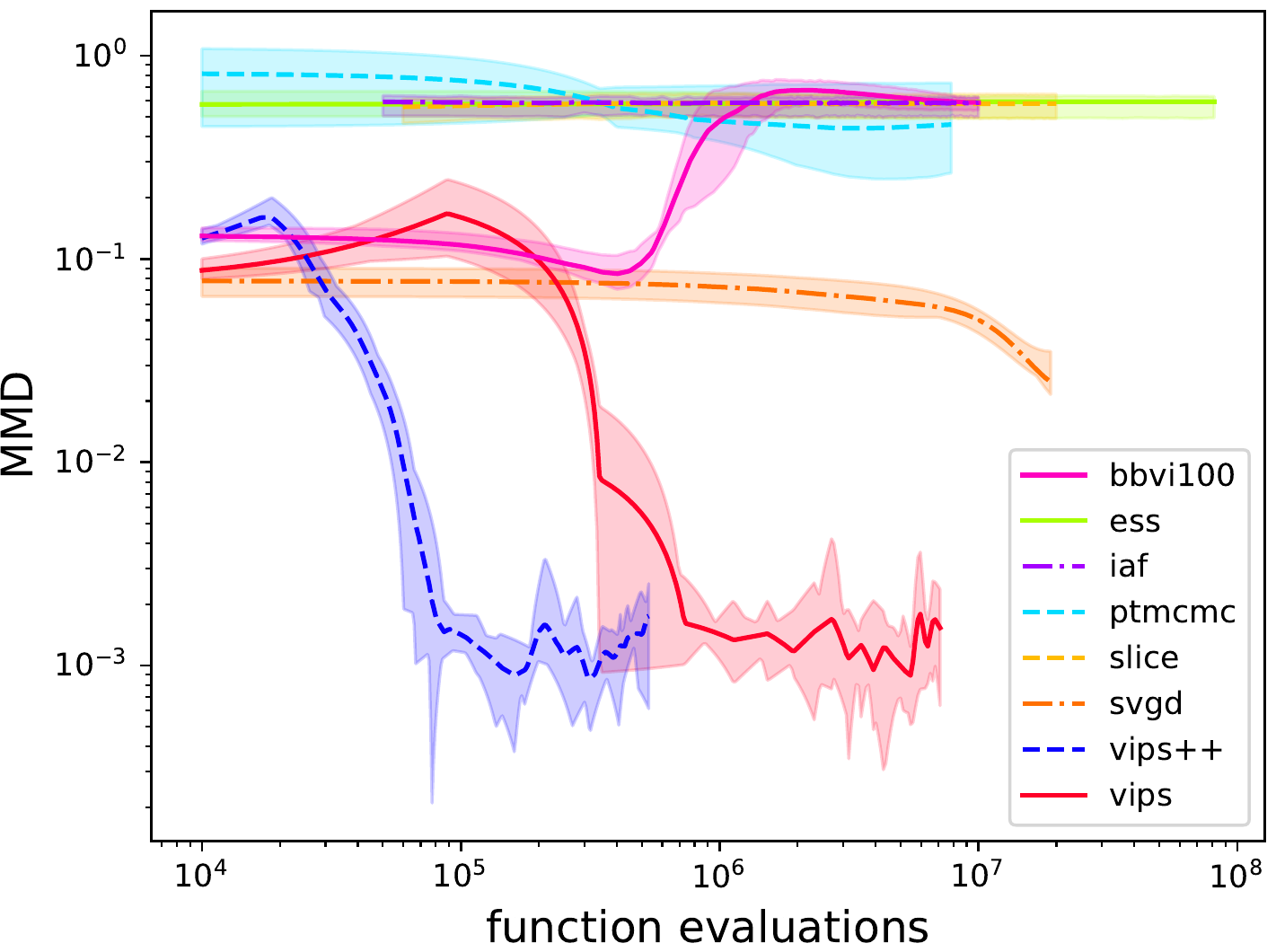}
         \caption*{GMM (20 dimensions)}
       \end{subfigure}        
        	\begin{subfigure}{.328\textwidth}%
		\centering
		\includegraphics[width=\linewidth]{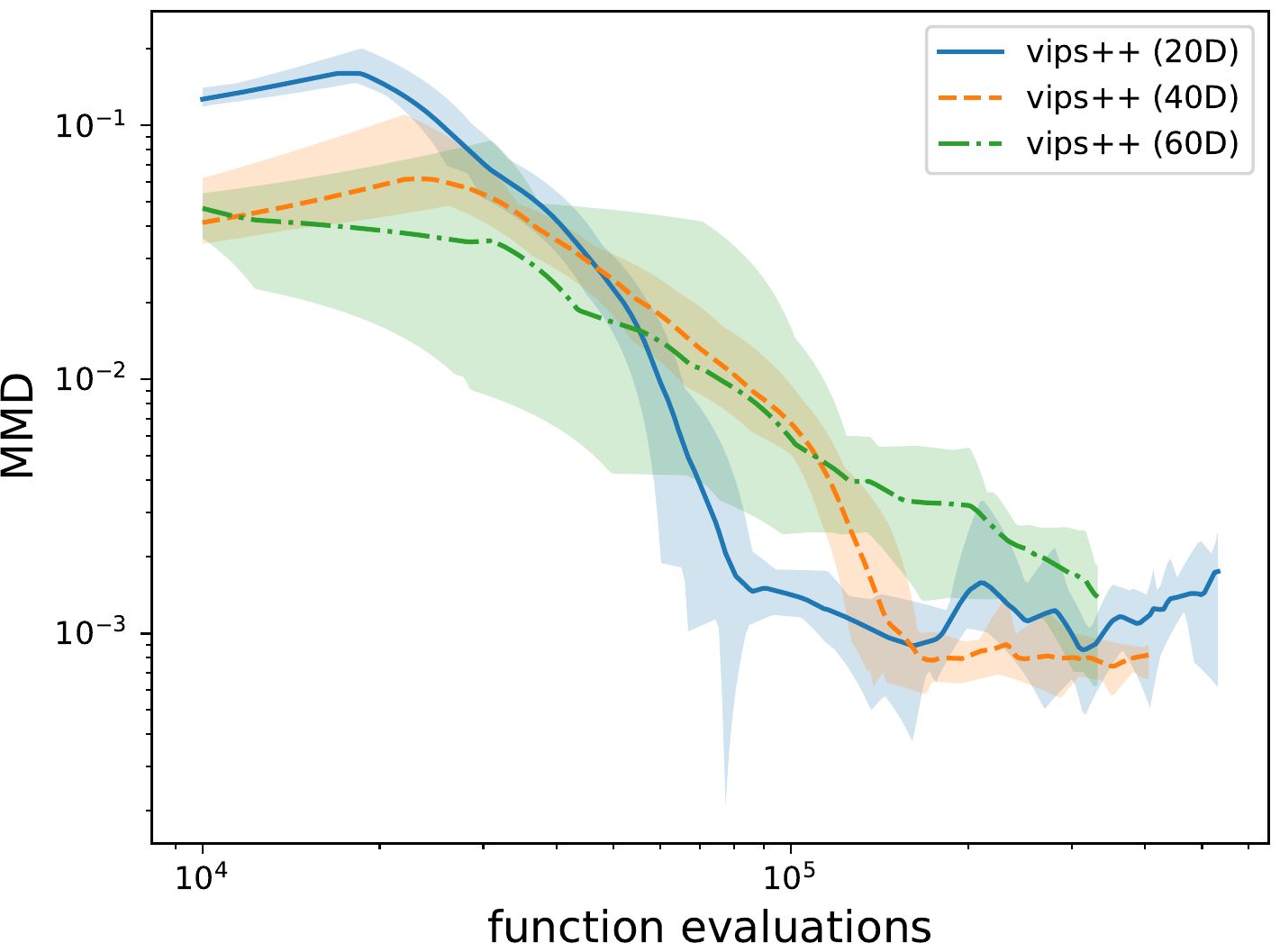}
               \caption*{GMM (different dimensions)}
        \end{subfigure}
        	\begin{subfigure}{.328\textwidth}
		\centering
		\includegraphics[width=\linewidth]{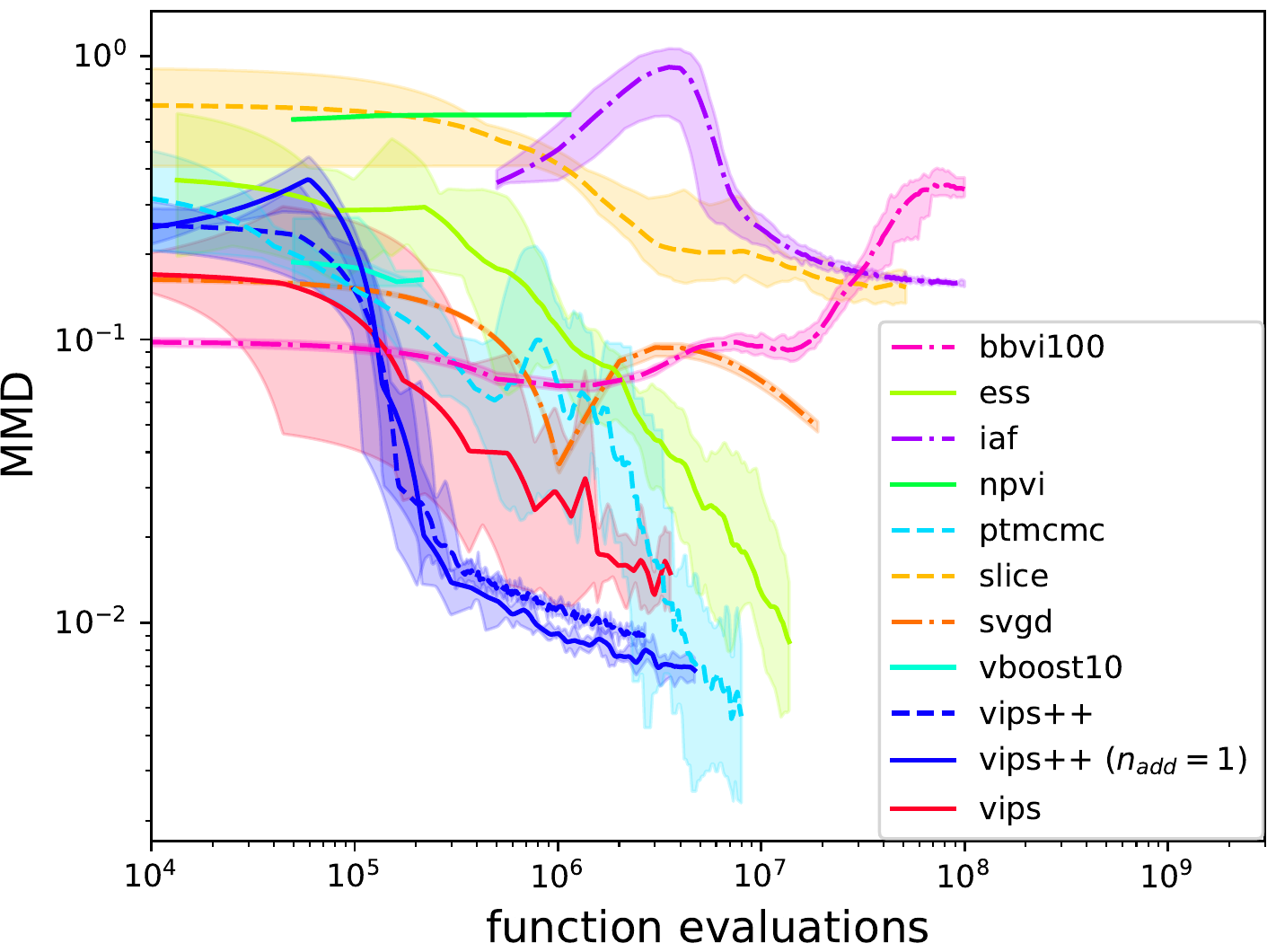}
         \caption*{planar robot (1 goal)}
        \end{subfigure}
        	\begin{subfigure}{.328\textwidth}
		\centering
		\includegraphics[width=\linewidth]{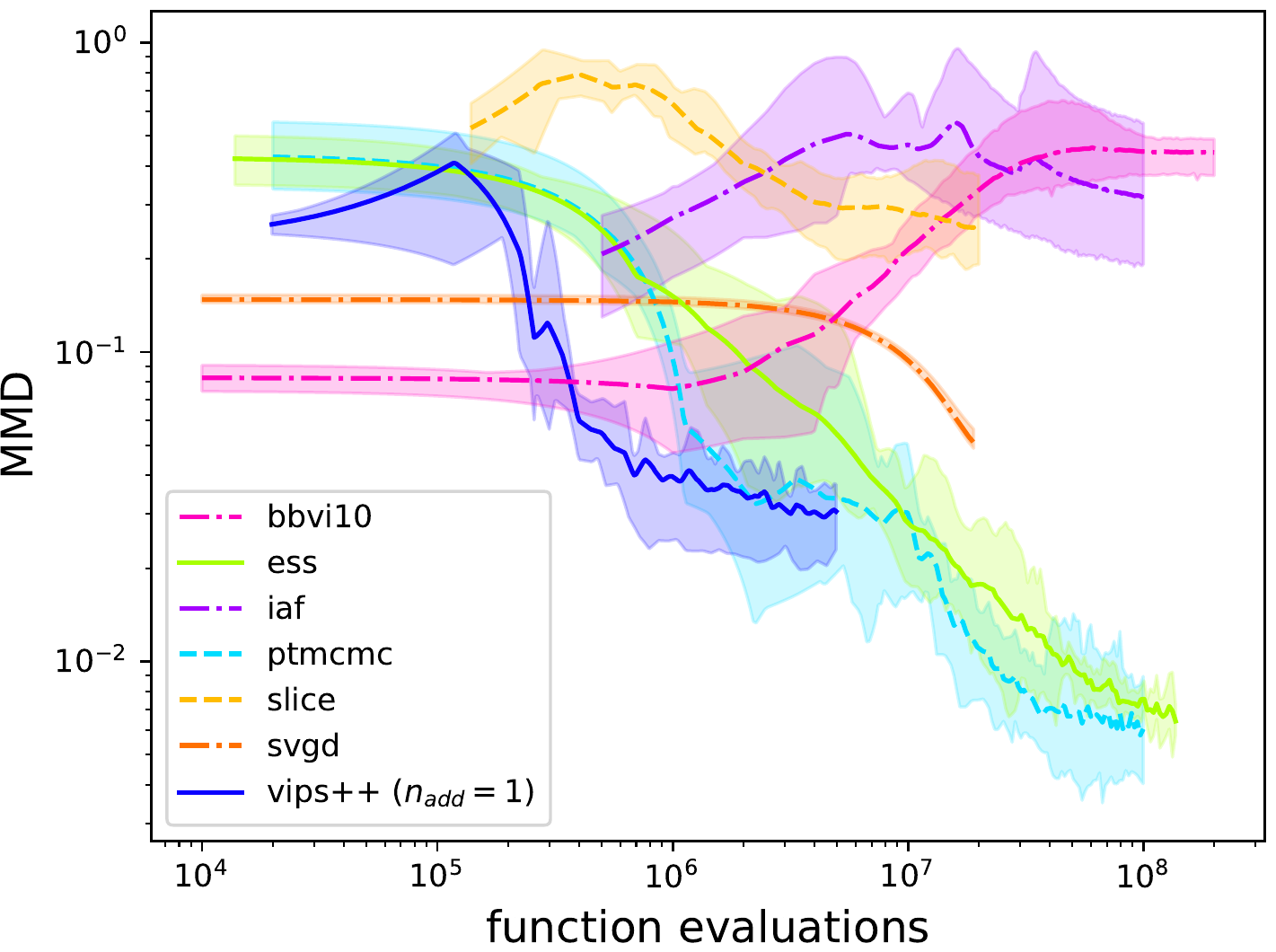}
           \caption*{planar robot (4 goals)}
        \end{subfigure}

        \caption{The maximum mean discrepancy with respect to ground-truth samples is plotted over the number of function evaluations on log-log plots for the different sampling problems in the test bed. {\sc{VIPS++}} achieves in most cases a sample quality that is on par with the best MCMC sampler while requiring up to three orders of magnitude fewer function evaluations.}
        \label{fig:quantitativeResults}
	\end{figure}
	
\begin{figure}
		\centering
		\begin{subfigure}{.328\textwidth}
			\centering
			\includegraphics[width=\linewidth]{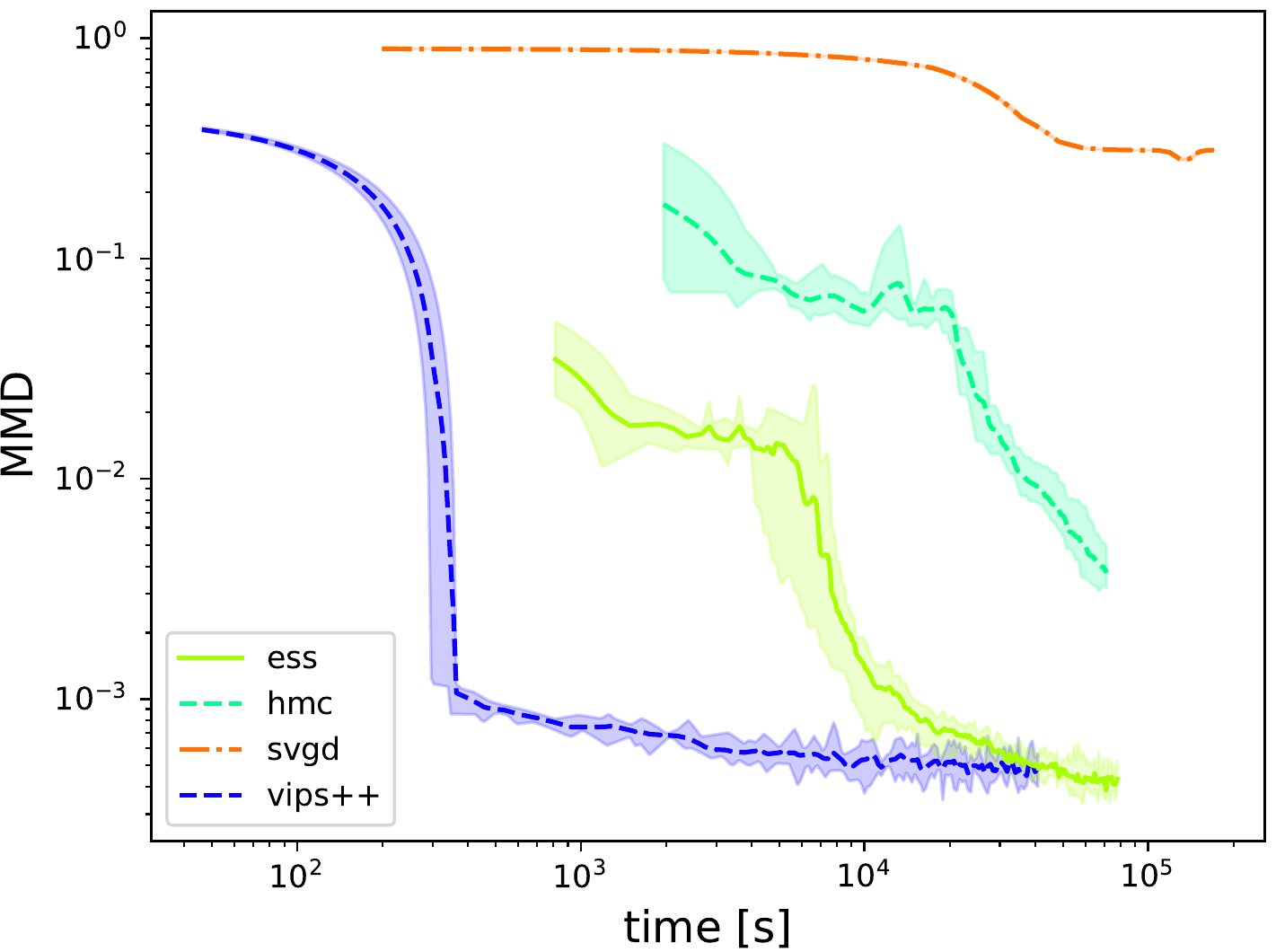}
			\caption*{ionosphere (GP Regression)}
		\end{subfigure}        
		\begin{subfigure}{.328\textwidth}
			\centering
			\includegraphics[width=\linewidth]{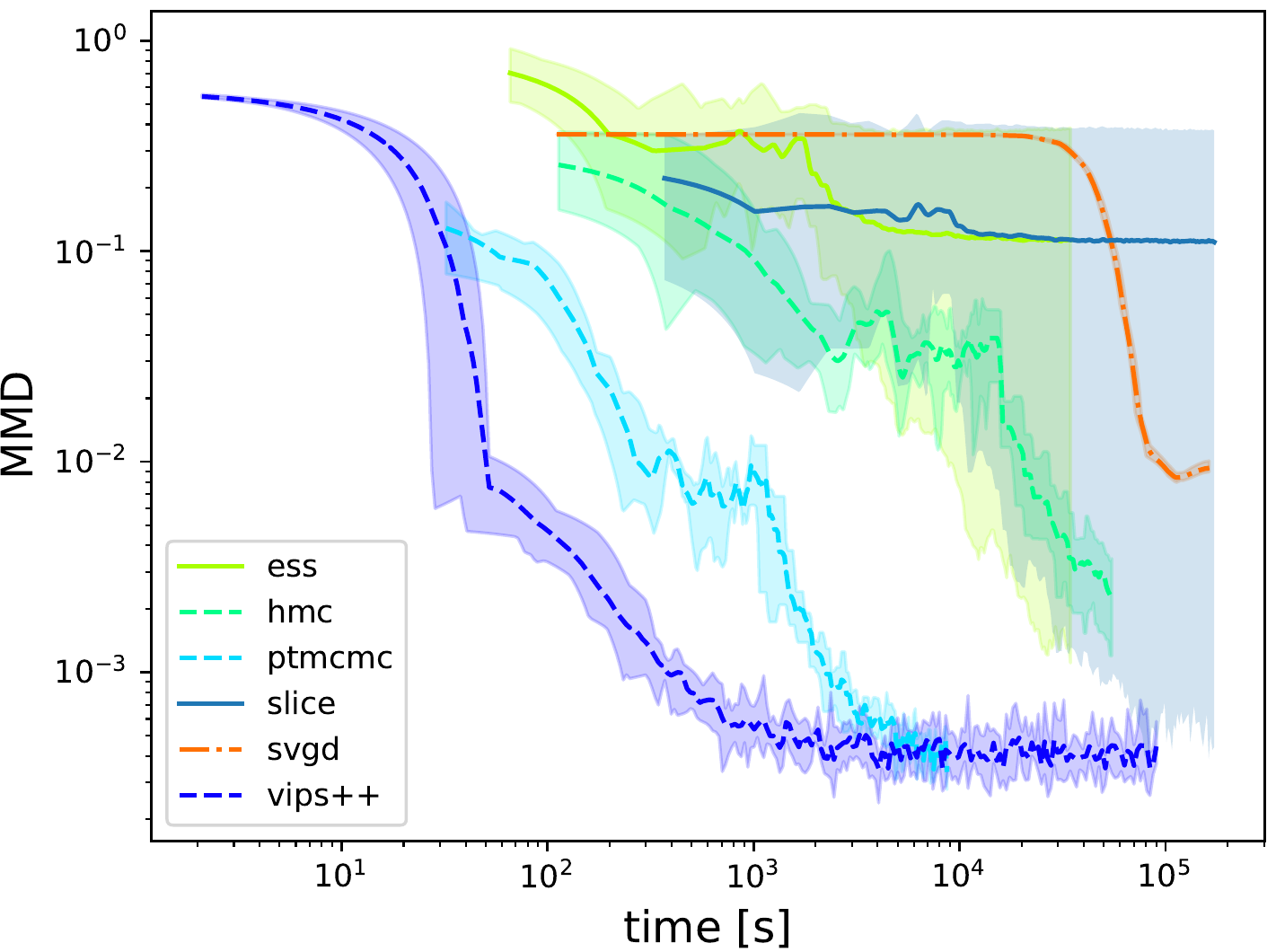}
			\caption*{Goodwin model (ODE)}
		\end{subfigure}
		\begin{subfigure}{.328\textwidth}
			\centering
			\includegraphics[width=\linewidth]{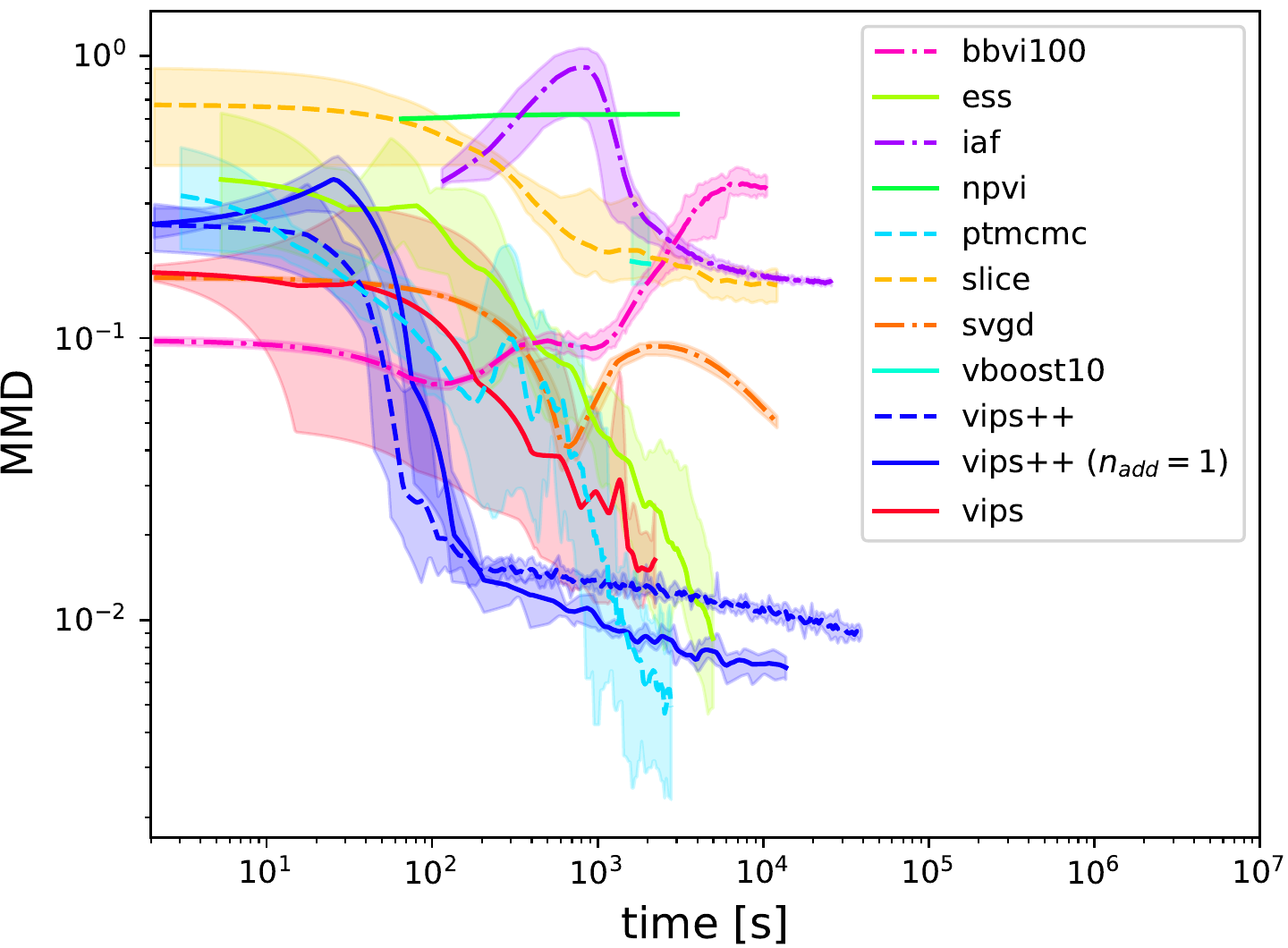}
			\caption*{planar robot (1 goal)}
		\end{subfigure}		
		\caption{Evaluating the methods with respect to computational time yields comparable results as evaluating with respect to the number of function evaluations. These results show that {\sc{VIPS++}} can also be competitive to MCMC in terms of computational time.}
		\label{fig:quantitativeResults_time}
	\end{figure}

Figure~\ref{fig:quantitativeResults} shows plots of the MMD over the number of function evaluations for the different sampling problems in the test bed. We perform five runs for each method and linearly interpolate the MMD values to produce continuous curves. The plots show the mean of these curves, as well as the smallest and largest value as shaded area. The tested methods are apparent from the legends. {\sc{VBOOST}} can make use of low-rank approximations for learning the covariance matrices and we indicate the chosen ranks in the legends. The \textit{German credit}, \textit{breast cancer}, \textit{stop-and-frisk} and the \num{20}-dimensional \textit{GMM} experiment, as well as the \textit{planar robot} experiment with a single goal position were also used in our previous work~\citep{Arenz2018} and we use some of the previous results. For example, we directly compare {\sc{VIPS++}} with the previously published results of {\sc{VIPS}}. Unlike {\sc{VIPS++}}, {\sc{VIPS}} bounds the maximum number of components by stopping to add new components if the current number of components matches a given threshold. This threshold is indicated in the respective legends. 
Figure~\ref{fig:quantitativeResults_time} presents the results with respect to computational time for the \textit{ionosphere} and \textit{Goodwin model} experiment as well as the \textit{planar robot} experiment with a single goal position. As the results are similar compared to the evaluations with respect to the number of function evaluations, we show the remaining plots in Appendix~\ref{app:computationalTime}.

Furthermore, for our comparisons with the variational inference methods {\sc{BBVI}} and {\sc{IAF}} we also present learning curves regarding the ELBO in Appendix~\ref{app:elbos}.

\subsubsection{Discussion}
\label{sec:exp:discussion}



        \begin{figure}
    \centering
	\begin{subfigure}{.494\textwidth}
		\centering
		\includegraphics[width=.99\linewidth]{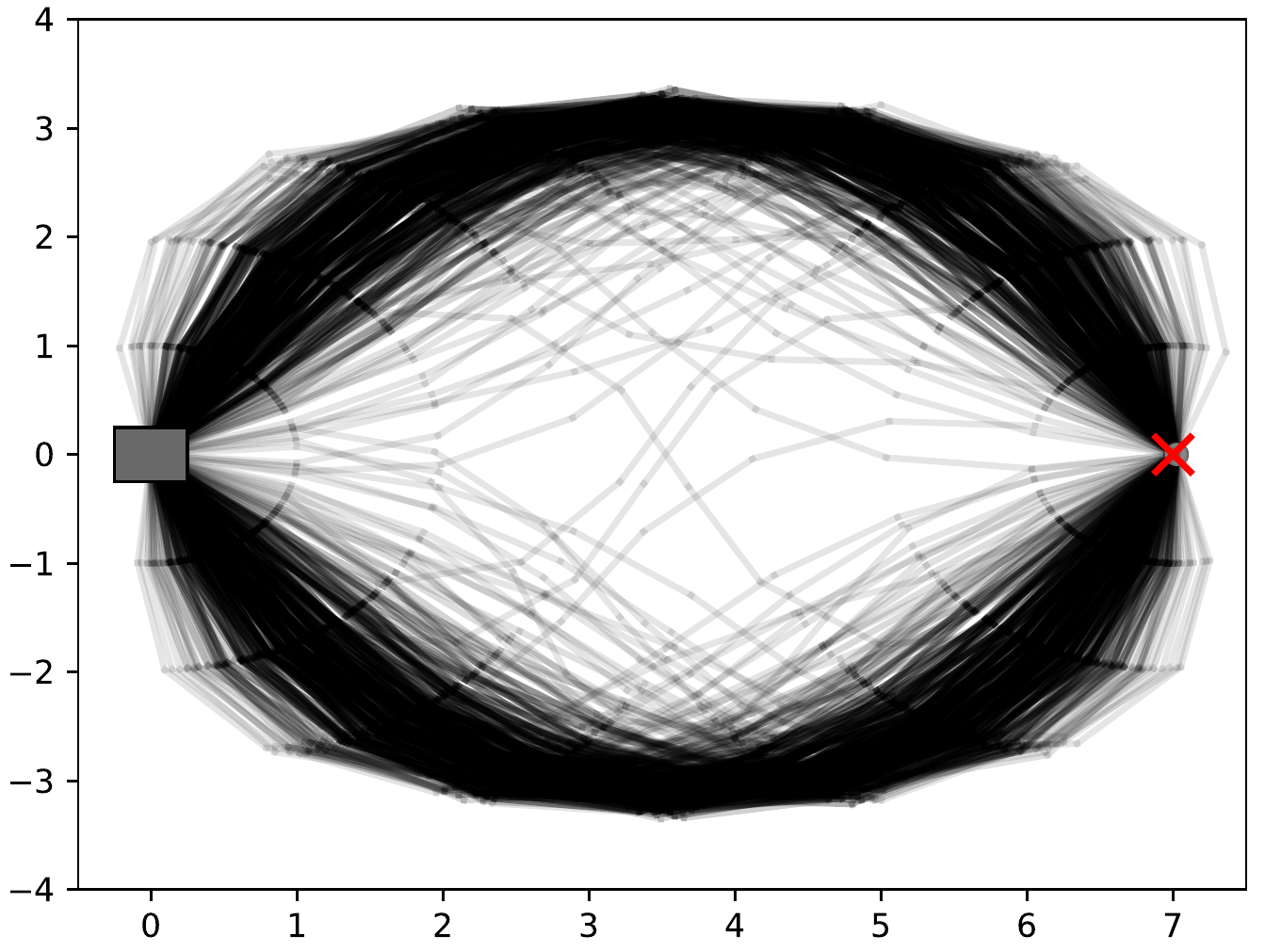}
	\end{subfigure}
	\begin{subfigure}{.495\textwidth}
		\centering
		\includegraphics[width=.99\linewidth]{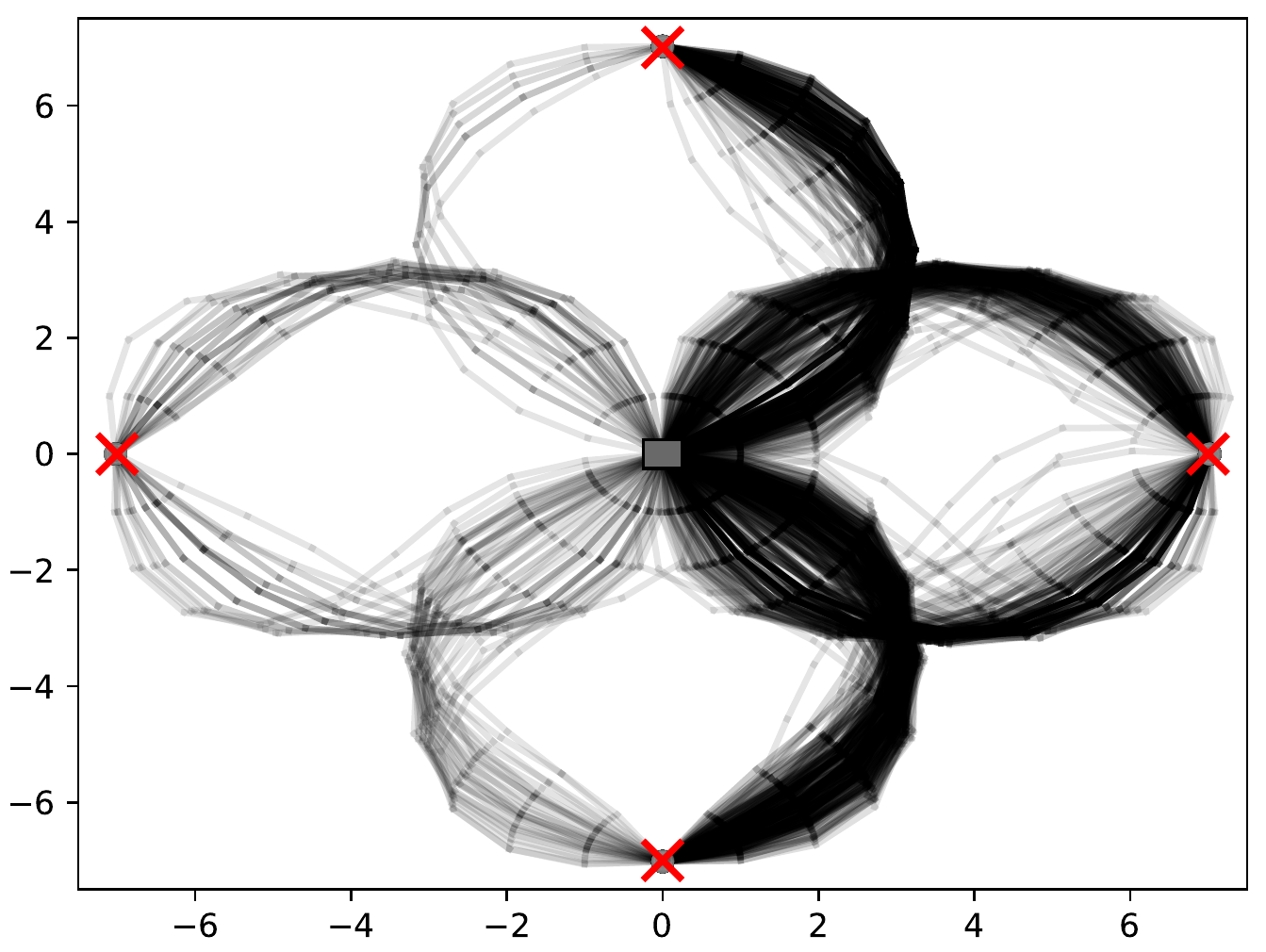}
	\end{subfigure}%
		\caption{The plots visualize the weights and means of the mixture models learned by VIPS++ for each of the planar robot experiments when adding new components with adding rate $n_\text{add}=1$. The gray box indicates the base of the robot; the red crosses indicate the goal positions. Components with larger weight are drawn darker. The visualized mixture models comprise of \num{333} and \num{360} components for the experiments with one goal position (left) and four goal positions (right), respectively. }
		\label{fig:planar_learned}
	\end{figure}
The sample quality achieved by {\sc{VIPS}} is unmatched by any variational inference method on all considered experiments and in most cases on par with the best MCMC sampler. {\sc{VIPS}} requires significantly fewer function evaluations and computational resources for producing such high quality samples. {\sc{VIPS++}} is approximately one order of magnitude more efficient than {\sc{VIPS}} and two to three orders of magnitude more efficient than the remaining methods.  {\sc{VIPS}} and  {\sc{VIPS++}} were also the only methods that could produce good results on the \num{20}-dimensional \textit{GMM} experiment, where they were able to reliably discover and approximate all ten modes of the target distribution. We therefore only evaluated {\sc{VIPS++}} on the higher-dimensional \textit{GMM} experiments where it also approximated the target distribution with high accuracy.
However, on the \textit{planar robot} experiment with four goal positions {\sc{ESS}} and {\sc{PTMCMC}} could produce significantly better samples than {\sc{VIPS++}}. We believe that learning highly accurate GMM approximations would require a very large number of components for this experiment. Already on the \textit{planar robot} experiment with a single goal position, we could slightly improve the learned approximations by adding new components more frequently. Compared to the default adding rate, which learned GMMs with approximately \num{150} components, the faster adding rate resulted in GMMs with approximately \num{350} components. We believe that {\sc{VIPS++}} would require significantly more components to achieve comparable sample quality to the MCMC samplers on the more challenging \textit{planar robot} experiment. However, learning very large mixture models can become infeasible, because computing the (log-)responsibilities $\log \responsibilities$ exactly can become prohibitive. Figure~\ref{fig:planar_learned} visualizes the weights and means of the learned approximation of the first run for both \textit{planar robot} experiments when adding new components at every iteration. We can see that the learned components are still of very good quality. Samples from the learned models are shown in Appendix~\ref{app:planarSamples} and compared to those obtained by BBVI, IAF, PTMCMC.

\section{Conclusion and Future Work}
We proposed {\sc{VIPS++}}, a method for learning GMM approximations of intractable probability distributions that exploits the connection between variational inference and policy search. We introduced a variant of {\sc{MORE}}~\citep{Abdolmaleki2015} that can be efficiently used for learning Gaussian variational approximations. We further derived a lower bound on the I-projection to latent variable models that can be used for learning a local optimum of the true objective, similar to expectation-maximization. By applying this decomposition to Gaussian mixture models, the I-projection can be performed independently for each component, allowing us to improve the GMM approximation by independently updating the components using our variant of {\sc{MORE}}. 
We argue that a good trade-off between exploration and ex\-ploita\-tion is essential for efficiently learning accurate multimodal approximations. 
We tackle the exploration-ex\-ploita\-tion dilemma locally for each component by updating them using information-geometric trust regions. For global exploration, we dynamically add new components at interesting regions.

For target distributions that can be well approximated with a small number of components, {\sc{VIPS}} does not only outperform existing methods for variational inference, but is also several orders of magnitude more efficient than Markov chain Monte Carlo at drawing samples. We also showed that {\sc{VIPS}} can learn large mixture models comprising several hundred components. However, learning very large GMMs is computationally expensive and {\sc{MCMC}} methods can be more efficient at drawing samples.

Learning Gaussian components with full covariance matrices can become intractable for high dimensional problems, and we thus applied {\sc{VIPS}} only for medium-scaled problems with up to 60 dimensions. For significantly higher-dimensional problems, learning low-rank approximations and using gradient information for the component updates are interesting routes of future work. 
It is also interesting to further investigate the strong ties between variational inference and policy search. Using our decomposition we can learn GMMs of policy parameters for the black-box reinforcement learning setting where time-series data is not assumed and exploited. In order to apply {\sc{VIPS}} for multimodal reinforcement learning with time-series data, we aim to contextualize the GMM parameterization on the state of an MDP to directly learn GMM policies.
Furthermore, it is interesting to investigate how our decomposition can be applied to different problems such as clustering or density estimation, or to other latent variable models.

\newpage 
\section*{Acknowledgements}
This project has received funding from the European Union’s Horizon 2020 research and innovation programme under grant agreement No 645582 (RoMaNS). Calculations for this research were conducted on the Lichtenberg high performance computer of the TU Darmstadt.

\renewcommand{\theHsection}{A\arabic{section}}
\appendix
\section{VIPS1 Derivations}
\label{app:Vips1Derivations}
For each update we wish to solve the optimization problem
\begin{equation*}
   \begin{alignedat}{2}
& \;\; \underset{ q(\mathbf{x}) }{\max}& \;\; & \int_{\mathbf{x}} q(\mathbf{x})  \rewardSurrogate d\mathbf{x} + \entropy(q(\mathbf{x})),\\
&\text{subject to} &      &  \textrm{KL}\Big(q(\mathbf{x})||q^{(i)}(\mathbf{x})\Big) \le \epsilon, \\
& & & \int_\mathbf{x} q(\mathbf{x}) d\mathbf{x} = 1,
   \end{alignedat}
\end{equation*}
where we recall that the the reward surrogate \rewardSurrogate is a quadratic function and the variational approximation of the previous iteration, $q^{(i)}(\mathbf{x})$, is Gaussian. We formulate the optimization for general distributions $q(\mathbf{x})$, but we will see that the optimal solution is also Gaussian.
Using the definition of the Shannon entropy and Kullback-Leibler divergence and introducing the Lagrangian multipliers $\eta$ and $\lambda$, the Lagrangian function is given by
\begin{align*}
    \mathcal{L}(q,\eta, \lambda) =& \int_\mathbf{x} q(\mathbf{x}) \big(\rewardSurrogate - \log q(\mathbf{x}) \big) d\mathbf{x} + \eta \left( \epsilon - \int_\mathbf{x} q(\mathbf{x}) \big( \log q(\mathbf{x}) - \log q^{(i)}(\mathbf{x}) \big) d\mathbf{x} \right) \\
    &+ \lambda \big( 1 - \int_\mathbf{x} q(\mathbf{x}) d\mathbf{x} \big) \\
    =& \int_\mathbf{x} q(\mathbf{x}) \big( \rewardSurrogate - (1+\eta) \log q(\mathbf{x}) + \eta \log q^{(i)}(\mathbf{x}) - \lambda \big) d\mathbf{x} + \eta \epsilon + \lambda.
\end{align*}
The optimum $q^\star(\mathbf{x})$ occurs where the partial derivative $\frac{\partial \mathcal{L}(q,\eta,\lambda)}{\partial q(\mathbf{x})}$ is equal to zero, that is, 
\begin{align}
\label{eq:optimalQ}
\frac{\partial \mathcal{L}(q^\star,\eta,\lambda)}{\partial q(\mathbf{x})} = \rewardSurrogate - (1+\eta) \log q^\star(\mathbf{x};\eta,\lambda) - (1+\eta) + \eta \log q^{(i)}(\mathbf{x}) - \lambda  \overset{!}{=} 0 \nonumber \\
\Rightarrow q^\star(\mathbf{x};\eta,\lambda) = \exp\Big(- \frac{\lambda+1+\eta}{1+\eta} \Big) \exp \Big( \frac{\rewardSurrogate + \eta \log q^{(i)}(\mathbf{x})}{1+\eta} \Big).
\end{align}
The Lagrange dual function is, thus, given by 
\begin{align*}
    \mathcal{G}(\eta,\lambda) =& \mathcal{L}(q^\star,\eta,\lambda) \\
    =& \int_\mathbf{x} q^\star(\mathbf{x};\eta,\lambda) \big( \rewardSurrogate - \big( - \lambda - 1 - \eta + \rewardSurrogate + \eta \log q^{(i)} \big) + \eta \log q^{(i)}(\mathbf{x}) - \lambda \big) d\mathbf{x} \\
    &+ \eta \epsilon + \lambda \\
    =& (1+\eta) \int_\mathbf{x} q^\star(\mathbf{x};\eta,\lambda) d\mathbf{x} + \eta \epsilon + \lambda.
\end{align*}
As strong duality holds due to Slater's condition~\citep{Boyd2004}, we can find the optimal distribution $q^\star(\mathbf{x};\eta,\lambda)$ by minimizing the dual function with respect to $\eta$ and $\lambda$ and then using the optimal step size $\eta^\star$ and Lagrangian multiplier $\lambda^\star$ to compute $q^\star(\mathbf{x};\eta^\star,\lambda^\star)$ according to Equation~\ref{eq:optimalQ}.
The partial derivatives are given by
\begin{align*}
    \frac{\partial \mathcal{G}(\eta,\lambda)}{\partial \eta} &= \epsilon + \int_\mathbf{x} q^\star(\mathbf{x};\eta,\lambda) d\mathbf x + (1+\eta) \int_\mathbf{x} q^\star(\mathbf{x};\eta,\lambda) \Big( \frac{\log q^{(i)}(\mathbf{x}) - 1}{1+\eta} - \frac{\log q^\star(\mathbf{x};\eta,\lambda)}{(1+\eta)} \Big) d\mathbf{x} \\
    &= \epsilon - \int_\mathbf{x} q^\star(\mathbf{x};\eta,\lambda) \big( \log q^\star(\mathbf{x};\eta,\lambda) - \log q^{(i)}\big) d\mathbf{x}
\end{align*}
and 
\begin{align*}
    \frac{\partial \mathcal{G}(\eta,\lambda)}{\partial \lambda} &= -\int_\mathbf{x} q^\star(\mathbf{x};\eta,\lambda) d\mathbf{x} + 1,
\end{align*}
where the optimal Lagrangian multiplier $\lambda^\star(\eta)$ for a given $\eta$ normalizes $q^\star(\mathbf{x};\eta,\lambda^\star)$, that is,
\begin{align*}
    \frac{\partial \mathcal{G}(\eta,\lambda^\star)}{\partial \lambda} &= 0 \Leftrightarrow \int_\mathbf{x} q^\star(\mathbf{x};\eta,\lambda^\star) d\mathbf{x} = 1.
\end{align*}
Hence, we can perform coordinate descent by alternately updating $\eta$ along its partial derivative and computing the optimal $\lambda$. Such procedure corresponds to optimizing the dual
\begin{align}
    \label{eq:dualOfEta}
    \mathcal{G}(\eta) &= (1+\eta) \int_\mathbf{x} q^\star(\mathbf{x};\eta,\lambda^\star) d\mathbf{x} + \eta \epsilon + \lambda^\star(\eta) = 1 + \eta + \eta \epsilon + \lambda^\star(\eta)
\end{align}
based on the gradient
\begin{align*}
    \frac{\partial \mathcal{G}(\eta)}{\partial \eta} = \epsilon - \text{KL}(q^\star(\mathbf{x};\eta,\lambda^\star)||q^{(i)}(\mathbf{x})).
\end{align*}

We will now express the approximation of the previous iteration  $q^{(i)}(\mathbf{x})$ in terms of its natural parameters $\mathbf{Q}^{(i)}$ and $\mathbf{q}^{(i)}$ and the reward surrogate as
\begin{equation*}
\rewardSurrogate = -\frac{1}{2} \mathbf{x}^\top \mathbf{R}^{(i)} \mathbf{x} + \mathbf{x}^\top \mathbf{r}^{(i)}.
\end{equation*}
Then, according to Equation~\ref{eq:optimalQ}, the optimal distribution
\begin{equation}
\label{eq:QofEta}
    \begin{aligned}
    q^\star(\mathbf{x}; \eta) =& \exp{\big(  \frac{\eta \log Z(\mathbf{Q}^{(i)}, \mathbf{q}^{(i)}) - \lambda^\star(\eta)-1 -\eta}{1+\eta} \big)} \\
    &\cdot \exp{\big( - \frac{1}{2} \mathbf{x}^\top \frac{\mathbf{R}^{(i)} + \eta \mathbf{Q}^{(i)})}{1+\eta}\mathbf{x} + \mathbf{x}^\top \frac{\mathbf{r}^{(i)} + \eta \mathbf{q}^{(i)})}{1+\eta} \big)}
\end{aligned}
\end{equation}
is Gaussian with natural parameters\\
\noindent\begin{minipage}{.44\linewidth}
\begin{equation*}
\mathbf{Q}(\eta) = \frac{\eta}{\eta + 1} \mathbf{Q}^{(i)} + \frac{1}{\eta + 1} \mathbf{R}^{(i)}, 
\end{equation*}
\end{minipage}
\begin{minipage}{.1\linewidth}
\hspace{.1\linewidth}
\end{minipage}
\begin{minipage}{.44\linewidth}
\begin{equation*}
\mathbf{q}(\eta) = \frac{\eta}{\eta + 1} \mathbf{q}^{(i)} + \frac{1}{\eta + 1} \mathbf{r}^{(i)}. 
\end{equation*}
\end{minipage}\\

\noindent Further, we can see from Equation~\ref{eq:QofEta} and the optimality condition $\int_\mathbf{x} q(\mathbf{x};\eta,\lambda^\star)=1$, that 
    \begin{align}
    \label{eq:optimalLambda}
    \lambda^\star(\eta) =& -(1+\eta) \log \int_\mathbf{x} \exp{\big( - \frac{1}{2} \mathbf{x}^\top \frac{\mathbf{R}^{(i)} + \eta \mathbf{Q}^{(i)})}{1+\eta}\mathbf{x} + \mathbf{x}^\top \frac{\mathbf{r}^{(i)} + \eta \mathbf{q}^{(i)})}{1+\eta} \big)} d\mathbf{x} - 1 - \eta \nonumber \\
    &+ \eta \log Z(\mathbf{Q}^{(i)}, \mathbf{q}^{(i)}) \nonumber \\
    =& \eta \log Z(\mathbf{Q}^{(i)}, \mathbf{q}^{(i)}) - (1+\eta) \log Z(\mathbf{Q}(\eta), \mathbf{q}(\eta)) -1 - \eta.
    \end{align}
Using Equation~\ref{eq:optimalLambda} and Equation~\ref{eq:dualOfEta}, the dual function can be expressed as
\begin{align}
    \mathcal{G}(\eta) = \eta \epsilon + \eta \log Z(\mathbf{Q}^{(i)}, \mathbf{q}^{(i)}) - (1+\eta) \log Z(\mathbf{Q}(\eta), \mathbf{q}(\eta)).
\end{align}

\section{Effects of Different Dissimilarity Measures for Sample Selection}
\label{app:sampleSelection}
\begin{figure}
	\centering
			\begin{subfigure}{.244\textwidth}
		\centering
		\includegraphics[width=\linewidth]{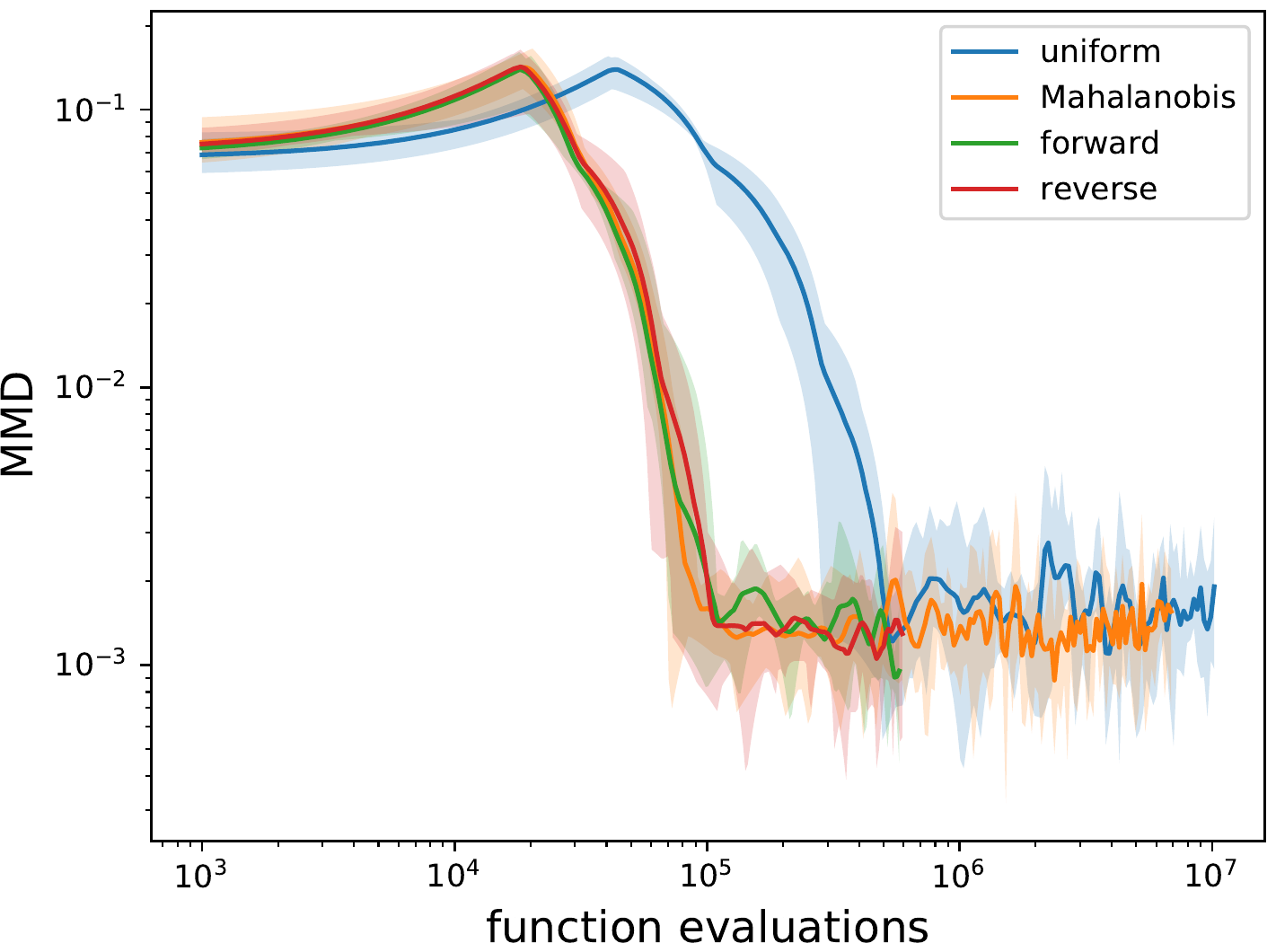}
		\caption{GMM (20D)}
\end{subfigure}
			\begin{subfigure}{.244\textwidth}
		\centering
		\includegraphics[width=\linewidth]{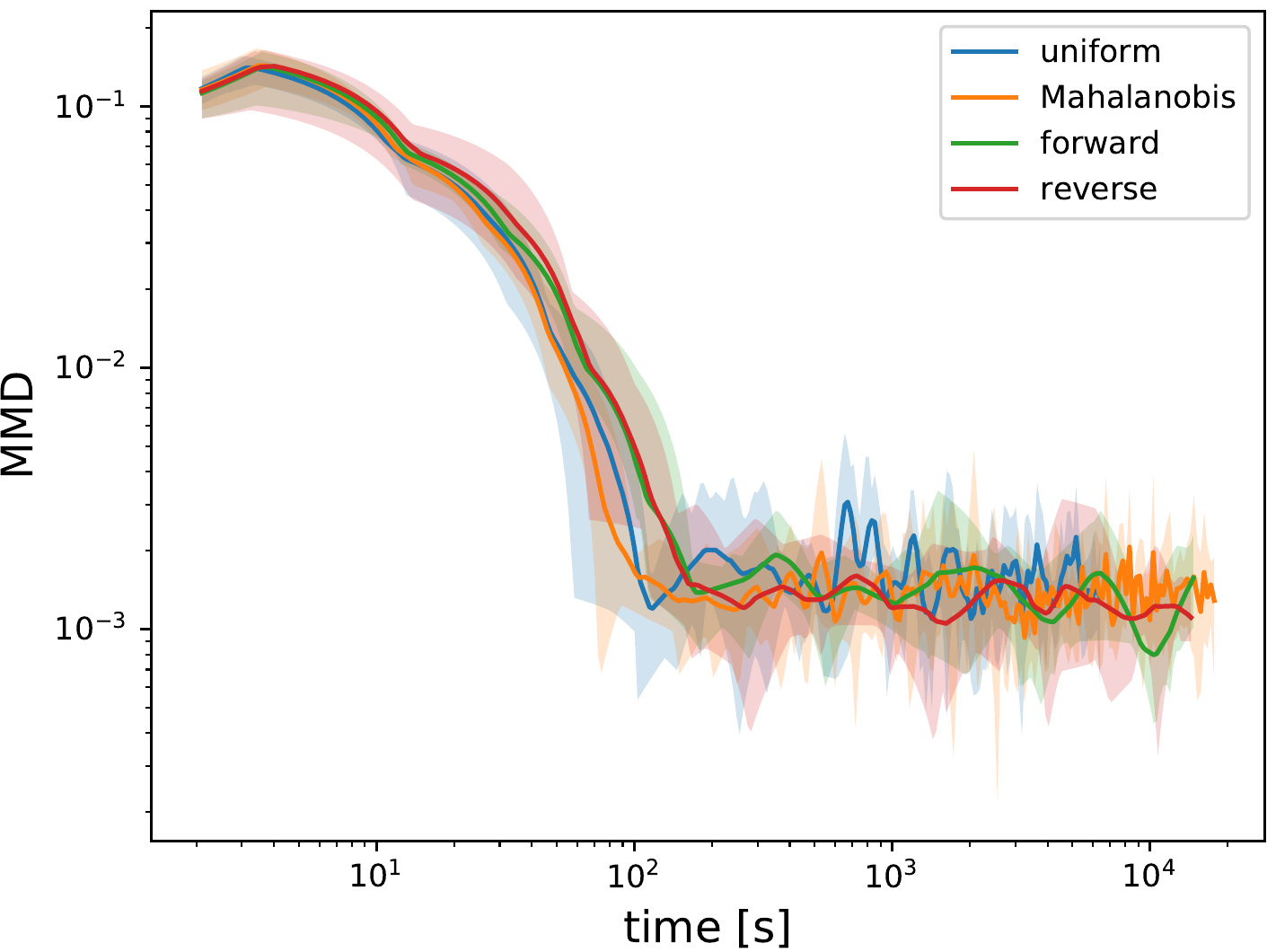}
		\caption{GMM (20D)}
\end{subfigure}  
			\begin{subfigure}{.244\textwidth}
		\centering
		\includegraphics[width=\linewidth]{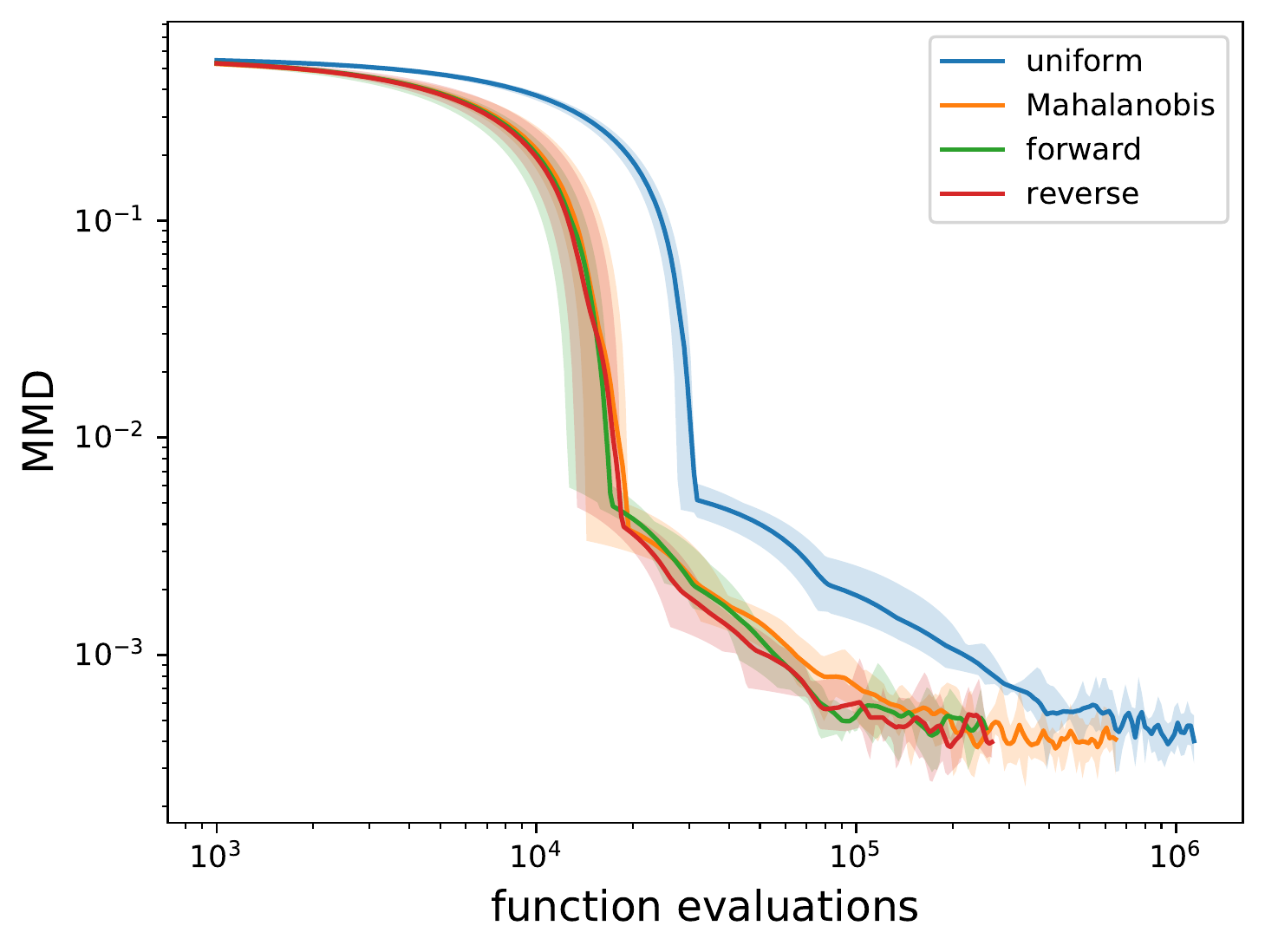}
	\caption{Goodwin}
\end{subfigure}
			\begin{subfigure}{.244\textwidth}
		\centering
		\includegraphics[width=\linewidth]{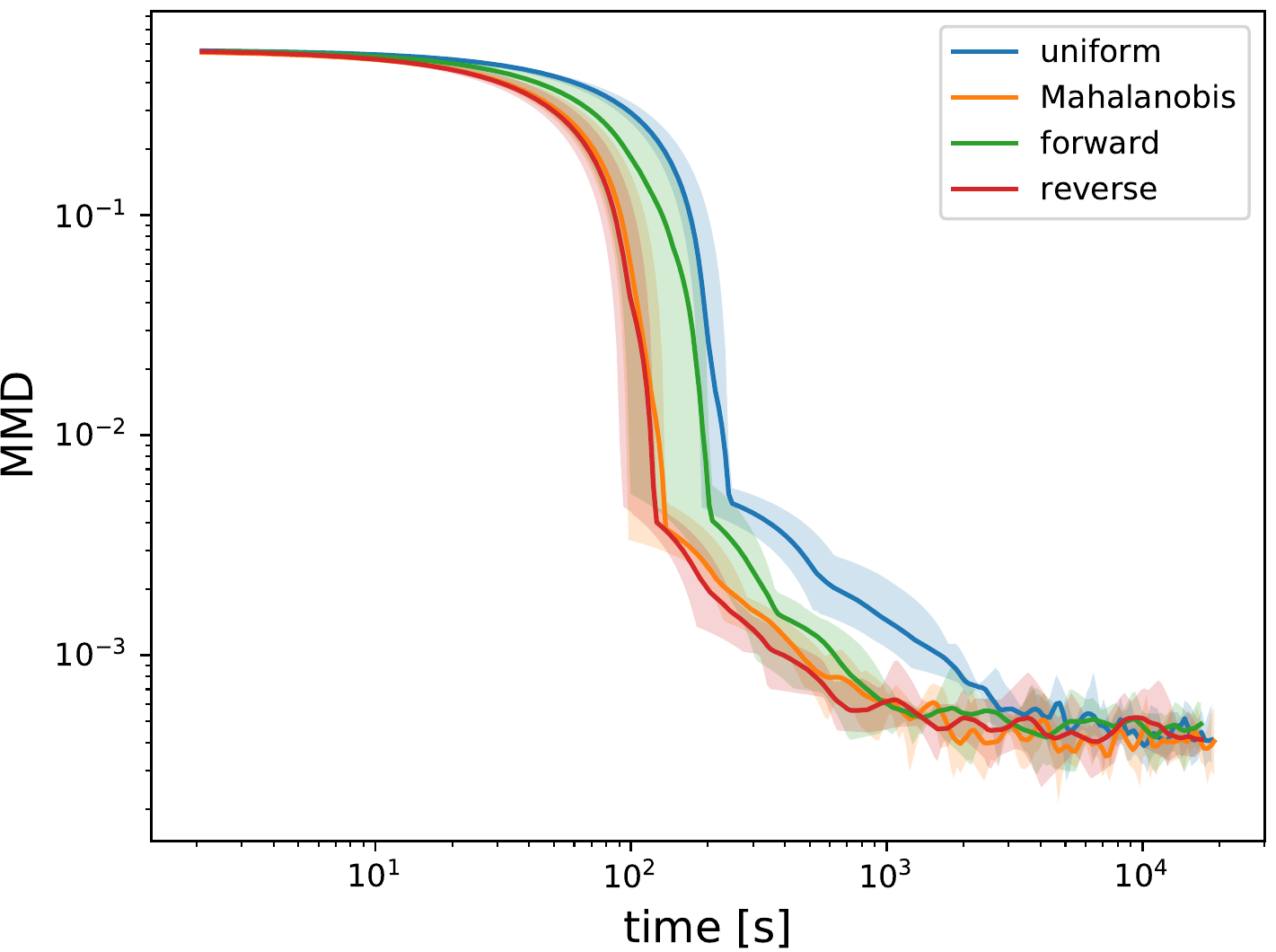}
			\caption{Goodwin}
\end{subfigure}
\caption{Using the Mahalanobis distance as dissimilarity measure results in similar sample efficiency compared to using the KL divergence while adding less computational overhead.}
\label{fig:SampleReusageDistances}
\end{figure}

{\sc{VIPS++}} uses the Mahalanobis distance to the mean of the distributions in the sample database as dissimilarity measure for sample selection according to Equation~\ref{eq:reuseDistribution}. We compared this choice to different dissimilarity measures, namely $KL\left(q(\mathbf{x}|o)||\mathcal{N}_{\mathbf{x}_i}(\mathbf{x}) \right)$ (denoted as reverse KL) and $KL\left(\mathcal{N}_{\mathbf{x}_i}(\mathbf{x}||q(\mathbf{x}|o)) \right)$ (denoted as forward) and against using a uniform distribution instead of Equation~\ref{eq:reuseDistribution}. The results are shown in Figure~\ref{fig:SampleReusageDistances}.

\section{Pseudo-Code for Sample Selection}
\label{app:sample_selection}
The procedure for selecting relevant samples from the database is shown in Algorithm~\ref{alg:selectSamples}.
\begin{algorithm}
   \caption{Identifying relevant samples in the database}
   \label{alg:selectSamples}
\begin{algorithmic}[1]
\Require database $\mathcal{S}=\{(\mathbf{x}_0, \log \tilde{p}(\mathbf{x_0}), \mathcal{N}_\mathbf{x_0}),\dots,(\mathbf{x}_N, \log \tilde{p}(\mathbf{x_N}), \mathcal{N}_\mathbf{x_N}) \}$
\Require number of components in the approximation, $N_o$
\Require desired number of samples that should be reused per component, $n_\text{reuse}$ 
\Function{select\_samples}{}
 \State $\boldsymbol{\mathcal{X}}_\subset \gets \{\}$
 \For{$o = 1 \dots N_{o}$}
    \State $n_\textup{added} \gets 0$
    \State $h(\cdot,o) \gets$ compute for each distinct component in the database according to (\ref{eq:reuseDistribution})
    \While{$n_\textup{added} < n_\textup{reuse}$}
     	\State $i \sim h(\cdot,o)$ \Comment{choose a distribution by sampling $h(i,o)$}
     	\State $h(\cdot,o) \gets$ remove element $i$ and normalize
     	\For {\textbf{each} sample $\mathbf{x_j}$ of component $\mathcal{N}_i$} 
     	    \If {$\mathbf{x}_j \not\in \boldsymbol{\mathcal{X}}_\subset$}
     		    \State $\boldsymbol{\mathcal{X}}_\subset \gets \boldsymbol{\mathcal{X}}_\subset \cup {\mathbf{x}_j}$ 
     		\EndIf
    		\State $n_\textup{added} \gets n_\textup{added} + 1$ \Comment{also count $\mathbf{x}_j$ if it was already added}
			\If {$n_\textup{added} == n_\textup{reuse}$}
    			\State \textbf{break}
    		\EndIf
		\EndFor
    \EndWhile
 \EndFor
 \State \Return $\boldsymbol{\mathcal{X}}_\subset$
\EndFunction
\end{algorithmic}
\end{algorithm}

\section{Approximating the Initial Reward and Sensitivity Regarding its Hyper-parameter}
\label{app:deltaSensitivity}
We approximate the initial reward of a new component based on the approximation given by Equation~\ref{eq:addingHeuristicMax} because it is simpler and more efficient and unlikely to affect the selected candidate. Please note that the difference between the log-sum-exp and the maximum is numerically zero unless for candidates where the density of the current mixture model is close to the threshold. In such case the log-sum-exp can be larger by at most $\log(2)$. 

Furthermore, during our experiments we did not exploit that the initial entropy of the new component can already be computed before deciding on the mean of the new component. Hence, we estimated the density at its mean as 
\begin{equation}
    \label{eq:approximatedCenterDensity}
    q_{\mathbf{x}_s}(\mathbf{x}_s|o_n) \approx \max_{\mathbf{x}_i \in \mathcal{X}_{\text{total}}} \log q(\mathbf{x}_i),
\end{equation}
since the current approximation needs to be evaluated anyway on each candidate for the first operand of the maximum operator in Equation~\ref{eq:addingHeuristicMax}. 
In the main document we presented the more principled, exact computation of $q_{\mathbf{x}_s}(\mathbf{x}_s|o_n)$ to improve clarity. However, in practice the difference between the described heuristic and the implemented heuristic is negligible because the errors that are introduced by the approximation are small compared to the variations of the assumed log weight $\log q(o_n)$.

\begin{figure}
	\centering
	\includegraphics[width=0.5 \linewidth]{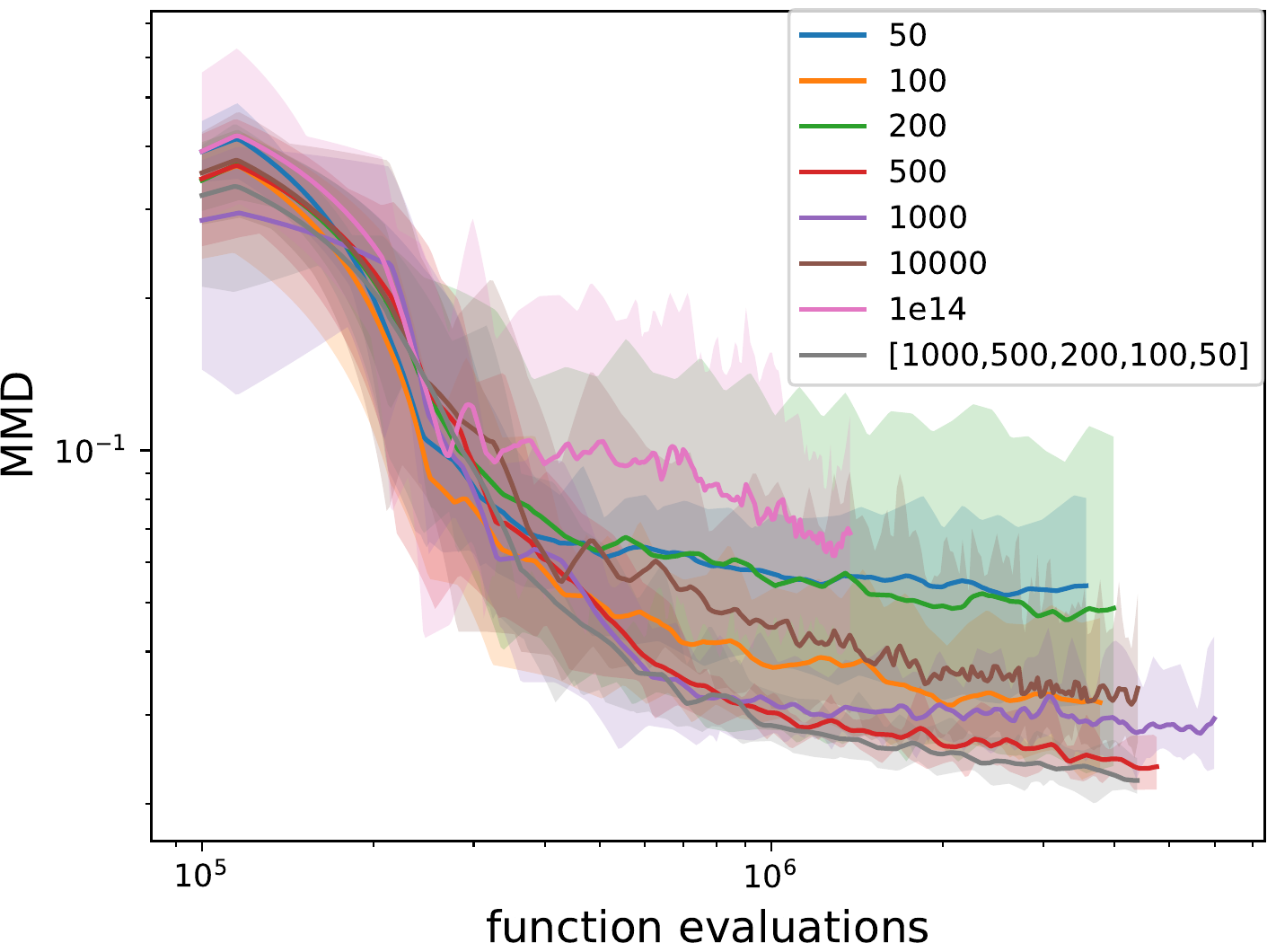}
	\caption{We evaluated the MMD for the \textit{planar robot} experiment with four goals for different fixed values $\Delta$ of the assumed initial log-weight (negated) as well as for varying values. Varying the value ({\sc{VIPS++}}) performed better than any fixed value that we tested. However, the experiment with $\Delta=500$ indicates that tuning a fixed value may also perform well.}
	\label{fig:planarDeltaAblations}
\end{figure}

\begin{figure}
	\centering
			\begin{subfigure}{.19\textwidth}
		\centering
		\includegraphics[width=\linewidth]{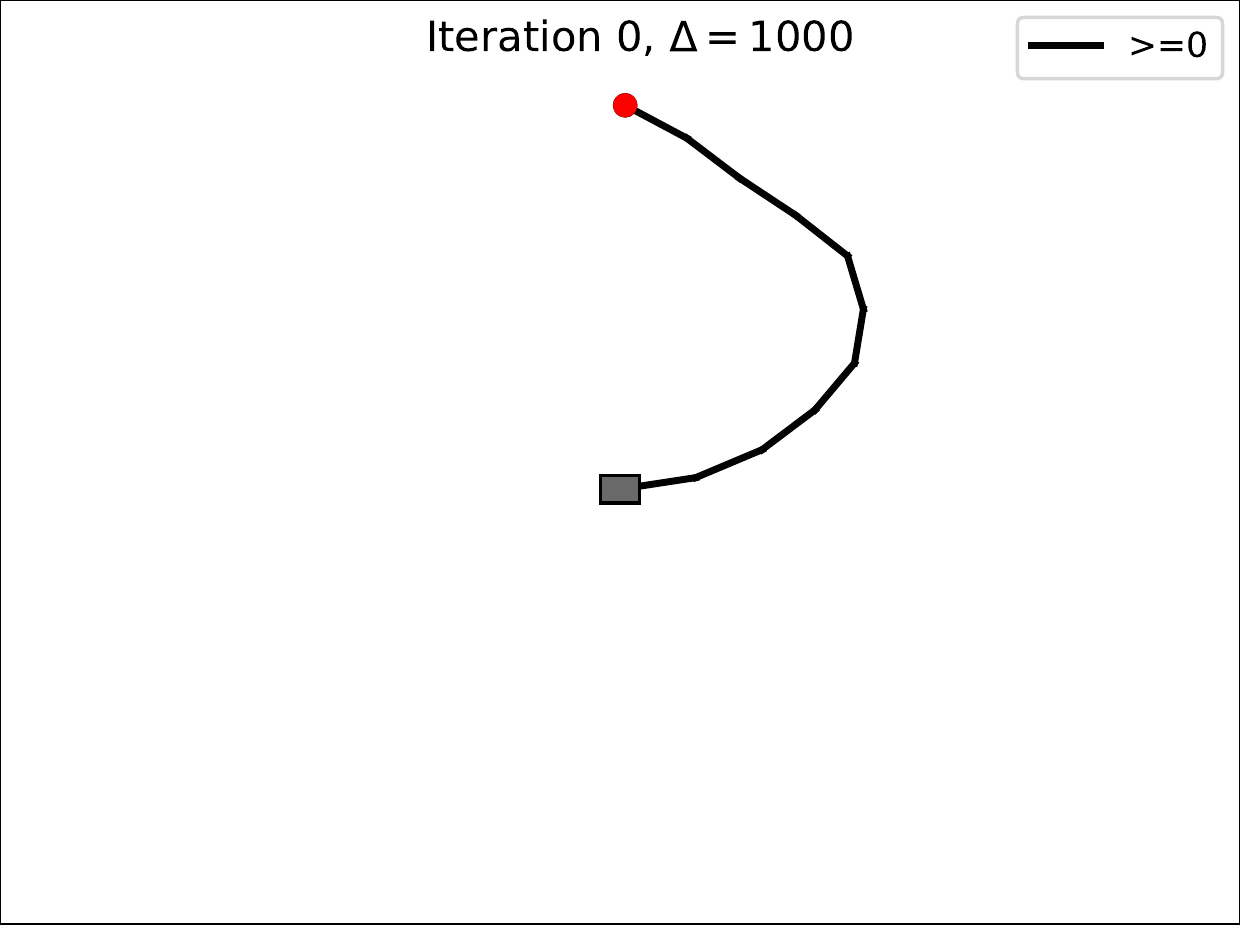}
	\end{subfigure}
	\begin{subfigure}{.19\textwidth}
		\centering
		\includegraphics[width=\linewidth]{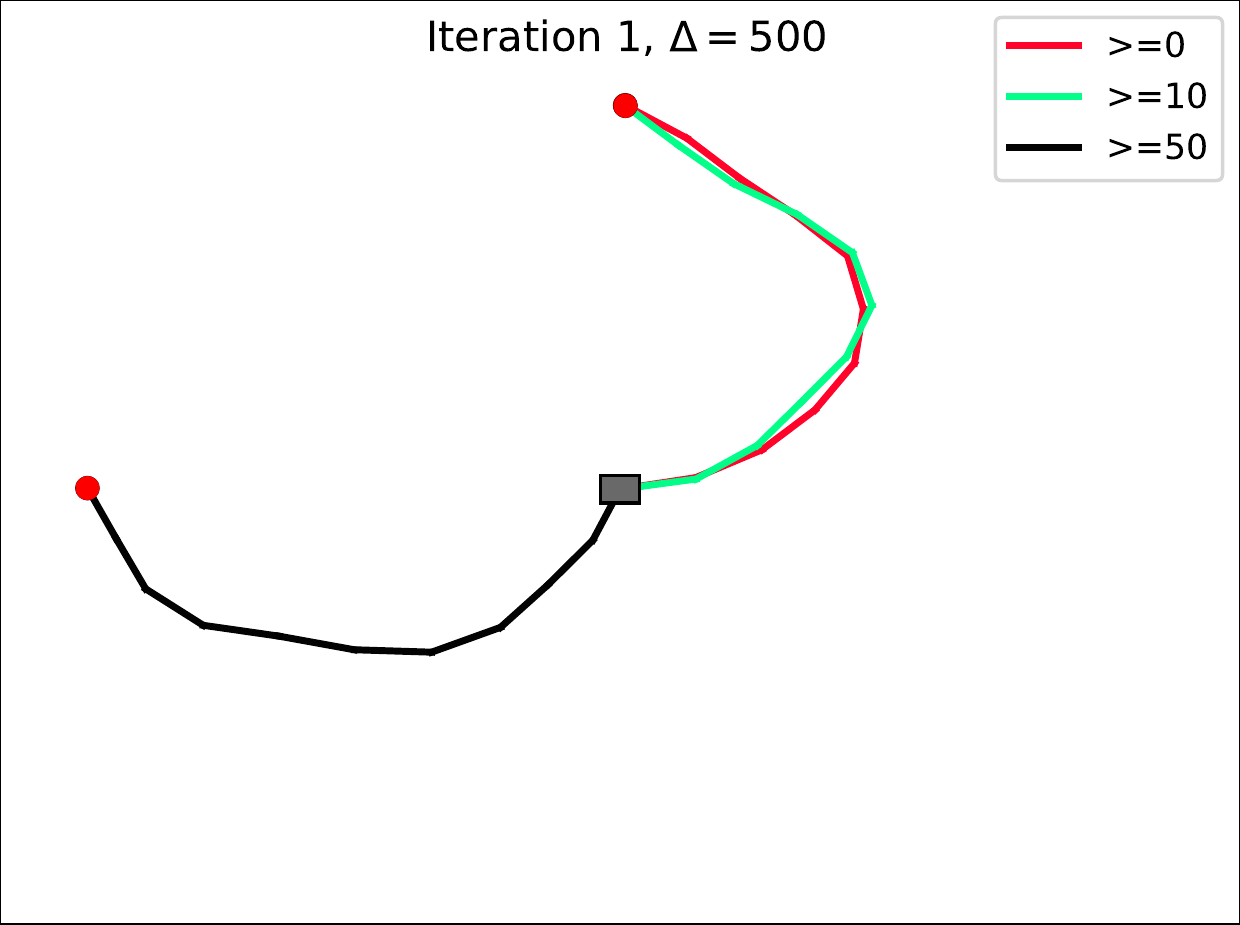}
	\end{subfigure}
	\begin{subfigure}{.19\textwidth}
		\centering
		\includegraphics[width=\linewidth]{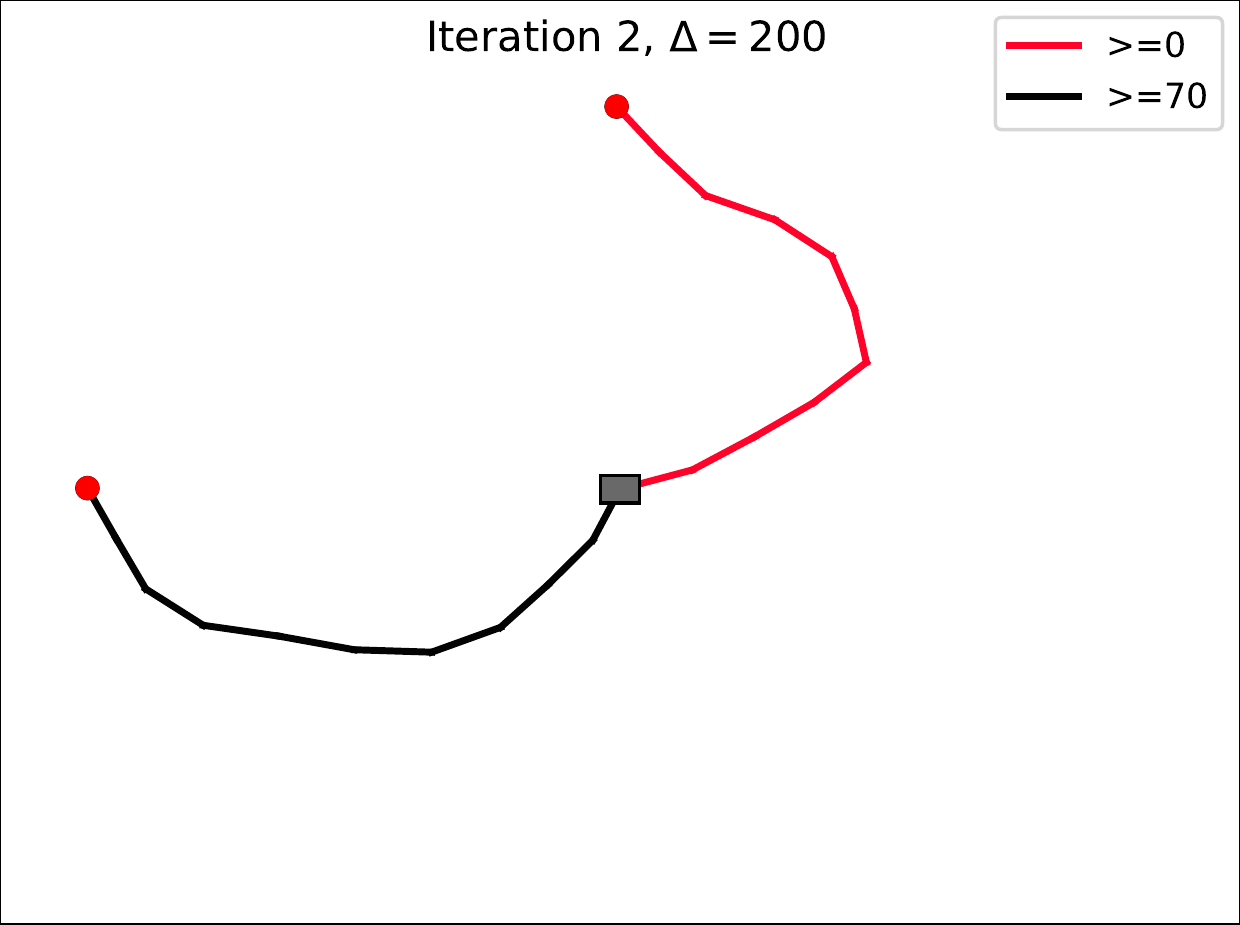}
	\end{subfigure}
	\begin{subfigure}{.19\textwidth}
		\centering
		\includegraphics[width=\linewidth]{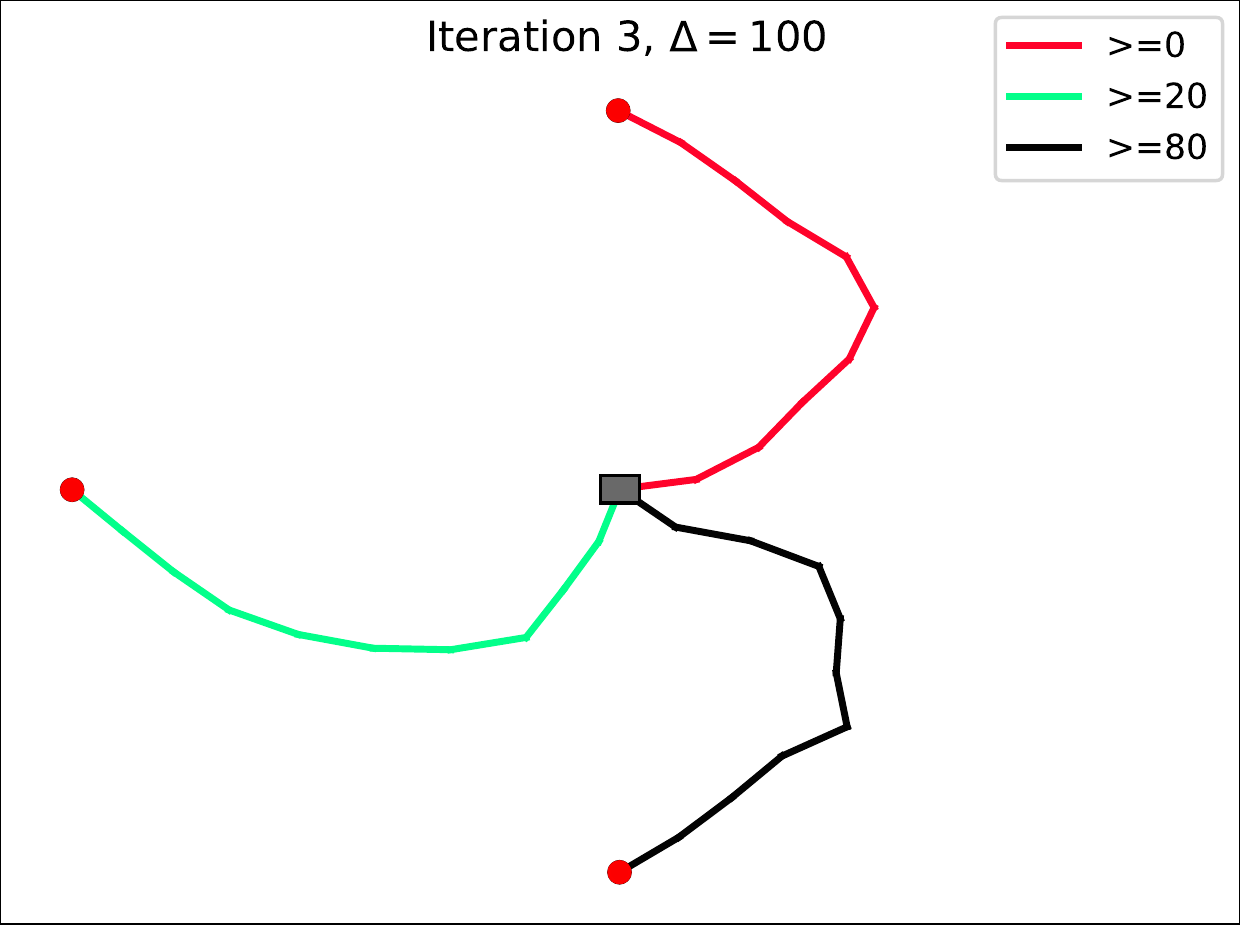}
	\end{subfigure}        
	\begin{subfigure}{.19\textwidth}%
		\centering
		\includegraphics[width=\linewidth]{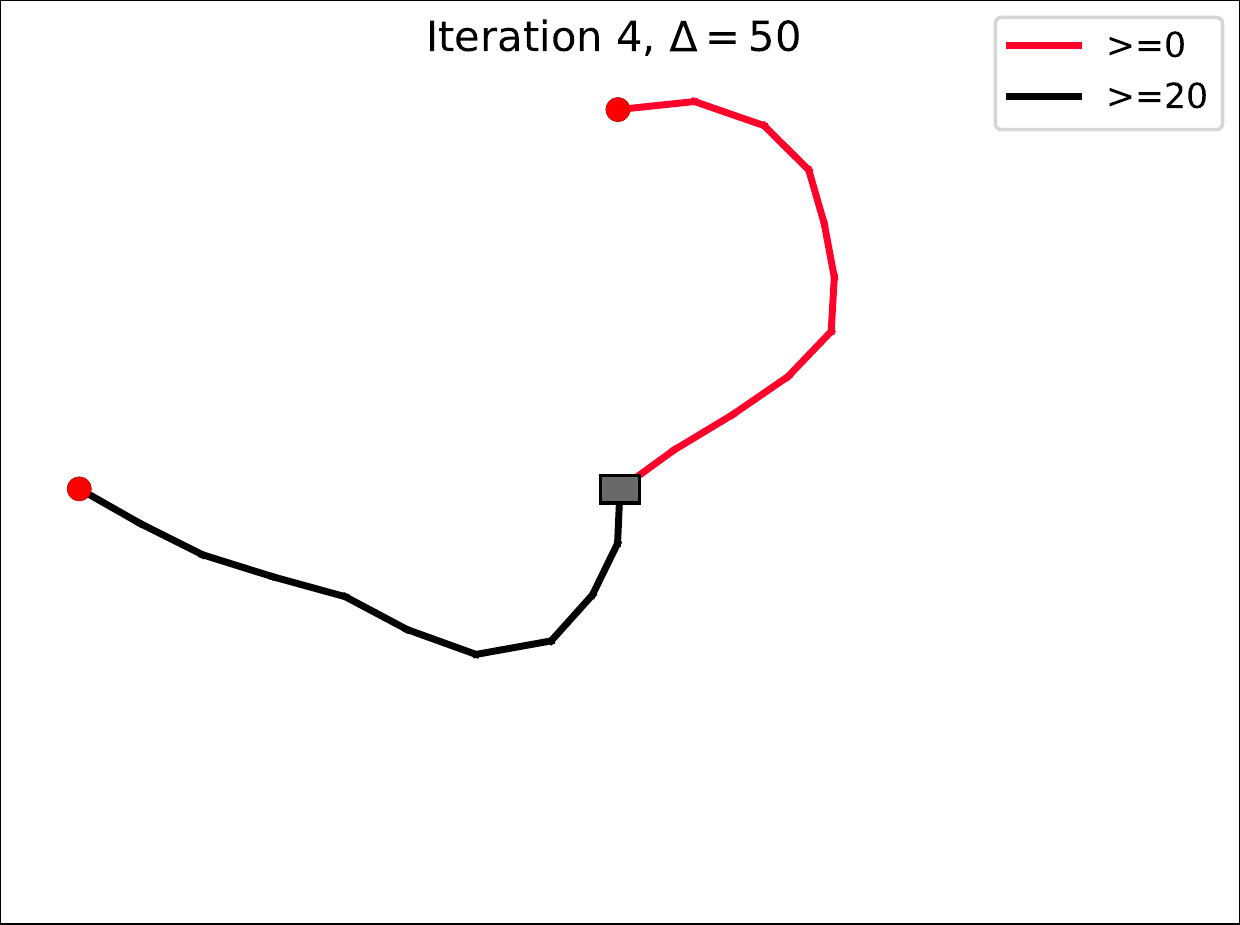}
	\end{subfigure}
	
		\begin{subfigure}{.19\textwidth}
		\centering
		\includegraphics[width=\linewidth]{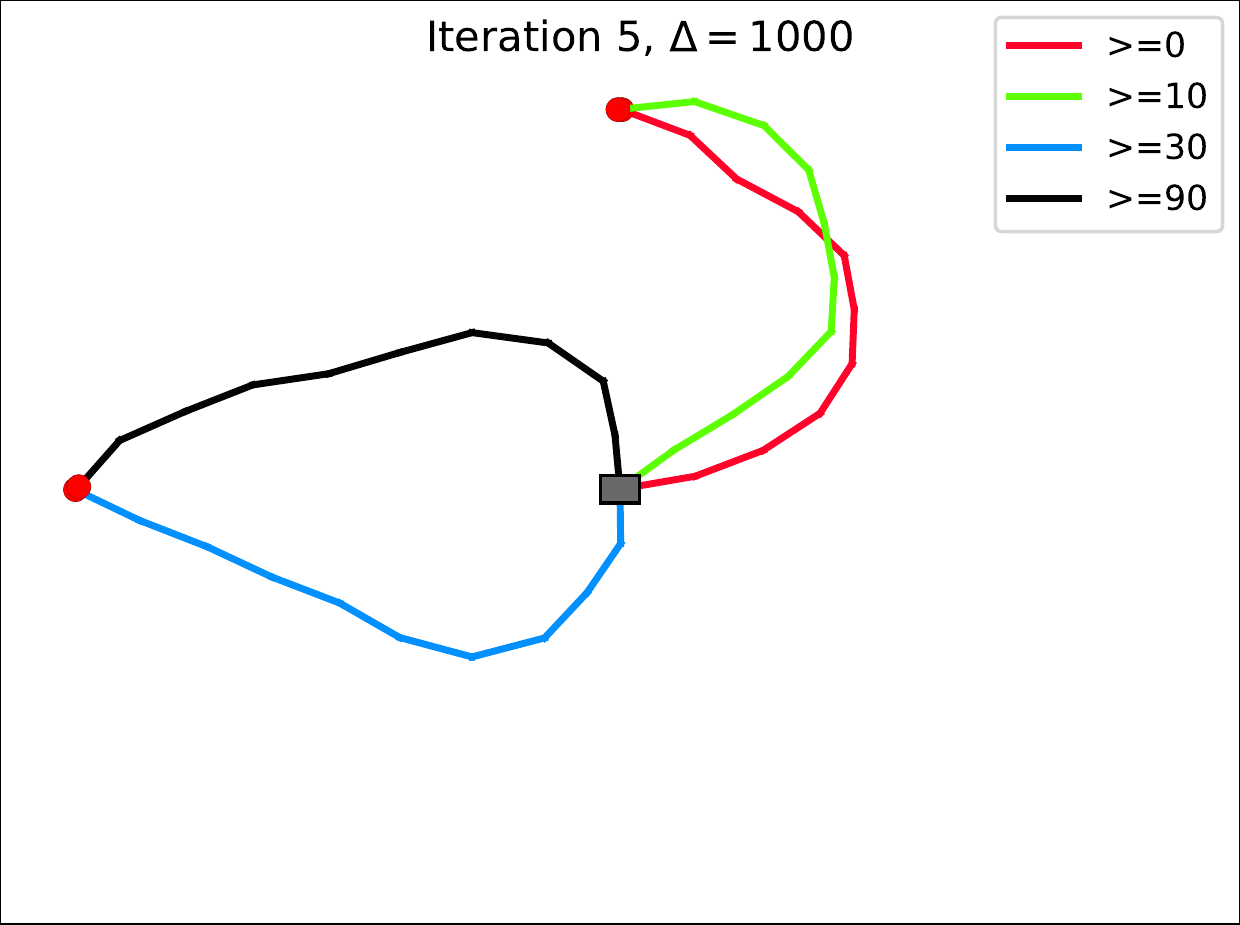}
	\end{subfigure}
	\begin{subfigure}{.19\textwidth}
		\centering
		\includegraphics[width=\linewidth]{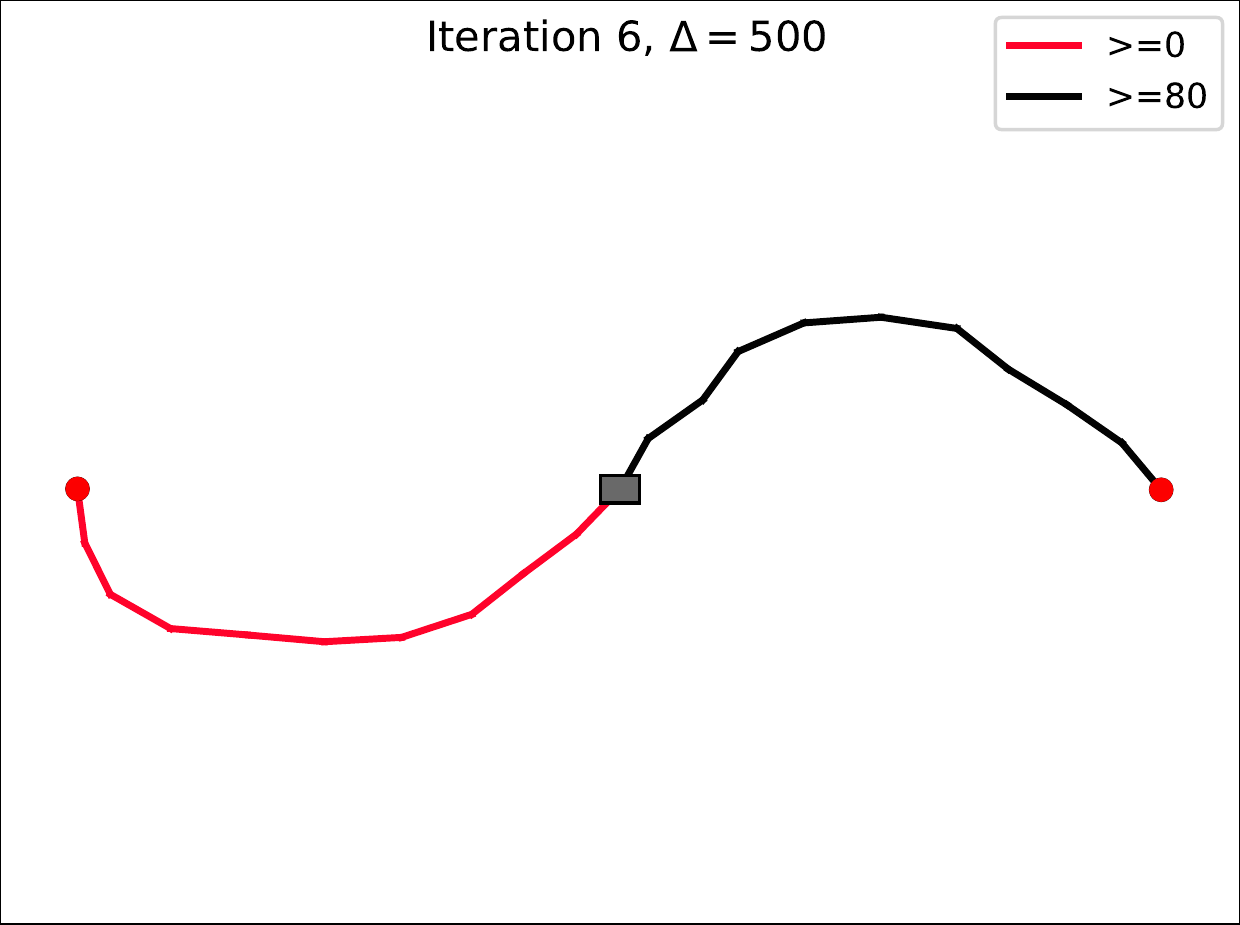}
	\end{subfigure}
	\begin{subfigure}{.19\textwidth}
		\centering
		\includegraphics[width=\linewidth]{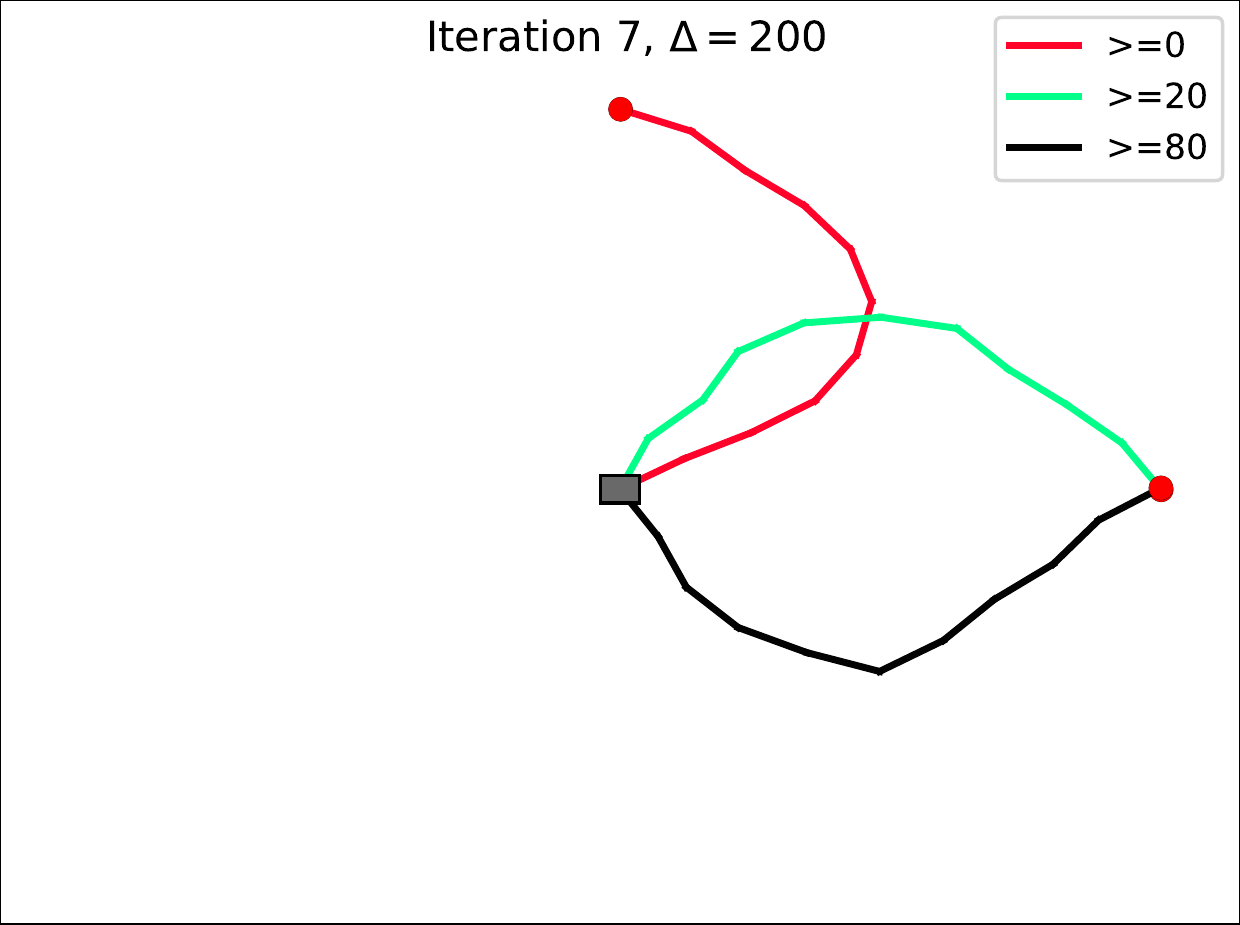}
	\end{subfigure}
	\begin{subfigure}{.19\textwidth}
		\centering
		\includegraphics[width=\linewidth]{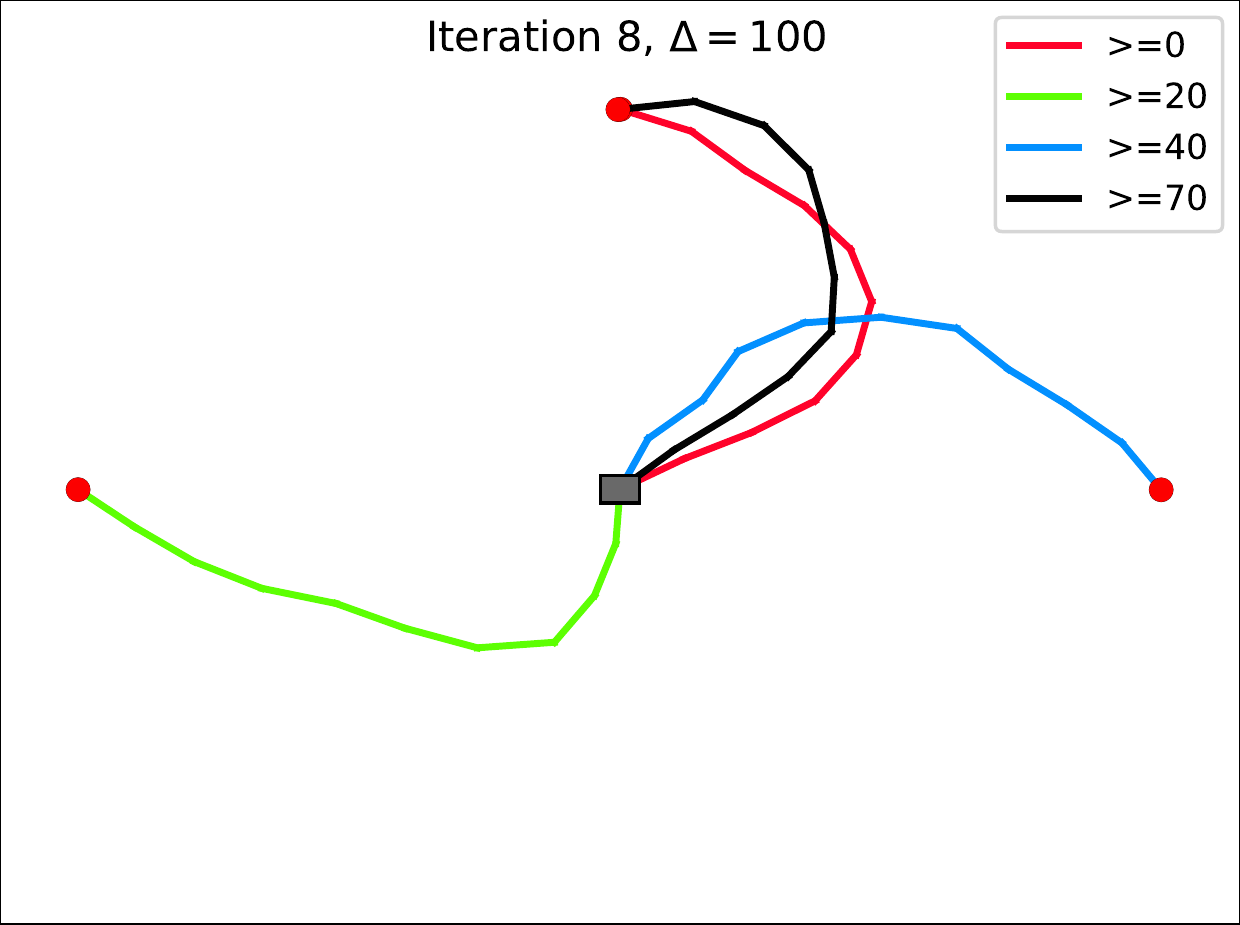}
	\end{subfigure}        
	\begin{subfigure}{.19\textwidth}%
		\centering
		\includegraphics[width=\linewidth]{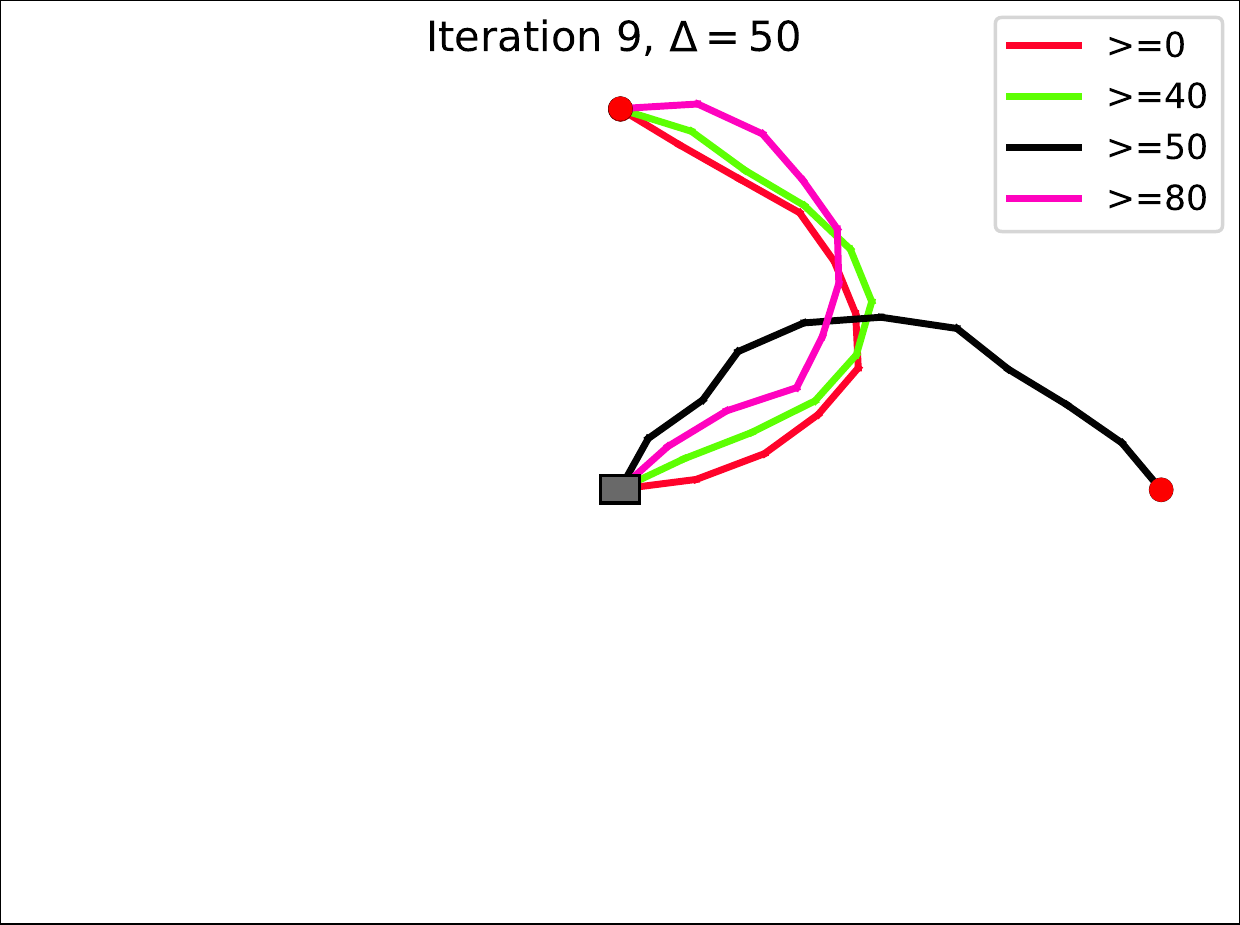}
	\end{subfigure}
	
			\begin{subfigure}{.19\textwidth}
		\centering
		\includegraphics[width=\linewidth]{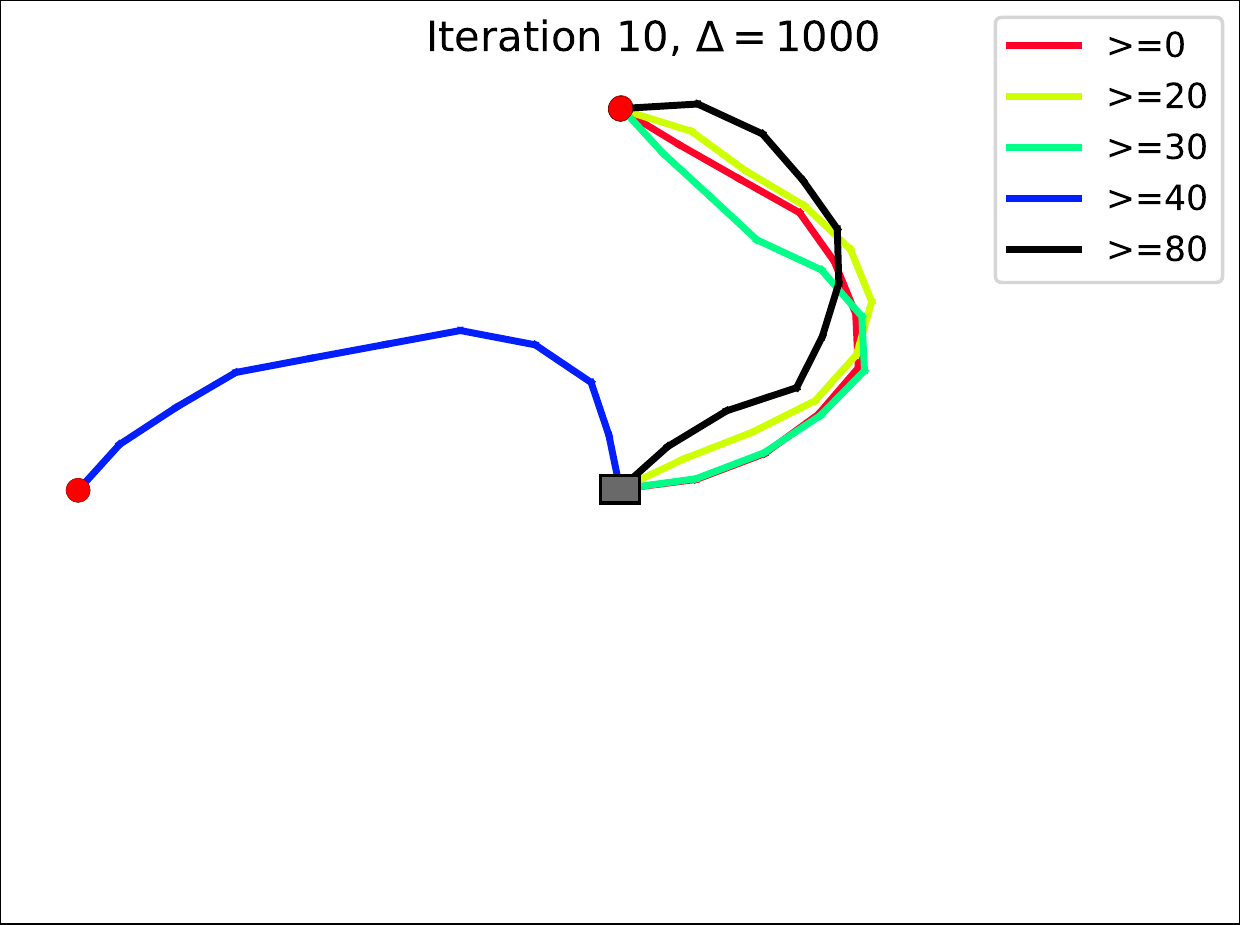}
	\end{subfigure}
	\begin{subfigure}{.19\textwidth}
		\centering
		\includegraphics[width=\linewidth]{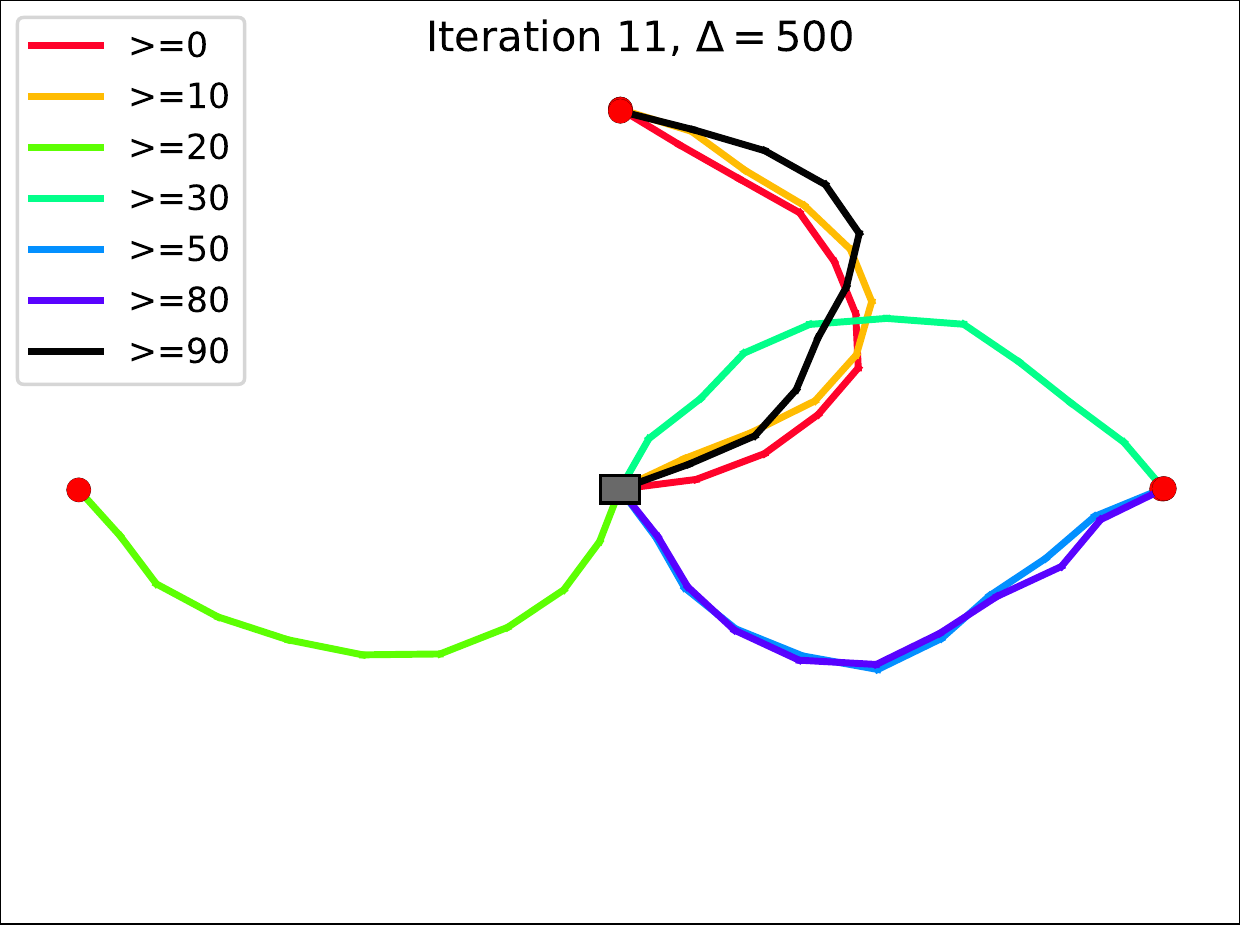}
	\end{subfigure}
	\begin{subfigure}{.19\textwidth}
		\centering
		\includegraphics[width=\linewidth]{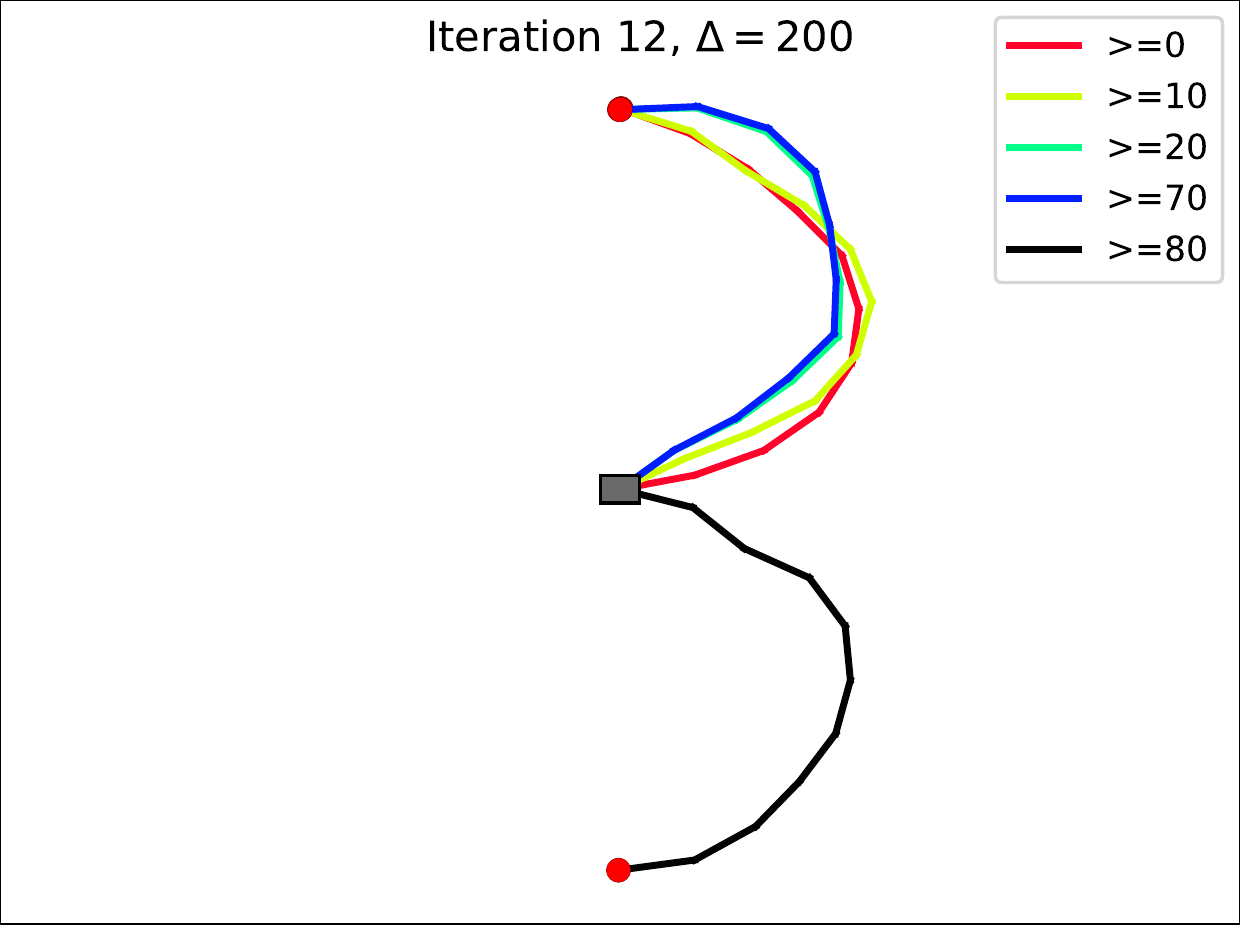}
	\end{subfigure}
	\begin{subfigure}{.19\textwidth}
		\centering
		\includegraphics[width=\linewidth]{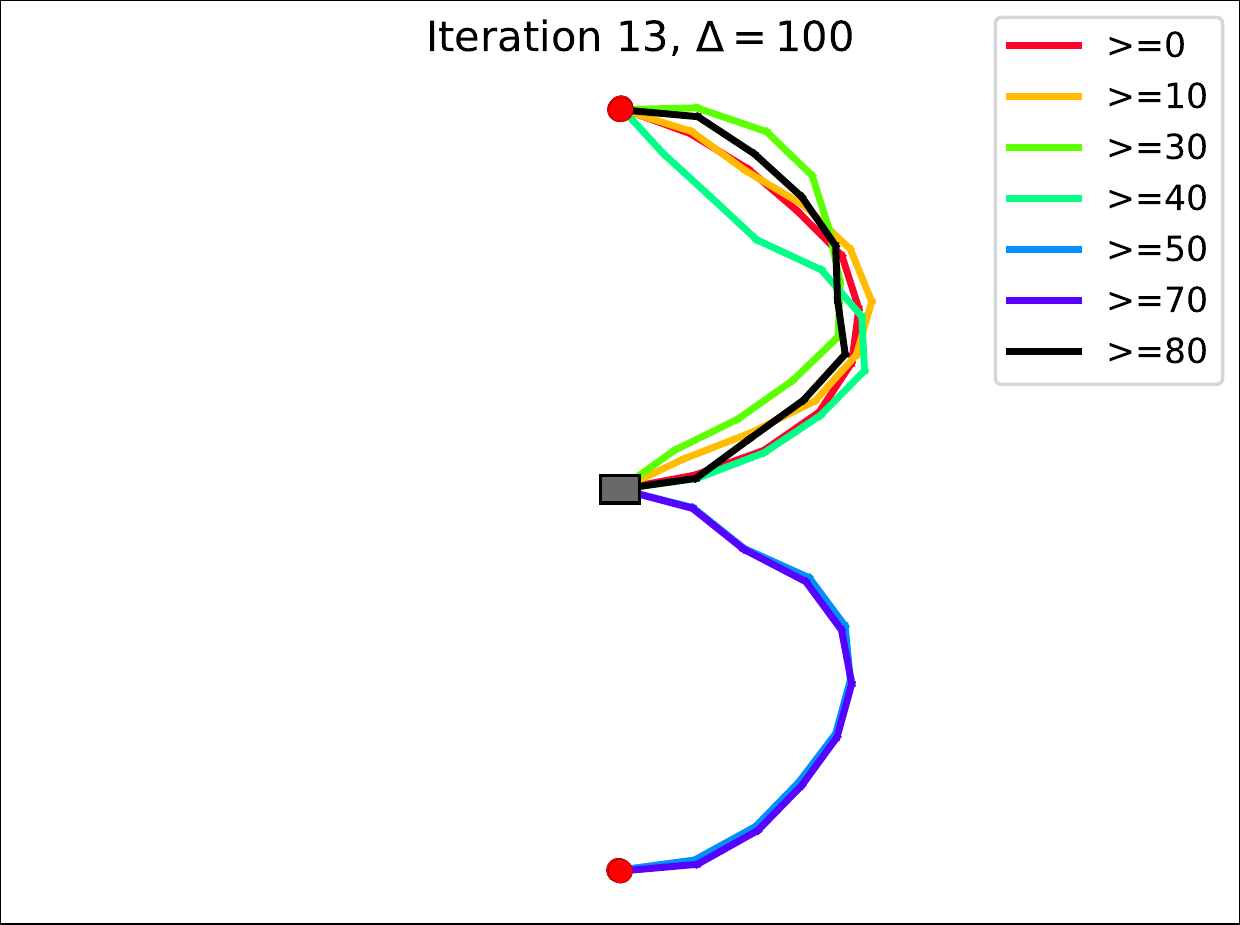}
	\end{subfigure}        
	\begin{subfigure}{.19\textwidth}%
		\centering
		\includegraphics[width=\linewidth]{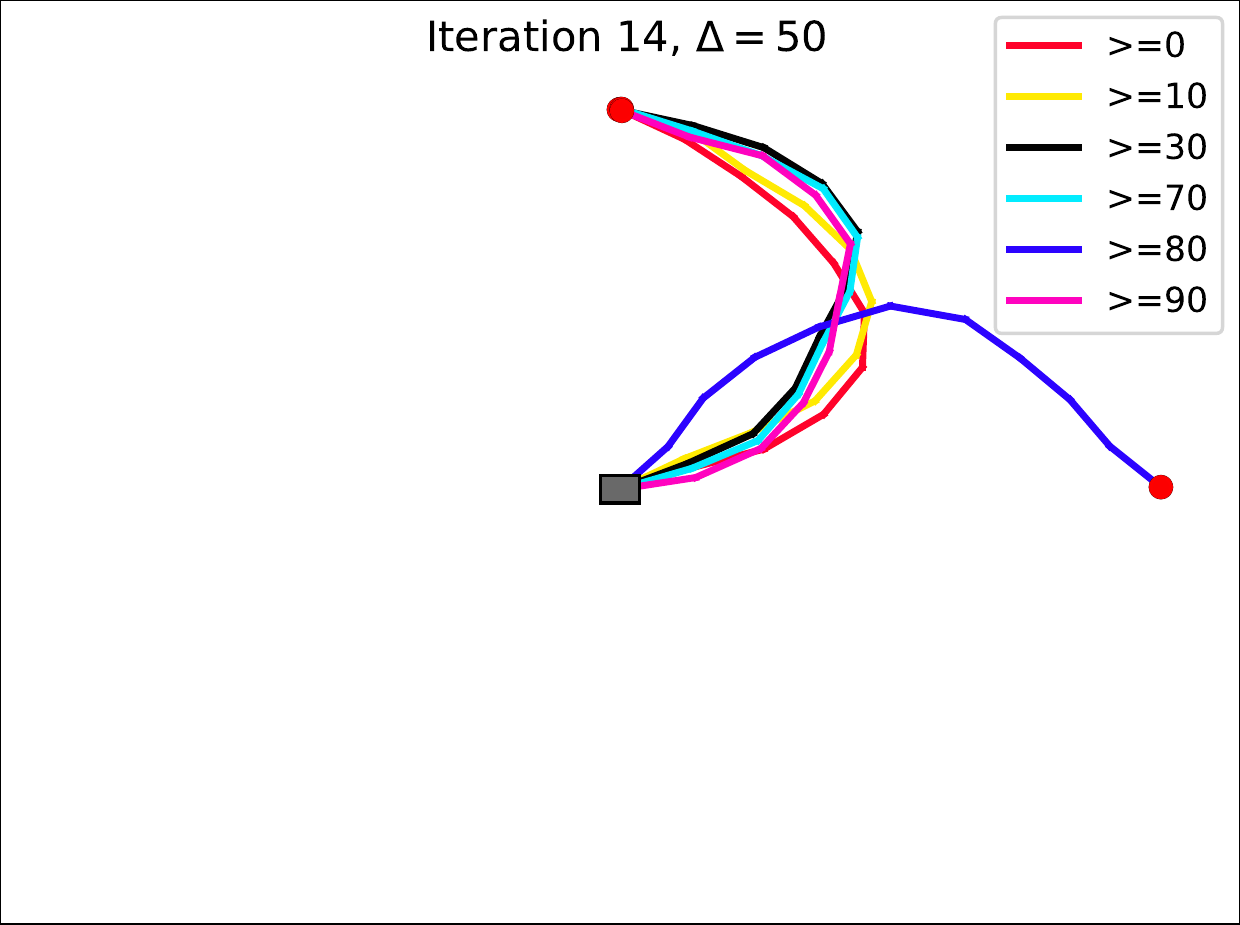}
	\end{subfigure}
	
		\begin{subfigure}{.19\textwidth}
		\centering
		\includegraphics[width=\linewidth]{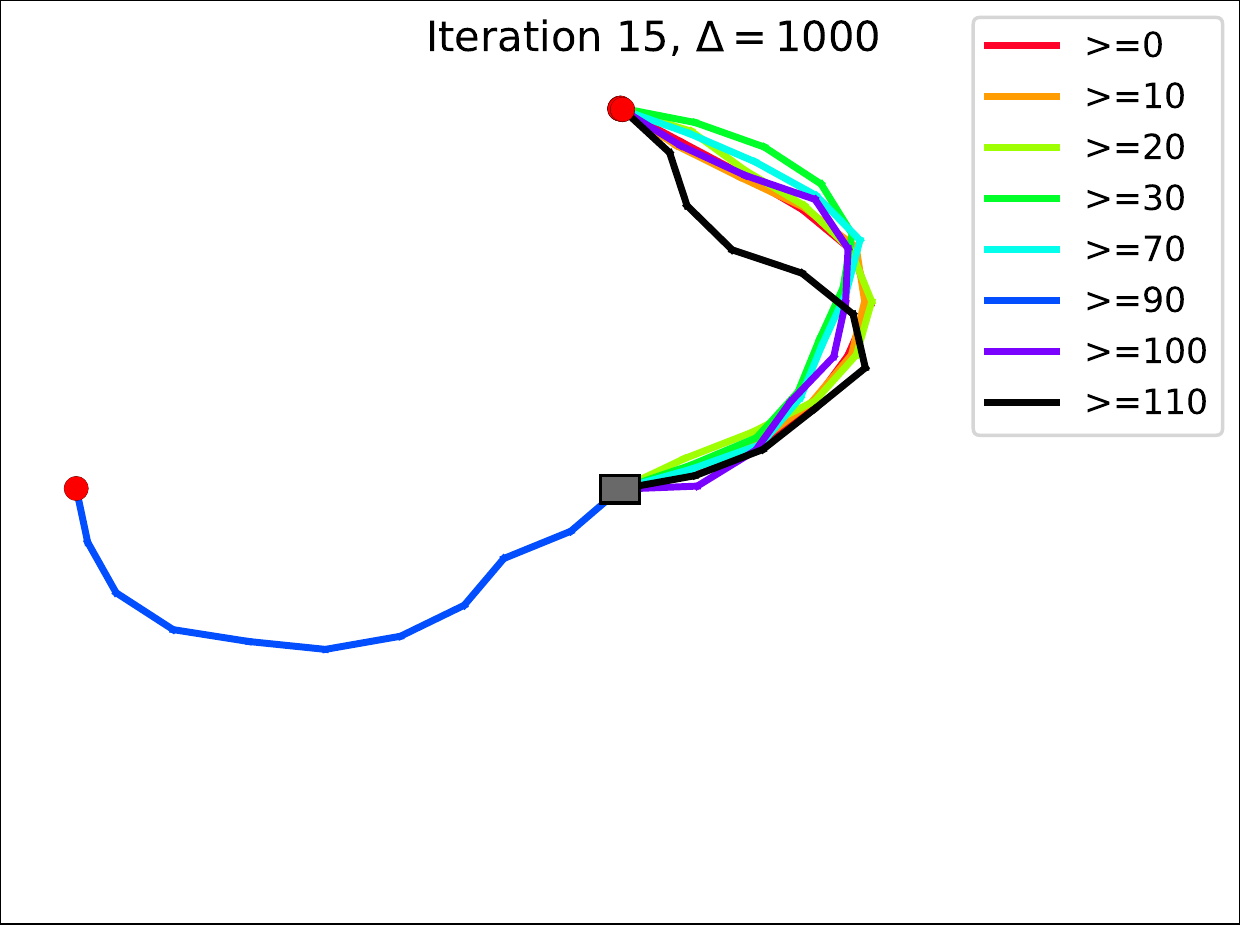}
	\end{subfigure}
	\begin{subfigure}{.19\textwidth}
		\centering
		\includegraphics[width=\linewidth]{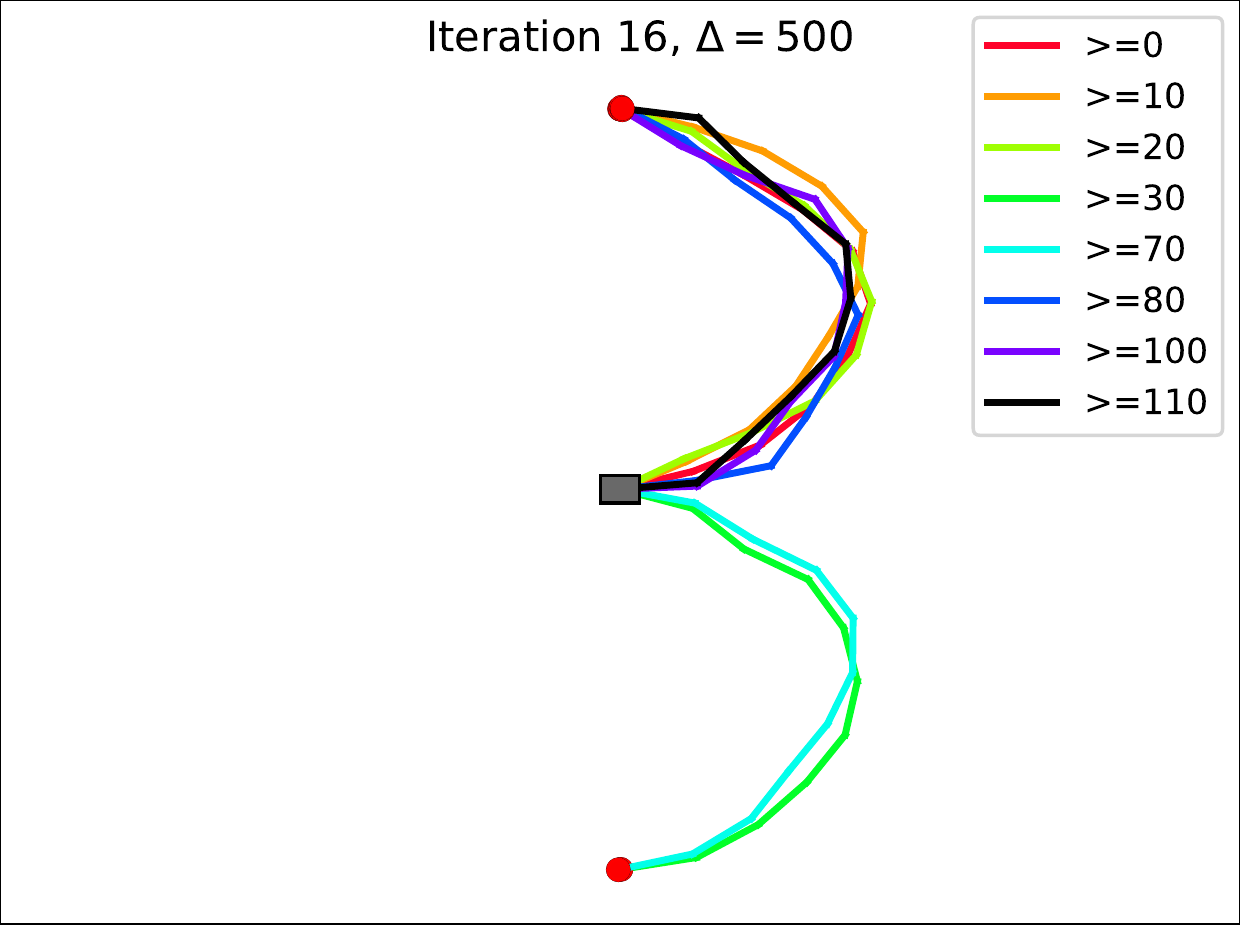}
	\end{subfigure}
	\begin{subfigure}{.19\textwidth}
		\centering
		\includegraphics[width=\linewidth]{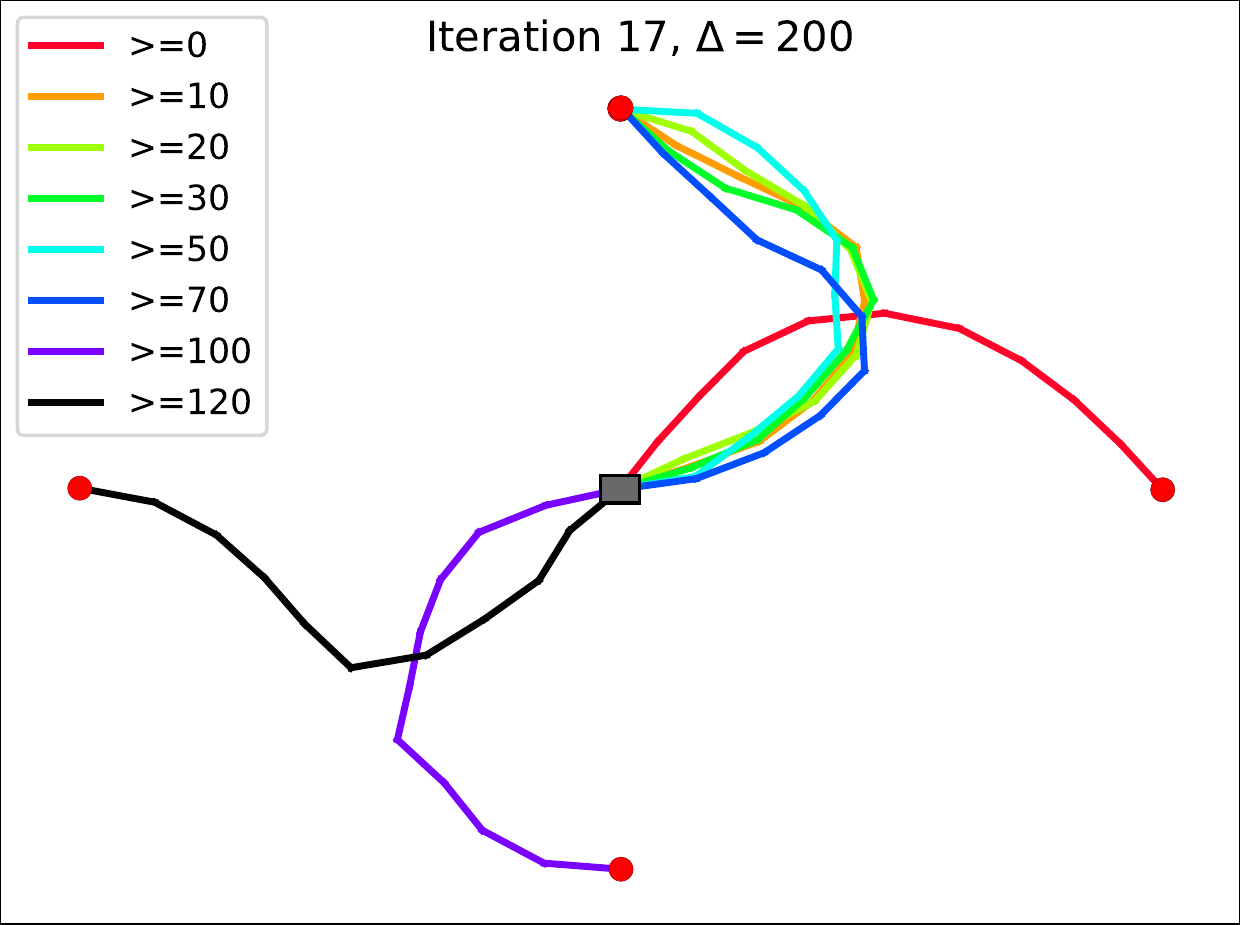}
	\end{subfigure}
	\begin{subfigure}{.19\textwidth}
		\centering
		\includegraphics[width=\linewidth]{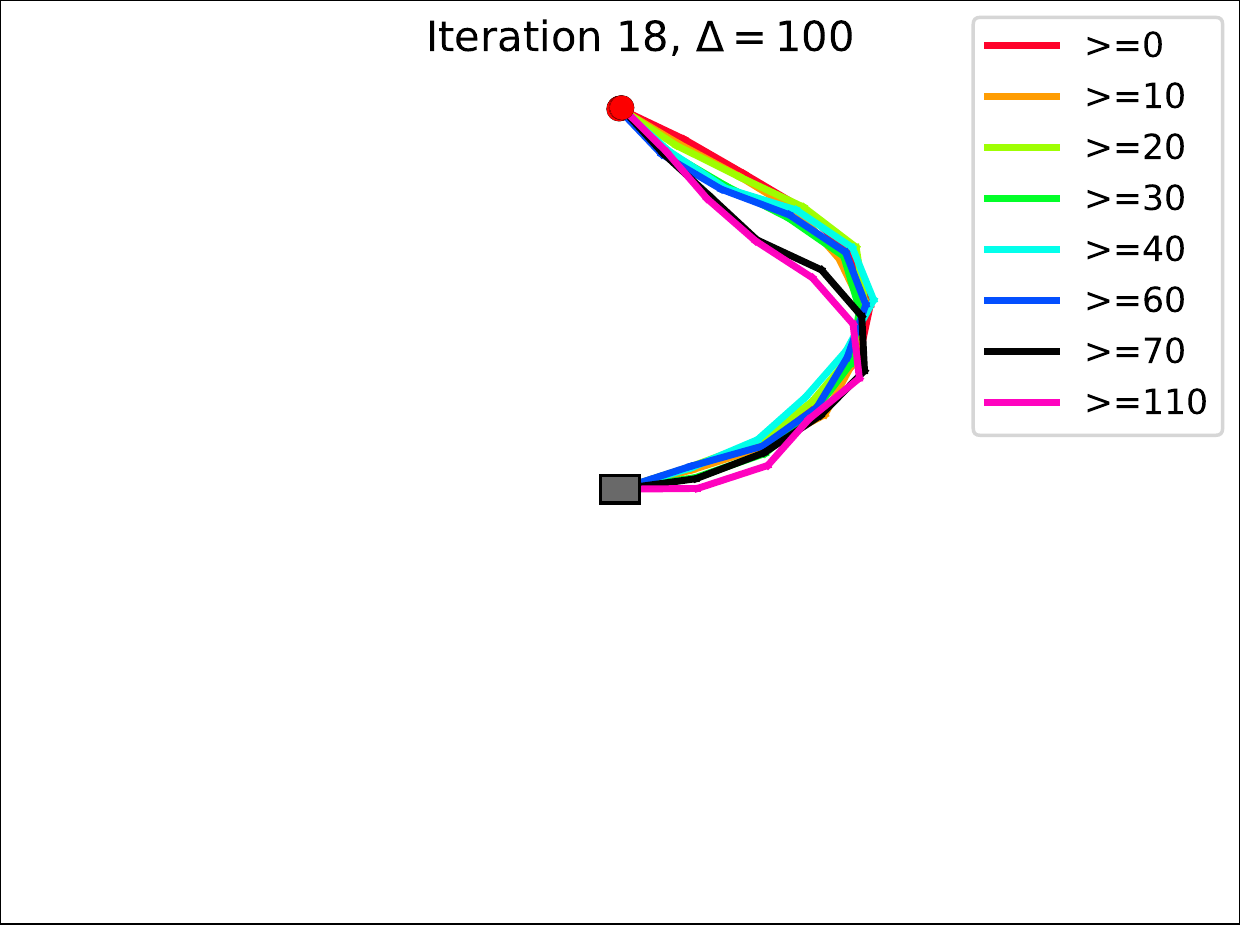}
	\end{subfigure}        
	\begin{subfigure}{.19\textwidth}%
		\centering
		\includegraphics[width=\linewidth]{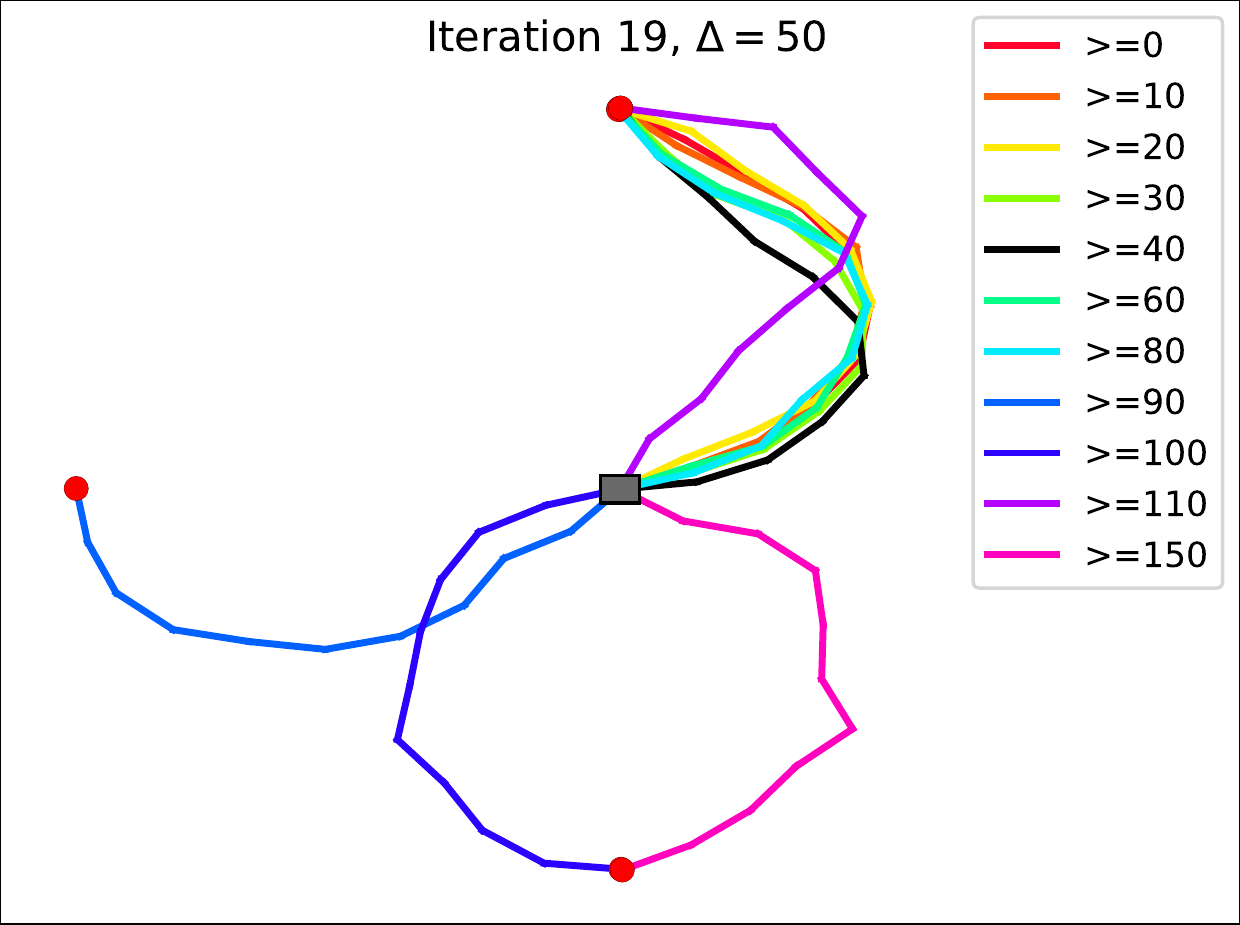}
	\end{subfigure}
	
				\begin{subfigure}{.19\textwidth}
		\centering
		\includegraphics[width=\linewidth]{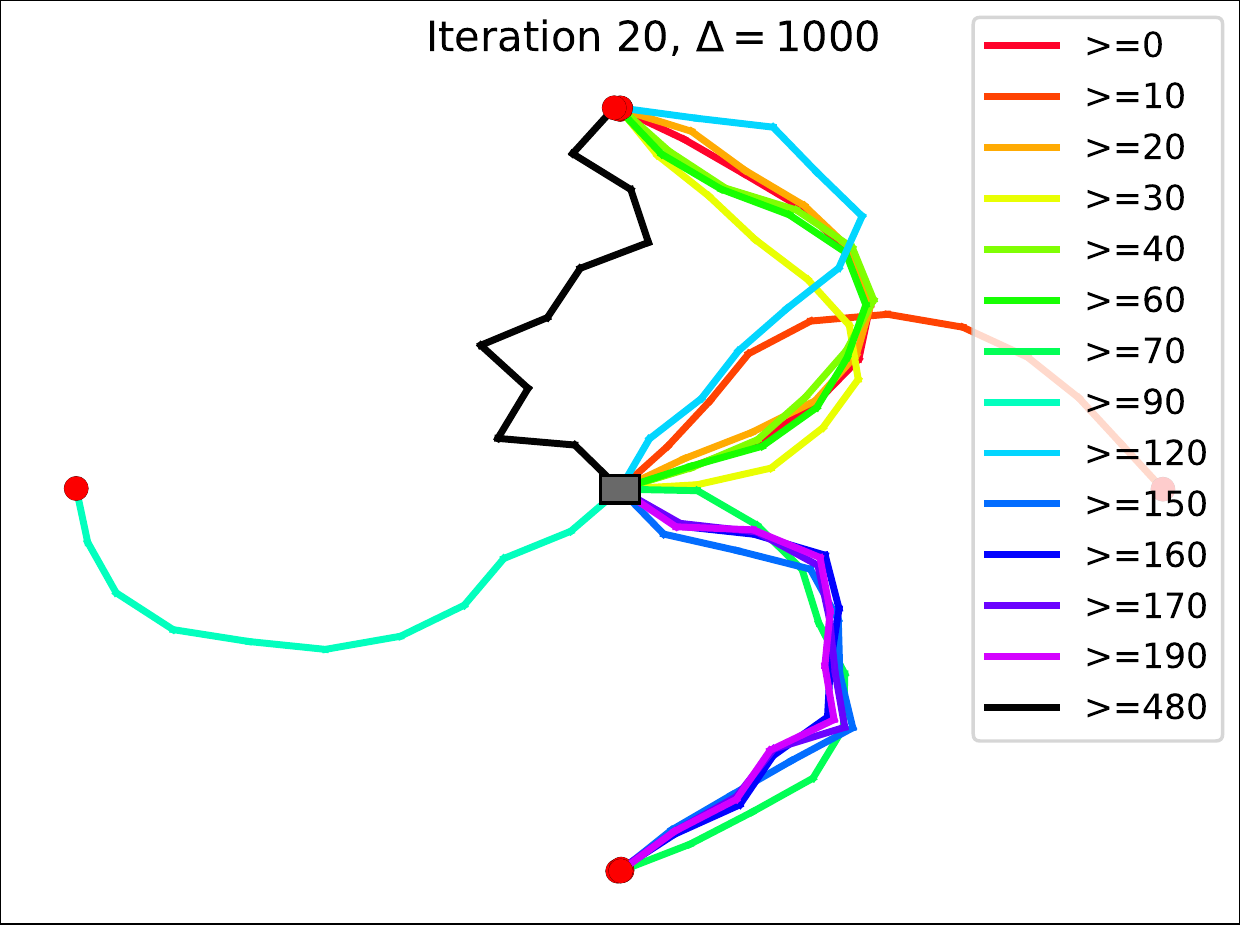}
	\end{subfigure}
	\begin{subfigure}{.19\textwidth}
		\centering
		\includegraphics[width=\linewidth]{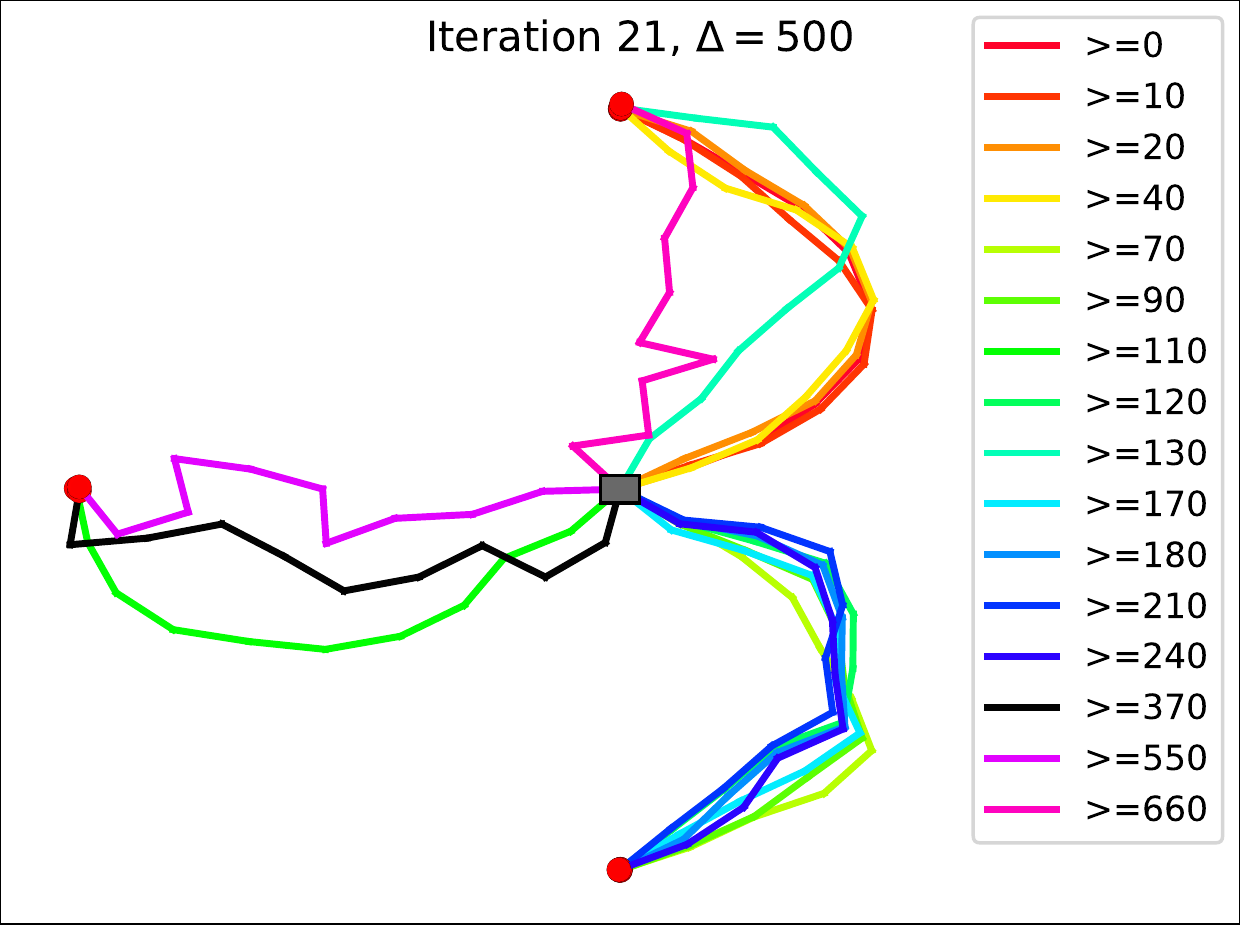}
	\end{subfigure}
	\begin{subfigure}{.19\textwidth}
		\centering
		\includegraphics[width=\linewidth]{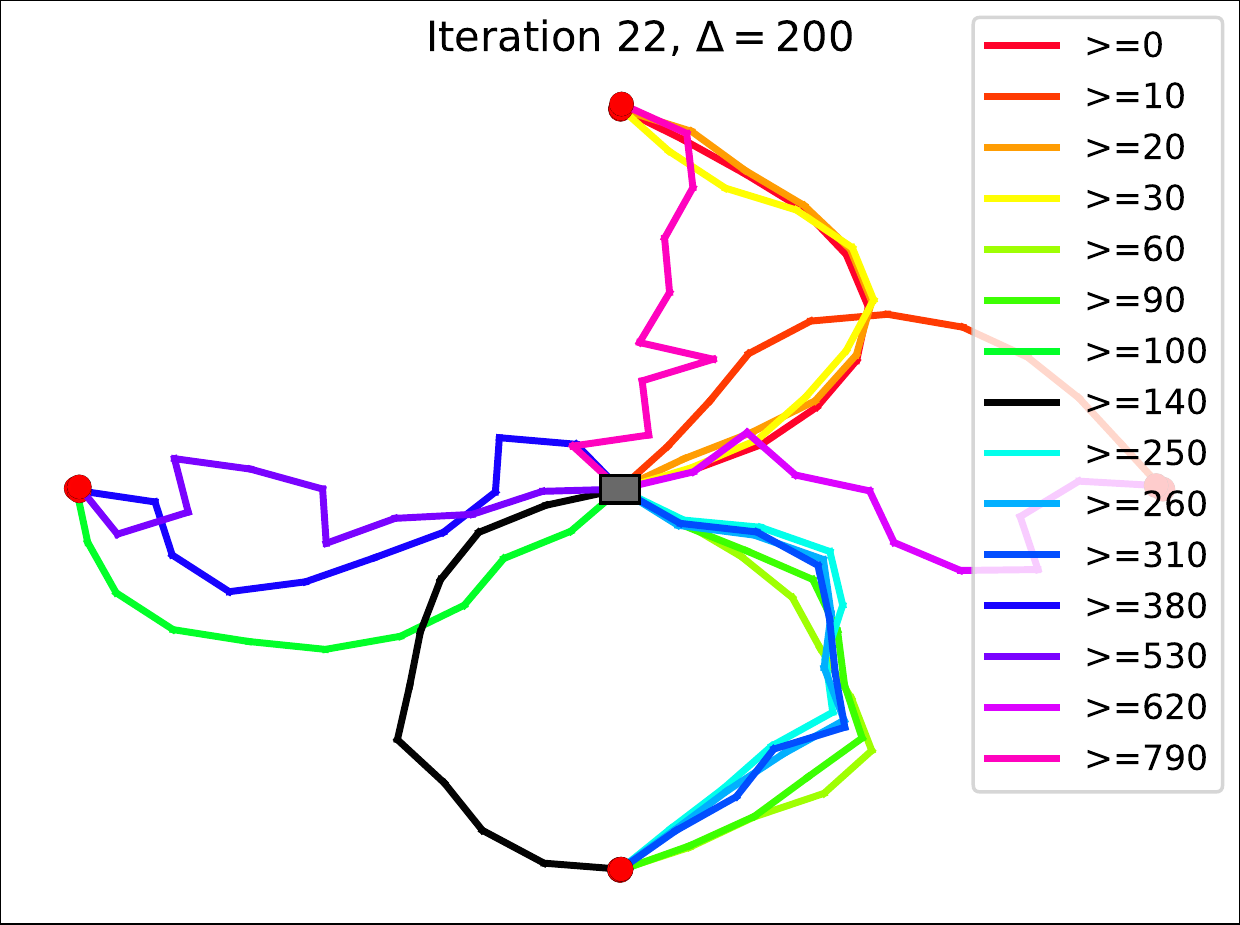}
	\end{subfigure}
	\begin{subfigure}{.19\textwidth}
		\centering
		\includegraphics[width=\linewidth]{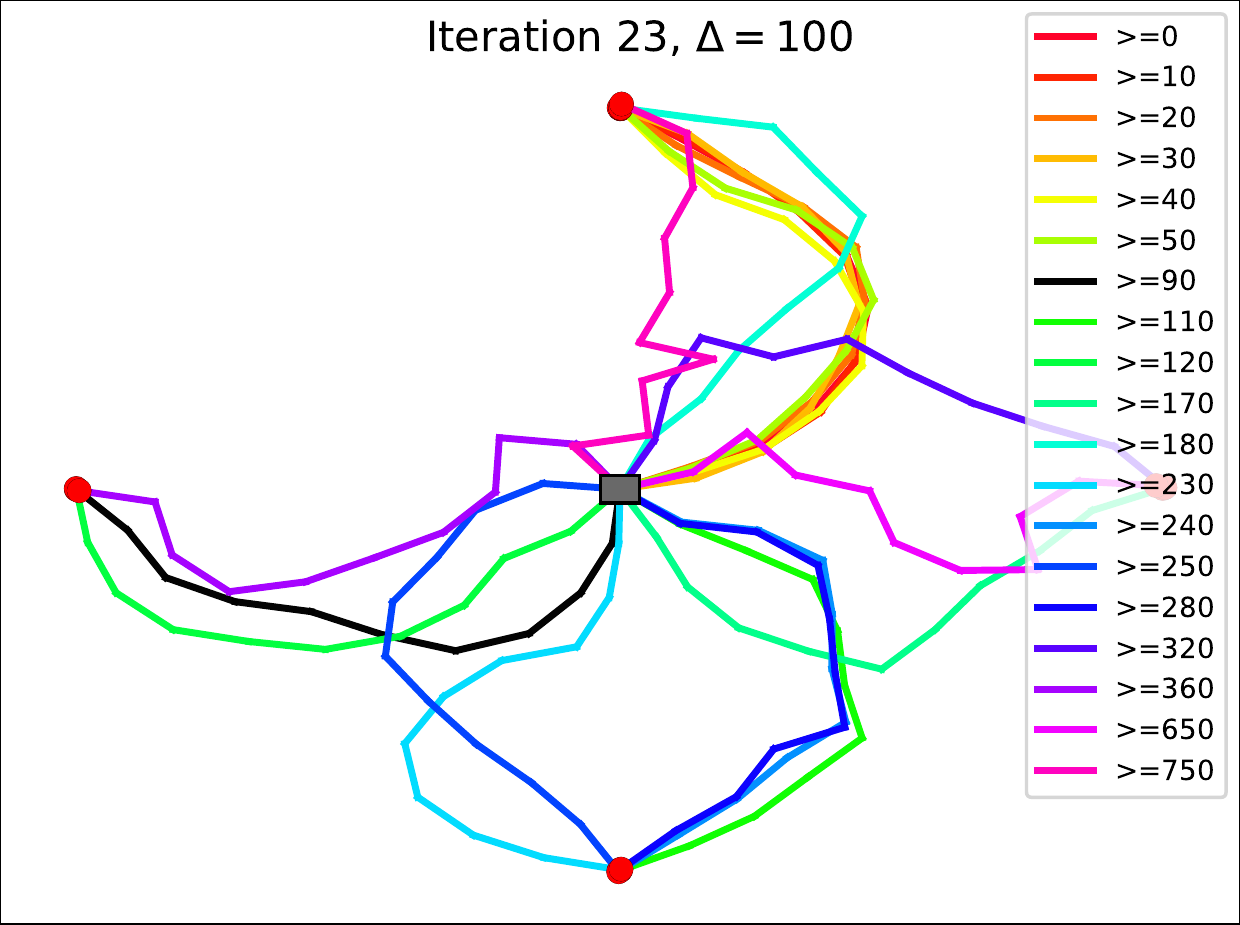}
	\end{subfigure}        
	\begin{subfigure}{.19\textwidth}%
		\centering
		\includegraphics[width=\linewidth]{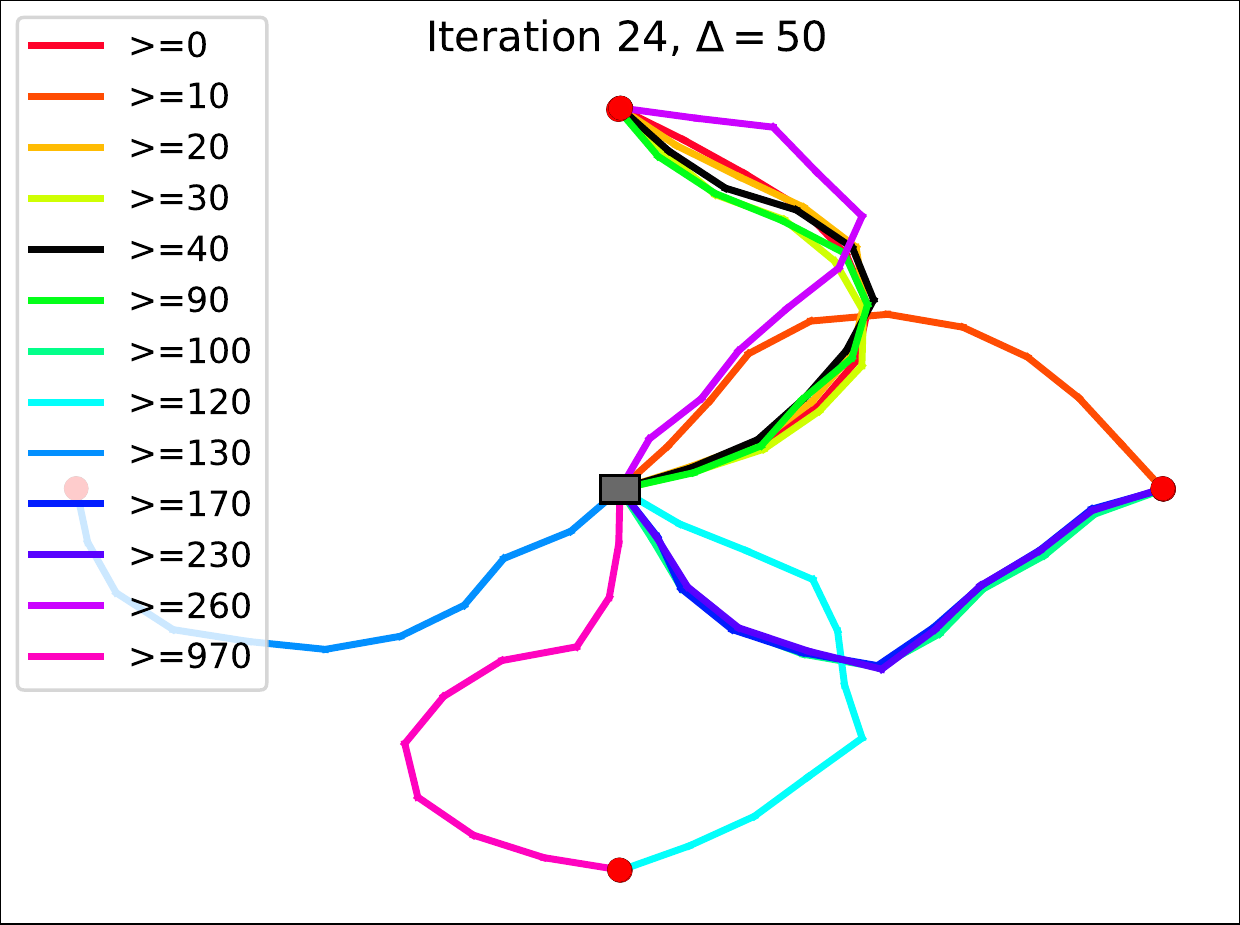}
	\end{subfigure}
	
		\begin{subfigure}{.19\textwidth}
		\centering
		\includegraphics[width=\linewidth]{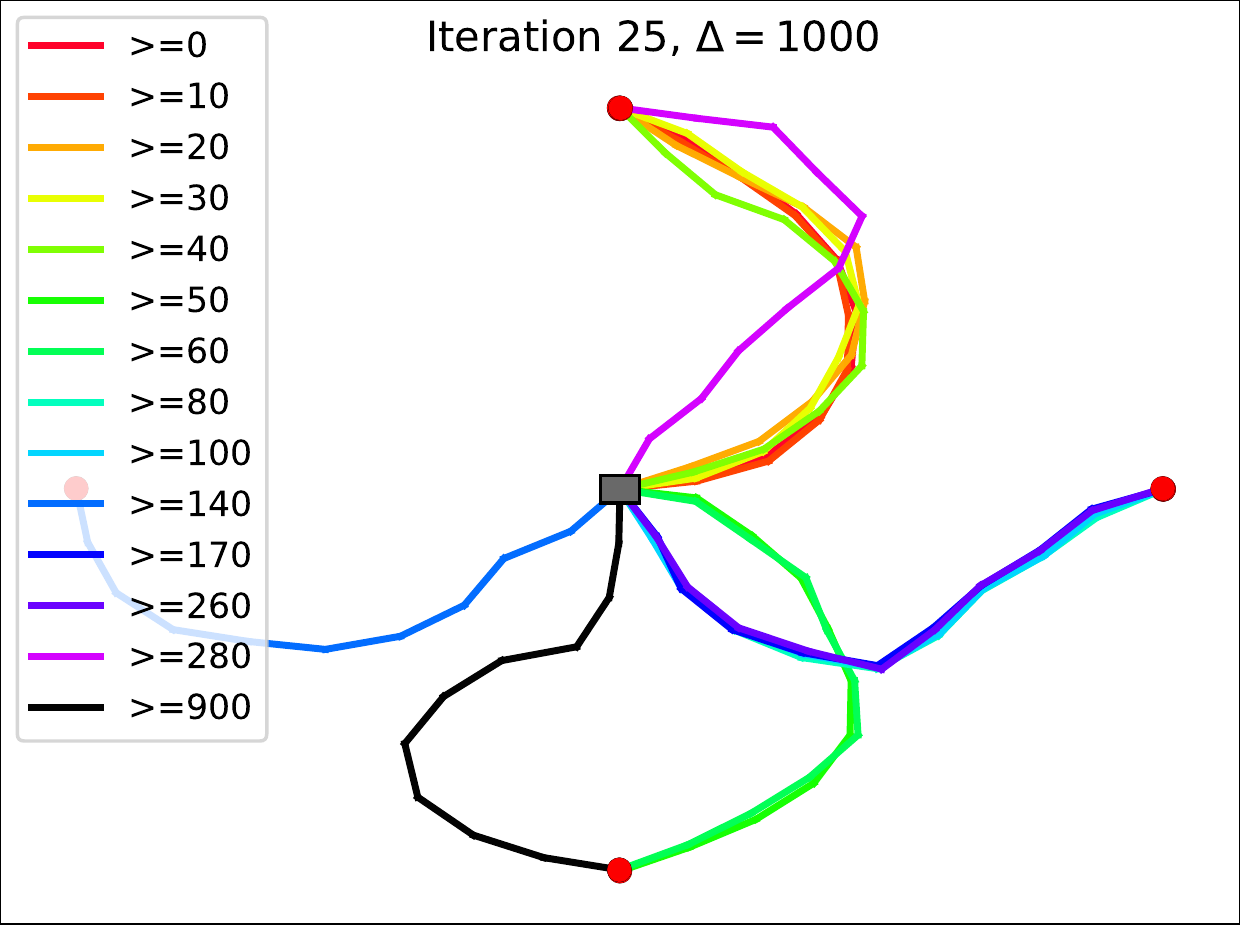}
	\end{subfigure}
	\begin{subfigure}{.19\textwidth}
		\centering
		\includegraphics[width=\linewidth]{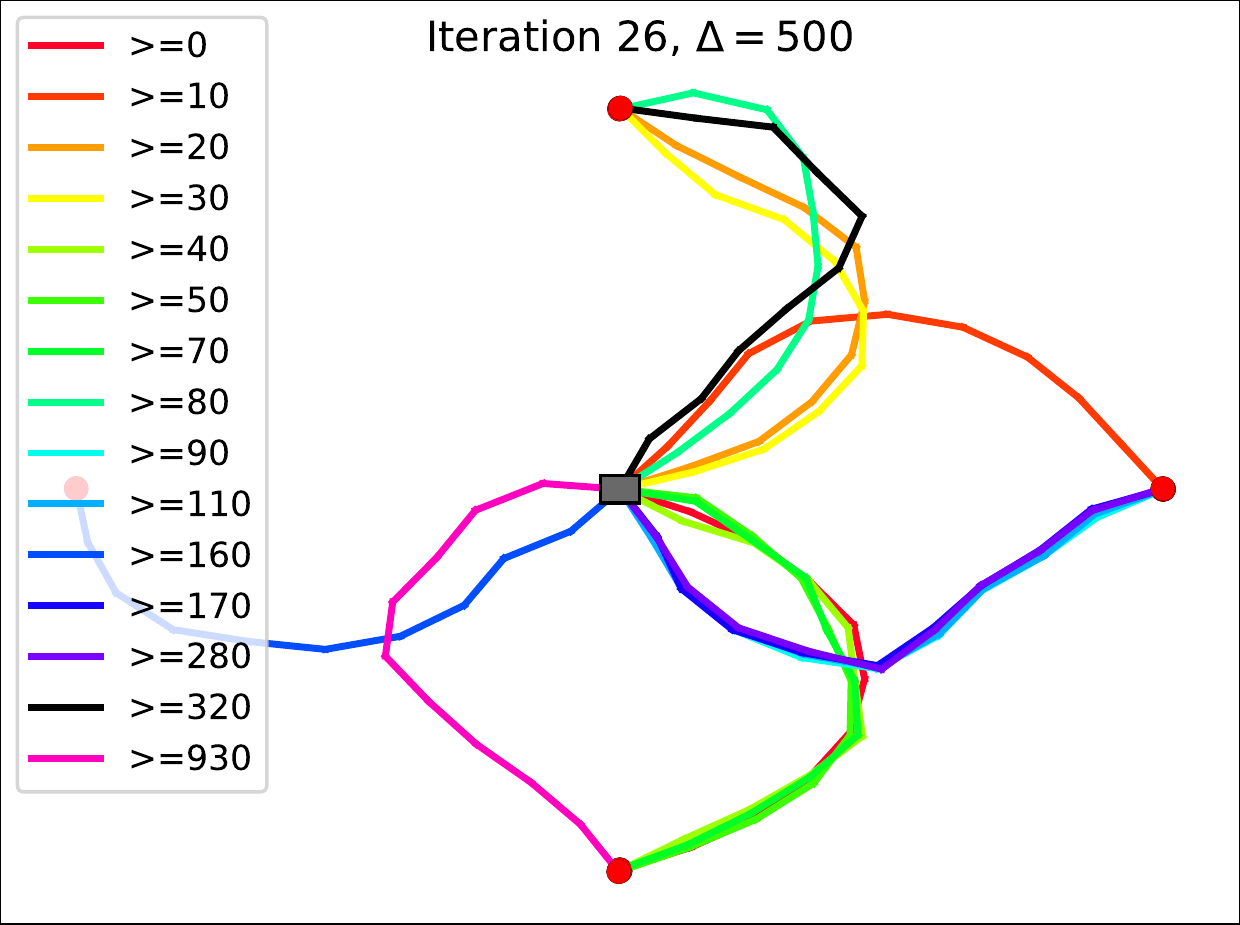}
	\end{subfigure}
	\begin{subfigure}{.19\textwidth}
		\centering
		\includegraphics[width=\linewidth]{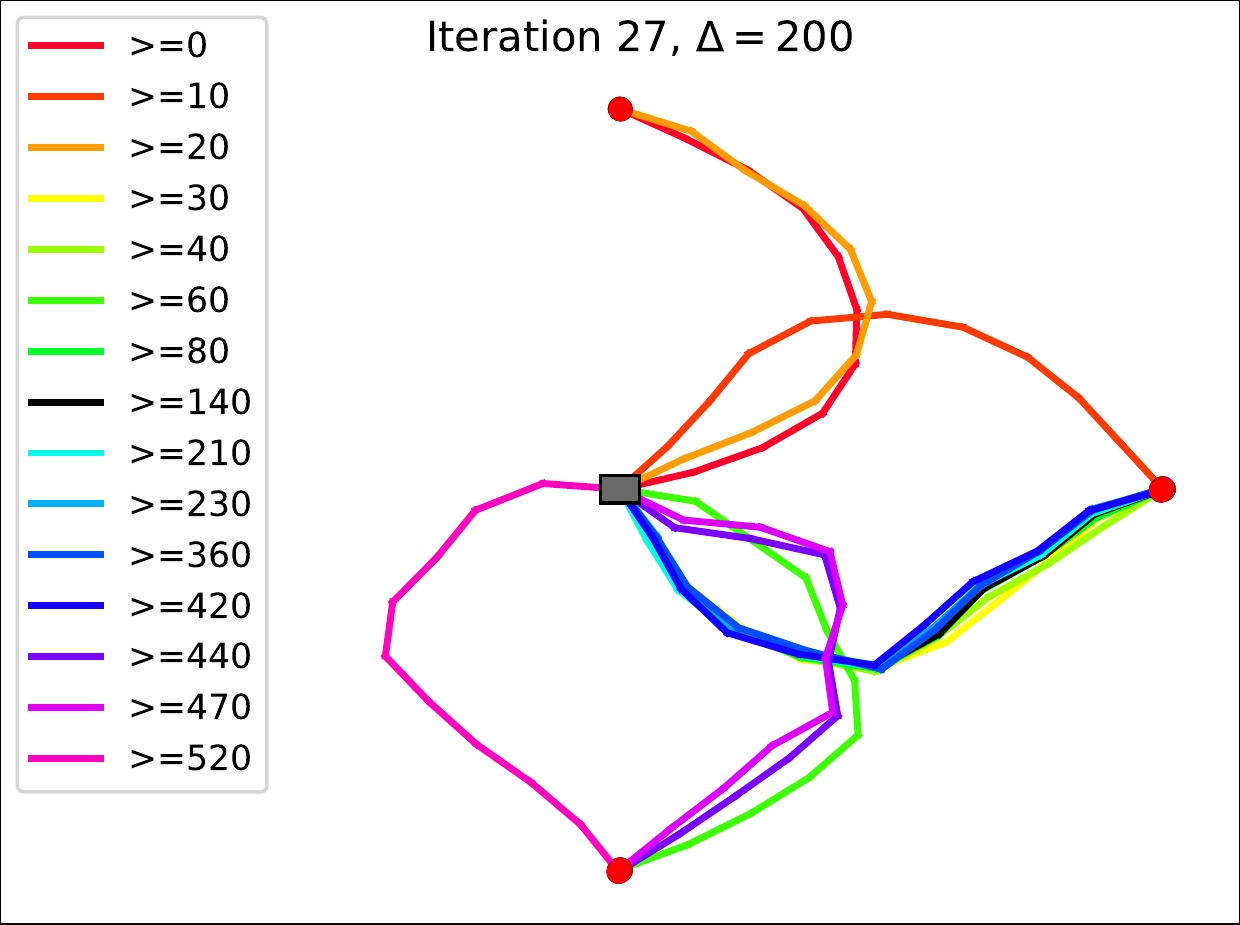}
	\end{subfigure}
	\begin{subfigure}{.19\textwidth}
		\centering
		\includegraphics[width=\linewidth]{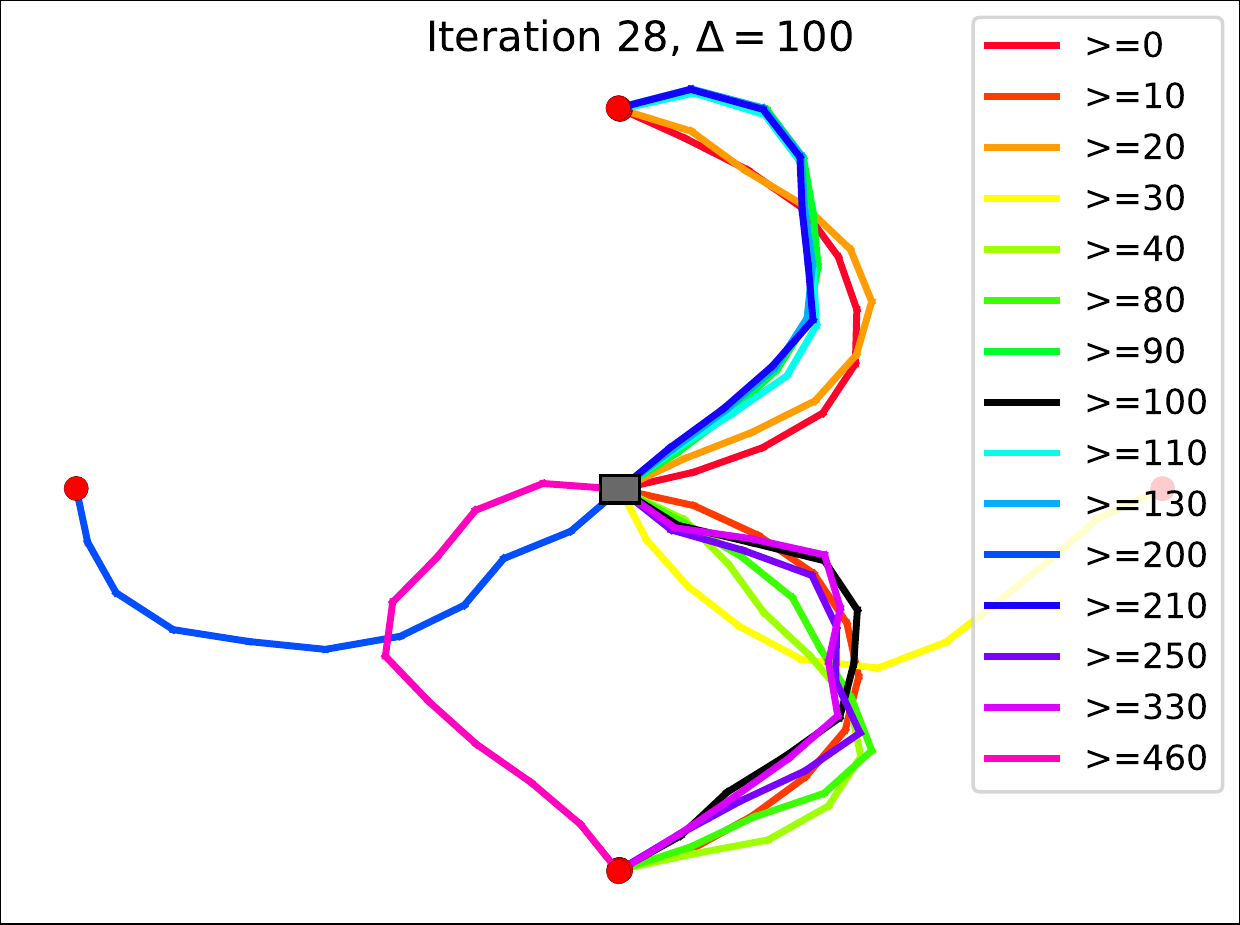}
	\end{subfigure}        
	\begin{subfigure}{.19\textwidth}%
		\centering
		\includegraphics[width=\linewidth]{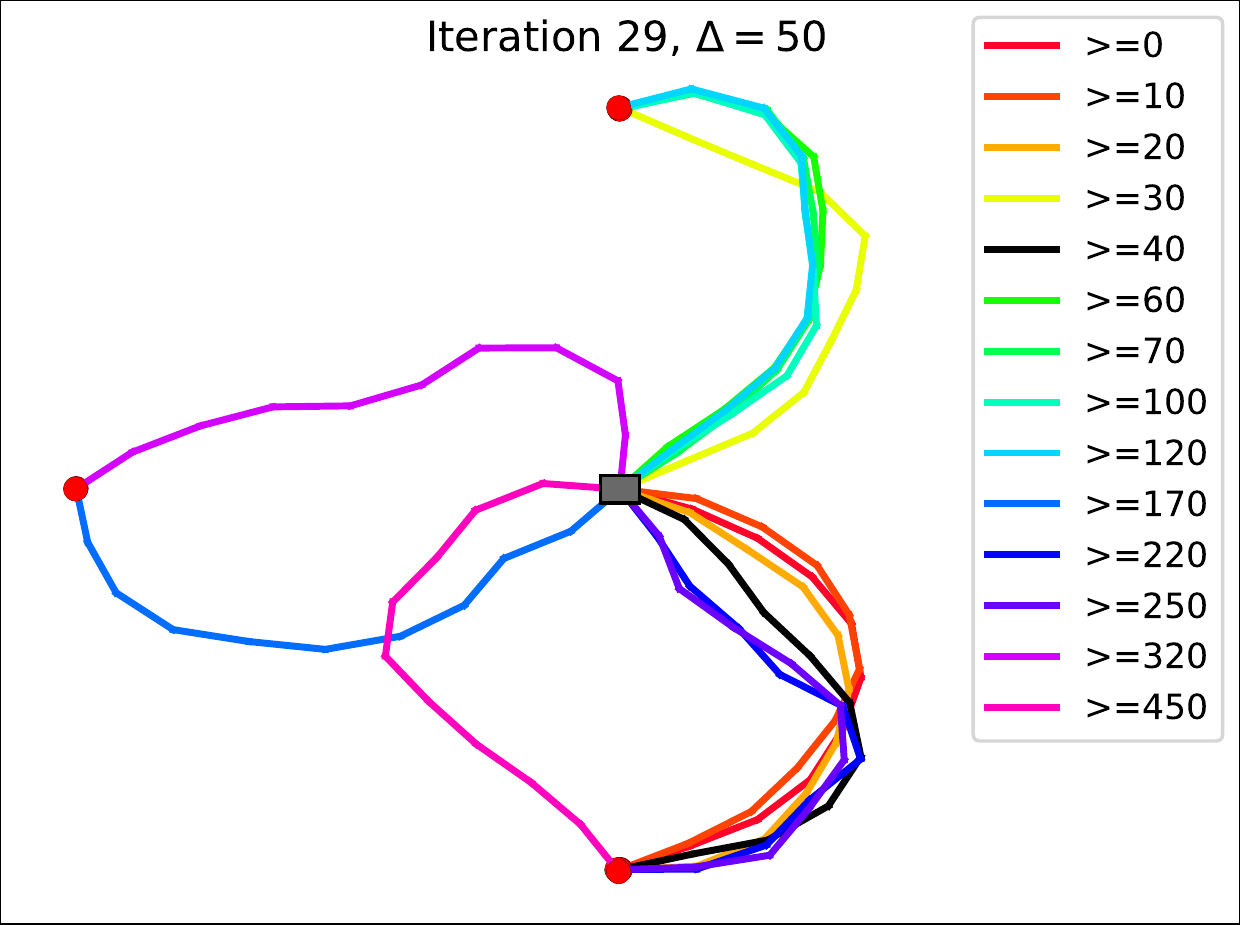}
	\end{subfigure}
	    \caption{The plots show the candidates selected by the heuristic (Equation~\ref{eq:addingHeuristicImpl}) for values of $\Delta$ from \num{0} to \num{1000} in steps of \num{10} on the \textit{planar robot} experiment with \num{4} goal positions for the first \num{30} iterations. At each iteration a new component is added based on the value shown in the title. These components are colored in black. Often the same candidate is selected for large ranges of $\Delta$. All selected candidates seem reasonable. However, although all candidates reach one of the desired goal positions, the configurations can be less smooth (resulting in low likelihood due to the prior) for large values of $\Delta$, which can be seen especially at iterations 20-23. While optimizing such components may require more iterations and samples, they are also more likely to discover a new mode. For example, the component added at iteration 20 is the first component that reaches the top goal position from the left side. 
	    }
	    \label{fig:deltaSensitivity}
\end{figure}

For varying the (negated) assumed initial weights we specify several different values in an array $\boldsymbol\Delta = [1000, 500, 200, 100, 50]$ and pick one of these values $\boldsymbol\Delta_j$ by cycling through this array. 
The adding heuristic is thus computed as 
\begin{equation}
    \label{eq:addingHeuristicImpl}
   \tilde{R}_{\mathbf{x}_s}(o_\text{n}) = R(\mathbf{x}_s) - \max \big(  \log q(\mathbf{x}_s), \max_{\mathbf{x}_i \in \mathcal{X}_{\text{total}}} \log q(\mathbf{x}_i) - \boldsymbol\Delta_j \big).
\end{equation}
Instead of pre-specifying the possible values for the assumed initial weights, it would also be possible to sample continuous values from a given distribution. However, for small changes in the assumed initial weight the heuristic would typically select qualitatively similar candidates and, thus, it is simpler to specify a few values that relate to different levels of exploration than to specify a distribution. It would also be possible to specify a single value $\Delta$, however, this would add a hyper-parameter that has to be tuned depending on the experiment. Furthermore, switching between different levels of exploration can be more efficient because we do not only want to add components close to missing modes, but also close to modes that are already covered in order to approximate them better. Figure~\ref{fig:planarDeltaAblations} shows learning curves for different values of $\Delta$ on the \textit{planar robot} experiment with four goals, which features several disconnected non-Gaussian modes. Here, varying the values performed better than any fixed assumed value for the initial weight. Figure~\ref{fig:deltaSensitivity} shows the different initial means that would have been chosen depending on the assumed initial weight. The selected candidates are sensible for a large range of $\Delta$.

\section{Scaling a Gaussian to Obtain a Desired Entropy}
\label{app:GaussianEntropy}
We want to find the scaling factor $c$ to obtain a desired entropy $\entropy_\text{init}$ for a Gaussian distribution with given covariance matrix $\boldsymbol{\Sigma}$ of order $n \times n$.
\begin{align*}
    \entropy(\boldsymbol{\Sigma};c) &= \frac{1}{2} \log | 2 \pi e c \boldsymbol{\Sigma} | \\
    &= \frac{1}{2} n \log(c) + \frac{1}{2} \log | 2 \pi e \boldsymbol{\Sigma} | \overset{!}{=} \entropy_\text{init} \Rightarrow c = \exp\big( \frac{1}{n} \left( 2 \entropy_\text{init} - \log |2 e \pi\boldsymbol{\Sigma}| \right) \big)
\end{align*}

\section{Goodwin Model}
\label{app:Goodwin}
The Goodwin model is defined as
\begin{equation}
\label{eq:Goodwin}
\begin{aligned}
\frac{dx_1}{dt} &= \frac{a_1}{1+a_2x_g^\rho} - \alpha x_1 \\
\frac{dx_2}{dt} &= k_1 x_1 - \alpha x_2 \\
\vdots\\
\frac{dx_g}{dt} &= k_{g-1} x_{g-1} - \alpha x_g,
\end{aligned}
\end{equation}
where $x_1$ represents the concentration of mRNA for a target gene, $x_2$ represents the corresponding protein product of the gene, and $x_3$ to $x_g$ are intermediate protein species that ultimately lead to a negative feedback, via $x_g$, on the rate at which mRNA is transcribed. We consider $g=9$ intermediate species and assume that the parameters $\rho=10$ and $\alpha=0.53$ are known. 
We put a Gamma prior with shape $2$ and rate $1$ on the remaining 10 parameters $a_1$, $a_2$ and $\kappa_1 \dots \kappa_{8}$ that need to be inferred. We use the prior also to randomly choose their true values.
For an initial condition $\mathbf{x}_0 = \mathbf{0}$, we create 81 noisy observations $\mathbf{o}_{1\dots81}$ of $x_1$ and $x_2$ using steps of $dt=1$. We assume Gaussian observation noise with zero mean and variance $\sigma^2=0.2$ and discard the first $40$ observations.
The posterior distribution is given by 
\begin{equation}
p(a_1,a_2,\kappa_1,\dots,\kappa_{8}|\mathbf{o}_{40\dots81}) = \frac{1}{Z} p(a_1) p(a_2) \prod_{i=1}^{8} p(\kappa_i) \prod_{t=40}^{81}p_t(\mathbf{o}_t|a_1,a_2,\kappa_1,\dots,\kappa_{8}),
\end{equation}
where $p_t(\mathbf{o}_t|a_1,a_2,\kappa_1,\dots,\kappa_{8})$ is a Gaussian distribution with variance $\sigma^2=0.2$ and a mean which is computed by numerically integrating the ODE~(Equation~\ref{eq:Goodwin}).

\section{Planar Robot Experiment}
\label{app:planarRobot}
The x and y coordinate of the end-effector are given by
    \begin{align*}
    x(\boldsymbol{\theta}) = \sum_{i=1}^{10} \cos\left( \sum_{j=1}^i \theta_j \right), && y(\boldsymbol{\theta}) = \sum_{i=1}^{10} \sin\left( \sum_{j=1}^i \theta_j \right). 
	 \end{align*}
     The target distribution is given as the product of two distributions,
     \begin{align*}
     p(\boldsymbol{\theta}) = \frac{1}{Z} p_\text{conf}(\mathbf{\boldsymbol{\theta}}) p_\text{cart}(\mathbf{\boldsymbol{\theta}}),
     \end{align*}
     where $p_\text{conf}(\mathbf{\boldsymbol{\theta}})$ enforces smooth configurations and $p_\text{cart}(\boldsymbol{\theta})$ penalizes deviations from the goal position. We model $p_\text{conf}(\mathbf{\boldsymbol{\theta}})$ as zero mean Gaussian distribution with diagonal covariance matrix, where the angle of the first joint has a variance of \num{1} and the remaining joints have a variance of \num{4e-2}. We consider two experiments that differ in the choice of goal positions. For the first experiment we specify a single goal position at position $(7,0)$ modeled by a Gaussian distribution in Cartesian space with variance \num{1e-4} in both directions, namely
     \begin{equation*}
     p_{\text{cart},1}(\boldsymbol{\theta}) = \mathcal{N} \left( \begin{bmatrix}x(\boldsymbol{\theta}) \\y(\boldsymbol{\theta}) \end{bmatrix} | \begin{bmatrix} 7 \\ 0 \end{bmatrix},  \begin{bmatrix}  \num{1e-4} && 0 \\0 && \num{1e-4} \end{bmatrix} \right).
     \end{equation*}
For the second experiment we specify four goal positions at positions $(7,0)$, $(0,7)$, $(-7,0)$ and $(0, -7)$. The likelihood $p_{\text{cart},2}$ is given by the maximum over the four respective Gaussian distributions. 

\section{Number of Components}
\label{app:numComponents}
\begin{figure}
    \centering
    \includegraphics[width=0.5\linewidth]{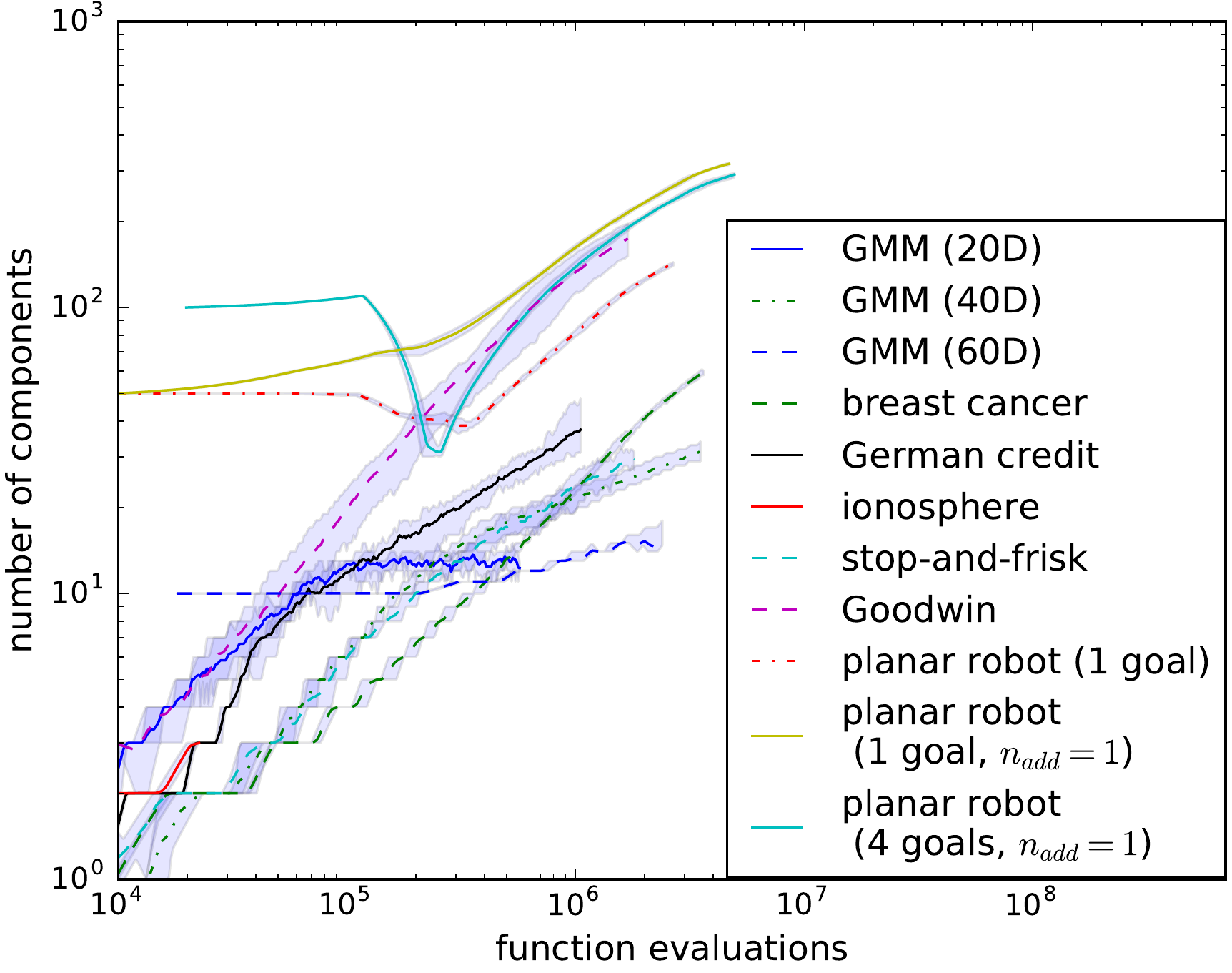}
	\caption{The average number of components learned by {\sc{VIPS++}} is plotted over function evaluations for all experiments in the test bed. When using the faster adding rate, $n_\text{add}=1$, {\sc{VIPS++}} learns GMMs with approximately \num{350} components.}
	\label{fig:numberOfComponents}
\end{figure}

The average number of components learned by {\sc{VIPS++}} is shown in Figure~\ref{fig:numberOfComponents}.

\section{Implementations}
\label{app:implementations}
For our comparisons we relied on open-source implementations, preferably by the original authors. 
\begin{itemize}
\item For {\sc{PTMCMC}}, we use an implementation by \citet{ellis2017} that uses adaptive proposal distributions for the individual chains. We roughly tuned the number of chains for each experiment. As we could not run this implementation on our cluster, we ran the experiments on a fast quad-core laptop and made use of multi-threading. We therefore report four times the actual wall-clock time. 

\item For {\sc{ESS}}, we use a Python implementation by \citet{BovyESS} that is based on the Matlab implementation by Iain Murray. If the target distribution decomposes into a product of a Gaussian prior and an arbitrary likelihood term, we directly provide this decomposition to the algorithm. If the target distribution does not use a Gaussian prior, we choose an appropriate Gaussian distribution $p_\text{prior}(\mathbf{x}) = \mathcal{N}(\mathbf{x}|\mathbf{0}, \alpha \mathbf{I})$ as prior and provide it along with the resulting likelihood $\log p_\text{likelihood}(\mathbf{x}) = \log \targetDistUnnormalized - \log p_\text{prior}(\mathbf{x})$, as described by~\citet{Nishihara2014}.

\item Our comparisons with {\sc{HMC}} are based on {\sc{pyhmc}}~\citep{PyHMC}. We tuned the step size and trajectory length for each experiment based on preliminary experiments. We also performed some experiments with {\sc{NUTS}}~\citep{Hoffman2014}, however, {\sc{HMC}} with tuned parameters always outperformed the automatically tuned parameters of {\sc{NUTS}}.

\item For slice sampling, we use a Python adaptation~\citep{SliceBlog} of a Matlab implementation by Iain Murray and tuned the step size based on preliminary experiments.

\item For {\sc{SVGD}}, we use the implementation of the original authors~\citep{Liu2016} and tune the step size based on preliminary experiments.

\item For Variational Boosting, we use the implementation of the original authors~\citep{Miller2017}. However, this implementation is not optimized with respect to the number of function evaluations and often uses an unnecessary large number of samples. We therefore modified the implementation slightly. We also use their implementation of {\sc{NPVI}} for our experiments.

\item For black-box variational inference and inverse autoregressive flows we used our own implementation based on tensorflow~\citep{tensorflow}. The code for conducting these experiments is available online\footnote{The implementation can be found at \url{https://github.com/OlegArenz/tensorflow\_VI}.}. For black-box variational inference, we tuned the learning rate as well as the number of samples per iteration (batch size). For inverse autoregressive  flows, we tuned the learning rate, the batch size, the number of flows and the (common) width of the two hidden layers of the autoregressive networks for each flow.
\end{itemize}

\section{Considered Algorithms and Experiments}
\label{app:consideredExperiments}
Table~\ref{table:algoExperiments} provides an overview about which algorithms have been evaluated on which experiments. 
\begin{itemize}
    \item Our implementations of IAF and BBVI use a different code base (based on Tensorflow \citep{tensorflow}) for which we only implemented a subset of the experiments. However, we ensured that the test bed includes simple, unimodal experiments (\textit{German credit} and \textit{breast cancer}) as well as the most challenging, multimodal experiments that we considered (\textit{planar robot} and \textit{GMM}).
    \item We did not evaluate PTMCMC on the simple test problems where parallel Markov chains would be wasteful.
    \item We did not evaluate HMC on the experiments with disconnected modes because we do not expect it to mix efficiently on such problems.
    \item We tried to evaluate VBOOST and NPVI on all test problems. However, we could not always obtain reliable results due to numerical problems that we could not fix without major changes to the implementation.
    \item We only evaluated {\sc{VIPS++}} on the higher-dimensional GMM experiments because it was the only method to solve the twenty-dimensional variant. 
\end{itemize}

\begin{table}
\begin{center}
\begin{small}
\begin{sc}
\begin{tabular}{lllllllllll}
\toprule
                 &   &   &   &   &   & V & P &   &   &   \\
                 &   &   &   &   &   & B & T & S &   &   \\
                 & V & S &   &   & N & O & M & L &   & B \\
                 & I & V & E & H & P & O & C & I & I & B \\
                 & P & G & S & M & V & S & M & C & A & V \\
                 & S & D & S & C & I & T & C & E & F & I \\
\midrule
German Credit    & X & X & X & X & X & X & - & \textbf{X} & \textbf{X} & \textbf{X} \\
Breast Cancer    & X & X & X & \textbf{X} & X & - & - & \textbf{X} & \textbf{X} & \textbf{X} \\
Frisk            & X & X & \textbf{X}  & X & X & X & - & \textbf{X} & - & - \\
GMM              & X & X & X & - & - & - & X & \textbf{X} & \textbf{X} & \textbf{X} \\
Planar (1 goal)  & X & X & X & - & X & X & X & X & \textbf{X} & \textbf{X} \\
Ionosphere       & \textbf{X}  & \textbf{X} & \textbf{X} & \textbf{X} & - & - & - & \textbf{X} & - & - \\
Goodwin          & \textbf{X}  & \textbf{X} & \textbf{X} & \textbf{X} & - & - & \textbf{X} & \textbf{X} & - & - \\
Planar (4 goal)  & \textbf{X}  & \textbf{X} & \textbf{X} & - & - & - & \textbf{X} & \textbf{X} & \textbf{X} & \textbf{X} \\
GMM (Higher Dim.)  & \textbf{X}  & - & - & - & - & - & - & - & - & - \\
\bottomrule
\end{tabular}
\end{sc}
\end{small}
\end{center}
\caption{The table shows which algorithms were applied to each test problem. New experiment compared to our previous work~\citep{Arenz2018} are marked in bold.}
\label{table:algoExperiments}
\end{table}

\section{Alternatives for Learning Gaussian Variational Approximations}
\label{app:GVAs}
\begin{figure}
	\centering
			\begin{subfigure}{.44\textwidth}
		\centering
		\includegraphics[width=\linewidth]{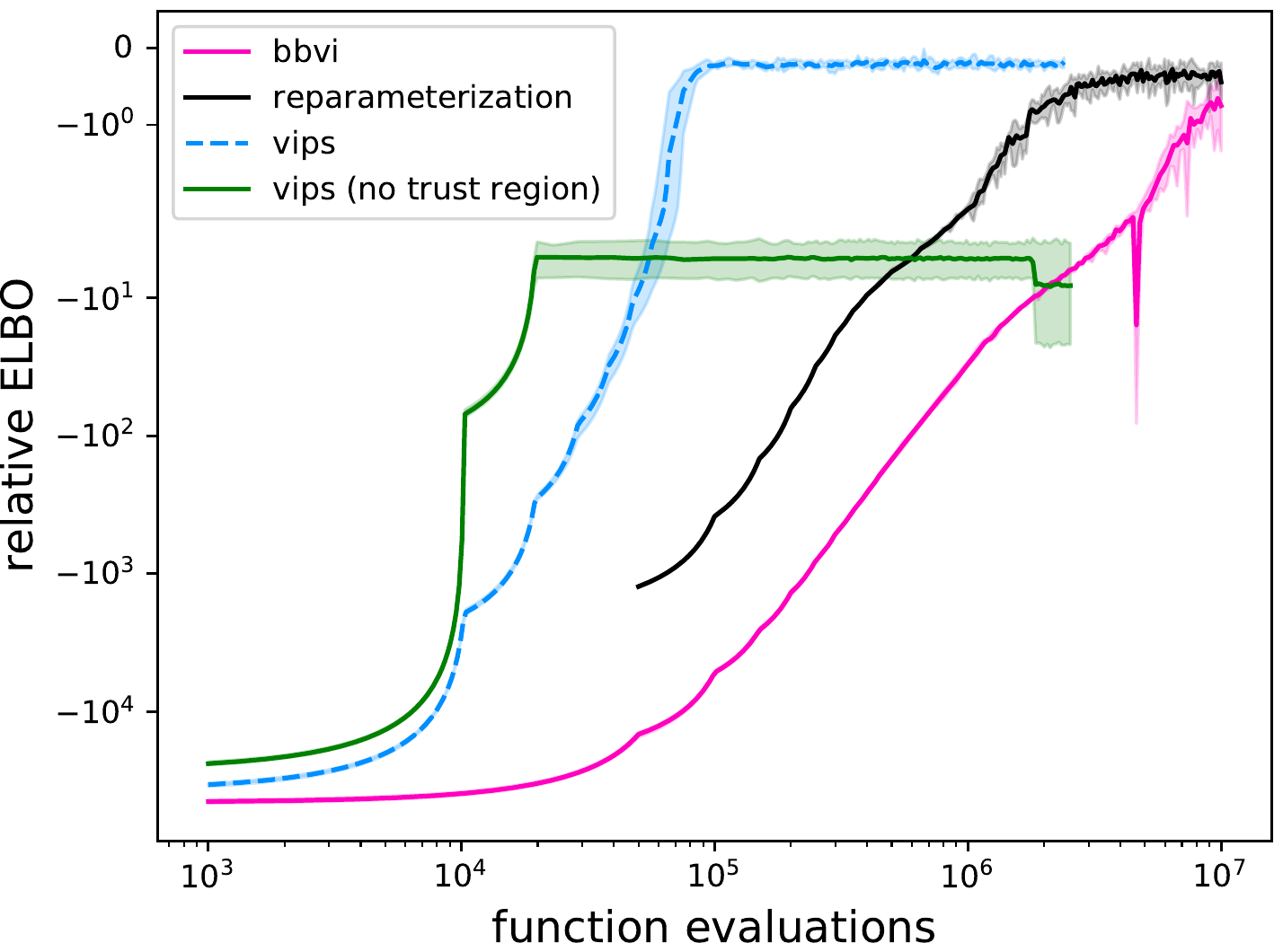}
		\caption{breast cancer}
\end{subfigure}
			\begin{subfigure}{.44\textwidth}
		\centering
		\includegraphics[width=\linewidth]{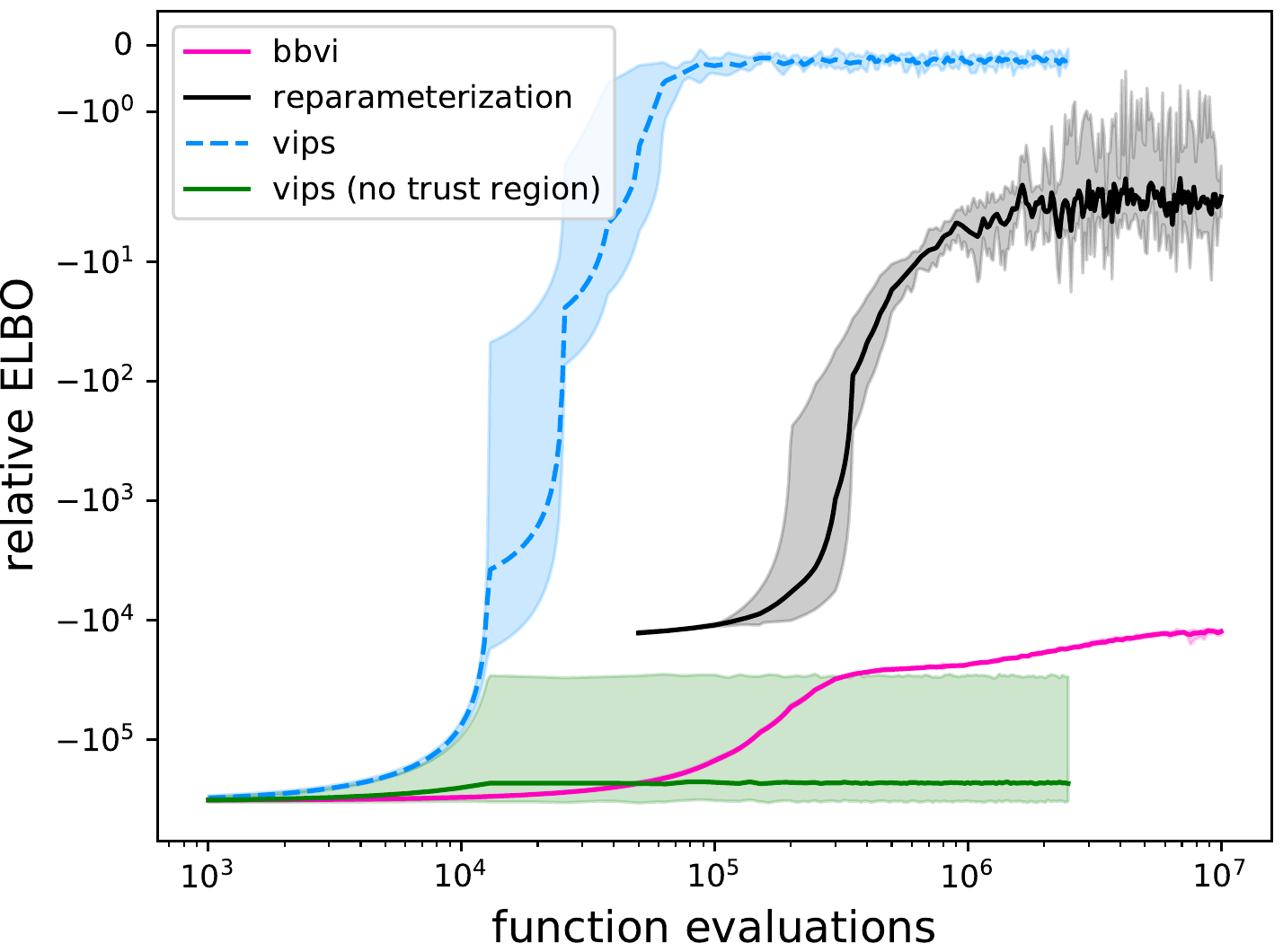}
		\caption{planar robot (1 goal)}
\end{subfigure}
\caption{The proposed variant of {\sc{MORE}} is significantly more efficient at optimizing Gaussian variational approximations compared to stochastic gradients using the reparameterization trick or black-box variational inference. By using locale surrogate objectives,  we require trust regions to ensure stable optimization.}
\label{fig:ELBOsGVA}
\end{figure}

{\sc{VIPS++}} uses a variant of {\sc{MORE}} (which we denote as {\sc{VIPS1}}) for learning Gaussian variational approximations. However, it would also be possible to update the individual components using black-box variational inference~\citep{Ranganath2014} or the reparameterization trick, which assumes that the target distribution is differentiable. We compared against these alternatives on \textit{breast cancer} experiment as well as on the \textit{planar robot} experiment with a single goal position. The learning curves of the ELBO are shown in Figure~\ref{fig:ELBOsGVA}.  For each experiment, we subtracted a constant offset from the ELBO such that the highest (approximated) ELBO on each plot equals zero. Such relative ELBO ensures high resolution in the vicinity of the best ELBO on each of the plots. Please note that we use the symmetric logarithm to scale the y-Axis. Remarkably, {\sc{VIPS1}} is significantly more efficient than the reparameterization trick even though we do not require the gradient of the target distribution. We also compared against a variant of {\sc{VIPS1}} that does not constrain the KL divergence between updates. Such optimization is unstable as it exploits model errors caused by the local surrogate.

\section{VIPS++ Hyper-Parameters}
\label{app:hyperparameters}
The hyper-parameters used for all experiments are given in Table~\ref{tab:hyperparameters}.
\begin{table}
\begin{center}
\begin{small}
\begin{sc}
\begin{tabular}{lcc}
\toprule
description & value \\
\midrule
KL bound for components & $ \num{1e-2} \le \epsilon(o) \le \num{5}$ \\
number of desired samples (per dimension, per component)& $20$ \\
number of reused samples (per dimension, per component) & $40$  \\
adding rate for components & $\num{30}$ or $\num{1}$  \\
deletion rate for components & $\num{10}$  \\
minimum weight & \num{1e-6}  \\
initial weight & \num{1e-29} \\
$\boldsymbol{\Delta}$ for adding-heuristic & [\num{1000}, \num{500}, \num{200}, \num{100}, \num{50}] \\ 
$\ell_2$-regularization for WLS & $ \num{1e-14} \le \kappa \le \num{1e-6} $ \\ 
\bottomrule
\end{tabular}
\end{sc}
\end{small}
\end{center}
\caption{The table shows the hyper-parameters of {\sc{VIPS++}} as well as their values used during the experiments. The bound on the KL-divergence and the coefficient for $\ell_2$-regularization when fitting the surrogates are automatically adapted within in the provided ranges.}
\label{tab:hyperparameters}
\end{table}
    
\section{Computing the Maximum Mean Discrepancy}
\label{app:MMD}
We approximate the MMD between two sample sets $\mathbf{X}$ and $\mathbf{Y}$ as
\begin{align*}
\textrm{MMD}(\mathbf{X},\mathbf{Y}) &= \frac{1}{m^2} \sum_{i,j}^m k(\mathbf{x}_i, \mathbf{x}_j) + \frac{1}{n^2} \sum_{i,j}^n k(\mathbf{y}_i, \mathbf{y}_j) \\ & - \frac{2}{mn} \sum_i^m \sum_j^n k(\mathbf{x}_i, \mathbf{y}_i).
\end{align*}
We use a squared exponential kernel given by
\begin{align*}
k(\mathbf{x},\mathbf{y}) = \exp\left(- \frac{1 }{\alpha}(\mathbf{x} - \mathbf{y})^\top \boldsymbol{\Sigma} (\mathbf{x} - \mathbf{y}) \right),
\end{align*}
where $\boldsymbol{\Sigma}$ is a diagonal matrix where each entry is set to the median of squared distances within the ground-truth set and the bandwidth $\alpha$ is chosen depending on the problem.
As true ground-truth samples are only available for the GMM experiment, we apply generalized elliptical slice sampling \citep{Nishihara2014} with large values for burn-in, thinning and chain lengths to produce baseline samples that are regarded as ground-truth for the remaining experiments. Note that obtaining these ground-truth samples is computationally very expensive, taking up to two days of computation time on 128 CPU cores. We estimate the MMD based on ten thousand ground-truth samples and two thousand samples from the given sampling method. For {\sc{MCMC}} methods, we choose the two thousand most promising samples by applying a sufficient amount of burn-in and using the largest thinning that keeps at least two thousand samples in the set.

\section{Evaluations with Respect to Computational Time}
\label{app:computationalTime}
\begin{figure}
	\centering
	\begin{subfigure}{.328\textwidth}
		\centering
		\includegraphics[width=\linewidth]{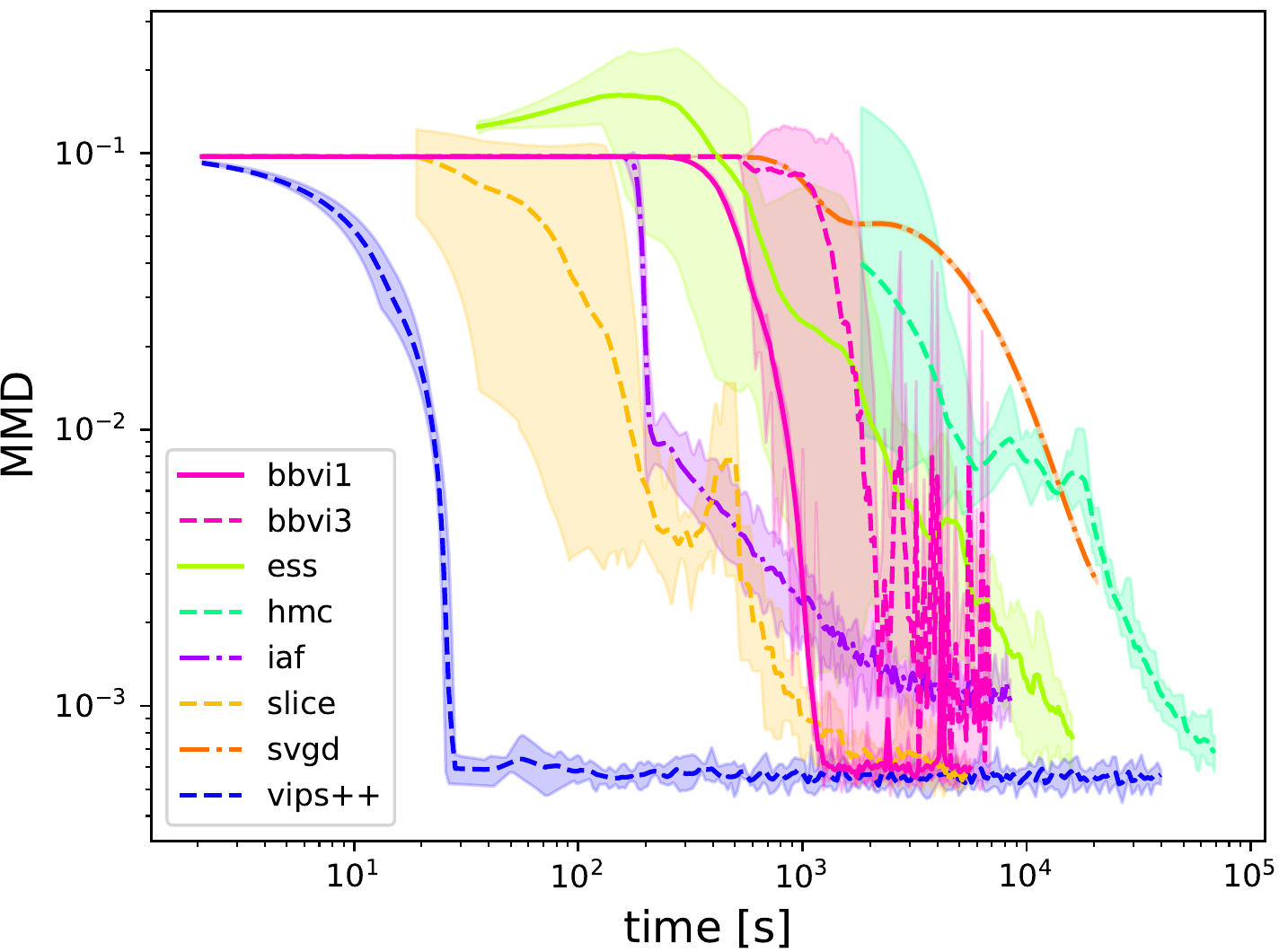}
		\caption*{German credit (Log. Reg.)}
	\end{subfigure}
	\begin{subfigure}{.328\textwidth}
		\centering
		\includegraphics[width=\linewidth]{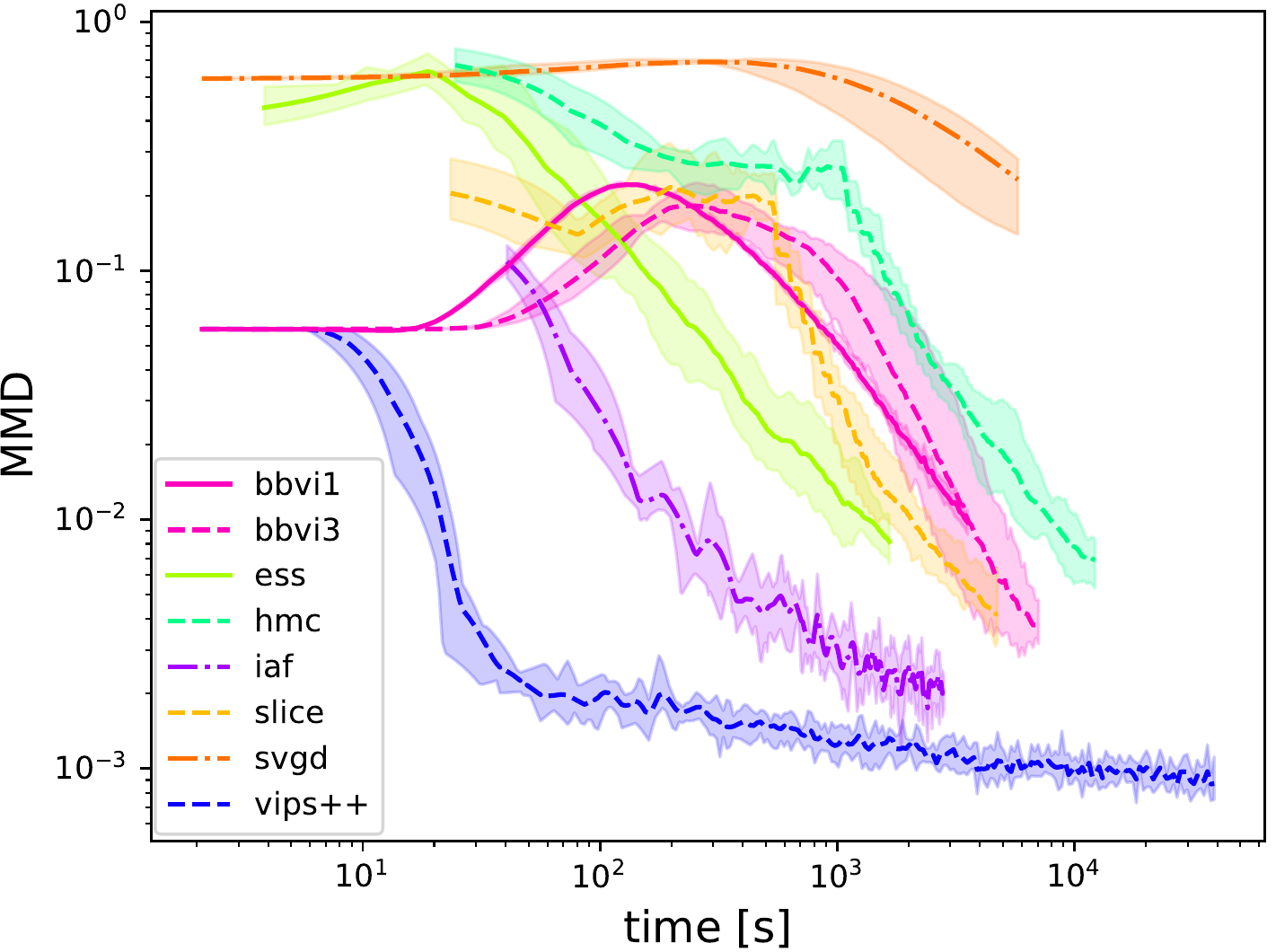}
		\caption*{breast cancer (Log. Reg.)}
	\end{subfigure}
	\begin{subfigure}{.328\textwidth}
		\centering
		\includegraphics[width=\linewidth]{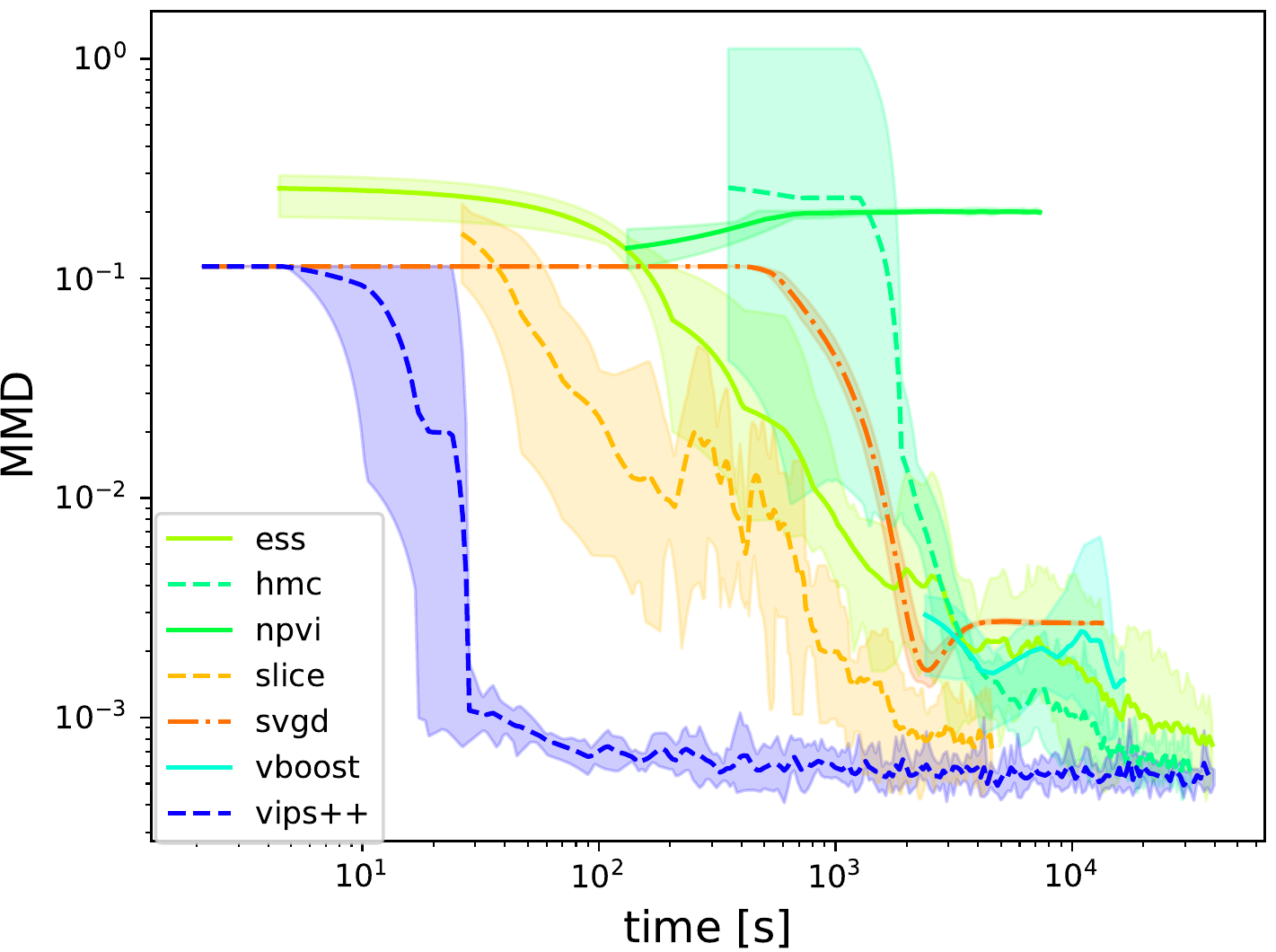}
		\caption*{stop-and-frisk (Poisson GLM)}
	\end{subfigure}
	\begin{subfigure}{.328\textwidth}
		\centering
		\includegraphics[width=\linewidth]{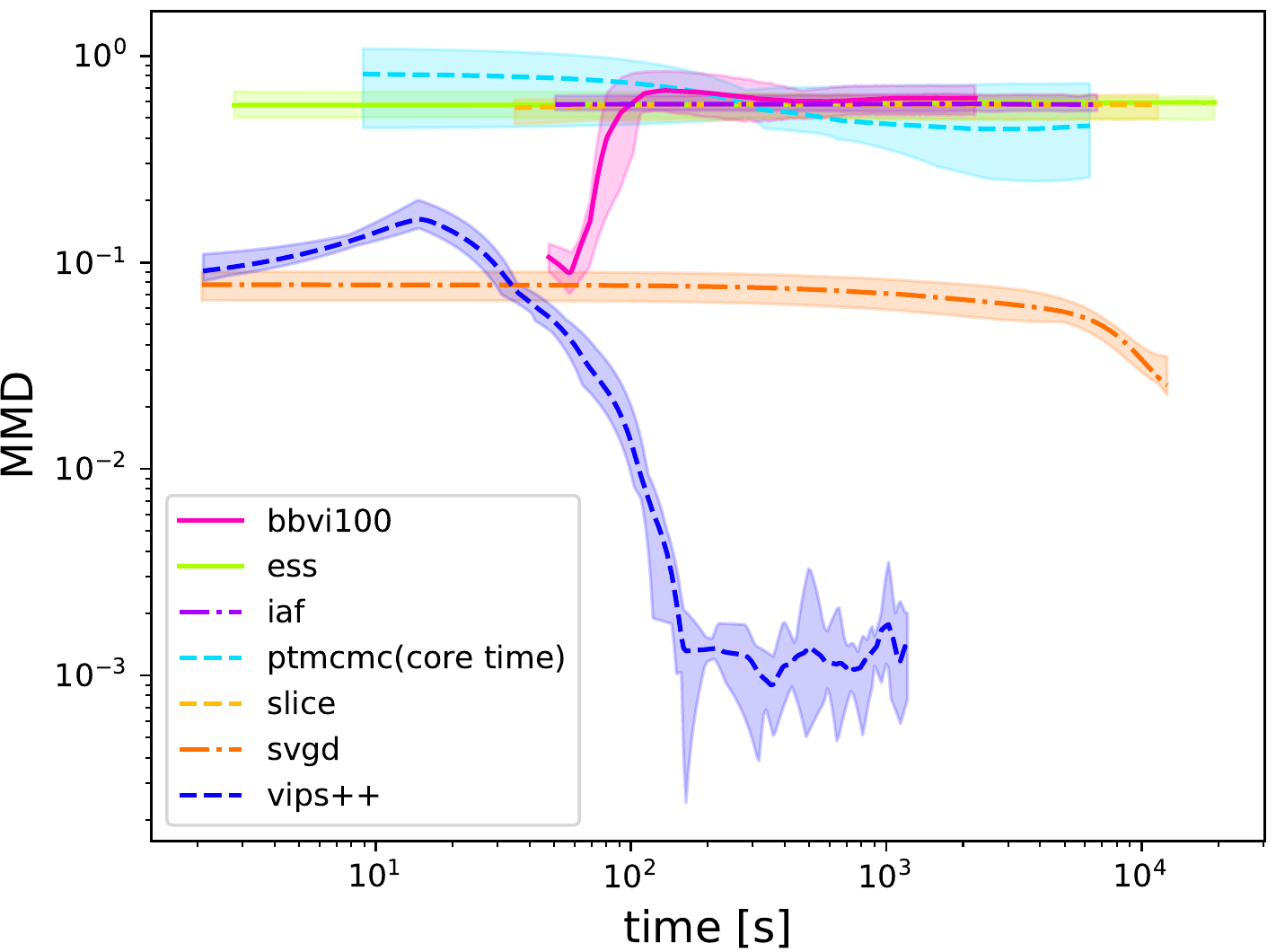}
		\caption*{GMM (20 dimensions)}
	\end{subfigure}        
	\begin{subfigure}{.328\textwidth}%
		\centering
		\includegraphics[width=\linewidth]{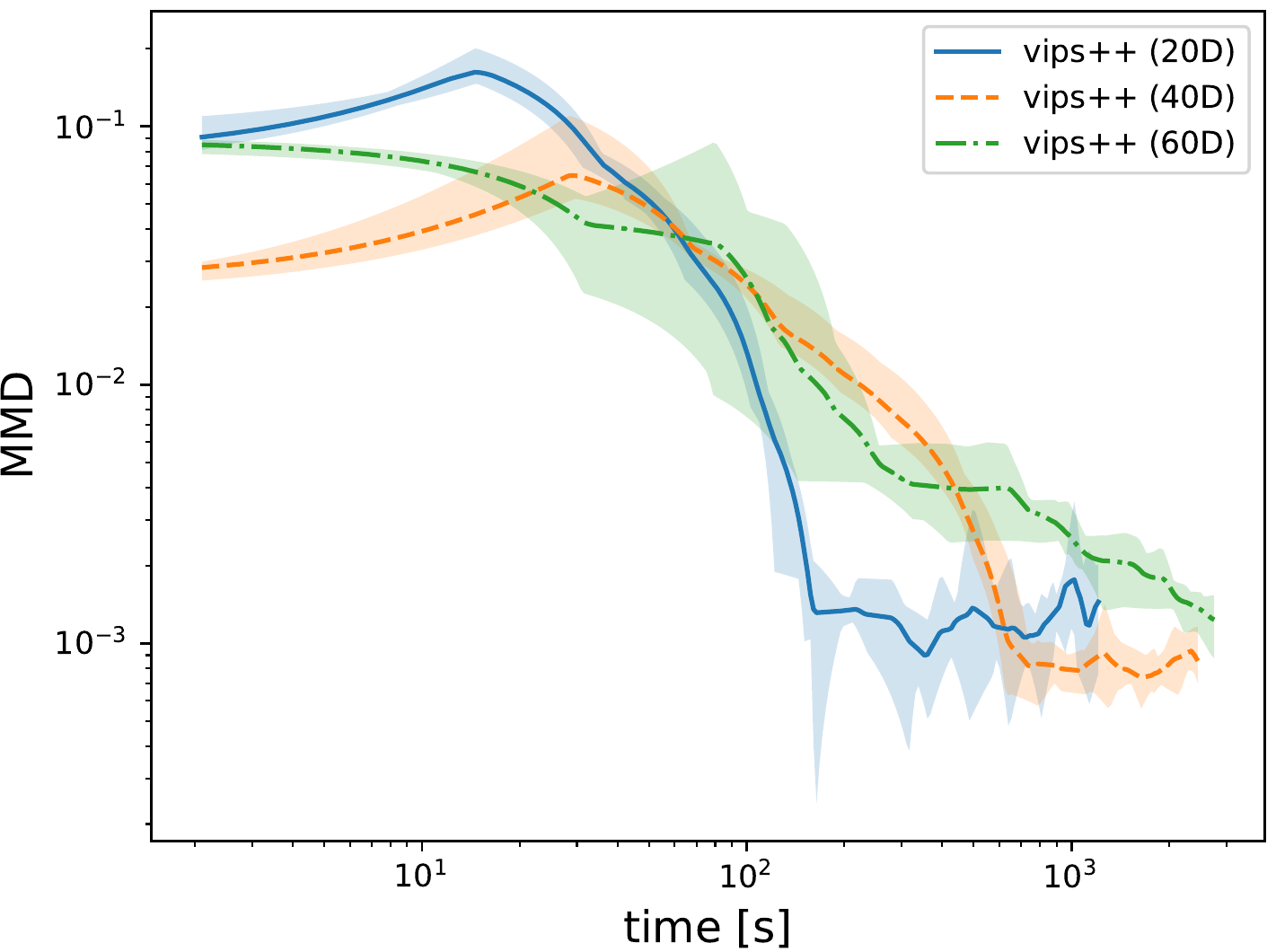}
		\caption*{GMM (different dimensions)}
	\end{subfigure}
	\begin{subfigure}{.328\textwidth}
		\centering
		\includegraphics[width=\linewidth]{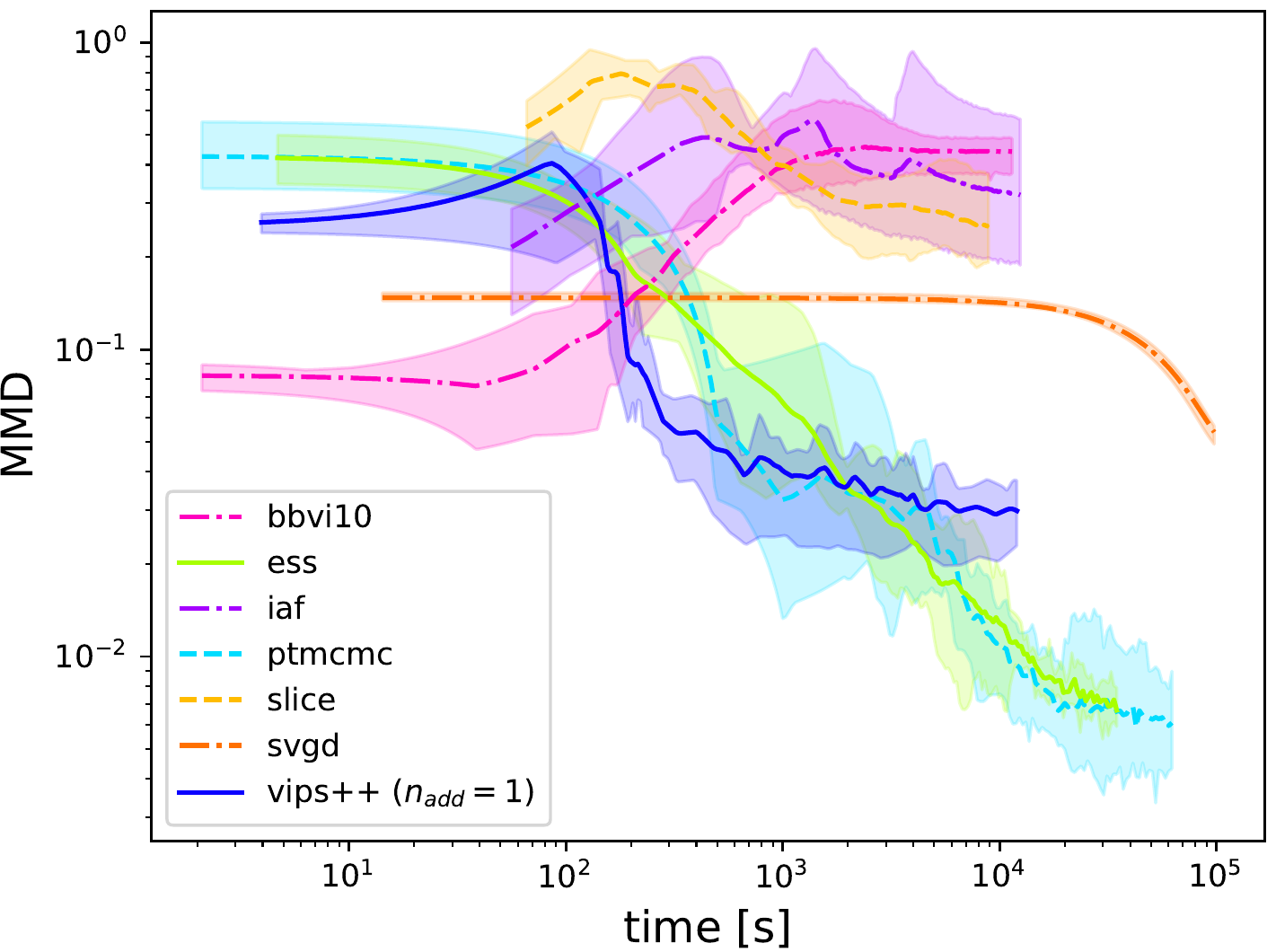}
		\caption*{planar robot (4 goals)}
	\end{subfigure}
	
	\caption{The maximum mean discrepancy with respect to baseline samples is plotted over computational time on log-log plots for the different sampling problems in the test bed.}
	\label{fig:quantitativeResults_time_app}
\end{figure}

Figure~\ref{fig:quantitativeResults_time_app} shows the achieved MMDs with respect to time for the experiments that have been omitted in the main document.

\section{Evaluations with respect to ELBO}
\label{app:elbos}
We also compared the achieved ELBO $L(\theta)$ between VIPS++, inverse autoregressive flows (IAF) and black-box variational inference (BBVI). We approximate the ELBO based on 2000 samples from the learned approximation. The respective learning curves are shown in Figure~\ref{fig:elbos} where we subtracted a constant offset as described in Appendix~\ref{app:GVAs}.

\begin{figure}
	\centering
	\begin{subfigure}{.328\textwidth}
		\centering
		\includegraphics[width=\linewidth]{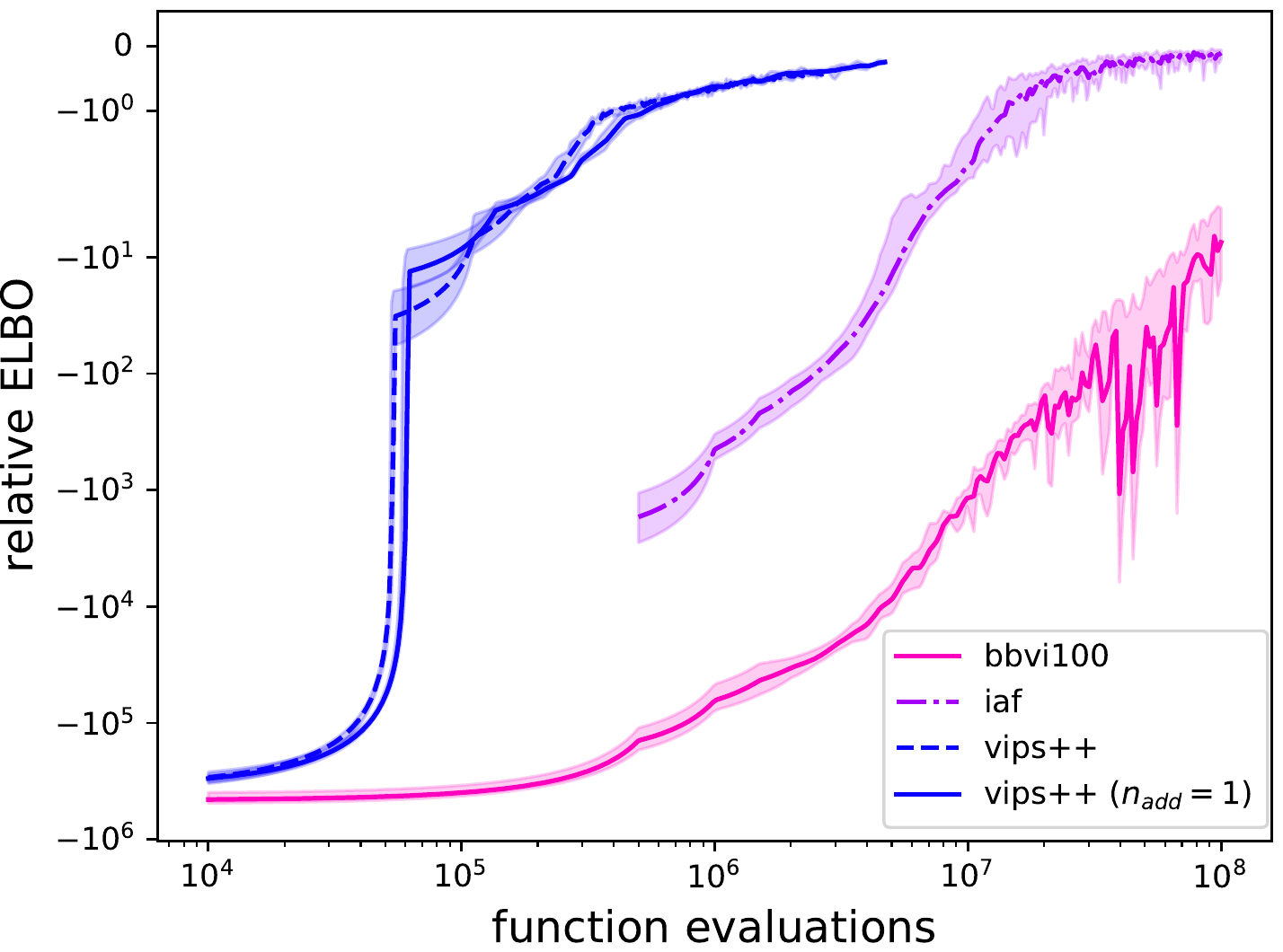}
		\caption*{planar robot (1 goal)}
	\end{subfigure}
	\begin{subfigure}{.328\textwidth}
		\centering
		\includegraphics[width=\linewidth]{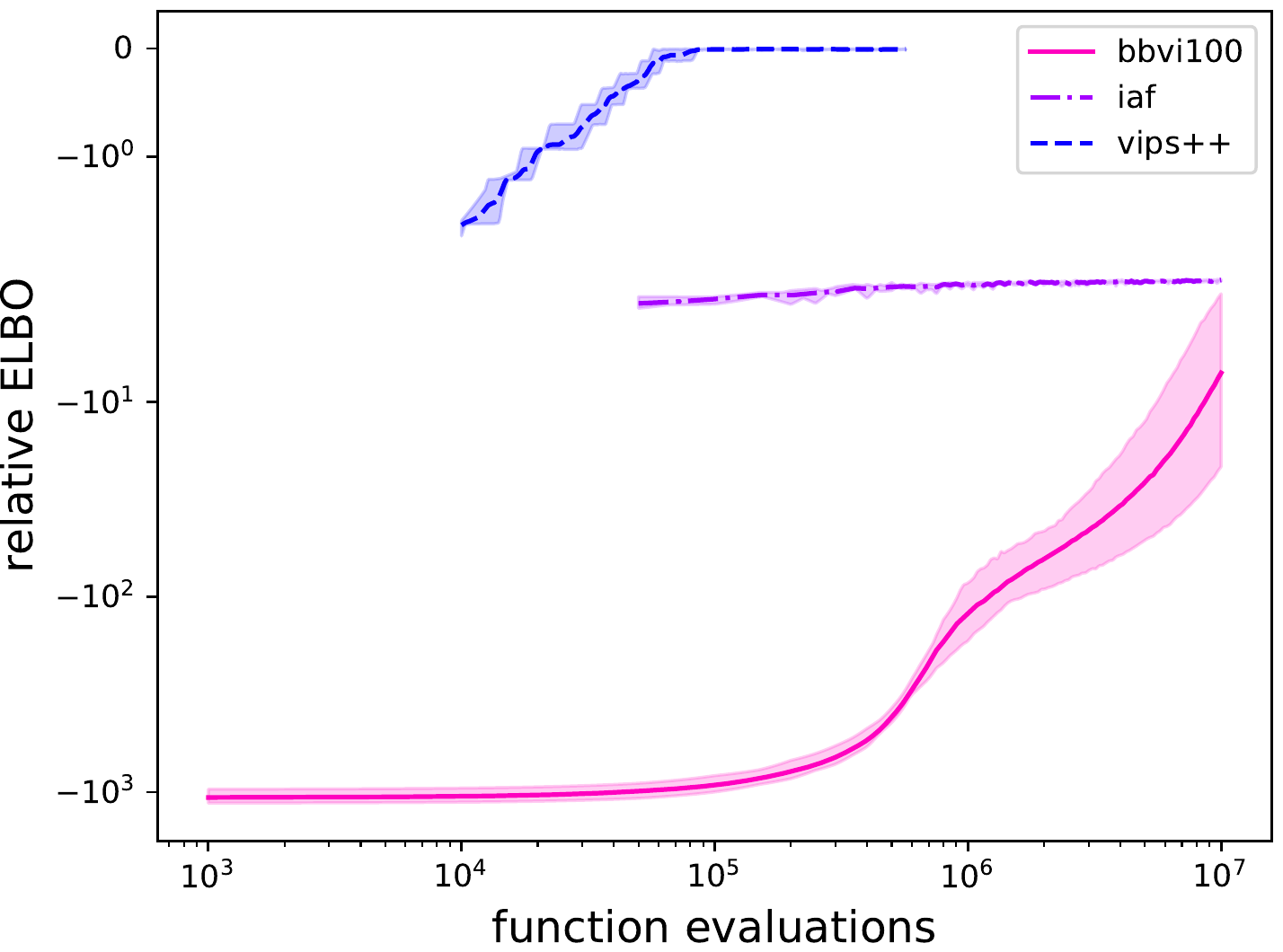}
		\caption*{GMM (20 dimensions)}
	\end{subfigure}
	\begin{subfigure}{.328\textwidth}
		\centering
		\includegraphics[width=\linewidth]{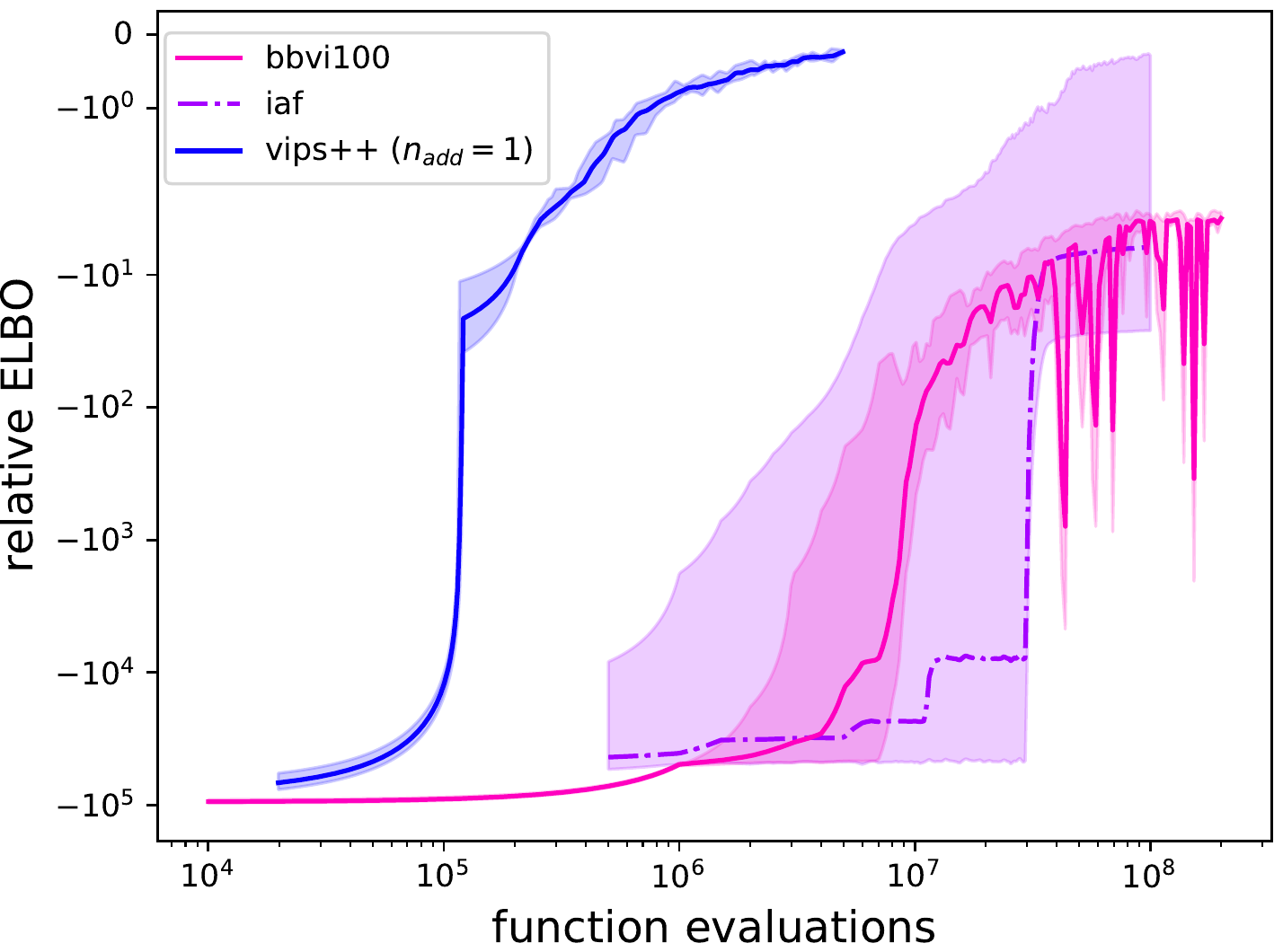}
		\caption*{planar robot (4 goals))}
	\end{subfigure}
	\begin{subfigure}{.328\textwidth}
		\centering
		\includegraphics[width=\linewidth]{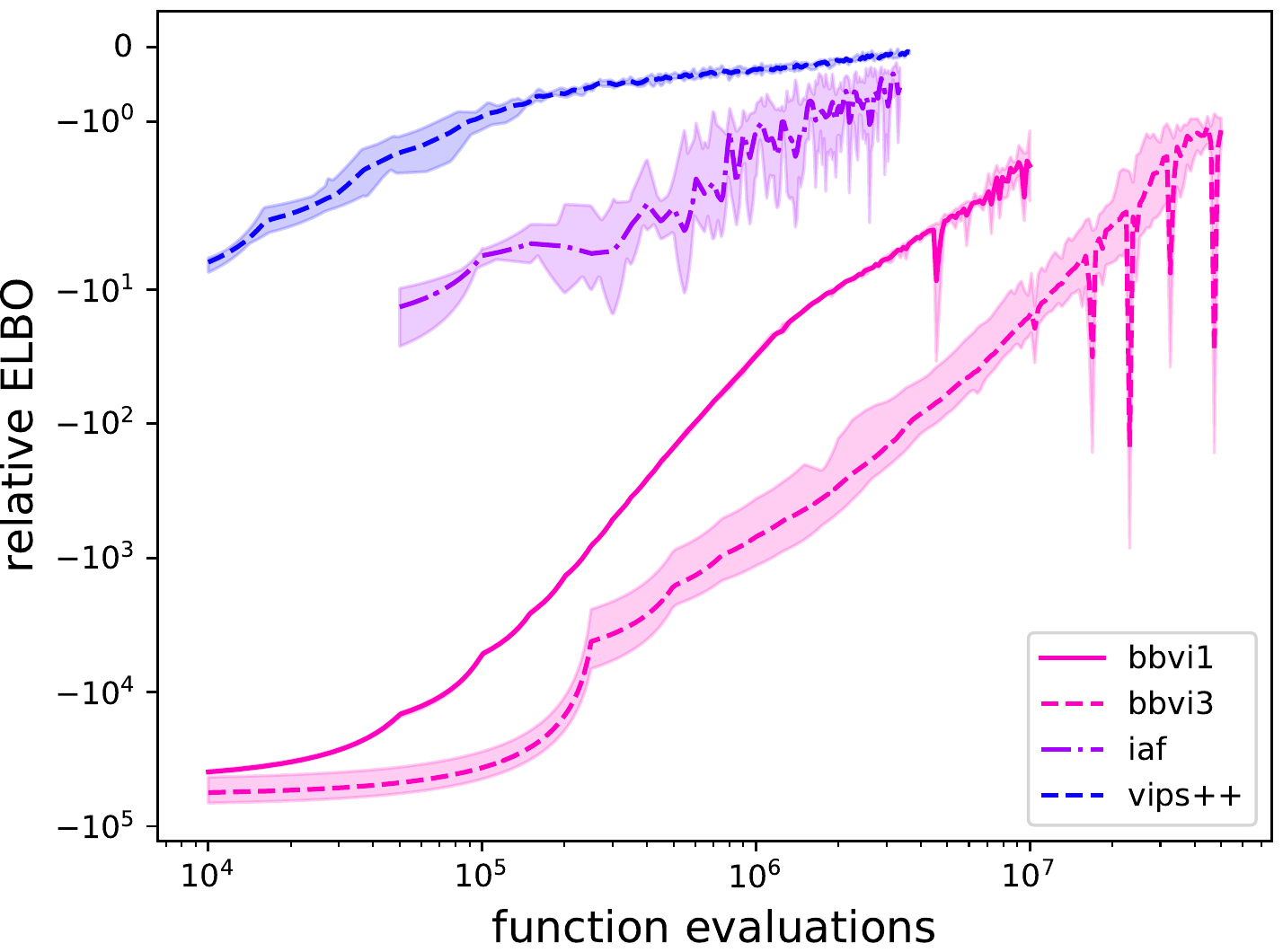}
		\caption*{breast cancer}
	\end{subfigure}        
	\begin{subfigure}{.328\textwidth}%
		\centering
		\includegraphics[width=\linewidth]{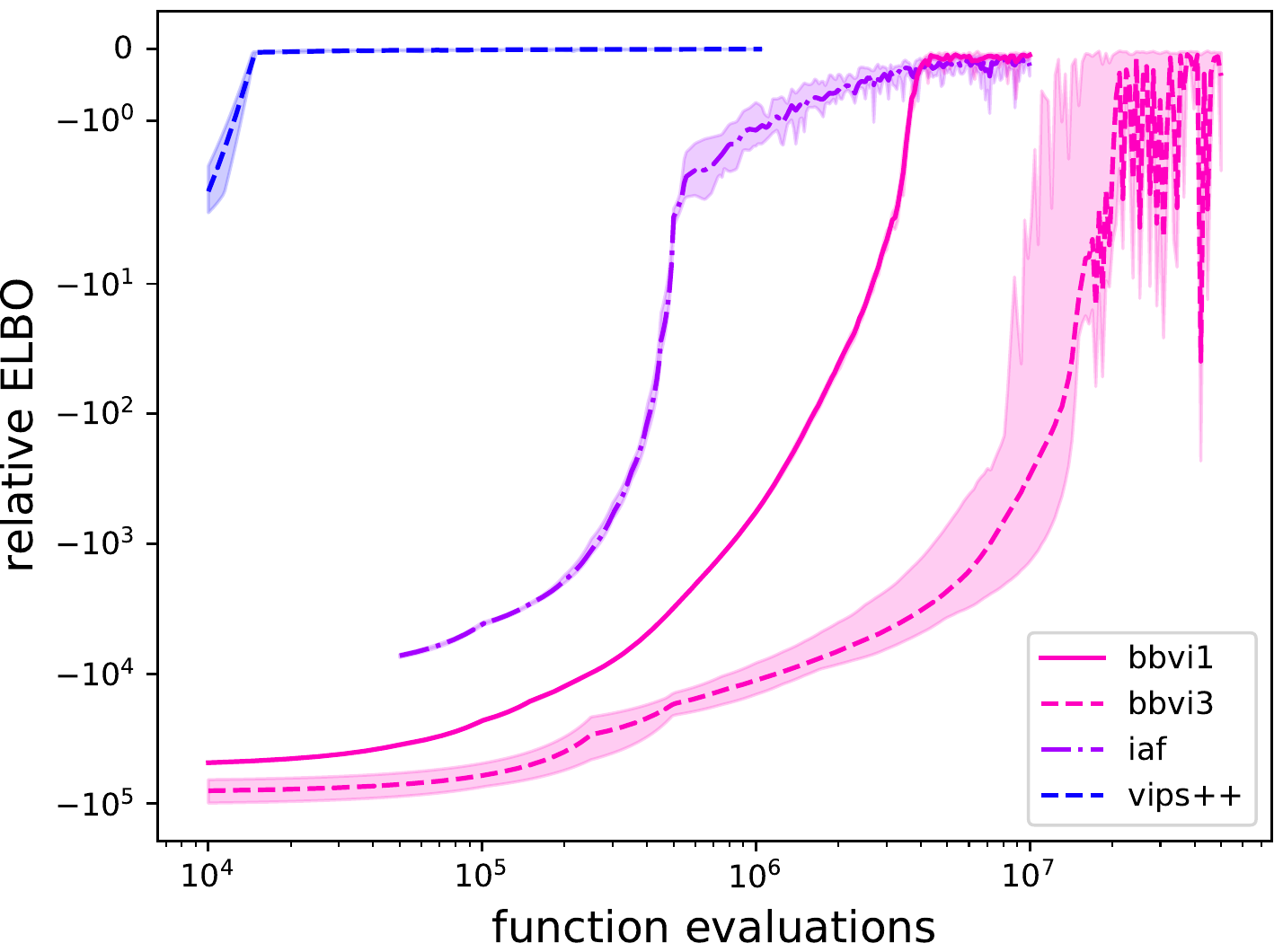}
		\caption*{German credit}
	\end{subfigure}
	    \caption{In contrast to the evaluation with respect to the MMD, all methods improve on the ELBO during learning, which is expected as the respective optimization problems aim to maximize the ELBO. Interestingly, IAF achieves a similar ELBO on the simpler planar robot experiment as VIPS++, although it performed significantly worse on the MMD. We verified that IAF achieves a similar approximated entropy as VIPS++, which is surprising since the learned approximation only sampled from one of the two main configurations (see Figure~\ref{fig:10link_samples}). We hypothesize that even the large GMMs learned by VIPS++ are not able to cover the modes as well as the normalizing flows.}
	    \label{fig:elbos}
\end{figure}

\section{Visualization of Samples for planar robot experiments}
\label{app:planarSamples}
\begin{figure}
	\centering
		\begin{subfigure}{.24\textwidth}
		\centering
		\includegraphics[width=\linewidth]{\detokenize{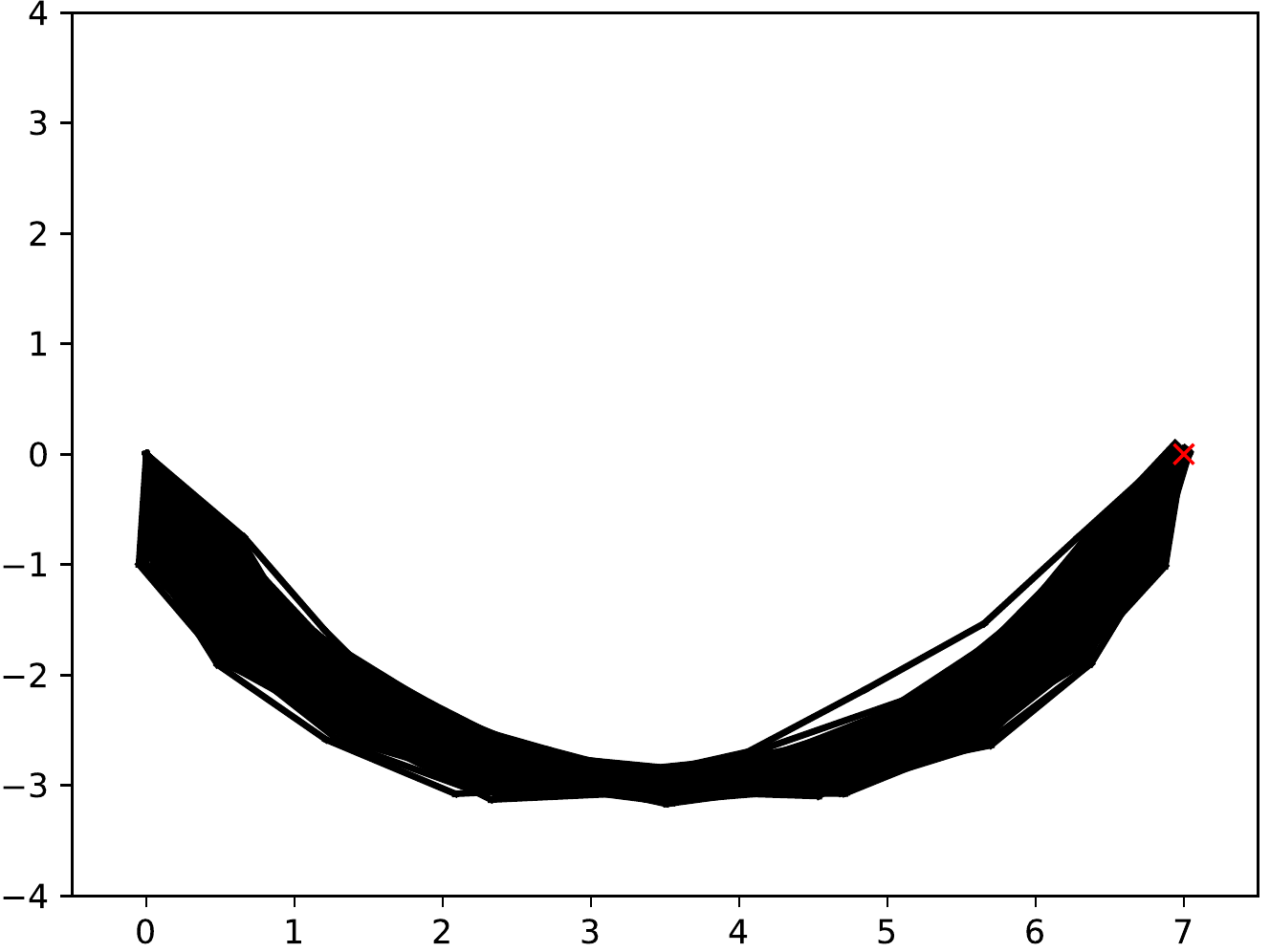}}
		\caption*{BBVI}
	\end{subfigure}
		\begin{subfigure}{.24\textwidth}
		\centering
		\includegraphics[width=\linewidth]{\detokenize{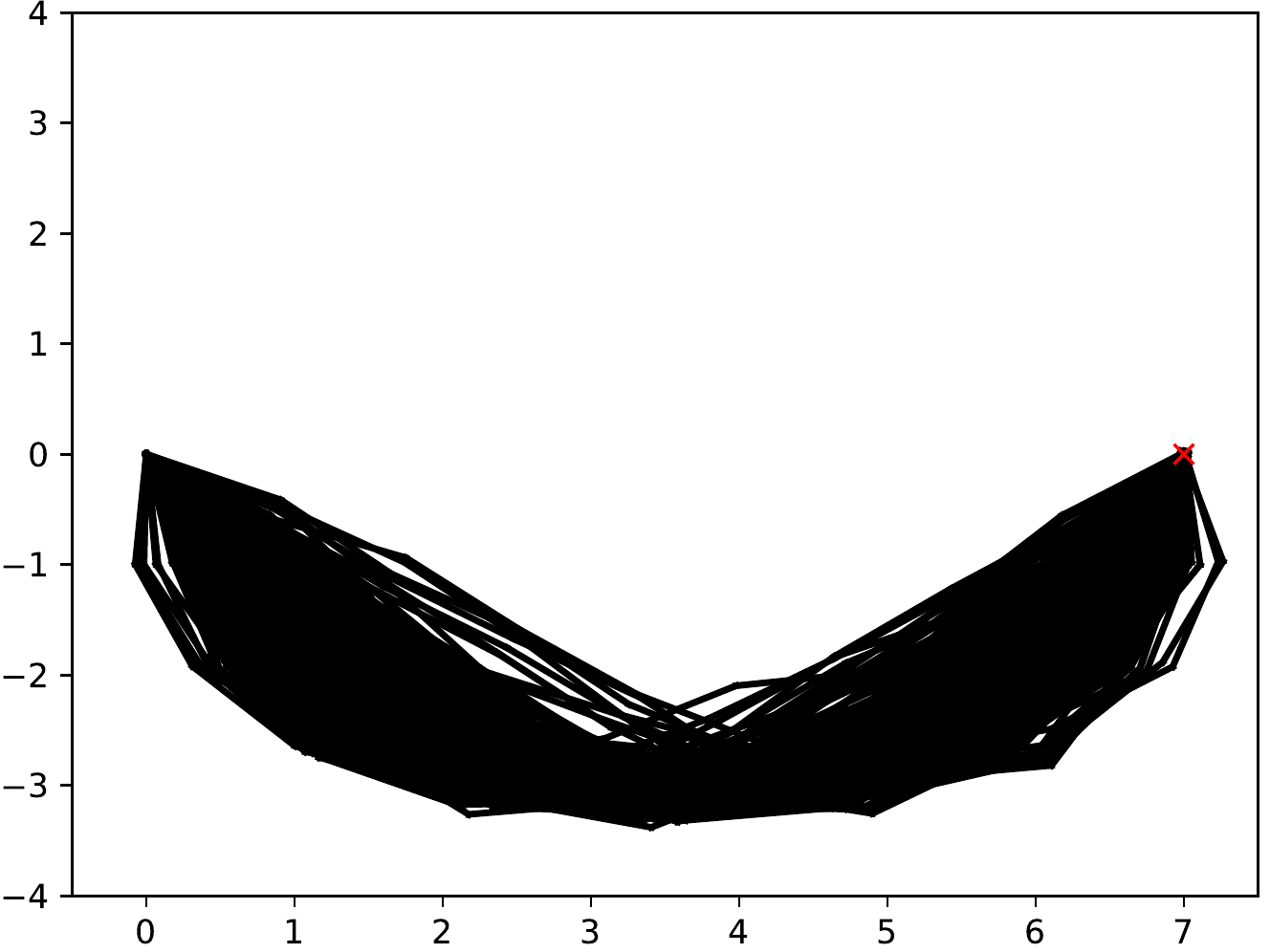}}
		\caption*{IAF}
	\end{subfigure}
		\begin{subfigure}{.24\textwidth}
		\centering
		\includegraphics[width=\linewidth]{\detokenize{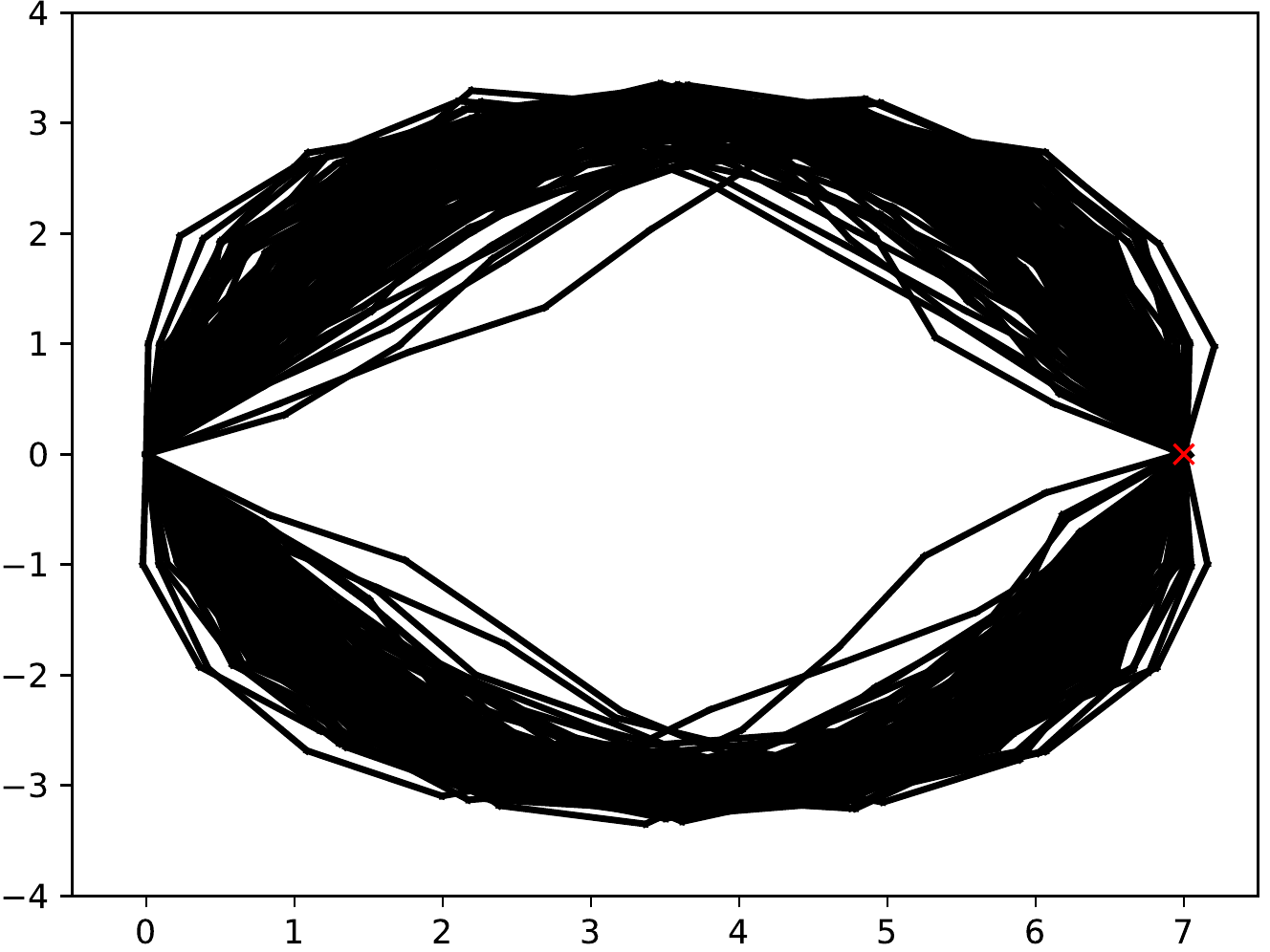}}
		\caption*{PTMCMC}
	\end{subfigure}
		\begin{subfigure}{.24\textwidth}
		\centering
		\includegraphics[width=\linewidth]{\detokenize{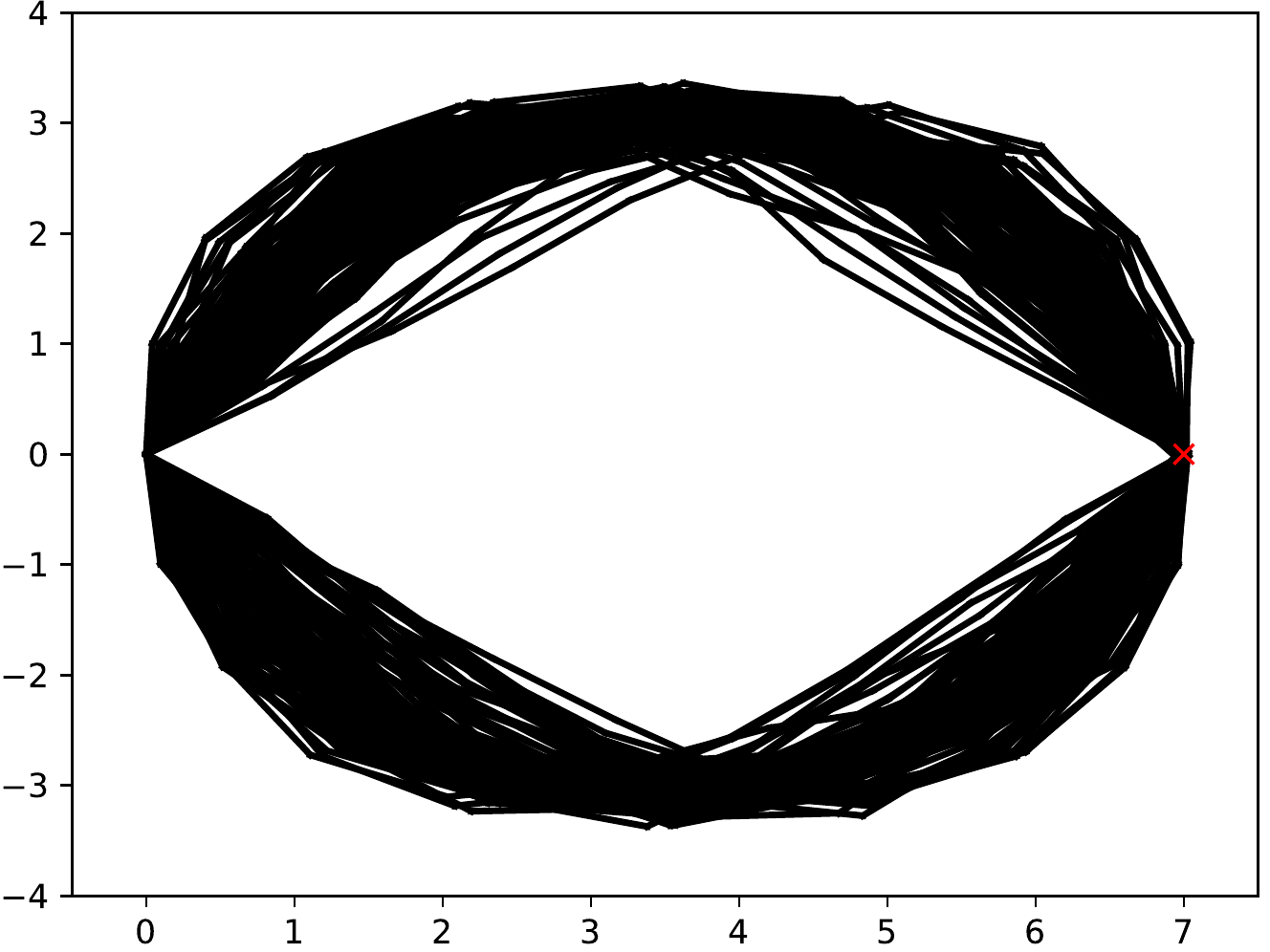}}
		\caption*{VIPS++}
	\end{subfigure}
	    \caption{ 200 sampled configurations are shown for the first training run for the planar robot experiment with a single goal.
	    For the variational inference methods BBVI, IAF and VIPS++, the plots show samples of the final learned model. For PTMCMC, the plots show the 200 most promising samples, which are obtained by applying a sufficient amount of burn-in and using the largest thinning that keeps at least 200 samples in the set.}
	    \label{fig:10link_samples}
\end{figure}

\begin{figure}
	\centering
			\begin{subfigure}{.24\textwidth}
		\centering
		\includegraphics[width=\linewidth]{\detokenize{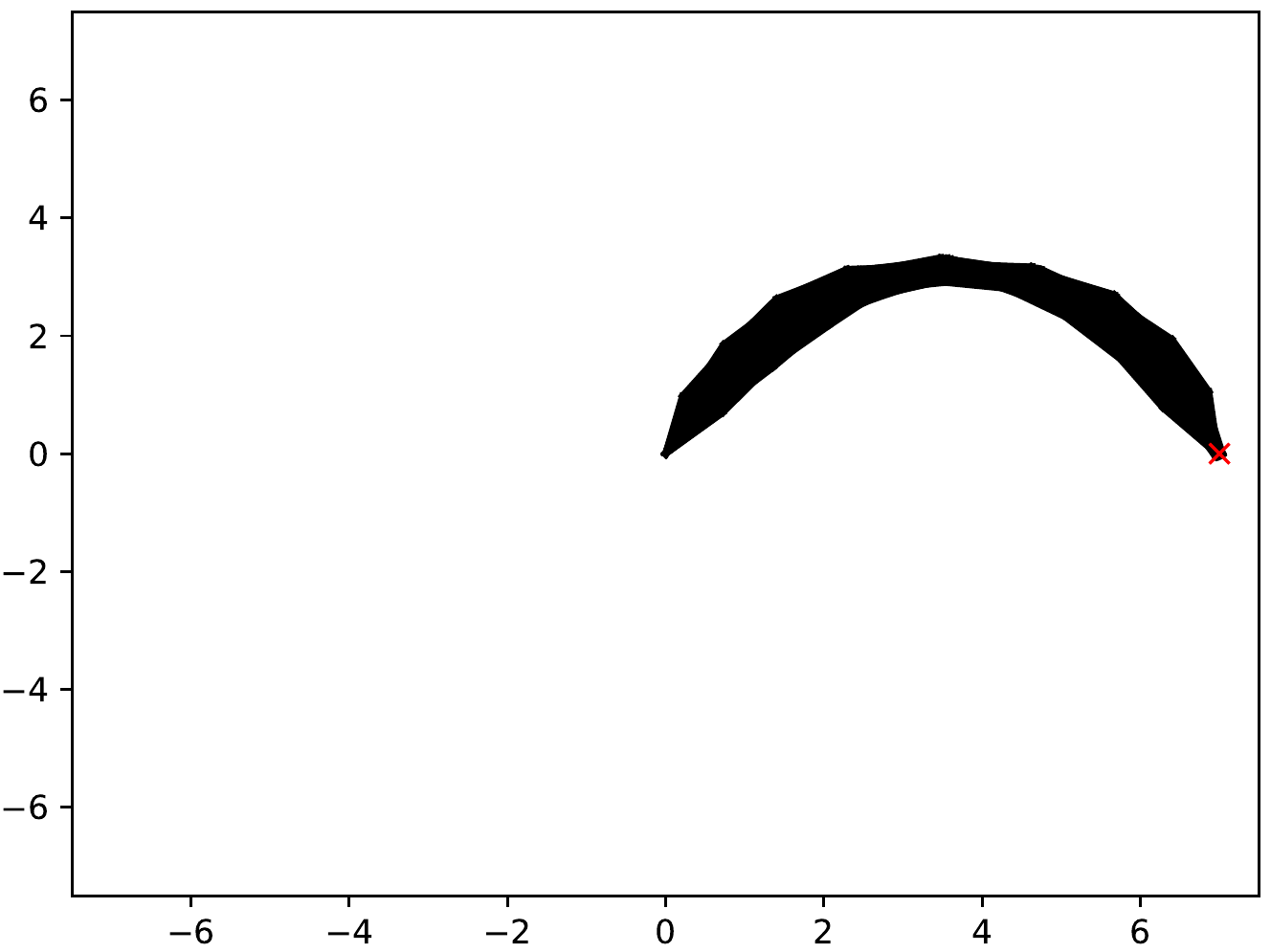}}
		\caption*{BBVI}
	\end{subfigure}
		\begin{subfigure}{.24\textwidth}
		\centering
		\includegraphics[width=\linewidth]{\detokenize{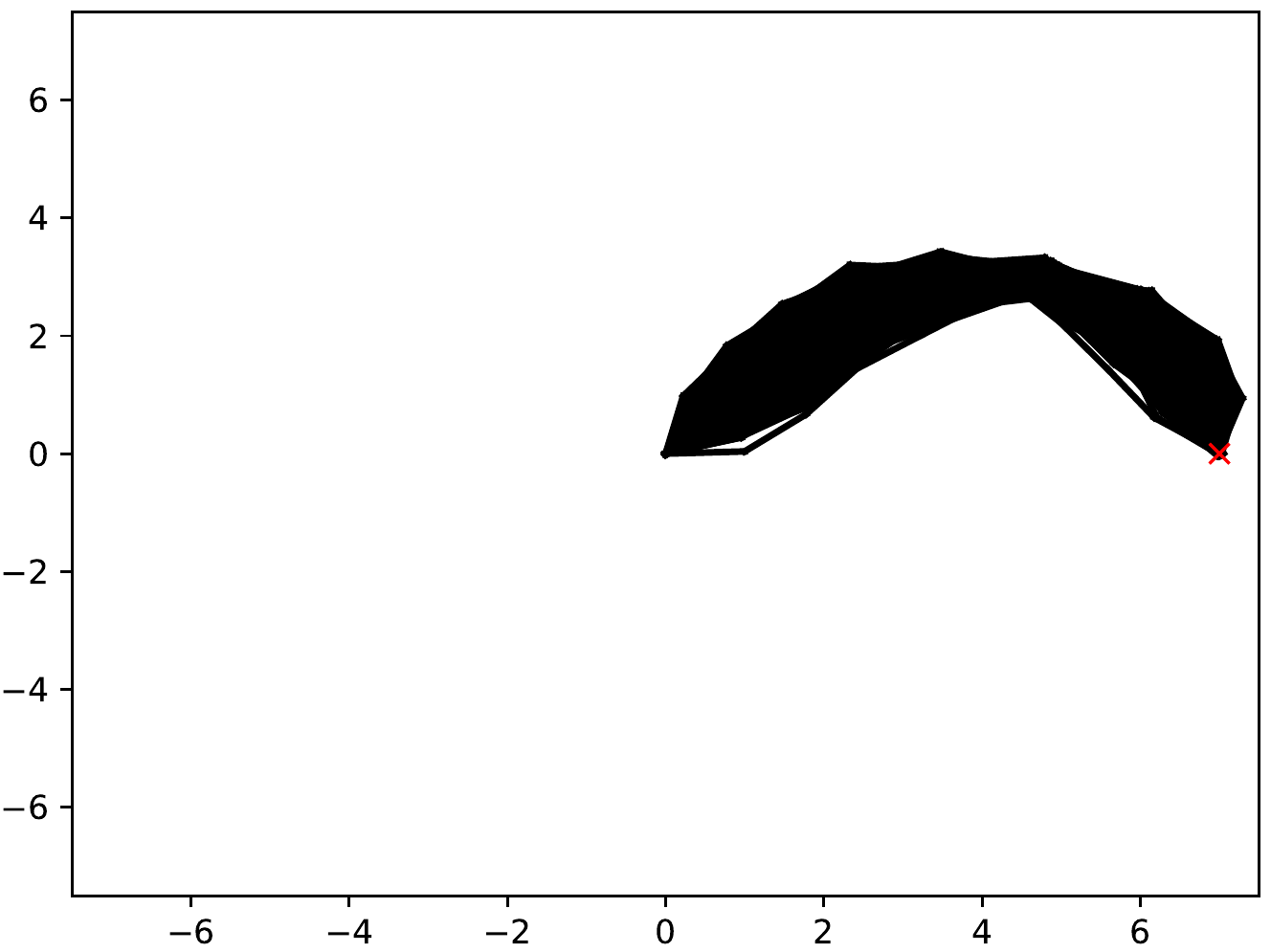}}
		\caption*{IAF}
	\end{subfigure}
	\begin{subfigure}{.24\textwidth}
		\centering
		\includegraphics[width=\linewidth]{\detokenize{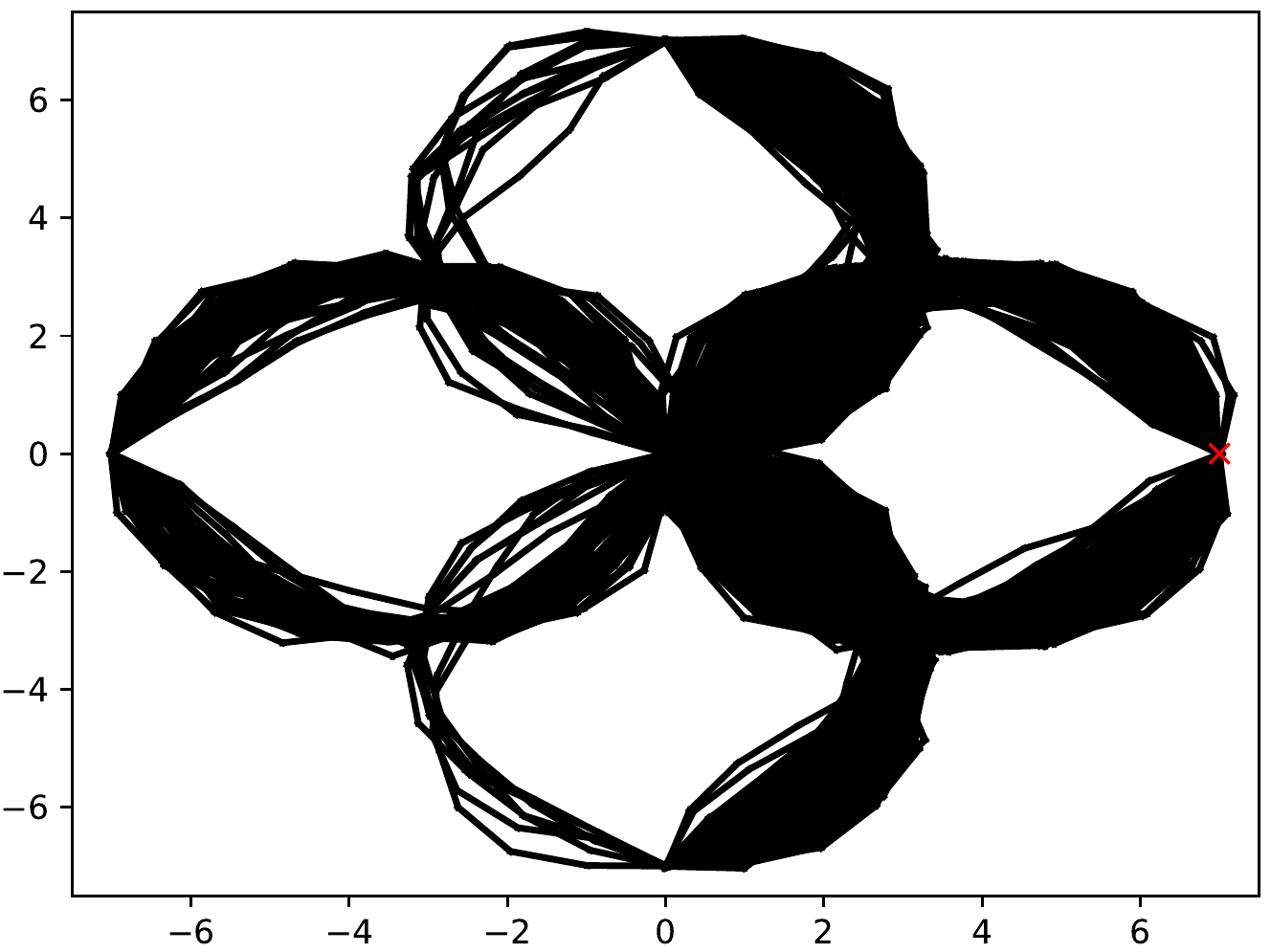}}
		\caption*{PTMCMC}
	\end{subfigure}
		\begin{subfigure}{.24\textwidth}
		\centering
		\includegraphics[width=\linewidth]{\detokenize{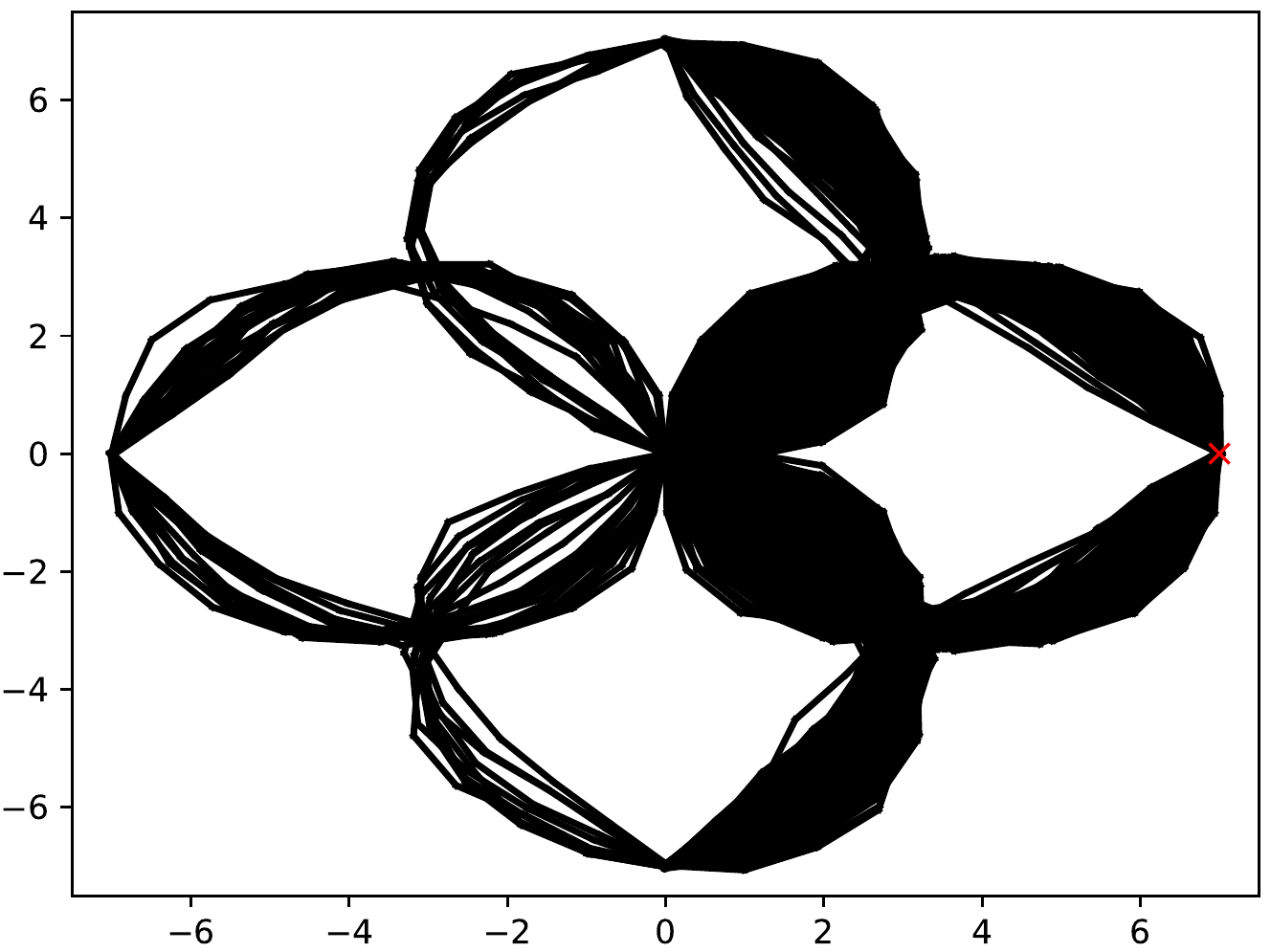}}
		\caption*{VIPS++}
	\end{subfigure}
	    \caption{
	    A thousand sampled configurations are shown for the first training run for the planar robot experiment with four goals.
	    For the variational inference methods BBVI, IAF and VIPS++, the plots show samples of the final learned model. For PTMCMC, the plots show the thousand most promising samples, which are obtained by applying a sufficient amount of burn-in and using the largest thinning that keeps at least thousand samples in the set.}
	    \label{fig:10link4_samples}
\end{figure}

Samples obtained by BBVI, IAF, PTMCMC and VIPS++ for the planar robot experiment with one goal and four goals are shown in Figure~\ref{fig:10link_samples} and Figure~\ref{fig:10link4_samples}, respectively.

\clearpage
\newpage

\end{document}